\documentclass[a4paper, 10pt, reqno]{amsart}
\usepackage[T1]{fontenc}
\usepackage{lmodern}
\usepackage[utf8]{inputenc}

\usepackage{amsmath,amssymb,graphicx,mathtools} 
\usepackage{amsthm}
\usepackage[foot]{amsaddr}
\usepackage[a4paper,left=2.6cm,right=2.6cm,top=3cm,bottom=3.5cm]{geometry}
\usepackage{enumerate}
\usepackage{enumitem}
\usepackage{tikz} 
\usetikzlibrary{arrows,decorations.markings}
\usepackage[colorlinks, linkcolor=red, urlcolor=blue, citecolor=blue]{hyperref}
\hypersetup{breaklinks=true}
\usepackage{url}

\usepackage[
    giveninits=true,
    url=false,
    doi=false,
    isbn=false,
    eprint=true,
    datamodel=mrnumber,
    sorting=nyt, 
    sortcites=false, 
    maxbibnames=99,
    maxcitenames=4,
    backref=false,
    block=space,
    backend=biber,
    style=phys,
    biblabel=brackets
]{biblatex}
\AtEveryBibitem{\clearfield{month}}
\AtEveryCitekey{\clearfield{month}} 
\renewbibmacro{in:}{}
\DeclareFieldFormat[article]{title}{\emph{#1}} 
\DeclareFieldFormat{mrnumber}{\ifhyperref{\href{http://www.ams.org/mathscinet-getitem?mr=#1}{\nolinkurl{MR#1}}}{\nolinkurl{#1}}}
\DeclareFieldFormat{pmid}{\ifhyperref{\href{https://www.ncbi.nlm.nih.gov/pubmed/#1}{\nolinkurl{PMID#1}}}{\nolinkurl{#1}}}
\DeclareFieldFormat{eprint}{\ifhyperref{\href{https://arxiv.org/abs/#1}{\nolinkurl{arXiv:#1}}}{\nolinkurl{#1}}}
\renewbibmacro*{doi+eprint+url}{%
  \iftoggle{bbx:doi}{\printfield{doi}}{}
  \newunit\newblock%
  \printfield{mrnumber}%
  \newunit\newblock%
  \printfield{pmid}%
  \newunit\newblock%
  \printfield{eprint}%
  \iftoggle{bbx:url}{\usebibmacro{url+urldate}}{}}
\bibliography{refs}
\usepackage{caption}
\usepackage{subcaption}

\DeclareMathOperator{\Tr}{Tr}

\newcommand{\norm}[1]{\lVert#1\rVert}
\newcommand{\R}{\mathbb{R}} 
\newcommand{\E}{\mathbb{E}}
\newcommand{\N}{\mathbb{N}} 

\theoremstyle{plain}
\newtheorem{thm}{Theorem}[section]
\newtheorem{cor}[thm]{Corollary}
\newtheorem{lem}[thm]{Lemma}
\newtheorem{prop}[thm]{Proposition}

\theoremstyle{definition}
\newtheorem{defn}[thm]{Definition}

\theoremstyle{remark}
\newtheorem{rmk}[thm]{Remark}

\numberwithin{equation}{section}

\makeatletter
\renewcommand\subsection{\@startsection{subsection}{2}%
  \z@{-.5\linespacing\@plus-.7\linespacing}{.5\linespacing}%
  {\normalfont\scshape}}
\renewcommand\subsubsection{\@startsection{subsubsection}{3}%
  \z@{.5\linespacing\@plus.7\linespacing}{-.5em}%
  {\normalfont\scshape}}
\makeatother

\makeatletter
\@namedef{subjclassname@2020}{%
  \textup{2020} Mathematics Subject Classification}
\makeatother

\usepackage[format=plain,
            labelfont=sc]{caption}
\usepackage[toc,page]{appendix} 
\usepackage{scalerel,stackengine}
\stackMath%
\newcommand\reallywidehat[1]{%
\savestack{\tmpbox}{\stretchto{%
  \scaleto{%
    \scalerel*[\widthof{\ensuremath{#1}}]{\kern.1pt\mathchar"0362\kern.1pt}%
    {\rule{0ex}{\textheight}}
  }{\textheight}%
}{2.4ex}}%
\stackon[-6.9pt]{#1}{\tmpbox}%
}

\makeatletter
\newsavebox\myboxA
\newsavebox\myboxB
\newlength\mylenA
\newcommand*\xoverline[2][0.75]{%
    \sbox{\myboxA}{$\m@th#2$}%
    \setbox\myboxB\null
    \ht\myboxB=\ht\myboxA%
    \dp\myboxB=\dp\myboxA%
    \wd\myboxB=#1\wd\myboxA
    \sbox\myboxB{$\m@th\overline{\copy\myboxB}$}
    \setlength\mylenA{\the\wd\myboxA}
    \addtolength\mylenA{-\the\wd\myboxB}%
    \ifdim\wd\myboxB<\wd\myboxA%
       \rlap{\hskip 0.5\mylenA\usebox\myboxB}{\usebox\myboxA}%
    \else
        \hskip -0.5\mylenA\rlap{\usebox\myboxA}{\hskip 0.5\mylenA\usebox\myboxB}%
    \fi}
\makeatother    

\usepackage{fancyhdr}

\newcommand\shortitle{Langevin dynamics for high-dimensional optimization}
\newcommand\name{Gérard Ben Arous, Cédric Gerbelot, and Vanessa Piccolo}

\begin{document}
\title{Langevin dynamics for high-dimensional optimization: the case of multi-spiked tensor PCA}
\author{Gérard Ben Arous\(^1\)}
\author{Cédric Gerbelot\(^{1,2}\)}
\address{\(^1\)Courant Institute of Mathematical Sciences, New York University}
\email{benarous@cims.nyu.edu}
\email{cedric.gerbelot-barrillon@ens-lyon.fr}
\author{Vanessa Piccolo\(^2\)}
\address{\(^2\)Unité de Mathématiques Pures et Appliquées (UMPA), ENS Lyon}
\email{vanessa.piccolo@ens-lyon.fr}
\subjclass[2020]{68Q87, 62F30, 60G44, 60H30} 
\keywords{High-dimensional optimization, Multi-spiked tensor PCA, Langevin dynamics, Spin glasses, Exact recovery, Subspace recovery}
\date{\today}

\begin{abstract}
We study nonconvex optimization in high dimensions through Langevin dynamics, focusing on the multi-spiked tensor PCA problem. In this tensor estimation model, the goal is to recover a finite number of hidden signal vectors, or spikes, from noisy Gaussian tensor observations using maximum likelihood estimation. We characterize the number of samples required for Langevin dynamics to efficiently recover the spikes and identify the separation conditions on the signal-to-noise ratios (SNRs) needed for exact recovery. In particular, we show that the sample complexity required to recover the spike associated with the largest SNR matches the well-known algorithmic threshold for the single-spike case, whereas the threshold degrades when recovering all spikes. A key ingredient is a precise low-dimensional description of the Langevin trajectory through its correlations with the spikes, which captures both the high-dimensional dynamics and the interactions among competing signal directions.
\end{abstract}

\maketitle	
{
\hypersetup{linkcolor=black}
\tableofcontents
}
\section{Introduction} \label{section:introduction}

Understanding optimization and sampling in high dimensions is a central question across applied mathematics, theoretical physics, and computer science.  The recent empirical success of deep learning methods in data science~\cite{lecun1998gradient, bottou2003large} has challenged our understanding of the efficiency of gradient-based algorithms in navigating nonconvex landscapes. Such landscapes are often random functions composed of a signal component and a noise component. The noise typically introduces an exponential number of critical points (see, e.g.,~\cite{ABC,nica2019,ros2019,piccolo2023}), potentially trapping gradient-based algorithms on exponential time scales. However, when the signal is sufficiently strong, a \emph{topology trivialization transition} occurs, allowing optimization algorithms to efficiently converge to meaningful solutions within polynomial time. One compelling explanation for the remarkable effectiveness of gradient-based methods in high-dimensional, nonconvex problems is that real-world data often possess intrinsic low-dimensional structures. In many applications, the relevant information lies on a manifold of much lower dimension than the ambient space. This structure effectively reduces the complexity of the optimization landscape, allowing algorithms to circumvent the \emph{curse of dimensionality}---the exponential growth of computational difficulty with increasing dimension. Quantitatively understanding how optimization dynamics exploit these properties to reach near-optimal regions of the landscape within feasible time scales remains a fundamental question.

In this context, \emph{Tensor Principal Component Analysis (Tensor PCA)} problem serves as a canonical model for studying such questions. Tensor PCA involves estimating an unknown vector on the \((N-1)\)-dimensional sphere from noisy Gaussian observations of an associated tensor. The problem was first formulated for matrices by Johnstone~\cite{Johnstone} and later generalized to higher-order tensors by Richard and Montanari~\cite{MontanariRichard}, where tensors of order \(2\) correspond to matrices. Recent work by the first author, in collaboration with Gheissari and Jagannath~\cite{benarousneurips, ben2022effective}, showed that for a broad class of problems, the high-dimensional dynamics of stochastic gradient descent (SGD) can be effectively reduced to low-dimensional, autonomous \emph{effective dynamics} governed by a set of summary statistics. This builds on earlier work by the same authors~\cite{ben2020bounding, arous2020algorithmic} analyzing Langevin dynamics in the single-spike tensor PCA model, where they showed that the high-dimensional dynamics reduce to a one-dimensional dynamical system capturing the evolution of the inner product between the estimator and the true signal.

Building on these advances, the present work extends the analysis to the \emph{multi-spike setting}, providing new insights into the behavior of multiple interacting summary statistics. Specifically, we consider the task of recovering \(r\) hidden signal vectors on the sphere \(\mathbb{S}^{N-1}(\sqrt{N})\) from noisy Gaussian tensor observations. Our goal is to recover these vectors via maximum likelihood estimator (MLE) and to determine the sample complexity required for Langevin dynamics to achieve successful recovery, namely convergence to the global minimizer of the nonconvex loss landscape. To simplify the analysis, we assume the signal vectors are orthogonal, so that the MLE optimization naturally takes place on the normalized \emph{Stiefel manifold}, i.e., the set of \(N \times r\) matrices with orthogonal columns of norm \(\sqrt{N}\). 

We study Langevin dynamics as a valuable proxy for SGD. From an estimation standpoint, it closely mirrors key statistical features of SGD algorithms in high-dimensional nonconvex problems, while remaining significantly more amenable to precise analysis. Prior works~\cite{arous2020algorithmic, saraomannelli2020, liang2023} have emphasized that Langevin dynamics provides a mathematically tractable model for understanding the dynamics of gradient-based algorithms. Although algorithms achieving optimal sample complexity are known for tensor PCA (e.g., approximate message passing, tensor power iteration, sum-of-squares-based methods; see Subsection~\ref{related works}), these approaches typically exploit the specific algebraic structure of the tensor and differ significantly from the gradient-based methods commonly used in practice. By contrast, Langevin dynamics serves as a generic diffusion-based analogue of SGD and therefore offers insight into the behavior of practical optimization procedures. Moreover, from a physics perspective, Langevin dynamics corresponds to the natural relaxation dynamics of the system toward thermal equilibrium for the Hamiltonian of the (spiked) \(p\)-spin glass model underlying tensor PCA. Its stochastic fluctuations allow the process to explore the energy landscape and reveal how geometric and thermodynamic properties influence the optimization trajectory. This viewpoint provides a useful bridge between stochastic analysis and statistical physics for studying nonconvex estimation problems. Building on this perspective, we extend the \emph{bounding flows method}, originally developed in~\cite{ben2020bounding} and applied to the single-spike case in~\cite{arous2020algorithmic}, to the setting of diffusions on the Stiefel manifold. This extension yields a full characterization of the resulting \(r^2\)-dimensional effective dynamics. In the multi-spike regime, we uncover a substantially richer phenomenology than in the single-spike setting, arising from the intricate interactions among the trajectories of the inner products between the estimators and the true spikes. These interactions generate complex dynamical behaviors, as the estimators compete to align with different signal directions, and this competition fundamentally shapes the recovery process. More precisely, in the single spike case, the aforementioned prior works~\cite{arous2020algorithmic, saraomannelli2020, liang2023} show that recovery occurs when the drift induced by the signal outweighs the Brownian fluctuations. In contrast, in our multi-spike setting, we show that, beyond determining the recovery threshold, we also need to impose a separation condition between successive signal-to-noise ratios (SNRs) to ensure that the ordering of the hidden vectors is preserved throughout the dynamics despite the influence of Brownian motion.\\

This article is part of a series of three companion papers studying high-dimensional optimization in the multi-spiked tensor PCA model. The present work focuses on Langevin dynamics, while~\cite{benarous2024} studies the discrete-time online SGD algorithm, and~\cite{benarous2024gradientflow} analyzes gradient flow, which can be viewed as the zero-temperature limit of Langevin dynamics. 

Our main contributions are as follows. We determine the sample complexity and the precise SNR conditions under which Langevin dynamics achieves \emph{exact recovery} of all \(r\) hidden spikes, for both tensors (\(p \ge 3\)) and matrices (\(p=2\)). If \(\boldsymbol{v}_1, \ldots, \boldsymbol{v}_r\) denote the unknown orthogonal vectors and \(\boldsymbol{x}_1, \ldots, \boldsymbol{x}_r\) their respective Langevin estimators, our goal is to reach
\[
\langle \boldsymbol{v}_i, \boldsymbol{x}_i \rangle = 1-o(1),
\] 
with high probability in the large-\(N\) limit (see Definition~\ref{def: exact recovery}). We therefore identify the exact threshold on the number of samples and the necessary separation between successive SNRs ensuring exact recovery. Our key finding is that the recovery process exhibits a \emph{sequential elimination phenomenon} (see Definition~\ref{def: sequential elimination}). The hidden spikes are recovered one after another: once \(\langle \boldsymbol{v}_i, \boldsymbol{x}_i \rangle\) becomes macroscopic, the cross-correlations \(\langle \boldsymbol{v}_i, \boldsymbol{x}_j \rangle\) and \(\langle \boldsymbol{v}_j, \boldsymbol{x}_i \rangle\) for \(j > i\) quickly become negligible, which in turn enables the subsequent growth of \(\langle \boldsymbol{v}_{i+1}, \boldsymbol{x}_{i+1} \rangle\). This mechanism reveals a hierarchical structure in the dynamics and provides a refined picture of high-dimensional estimation, where individual correlations evolve in a strongly coupled manner. For the matrix case (\(p=2\)), we further analyze the regime in which all SNRs are identical. In this setting, the individual correlations cannot be separated dynamically, and the appropriate notion of recovery becomes that \emph{subspace recovery}.

A distinctive feature of Langevin dynamics is the presence of a continuous noise term modeled by Brownian motion, which introduces additional analytical challenges---particularly when the SNRs are not well separated. In contrast, the discrete-time SGD dynamics studied in~\cite{benarous2024} and the continuous-time gradient flow dynamics analyzed in~\cite{benarous2024gradientflow} do not include such stochastic noise, allowing for finer control of coordinate-level fluctuations. In these two companion works, we establish recovery of the spikes \emph{up to a permutation} without assuming any separation condition on the SNRs. Despite these differences, we show that the recovery of the first spike under Langevin dynamics occurs with the same sample complexity as in the single-spike case (\(r=1\)), demonstrating that the presence of additional components does not hinder the recovery of the leading signal direction. However, achieving recovery of all spikes is slightly suboptimal compared to the discrete-time SGD results that we obtain in~\cite{benarous2024}, which differ only by logarithmic factors in the sample complexity compared to the single-spike case. We emphasize that this suboptimality arises from technical limitations of our current proof strategy rather than any intrinsic difference between the underlying dynamics; a more detailed discussion is provided in Subsection~\ref{outline proofs}. Finally, the estimates derived here for Langevin dynamics form the analytical backbone for our companion papers~\cite{benarous2024, benarous2024gradientflow}. In particular, the results established in this work provide a complete characterization of the continuous-time population dynamics, which serve as the foundational step for the analysis of both the discrete-time and noiseless (gradient flow) settings.

\subsection{Setting}

We formalize the multi-rank spiked covariance and tensor model as follows. Our goal is to recover the underlying signal directions from noisy observations using a likelihood-based approach, and to analyze the associated optimization dynamics through the lens of Langevin dynamics. 

Let \(p \ge 2\) and \(r \ge 1\) be fixed integers. Suppose we observe \(M\) independent Gaussian \(p\)-tensors \(\boldsymbol{Y}^1, \ldots, \boldsymbol{Y}^M\) of dimension \(N\), each of the form
\begin{equation}\label{eq: spiked tensor model}
\boldsymbol{Y}^\ell =  \sum_{i=1}^r \lambda_i \sqrt{N} \left( \frac{\boldsymbol{v}_i}{\sqrt{N}} \right)^{\otimes p} + \boldsymbol{W}^\ell,
\end{equation}
where the parameters \(\lambda_1 \geq \ldots \geq \lambda_r \geq 0\) are the signal-to-noise ratios (SNRs) and the noise tensors \(\boldsymbol{W}^\ell\) have i.i.d.\ entries \(W^\ell_{i_1, \ldots, i_p} \sim \mathcal{N}(0,1)\). The unknown signal vectors \(\boldsymbol{v}_1, \ldots,\boldsymbol{v}_r\) are assumed to be \emph{orthogonal} vectors and to lie on the \((N-1)\)-dimensional unit sphere of radius \(\sqrt{N}\). Equivalently, the matrix \(\boldsymbol{V} = [\boldsymbol{v}_1, \ldots, \boldsymbol{v}_r] \in \R^{N \times r}\) lies on the \emph{normalized Stiefel manifold}
\[
\mathcal{S}_{N,r} \coloneqq \{\boldsymbol{X} \in \R^{N \times r} \colon \boldsymbol{X}^\top \boldsymbol{X} = N \boldsymbol{I}_r \}.
\]
Our goal is to estimate the unknown directions \(\boldsymbol{v}_1, \ldots, \boldsymbol{v}_r\) via empirical risk minimization. In particular, we choose the most common method for Gaussian variables, namely maximum likelihood estimation. This involves minimizing the empirical risk \(\hat{\mathcal{R}}_{N,r}(\boldsymbol{X})\) 
\begin{equation} \label{empirical risk minimization}
\hat{\mathcal{R}}_{N,r}(\boldsymbol{X}) = \frac{1}{M} \sum_{\ell=1}^M \mathcal{L}_{N,r}(\boldsymbol{X};\boldsymbol{Y}^\ell),
\end{equation}
over the constraint set \( \mathcal{S}_{N,r}\), where the loss function \(\mathcal{L}_{N,r} \colon \mathcal{S}_{N,r} \times (\R^N)^{\otimes p} \to \R\) is chosen as the \(\log\)-likelihood function and takes the form
\begin{equation} \label{eq: loss function}
\begin{split}
\mathcal{L}_{N,r}(\boldsymbol{X}; \boldsymbol{Y}^\ell) & = - \sum_{i=1}^r \lambda_i \sqrt{N} \left \langle \boldsymbol{Y}^\ell, \left ( \frac{\boldsymbol{x}_i}{\sqrt{N}}\right )^{\otimes p} \right \rangle, \\
& = - N^{- \frac{p-1}{2}} \sum_{i=1}^r \lambda_i \langle\boldsymbol{W}^\ell, \boldsymbol{x}_i^{\otimes p} \rangle - \sum_{1 \le i,j \le r} N \lambda_i \lambda_j \left( m_{ij}^{(N)}(\boldsymbol{X}) \right)^p,
\end{split}
\end{equation}
where \(m_{ij}^{(N)}(\boldsymbol{X}) \coloneqq N^{-1} \langle \boldsymbol{v}_i, \boldsymbol{x}_j \rangle\) denotes the \emph{correlation} between \(\boldsymbol{v}_i\) and \(\boldsymbol{x}_j\). We remark that the MLE loss naturally incorporates the SNRs. In contrast, in the single-spike case the prefactor \(\lambda\) simply rescales the objective, so minimizing \(\lambda \sqrt{N} \langle \boldsymbol{Y}^\ell, (N^{-1} \boldsymbol{x})^{\otimes p} \rangle\) is equivalent to minimizing the unweighted term \(\langle \boldsymbol{Y}^\ell, (N^{-1} \boldsymbol{x})^{\otimes p} \rangle\).

We note from~\eqref{empirical risk minimization} that optimizing the empirical risk \(\hat{\mathcal{R}}_{N,r}\) is equivalent in law to minimizing
\[
\mathcal{R} (\boldsymbol{X}) = \frac{1}{\sqrt{M}} H_0 (\boldsymbol{X}) - \sum_{1 \le i,j \le r} N \lambda_i \lambda_j \left( m_{ij}^{(N)}(\boldsymbol{X}) \right)^p,
\]
where 
\begin{equation} \label{eq: Hamiltonian H0}
H_0(\boldsymbol{X}) = N^{-\frac{p-1}{2}} \sum_{i=1}^r \lambda_i  \langle \boldsymbol{W}, \boldsymbol{x}_i^{\otimes p} \rangle.
\end{equation}
The noise term \(H_0\) is a centered Gaussian process with covariance of the form 
\[
\E \left[H_0 (\boldsymbol{X})H_0(\boldsymbol{Y})\right] = N \sum_{1\leq i,j\leq r} \lambda_i \lambda_j \left(\frac{\langle \boldsymbol{x}_i, \boldsymbol{y}_j \rangle}{N}\right)^p.
\]
In our case, it is convenient to rescale the empirical risk by \(\sqrt{M}\) and introduce the corresponding Hamiltonian \(H \colon \mathcal{S}_{N,r} \to \R\):
\begin{equation} \label{eq: Hamiltonian}
H (\boldsymbol{X}) = H_0(\boldsymbol{X}) - \sum_{1 \leq i,j \leq r} N \sqrt{M} \lambda_i \lambda_j \left (m_{ij}^{(N)}(\boldsymbol{X}) \right )^p.
\end{equation}
Minimizing the Hamiltonian~\eqref{eq: Hamiltonian} is thus equivalent, in law, to minimizing the empirical risk~\eqref{empirical risk minimization}, while the scaling by \(\sqrt{M}\) allows the corresponding dynamics to evolve on an $\mathcal{O}(1)$ time scale.

\subsection{Langevin dynamics}
We now define the Langevin dynamics with Hamiltonian \(H\). For \(\beta \in (0,\infty)\) and an initial point \(\boldsymbol{X}_0 \in \mathcal{S}_{N,r}\), let \(\boldsymbol{X}_t^\beta \in \mathcal{S}_{N,r}\) solve the stochastic differential equation given by 
\begin{equation} \label{eq: langevin dynamics}
d \boldsymbol{X}^\beta_t = \sqrt{2} d \boldsymbol{B}_t - \beta \nabla H(\boldsymbol{X}_t^\beta) dt,
\end{equation}
where \(\boldsymbol{B}_t \in \R^{N \times r}\) is a Brownian motion on \(\mathcal{S}_{N,r}\) and \(\nabla\) denotes the Riemannian gradient on \(\mathcal{S}_{N,r}\). Specifically, for any smooth function \(f \colon \mathcal{S}_{N,r} \to \R\), 
\begin{equation}  \label{eq: riemannian gradient on normalized stiefel}
\nabla f(\boldsymbol{X}) = \hat{\nabla} f(\boldsymbol{X}) - \frac{1}{2N} \boldsymbol{X} \left (\boldsymbol{X}^\top \hat{\nabla} f (\boldsymbol{X}) + \hat{\nabla} f (\boldsymbol{X})^\top \boldsymbol{X} \right ),
\end{equation}
where \(\hat{\nabla}\) denotes the Euclidean gradient. The generator \(L_\beta\) of this process is given by
\begin{equation} \label{eq: generator Langevin}
L_\beta = \Delta- \beta \langle \nabla H, \hat{\nabla}\cdot \rangle,
\end{equation}
where \(\Delta\) denotes the Laplace-Beltrami operator on \(\mathcal{S}_{N,r}\) (endowed with the Euclidean metric) and is given according to~\cite{YuanLaplaceStiefel} by
\begin{equation} \label{eq: stiefel laplace}
\Delta f (\boldsymbol{X})= \hat{\Delta} f (\boldsymbol{X}) - \frac{N-1}{N} \Tr (\boldsymbol{X}^\top \hat{\nabla} f (\boldsymbol{X})) - \frac{1}{N} \sum_{i,k=1}^r \sum_{j,\ell=1}^N (x_i)_j (x_k)_\ell \frac{\partial^2}{ \partial (x_i)_j \partial (x_k)_\ell} f (\boldsymbol{X}) ,
\end{equation}
for any function \(f \colon \mathcal{S}_{N,r} \to \R\). Here, \(\hat{\Delta}\) denotes the Euclidean Laplacian. We also write \(L_{0,\beta}\) for the generator of Langevin dynamics related to the noise Hamiltonian \(H_0\), i.e.,
\begin{equation} \label{eq: generator Langevin noise}
L_{0,\beta} = \Delta - \beta \langle \nabla H_0, \hat{\nabla}\cdot \rangle.
\end{equation}

\begin{rmk}
Langevin dynamics reduces to gradient flow when \(\beta = \infty\). Accordingly, all results stated for Langevin dynamics in this paper also hold under gradient flow dynamics. In our companion paper~\cite{benarous2024gradientflow}, we analyze the gradient flow case without assuming any separation among the SNRs, and show that gradient flow achieves recovery of all spikes up to permutation with the same sample complexity.
\end{rmk}

\subsection{Main results} \label{subsection: main asymptotic results}

Our main results concern the \emph{sample complexity}, i.e., the number \(M\) of observations required to efficiently recover the unknown orthogonal vectors \(\boldsymbol{v}_1, \ldots, \boldsymbol{v}_r\) via Langevin dynamics~\eqref{eq: langevin dynamics}, under the assumption that the SNRs \(\lambda_1 \ge \cdots \ge \lambda_r \ge 0\) are fixed constants of order one. Throughout, we consider the process \((\boldsymbol{X}_t^\beta)_{t \ge 0}\) defined by~\eqref{eq: langevin dynamics}, initialized from the uniform distribution \(\mu_{N \times r}\) on the Stiefel manifold \(\mathcal{S}_{N,r}\). The measure \(\mu_{N \times r}\) is the unique probability measure on \(\mathcal{S}_{N,r}\) that is invariant under both the left- and right-orthogonal transformations. Sampling from this distribution is straightforward: if \(\boldsymbol{Z} \in \R^{N \times r}\) is a random matrix filled with i.i.d.\ standard normal entries, then \(\boldsymbol{Z} \left ( \frac{1}{N} \boldsymbol{Z}^\top \boldsymbol{Z} \right)^{-1/2}\) is uniformly distributed on \(\mathcal{S}_{N,r}\)~\cite{chikuse2012statistics}. We denote by \((\Omega, \mathcal{F}, \mathbb{P})\) the probability space on which the random tensors \((\boldsymbol{W}^\ell)_\ell\) are defined, and by \(\mathbb{Q}_{\boldsymbol{X}_0}\) the law of the Langevin process \((\boldsymbol{X}_t^\beta)_{t \ge 0}\) initiated at \(\boldsymbol{X}_0 \sim \mu_{N \times r}\). \\
\emph{Notations.} For a positive integer \(n \in \N\), we write \([n] \coloneqq \{1, \ldots, n\}\). For two sequences \(x_N\) and \(y_N\), we write \(x_N \gtrsim y_N\) to indicate that \(x_N \ge C y_N\) for a universal constant \(C>0\).\\

Our main results address both the recovery of the leading spike \(\boldsymbol{v}_1\) and the \emph{exact recovery} of all spikes, under suitable separation conditions on the SNRs. We begin by formalizing the notion of recovery under Langevin dynamics.

\begin{defn}[Exact recovery] \label{def: exact recovery}
Let \(m \le r\) and \(\xi \in [0,1]\). We say that Langevin dynamics \emph{exactly recovers} the first \(m\) signal vectors \(\boldsymbol{v}_1, \ldots, \boldsymbol{v}_m\) with \emph{asymptotic success rate} \(\xi\) if, for every \(\varepsilon > 0\), there exists \(T_0 = T_0 (\varepsilon) > 0\) such that for every \(T \ge T_0\) and every \(i \in [m]\),
\[
\lim_{N \to \infty} \int_{\mathcal{S}_{N,r}} \mathbb{Q}_{\boldsymbol{X}_0} \left ( \min_{t \in [T_0,T]} |m_{ii}^{(N)} (\boldsymbol{X}_t^\beta)| \ge 1 - \varepsilon \right ) d \mu_{N \times r} = \xi, \quad \mathbb{P}\textnormal{-a.s.}
\]
We say that Langevin dynamics \emph{(strongly) recovers the leading spike} \(\boldsymbol{v}_1\) when \(m=1\).
\end{defn}

The number of spikes that can be recovered depends on both parity of the tensor order \(p\) and on the sign of the correlations at initialization. Define 
\[
r_\textnormal{c} \coloneqq \sum_{i=1}^r \mathbf{1}_{\left \{ \left ( m_{ii}^{(N)} (\boldsymbol{X}_0) \right)^{p-2}>0 \right \}},
\]
which counts the number of signal directions for which the initial correlation has the correct sign to allow recovery. Since \(\boldsymbol{X}_0\) is random, \(r_c\) is a random variable which depends on the initialization \(\boldsymbol{X}_0\). If \(p\) is even, or if \(m_{ii}^{(N)} (\boldsymbol{X}_0) >0\) for all \(i \in [r]\), then \(r_\textnormal{c} =r\). 

We now present our asymptotic recovery results, starting with the case \(p \ge 3\). A nonasymptotic version, with explicit constants and convergence rates, is provided later in Section~\ref{section: Langevin}.

\begin{thm}[Exact recovery for \(p\geq 3\)] \label{thm: exact recovery Langevin p>2 asymptotic}
Fix any \(p \ge 3\) and \(\beta \in (0,\infty)\). 
\begin{itemize}
\item[(a)] \textbf{Recovery of the leading spike.} If \( \left ( m_{11}^{(N)} (\boldsymbol{X}_0) \right)^{p-2}>0\), \(M = N^\alpha\) with \(\alpha > p-2\), and there exists \(\eta > 1\) with
\[
\lambda_1 > C \left (\eta\sqrt{\log(\eta)} \right )^{p-2}  \lambda_2 ,
\]
for some universal constant \(C>0\), then Langevin dynamics strongly recovers the spike \(\boldsymbol{v}_1\) with asymptotic success rate \(\xi = 1 - \frac{1}{\eta}\).
\item[(b)] \textbf{Exact recovery of multiple spikes.} Let \(I_\textnormal{c} \coloneqq \{i \in [r] \colon (m_{ii}^{(N)} (\boldsymbol{X}) )^{p-2} >0\}\), and let \(i_1, \ldots, i_{r_\textnormal{c}}\) denote these indices so that \(\lambda_{i_1} \ge \cdots \ge \lambda_{i_{r_\textnormal{c}}}\). Then, if \(M \gtrsim N^{p-1}\) and there exists \(\eta > 1\) with
\[
\lambda_{i_k} > C \left (\eta\sqrt{\log(\eta)} \right)^{p-2}  \lambda_{i_{k+1}}, \quad \textnormal{for all} \enspace 1 \le k \le r_\textnormal{c}-1
\]
for some universal constant \(C>0\), then Langevin dynamics exactly recovers the spikes \(\boldsymbol{v}_{i_1}, \ldots, \boldsymbol{v}_{i_{r_{\textnormal{c}}}}\) with asymptotic success rate \(\xi = 1 - \frac{1}{\eta}\).
\end{itemize}
\end{thm}

The first statement ensures recovery of the leading spike with probability arbitrarily close to \(1\), provided that \(\lambda_1\) is separated from \(\lambda_2\) by a large but \(N\)-independent factor. See Proposition~\ref{thm: strong recovery first spike Langevin p>2} for a detailed discussion of this separation condition. The sample complexity required to recover the leading spike \(\boldsymbol{v}_1\) is at least \(N^{p-2 + \delta}\) for any fixed \(\delta > 0\), consistent with the single-spike result~\cite{arous2020algorithmic}. It is conceivable that the slack \(N^\delta\) could be reduced to a polylogarithmic factor by sharpening several estimates; we leave this as an open question. We note that in the setting of tensor power iteration for the single spike case, Wu and Zhou~\cite{Wu24} recently improved the analogous \(N^\delta\) slack to a polylogarithmic factor by exploiting a reduction to a one-dimensional stochastic process, whose simple form enables a precise analysis. Extending such refinements to the multi-spike Langevin dynamics considered here appears substantially more challenging, due to the interacting evolution of the correlations and the presence of continuous stochastic noise.

Exact recovery of multiple spikes holds when \(M \ge c N^{p-1}\) for some constant \(c>0\), under the same SNR separation condition as in item (a). The gap between the \(N^{p-2 + \delta}\) complexity for the first spike and the \(N^{p-1}\) complexity for subsequent spikes has the following origin. At initialization the estimator \(\boldsymbol{X}_0\) is independent of the noise tensor \(\boldsymbol{W}\). Under this independence, the typical scale of the noise generator \(L_{0,\beta} m_{ij}^{(N)}(\boldsymbol{X})\) is of order \(N^{-1/2}\). In contrast, without such independence one only has an \(\mathcal{O}(1)\) bound. This \(N^{-1/2}\) scaling reduces the sample complexity needed for estimating the signal component of the generator from \(N^{p-1}\) down to \(N^{p-2}\), provided one can leverage the initialization independence over a sufficiently long time window to recover the first direction. However, once the first spike is recovered, our proof can no longer exploit this advantageous scaling, and the relevant estimates revert to the \(\mathcal{O}(1)\) bound, yielding the \(N^{p-1}\) complexity for subsequent spikes. A more detailed discussion of this mechanism is given in Section~\ref{outline proofs}.

\begin{rmk}
We do not currently have lower bounds excluding sample complexities strictly between \(N^{p-2}\) and \(N^{p-1}\) for exact recovery of all spikes, so the optimality of the \(N^{p-1}\) rate remains open. In our companion paper on the online SGD algorithm~\cite{benarous2024}, we obtain the sharper \(N^{p-2}\) threshold for recovery of all directions. The key difference is that in online SGD each step uses an independent sample \(\boldsymbol{W}^\ell\), independent of the previous iterate; this  preserves the \(N^{-1/2}\) scaling of the noise contribution to the generator at every time step, allowing the improved complexity to propagate beyond the first direction.
\end{rmk}

\begin{rmk}
When the tensor order \(p \ge 3\) is even, each estimator \(\boldsymbol{x}_i\) recovers \(\operatorname{sgn}(m_{ii}^{(N)}(\boldsymbol{X}_0))\,\boldsymbol{v}_i\).  
Thus, a positive initial correlation implies recovery of \(\boldsymbol{v}_i\), while a negative initial correlation leads to recovery of \(-\boldsymbol{v}_i\). In contrast, when \(p\) is odd, the dynamics only amplifies positive correlations: if \(m_{ii}^{(N)}(\boldsymbol{X}_0) > 0\), then \(\boldsymbol{x}_i\) converges to \(\boldsymbol{v}_i\); but if \(m_{ii}^{(N)}(\boldsymbol{X}_0) < 0\), the correlation remains trapped near the equator and does not grow, so recovery fails in that direction. This is why, without making any assumptions on the signs of the initial correlations, Langevin dynamics can guarantee exact recovery of only a subset of size \(r_\textnormal{c}\) of the spikes.
\end{rmk}

In the multi-spike regime, the Langevin dynamics does not recover all correlations simultaneously. Instead, the process exhibits a hierarchical behavior and the correlations \(\{m_{ii}^{(N)}(\boldsymbol{X}_t^\beta)\}_{i=1}^{r_\textnormal{c}}\) reach macroscopic values sequentially, one after another. Once a correlation \(m_{ii}^{(N)}\) becomes sufficiently large, all cross-correlations \(m_{ij}^{(N)}\) and \(m_{ji}^{(N)}\) for \(j > i\) start to decay and rapidly reach negligible values. The vanishing of these off-diagonal correlations allows the next diagonal term \(m_{i+1 \, i+1}^{(N)}\) to grow to a macroscopic level. We refer to this hierarchical recovery process as \emph{sequential elimination}, formalized below.

\begin{defn}[Sequential elimination] \label{def: sequential elimination}
Let \(m \le r\). We say that the correlations \(\{m_{ii}^{(N)}\}_{i=1}^m\) exhibit \emph{sequential elimination} if, for every \(\varepsilon, \varepsilon' > 0\), there exist stopping times \(T_1 \leq \cdots \leq T_m\) such that for every \(i \in [m]\) and every \(T \ge T_i\),
\[
| m_{ii}^{(N)}(\boldsymbol{X}_T^\beta)| \geq 1 - \varepsilon  \quad \textnormal{and} \quad |m_{ij}^{(N)}(\boldsymbol{X}_T^\beta)| \leq \varepsilon', |m_{ji}^{(N)}(\boldsymbol{X}_T^\beta)| \leq \varepsilon' \enspace \forall \, j > i.
\]
\end{defn}

The next result shows that, under the sample complexity \(M \gtrsim N^{p-1}\), Langevin dynamics satisfies this deterministic property with probability \(1\) in the large-\(N\) limit.

\begin{thm}[Sequential recovery of the spikes] \label{thm: sequential elimination}
If \(M \gtrsim N^{p-1}\), then with probability \(1\) in the large-\(N\) limit Langevin dynamics exhibits sequential elimination for the correlations \(\{m_{ii}^{(N)}(\boldsymbol{X}_t^\beta)\}_{i=1}^{r_\textnormal{c}}\). 
\end{thm}

We now turn to the case \(p=2\). In this setting, we obtain results analogous to those for \(p \ge 3\). However, the required separation between the SNRs is substantially less restrictive, as an order-\(1\) factor already suffices for exact recovery. 

\begin{thm}[Exact recovery for \(p=2\)] \label{thm: exact recovery Langevin p=2 asymptotic}
Fix \(p =2\) and any \(\beta \in (0,\infty)\). 
\begin{itemize}
\item[(a)] \textbf{Recovery of the leading spike.} Suppose that \(\lambda_1 = \lambda_2 (1+\kappa_1)\) for some constant \(\kappa_1 >0\) of order one. If \(M = N^\delta\) for any fixed \(\delta>0\), then Langevin dynamics strongly recovers the spike \(\boldsymbol{v}_1\) with asymptotic success rate \(\xi = 1\).
\item[(b)] \textbf{Exact recovery of multiple spikes.} Suppose that \(\lambda_i = \lambda_{i+1} (1+\kappa_i)\), where the constants \(\kappa_i >0\) are of order one. If \(M \ge N^{1 - \frac{\lambda_r^2}{\lambda_1^2} + \delta}\) for any fixed \(\delta>0\), then Langevin dynamics exactly recovers all \(r\) spikes with asymptotic success rate \(\xi = 1\).
\end{itemize}
\end{thm}

Statement (a) shows that Langevin dynamics strongly recovers the leading spike with sample complexity \(M=N^\delta\) for any fixed \(\delta >0\). Compared to Theorem~\ref{thm: exact recovery Langevin p>2 asymptotic}(a) for \(p \ge3\), this result shows that the separation condition required for the SNRs to ensure strong recovery of the leading spike is an order-\(1\) factor. Statement (b) further shows that exact recovery of all spikes is possible as soon as the SNRs are separated by order-\(1\) factors and the number of samples scales as \(N^\xi\), where \(\xi \in (0,1)\) depends on the ratio between the signal sizes. As in the case \(p \ge 3\), we again observe a difference in sample complexity between recovering the first spike and the subsequent ones. However, for \(p = 2\), once the first direction has been recovered, the subsequent correlations \(m_{ij}^{(N)}\) scale up to \(N^{\frac{\delta_{ij}}{2}}\), where \(\delta_{ij} \in (0,1)\) depends on the corresponding ratios of SNRs. This scaling can be exploited to show that the sample complexity required to recover the directions beyond the first degrades only by a factor \(\delta_{ij}\), rather than by a full order-of-magnitude factor. For a more detailed explanation, we direct the reader to Section~\ref{outline proofs}. \\

For \(p=2\), we further study Langevin dynamics in the regime where all signal strengths are identical. In this isotropic setting, the Hamiltonian \(H\) is invariant under both right- and left-orthogonal transformations, and the recovery task shifts from identifying each individual spike to recovering the subspace spanned by \(\boldsymbol{v}_1, \ldots, \boldsymbol{v}_r\). To analyze this, we introduce the matrix-valued function \(\boldsymbol{G} \colon \mathcal{S}_{N,r} \to \R^{r \times r}\) defined by 
\[
\boldsymbol{G} (\boldsymbol{X}) = \boldsymbol{M}^\top \boldsymbol{M}, \quad \textnormal{where} \enspace \boldsymbol{M} = \frac{1}{N}\boldsymbol{V}^\top \boldsymbol{X} = (m_{ij}^{(N)}(\boldsymbol{X}))_{1 \le i,j \le r}.
\]
Using the identity
\[
\frac{1}{N^2} \norm{\boldsymbol{X} \boldsymbol{X}^\top - \boldsymbol{V} \boldsymbol{V}^\top}_{\textnormal{F}}^2 = 2 (r - \Tr (\boldsymbol{G})),
\]
we see that achieving \(N^{-2} \norm{\boldsymbol{X} \boldsymbol{X}^\top - \boldsymbol{V} \boldsymbol{V}^\top}_{\textnormal{F}}^2  = o(1)\) is equivalent to requiring that all eigenvalues of \(\boldsymbol{G} (\boldsymbol{X})\) converge to \(1-o(1)\) with high probability. The following theorem provides the corresponding sample complexity for this subspace recovery task. We write \(\theta_1 (\boldsymbol{X}), \ldots, \theta_r(\boldsymbol{X}) \ge 0\) for the eigenvalues of the symmetric positive semi-definite matrix \(\boldsymbol{G}(\boldsymbol{X})\).

\begin{thm}[Subspace recovery] \label{thm: strong recovery isotropic GF asymptotic}
Fix \(p=2\) and any \(\beta \in (0,\infty)\). Suppose \(\lambda_1 = \cdots = \lambda_r\). If \(M = N^\delta\) for any fixed \(\delta >0\), then Langevin dynamics recovers the subspace spanned by the \(r\) spikes with high-probability as \(N \to \infty\). That is, for every \(\varepsilon>0\), there exists \(T_0 >0\) such that for every \(T \ge T_0\) and every \(i \in [r]\),
\[
\lim_{N \to \infty} \int_{\mathcal{S}_{N,r}} \mathbb{Q}_{\boldsymbol{X}_0} \left( \min_{t \in [T_0,T]} \theta_i (\boldsymbol{X}_t^\beta ) \geq 1-\varepsilon \right) d \mu_{N \times r} = 1,\quad \mathbb{P}\textnormal{-a.s.}.
\]
\end{thm}

Finally, we note that a third regime of SNRs remains unexplored in this work and in our companion paper on online SGD~\cite{benarous2024}. This intermediate regime corresponds to SNRs that are separated by a small \(\epsilon\)-factor, where \(\epsilon\) may depend on the dimension \(N\). The analysis of this regime is left for future work.
\subsection{Related works} \label{related works}

Single-spike tensor PCA is a prototypical example of a model exhibiting a \emph{statistical-to-computational gap} in high-dimensional estimation. Such gaps arise when the statistical threshold for successful estimation is significantly lower than the computational threshold achievable by efficient algorithms. The statistical properties of MLE are by now well understood~\cite{jagannath2020,chen2019,chen2021}: beyond an order-one critical threshold, the MLE can successfully distinguish signal from noise. However, computing the MLE is computationally challenging in practice. To address this challenge, a wide range of algorithmic approaches have been extensively studied to determine the sample complexity required for efficient recovery. Spectral methods and algorithms based on the Sum-of-Squares hierarchy achieve recovery with sample complexity scaling as \(N^{\frac{p-2}{2}}\)~\cite{Hopkins15, Hopkins16, Bandeira2017, Weinkikuchi, BenArousUnfolding}. In contrast, local methods---such as gradient flow, Langevin dynamics~\cite{arous2020algorithmic}, online SGD~\cite{arous2021online}, tensor power iteration~\cite{HuangPCA, Wu24}, and approximate message passing~\cite{MontanariRichard}---achieve recovery at the threshold of \(N^{p-2}\). In particular, Wu and Zhou~\cite{Wu24} recently showed that tensor power iteration succeeds with sample complexity \(N^{p-2} \log(N)^{-C}\), where \(C\) is a constant depending on the tensor order \(p\), which represents the current state-of-the-art bound for local algorithms.

These results suggest that the algorithmic efficiency of local methods is closely tied to the geometric structure of the underlying loss landscape. In particular, the signal-noise decomposition inherent to tensor PCA gives rise to a small set of low-dimensional quantities that govern the dynamics of optimization. Understanding how these quantities evolve provides a unifying perspective on both single- and multi-spiked models. The multi-spiked tensor PCA problem belongs to a broader class of inference models where the MLE objective consists of a signal term corrupted by Gaussian noise. In many such problems, the high-dimensional dynamics of the estimator can be described by a reduced system involving a few \emph{sufficient statistics} that evolve autonomously. This phenomenon---where the effective dynamics concentrate on a low-dimensional manifold---has been rigorously formalized under general conditions in~\cite{benarousneurips,ben2022effective}, which introduce autonomous quantities known as \emph{summary statistics}, of which sufficient statistics form a special case. Analyzing the evolution of summary statistics under gradient-based dynamics poses several challenges. First, the effects of stochasticity---such as Gaussian noise or Brownian motion in the case of Langevin dynamics---must be controlled over sufficiently long timescales to ensure that events of interest, such as achieving a nontrivial correlation with the hidden signal, occur with high probability. Second, the system of autonomous equations describing the summary statistics must remain simple enough to allow precise quantitative analysis. Finally, the projection of the dynamics onto the space of summary statistics should preserve as much of the regularity and structure of the original high-dimensional system as possible. In statistical physics, such low-dimensional quantities that capture the macroscopic behavior of complex systems are referred to as \emph{order parameters}, and we adopt this terminology in the following discussion.\\

In the statistical physics of disordered systems, classical examples of signal-plus-noise Hamiltonians are planted versions of spin glass Hamiltonians~\cite{nishimori2001statistical, zdeborova2016statistical}. In this context, the out-of-equilibrium dynamics of the order parameters are typically studied using low-dimensional integro-differential equations involving two-time autocorrelation functions. This approach, originally proposed in~\cite{sompolinsky1981dynamic,sompolinsky1982relaxational,crisanti1993spherical,cugliandolo1993analytical,bouchaud1998out} to model aging phenomena in spin glasses, is commonly referred to as \emph{dynamical mean-field theory} (DMFT). Recently, DMFT has been adapted in the statistical physics of learning to analyze Langevin dynamics in high-dimensional estimation problems. Examples include perceptron models~\cite{agoritsas2018out}, mixed matrix-tensor estimation~\cite{sarao2019,sarao2019bis,sarao2020marvels}, Gaussian mixture classification~\cite{mignacco2020dynamical}, and phase retrieval~\cite{mignacco2021stochasticity}. In these works, the resulting integro-differential equations are solved numerically to obtain phase diagrams for the order parameters over order-one timescales. However, a key challenge of this approach lies in the complexity of the integro-differential equations, which often hinders a complete analytical study. 

On a rigorous level, DMFT equations have been proven for mean-field spin glasses using large deviations techniques~\cite{grunwald1996sanov,arous2001aging,ben2006cugliandolo} and more recently for inference problems using iterative Gaussian conditioning~\cite{celentano2021high,gerbelot2024rigorous} and random matrix theory in~\cite{bodin2021rank,bodin2021model,liang2023high} for instances of spiked matrix models. In these works, quantitative statements (e.g. recovery thresholds) are only obtained due to simplifications specific to quadratic models, and extending these results to a multi-spike tensor model appears challenging. Furthermore, DMFT approaches for statistical estimation problems are always proven in the proportional limit of dimension and number of samples one order one timescales, making it difficult to study problems that require polynomial sample complexity or running time. Finally, these proofs also require an $\mathcal{O}(1)$ (akin to a warm start, see the discussion in~\cite{liang2023high}) initial correlation between the estimator and the hidden signal, whereas this correlation is typically of order $\mathcal{O}(N^{-1/2}$ when using uninformative initializations.\\

In statistical learning, the traditional approach to controlling noisy dynamics involves establishing a single uniform convergence bound (over the input space) for the empirical risk, or gradient thereof, and then directly analyzing the population dynamics under this control. Recently, this approach has been applied to problems involving low-dimensional population dynamics~\cite{dudeja2018learning, bietti2022learning}. Its main advantages include simplicity, robustness across different time horizons, sample complexity regimes, and initialization scales, as well as the tractability of the systems of ordinary differential equations (ODEs) derived from the noiseless problem. However, this method has significant limitations, particularly its inability to achieve sharp bounds. Specifically, it fails to leverage the independence between the initial condition and the noise tensor discussed earlier, resulting in a loss of a factor \(N^{1/2}\) in the recovery threshold. \\

To address these shortcomings, we adopt an alternative approach, first introduced in~\cite{ben2020bounding} for Langevin dynamics in spin glasses and subsequently applied to the single-spiked tensor PCA problem in~\cite{arous2020algorithmic}, where it was used to prove the conjectured recovery threshold for first-order methods proposed in~\cite{MontanariRichard}. In this work, we extend this method to the multi-spiked tensor PCA problem, showing that it overcomes the deficiencies of traditional approaches outlined above. It is worth noting, however, that while results obtained using DMFT are asymptotically exact, the bounding flows method provides inequalities rather than exact characterizations of the studied trajectories. The main difference between our work and~\cite{arous2020algorithmic} lies in the increased complexity of the resulting dynamical system, which involves the summary statistics $\left\{m_{ij}^{(N)} \right\}_{1 \leq i,j \leq r}$ as well as the time-dependent bounds for the quantities $\left\{\norm{L_{0,\beta} m_{ij}^{(N)}}_\infty \right\}_{1 \leq i,j \leq r}$.\\

We conclude this section by noting that the multi-spiked tensor PCA problem also serves as a benchmark for studying nonconvex optimization in high dimensions, a topic of significant importance in machine learning theory. For a detailed discussion of the relevance of our work in this context and additional references on this subject, we refer the reader to the related works section of our companion paper~\cite{benarous2024}.

\subsection{Overview}

An overview of the paper is given as follows. In Section~\ref{outline proofs}, we outline the proofs of our main results, providing a roadmap for the key arguments. The nonasymptotic formulation of our main results, including explicit constants and convergence rates, is presented in Section~\ref{section: Langevin}. Preliminary results necessary for the proofs are discussed in Section~\ref{preliminaries}. The detailed proofs of our main results are provided in Sections~\ref{section: proof recovery Langevin p>2} through~\ref{section: proof recovery Langevin isotropic p=2}. These sections address the cases \(p\ge 3\), \(p=2\) with distinct SNRs, and \(p=2\) with equal SNRs, respectively. Finally, Appendix~\ref{appendix: invariant measure} provides concentration properties of the uniform measure on the normalized Stiefel manifold \(\mathcal{S}_{N,r}\).\\ 

\textbf{Acknowledgements.}\ G.B.\ and C.G.\ acknowledge the support of NSF grant DMS-2134216. V.P.\ acknowledges the support of the ERC Advanced Grant LDRaM No.\ 884584.

\section{Outline of proofs} \label{outline proofs}

In this section, we outline our proofs, focusing on the rank-\(2\) spiked tensor model for simplicity. We first address the recovery of all spikes and then focus on subspace recovery, by assuming \(p=2\) and \(\lambda_1 = \lambda_2\). The core of the proof revolves around the simultaneous control of the noise and signal part of the dynamics. Once the noise is controlled, we describe the population dynamics, which is similar to the discrete time dynamical system obtained in our companion paper \cite{benarous2024}. The main difference between these two works lies in the control of the noise : in the case of Langevin dynamics, the drift induced by the Gaussian noise in the Hamiltonian is predictable (as opposed to a martingale noise in SGD), and more involved tools are needed to control the resulting correlations. Moreover, the effect of Brownian motion must be considered, which leads to a condition on the separation of the SNRs. 

\subsection{Full recovery of spikes}

We focus on the evolution of the correlations \(m_{ij}^{(N)}(\boldsymbol{X}_t^\beta)\) under Langevin dynamics. We assume an initial random start with a completely uninformative prior, specifically the uniform distribution on the normalized Stiefel manifold \(\mathcal{S}_{N,r}\). As a consequence, all correlations \(m_{ij}^{(N)} (\boldsymbol{X}_0)\) have the typical scale of order \(N^{-\frac{1}{2}}\) at initialization. For simplicity, we assume that all correlations are positive at initialization in the following discussion. Moreover, we write \(m_{ij}(t) \coloneqq m_{ij}^{(N)}(\boldsymbol{X}_t^\beta)\) to simplify notation slightly.

According to~\eqref{eq: langevin dynamics}, the evolution equation for \(m_{ij}(t)\) under Langevin dynamics is given by
\begin{equation} \label{eq: evolution correlation gradient flow informal}
dm_{ij} (t) = L_\beta m_{ij}(t) dt + dM_t^{m_{ij}},
\end{equation} 
where \(M_t^{m_{ij}} = \sqrt{2} \int_0^t \langle \nabla m_{ij}(s), d \boldsymbol{B}_s \rangle\) denotes the martingale part of the evolution. The generator \(L_\beta\) is given according to~\eqref{eq: generator Langevin}:
\[
L_\beta m_{ij}(t) dt = \Delta m_{ij}(t) - \beta \langle \nabla H, \hat{\nabla} m_{ij} \rangle.
\]
Recall that 
\[
H (\boldsymbol{X}) = H_0(\boldsymbol{X}) - \sum_{1 \leq i,j \leq r} N \sqrt{M} \lambda_i \lambda_j m_{ij}^p(\boldsymbol{X}) ,
\]
where \(H_0\) is given by~\eqref{eq: Hamiltonian H0}. A straightforward computation shows that 
\[
L_\beta m_{ij} = L_{0,\beta} m_{ij} + \beta\sqrt{M} p \lambda_i \lambda_j m_{ij}^{p-1}  -  \beta\sqrt{M}\frac{p}{2} \sum_{1 \leq k, \ell \leq r} \lambda_k m_{kj} m_{k \ell} m_{i \ell}  \left(\lambda_j m_{kj}^{p-2} + \lambda_\ell m_{k \ell}^{p-2}  \right) ,
\]
where \(L_{0,\beta}\) is the generator induced by the noise part (see~\eqref{eq: generator Langevin noise}). The population generator consists of two terms: the first term \(p \lambda_i \lambda_j m_{ij}^{p-1}\) represents the drift and dominates the dynamics, particularly near initialization, while the second term \(\frac{p}{2}\sum_{1 \leq k, \ell \leq r} \lambda_k m_{kj} m_{k \ell} m_{i \ell}\left(\lambda_j m_{kj}^{p-2} + \lambda_\ell m_{k \ell}^{p-2} \right) \) is the ``correction'' term due to the constraint of the estimator \(\boldsymbol{X}\) being on the orthogonal manifold. Analogous to the analysis for online SGD, the main challenge lies in balancing the noise and signal terms. At small scales, such as near initialization, the population drift predominates over the correction term and the generator \(L_\beta m_{ij}\) can be approximated by 
\[
L_\beta m_{ij}(t) \approx \Delta m_{ij} (t) - \beta \langle \nabla H_0, \hat{\nabla} m_{ij} (t) \rangle + \beta \sqrt{M} p \lambda_i \lambda_j m_{ij}^{p-1}(t).
\]
For the correlations \(m_{ij}\) to increase, the drift \( p \lambda_i \lambda_j m_{ij}^{p-1}\) at initialization needs to be larger than the noise part. At initialization, the correlations \(m_{ij}\) typically scale as \(N^{-\frac{1}{2}}\), so that \(m_{ij}^{p-1}\) typically scales as \(N^{\frac{p-1}{2}}\). The noise generator \(L_{0,\beta} m_{ij} =  \Delta m_{ij} (t) - \beta \langle \nabla H_0, \hat{\nabla} m_{ij} (t) \rangle \) with \( \Delta m_{ij} (t) = - \frac{N-1}{N}m_{ij}(t)\) according to~\eqref{eq: stiefel laplace}. The term \(- \beta \langle \nabla H_0, \hat{\nabla} m_{ij} (t) \rangle \) is also of order \(N^{-\frac{1}{2}}\) at initialization, thus a sample complexity of order \(N^{p-2}\) is sufficient for the population drift to dominate over the noise influence. Under this sample complexity, the dynamics during this first phase can described by the following simple stochastic differential equation:
\begin{equation} \label{eq: simple ODE}
d m_{ij}(t) \approx \left ( \beta \sqrt{M} p \lambda_i \lambda_j m^{p-1}_{ij} - \frac{N-1}{N}m_{ij}(t) \right) dt + dM_t^{m_{ij}}.
\end{equation}
At this point, we need to show that the drift term in the population dynamics keeps dominating the noise generator \(L_{0,\beta} m_{ij}\) over a sufficiently large time scale, allowing \(m_{ij}\) to \emph{escape mediocrity}. This is exactly what is achieved by the bounding flows method of~\cite{ben2020bounding, arous2020algorithmic}, providing a time dependent bound for the noise processes based on estimates of a Sobolev-type norm of \(H_0(\boldsymbol{X})\), while retaining the regularity of the initial condition. Once the first correlation reaches a critical threshold, the correction term in the population generator, which was negligible during the initial phase of dynamics, becomes relevant and the correlations begin to interact with each other. As explained below in the description of the population dynamics, the values of the SNRs and of the correlations at initialization lead to an ordering of the correlations at the microscopic scale that is crucial for the analysis of the recovery process. However, in the presence of Brownian motion, the fluctuations due to the martingale part $M_t^{m_{ij}}$ of the dynamics are comparable to the values of the correlations near initialization. In particular, unless the SNRs are sufficiently separated, we are unable to guarantee that this ordering is stable.   \\

\textbf{Analysis of the population dynamics.} The evolution of correlations under the population dynamics of gradient flow is close to the behavior described for online SGD in~\cite[Section 2]{benarous2024}. For the sake of completeness and the reader's understanding, we provide a brief explanation of this population analysis below.

Assume first \(p \ge 3\). The solution to the ODE~\eqref{eq: simple ODE} shows that the correlations \(m_{ij}\) are well approximated by 
\begin{equation} \label{eq: simple sol p>3}
m_{ij}(t) \approx m_{ij}(0)\left(1- \beta \sqrt{M} \lambda_i \lambda_j p(p-2) m_{ij}(0)^{p-2} t\right)^{-\frac{1}{p-2}},
\end{equation}
where \(m_{ij}(0) = \frac{\gamma_{ij}}{\sqrt{N}}\) for some order-\(1\) constant \(\gamma_{ij} > 0\). From~\eqref{eq: simple sol p>3}, it follows that the typical time for \(m_{ij}\) to reach a macroscopic threshold \(\varepsilon > 0\) is given by
\[
T_\varepsilon^{(ij)} \approx \frac{1- \left (\frac{\gamma_{ij}}{\varepsilon \sqrt{N}} \right)^{p-2}}{\beta \sqrt{M} \lambda_i \lambda_j p (p-2) \gamma_{ij}^{p-2}} N^{\frac{p-2}{2}}.
\]
In particular, the first correlation to become macroscopic is the one associated with the largest value of \(\lambda_i \lambda_j \gamma_{ij}^{p-2}\), where \((\gamma_{ij})_{1 \le i,j \le 2}\) approximately follow independent standard normal distributions. Recall that the true evolution of $m_{ij}(t)$ corresponds to a perturbation of \ref{eq: simple sol p>3}, whose sensitivity to the initial condition crucially determines the ordering of the correlations near initialization. We show that, using the sample complexity $M$, the perturbation due to $L_{0,\beta}$ becomes negligible as soon as $M \gtrsim N^{p-2}$ for some constant that is made explicit in the proof. To obtain a uniform control of the martingale part $M_t^{m_{ij}}$ over the time interval required to use the bounding flow method, we use Doob's maximal inequality to compare it with the initialization, which perturbes the ordering of the quantities $\lambda_i\lambda_j\gamma_{ij}^{p-2}$. To compensate this perturbation, we impose a separation condition on the $\left\{\lambda_i\right\}_{i=1}^{r}$. In our companion paper on gradient flow, we show that, due to the absence of the Brownian motion, a sharper analysis can be carried out without any separation assumption on the SNRs. We leave the elaboration of a more refined argument at finite temperature for future work.

Once the first correlation \(m_{ij}\) reaches some macroscopic value \(\varepsilon >0\), we can show that  the other correlations are still of order \(\frac{1}{\sqrt{N}}\). Next, we describe the evolution of \(m_{12}, m_{21}\) and \(m_{22}\) during this time interval. We first note that the correction part of the population generator, i.e.,  
\[
\sum_{1 \le k, \ell \le 2} \lambda_k m_{kj} m_{k \ell} m_{i \ell} (\lambda_j m_{kj}^{p-2} + \lambda_\ell m_{k \ell}^{p-2}),
\]
becomes dominant in the evolution of the correlations \(m_{12}, m_{21}\), as soon as \(m_{11}\) reaches the microscopic threshold \(N^{-{\frac{p-2}{2p}}}\). Therefore, until \(m_{11}\) reaches this threshold value, \(m_{12}\) and \(m_{21}\) are non-decreasing and they begin to decrease from this point onward. On the other hand, the correction part of the population generator may become dominant in the evolution of \(m_{22}\), as soon as \(m_{11}\) reaches another microscopic threshold of order \(N^{-{\frac{p-3}{2(p-1)}}}\), leading to a potential decrease of \(m_{22}\). Through a careful analysis, we show that this potential decrease is at most of order \(\frac{\log(N)}{N}\), so that \(m_{22}\) remains stable at the scale \(\frac{1}{\sqrt{N}}\) during the ascending phase of \(m_{11}\). Once \(m_{12}, m_{21}\) become sufficiently small, we show that the evolution of \(m_{22}\) can still be described by~\eqref{eq: simple sol p>3}, thus ensuring the recovery of the second direction. The phenomenon we observe here is referred to as \emph{sequential elimination phenomenon}, as introduced in Definition~\ref{def: sequential elimination}. The correlations increase sequentially, one after another. When the first correlation \(m_{11}\) exceeds a certain threshold, the correlations sharing a row and column index, i.e., \(m_{12}\) and \(m_{21}\)) start decreasing until they become sufficiently small to be negligible, thereby enabling the subsequent correlation \(m_{22}\) to increase and reach a macroscopic value. We refer to Figure 9 in our companion paper~\cite{benarous2024} to an illustration of this phenomenon.

We now consider \(p=2\). In this case, the solution to~\eqref{eq: simple ODE} is given by
\begin{equation} \label{eq: simple sol p=2}
m_{ij} \approx m_{ij}(0) \exp \left (2\lambda_i  \lambda_j t \right),
\end{equation}
with \(m_{ij}(0) = \gamma_{ij} N^{- \frac{1}{2}}\). This implies that the typical time for \(m_{ij}\) to reach a macroscopic threshold \(\varepsilon > 0\) is given by 
\[
T_{\varepsilon}^{(ij)} \approx \frac{\frac{1}{2} \log(N)-\log \left(\frac{\gamma_{ij}}{\varepsilon}\right)}{2\lambda_i \lambda_j}.
\]
Compared to the case where \(p \ge 3\), where the quantity \(\lambda_i \lambda_j \gamma_{ij}^{p-2}\) determines the correlation that first escapes mediocrity, we observe here that the initialization \(\gamma_{ij}\) has less influence and the differences in time for the correlations to reach a macroscopic threshold are dominated by the separation between the SNRs, which is less pronounced as the \(\gamma_{ij}\)'s are not in the denominator. When \(\lambda_1 > \lambda_2\), a sequential elimination phenomenon akin to the one observed for \(p \geq 3\) occurs. However, there is an important difference compared to the tensor case: once \(m_{11}\) reaches the macroscopic threshold \(\varepsilon >0\), the correlations \(m_{12}, m_{21}\) and \(m_{22}\) scale as \(N^{- \delta_1/2}, N^{-\delta_1/2}\) and \(N^{-\delta_2/2}\), respectively, where \(\delta_1 = 1 - \frac{\lambda_2}{\lambda_1}\) and \(\delta_2 = 1 - \frac{\lambda_2^2}{\lambda_1^2}\). We then show that \(m_{11}\) must reach approximately the critical value \((\frac{\lambda_2}{\lambda_1})^{1/2}\) for \(m_{12}, m_{21}\) to decrease. This value is achieved in an order one time once \(m_{11}\) has reached \(\varepsilon\), while \(m_{12}, m_{21}\) still require a time of order \(\log(N)\) to escape their scale of \(N^{-\frac{\delta_1}{2}}\). In addition, we show the stability of \(m_{22}\) during this first phase by controlling that it remains increasing throughout the ascending phase of \(m_{11}\) and until \(m_{12}, m_{21}\) begin to decrease. Note that, since $m_{22}$ is of order $N^{-\delta_{2}/2}$ as $m_{11}$ reaches $\varepsilon$, the sample complexity for the recovery of the second direction degrades by a factor $N^{1-\frac{\lambda^2_{r}}{\lambda^2_{1}}}$, as opposed to $\sqrt{N}$ in the tensor case.

\subsection{Subspace recovery}

As explained in our companion paper on SGD, the multi-spike matrix model becomes isotropic when the SNRs are equal, since any rotation of the hidden directions is a global minimizer. Thus, rather than focusing on the correlations $m_{ij}$ individually, we apply our bounding flow directly at the level of the eigenvalues of the matrix \(\boldsymbol{G} = \boldsymbol{M}^\top \boldsymbol{M}\). In particular, fluctuations of the initial condition (at the microscopic scale) have little effect in this case, so that the martingale part can be easily controlled. We then show that, when the noisy dynamics is sufficiently close to the population dynamics, the matrix $\mathbf{G}$ follows a constant coefficient algebraic Ricatti equation, i.e., 
\[
\dot{\boldsymbol{G}}(t) \approx \lambda^2 \boldsymbol{G}(t)(\mathbf{I}_r - \boldsymbol{G}(t)),
\]
where its eigenvalues \(\theta_1,\ldots, \theta_r\) evolve according to
\[
\dot{\theta_i}(t) \approx \lambda^2 \theta_i (t) \left ( 1-\theta_i (t)\right ).
\]
Since these equations are coordinate-wise autonomous and have monotone solutions, it is easy to prove efficient recovery once the estimator has exited the maximum entropy region of the landscape.
\section{Main results} \label{section: Langevin}

In this section, we present our main results of Subsection~\ref{subsection: main asymptotic results} in nonasymptotic form. These nonasymptotic versions are stronger, as they explicitly provide all constants and convergence rates. Moreover, the asymptotic results from Subsection~\ref{subsection: main asymptotic results} follow directly as corollaries of these nonasymptotic statements.

\subsection{Initial conditions}

Although our main asymptotic results are presented for Langevin dynamics initialized from a random point \(\boldsymbol{X}_0\) drawn according to the uniform measure \(\mu_{N \times r}\) on \(\mathcal{S}_{N,r}\), our recovery guarantees extend beyond the uniform initialization to a broader class of random initial data, provided certain natural conditions are satisfied. If \(\mathcal{M}_1 (\mathcal{S}_{N,r})\) denotes the space of probability measures on \(\mathcal{S}_{N,r}\), a choice of initialization corresponds to a choice of measures \(\mu_N \in \mathcal{M}_1 (\mathcal{S}_{N,r})\). We now introduce the conditions under which our guarantees hold. 

The first condition imposes a regularity assumption on \(L_{0,\beta} m_{ij}^{(N)} (\boldsymbol{X})\). It ensures that successive applications of \(L_{0,\beta}\) remain uniformly bounded with high probability.

\begin{defn}[Condition 0 at level \(n\)] \label{def: condition 0}
Fix \(\beta >0\). For every \(N \in \N\), \(\gamma_0 >0\), and \(n\geq 1\), define 
\begin{equation*} \label{eq: regularity noise generator}
\mathcal{C}_{0,\beta}^{(N)} (n,\gamma_0) = \bigcap_{k=0}^{n-1} \left \{ \boldsymbol{X} \in \mathcal{S}_{N,r} \colon |L_{0,\beta}^k m_{ij}^{(N)}(\boldsymbol{X})| \leq \frac{\gamma_0}{\sqrt{N}}  \enspace \text{for every} \enspace 1 \leq i,j \leq r\right \}.
\end{equation*}
We say that a sequence \((\mu_N)_{N \ge 1}\) of probability measures on \(\mathcal{M}_1(\mathcal{S}_{N,r})\) satisfies \emph{Condition 0 at level \(n\)} for inverse temperature \(\beta\) if there exist absolute constants \(C,c >0\) such that, for every \(N \in \N\) and \(\gamma_0 > 0\), 
\begin{equation*} 
\mu_N \left ( \mathcal{C}_{0,\beta}^{(N)} (n,\gamma_0)^\textnormal{c} \right) \leq C e^{-c\gamma_0^2}.
\end{equation*}
\end{defn}

\begin{defn}[Condition 0 at level \(\infty\)] \label{def: condition 0 infty}
Fix \(\beta >0\). For every \(N \in \N\), \(\gamma_0 >0\), and \(T >0\), define
\begin{equation*} \label{eq: regularity initial data infty}
\mathcal{C}_{0,\beta}^{(N)}(T,\gamma_0) = \left \{ \boldsymbol{X} \in \mathcal{S}_{N,r} \colon \sup_{t \le T} |e^{t L_{0,\beta}} L_{0,\beta} m_{ij}^{(N)}(\boldsymbol{X})| \leq \frac{\gamma_0}{\sqrt{N}}  \enspace \text{for every} \enspace 1 \leq i,j \leq r\right \},
\end{equation*}
where \(e^{t L_{0,\beta}}\) denotes the semigroup generated by \(L_{0,\beta}\). We say that a sequence \((\mu_N)_{N \ge 1}\) of probability measures on \(\mathcal{M}_1(\mathcal{S}_{N,r})\) weakly satisfies \emph{Condition 0 at level \(\infty\)} if there exist constants \(C,c>0\) such that, for every \(N \in \N\), \(\gamma_0 > 0\), and \(T>0\),
\begin{equation*} 
\mu_N \left (\mathcal{C}_{0,\beta}^{(N)}(T,\gamma_0) ^\textnormal{c} \right) \leq C \sqrt{N}T e^{-c\gamma_0^2}.
\end{equation*}
\end{defn}

The second condition requires that the initial correlations are of typical order \(N^{-\frac{1}{2}}\), consistent with fluctuations under the uniform measure on \(\mathcal{S}_{N,r}\).
    
\begin{defn}[Condition 1]\label{def: condition 1}
For every \(N \in \N\) and \(\gamma_1 > \gamma_2 >0\), define
\[
\mathcal{C}_1^{(N)} (\gamma_1,\gamma_2) = \left \{ \boldsymbol{X} \in \mathcal{S}_{N,r} \colon \frac{\gamma_2}{\sqrt{N}} \leq m_{ij}^{(N)}(\boldsymbol{X}) < \frac{\gamma_1}{\sqrt{N}}\enspace \text{for every} \enspace 1 \leq i,j \leq r\right \}.
\]
We say that a sequence \((\mu_N)_{N \ge 1}\) of probability measures on \(\mathcal{M}_1(\mathcal{S}_{N,r})\) satisfies \emph{Condition 1} if there exist absolute constants \(C_1, c_1, C_2, c_2, C_3>0\) such that, for every \(N \in \N\) and \(\gamma_1 > \gamma_2 > 0\),
\[ 
\mu_N \left ( \mathcal{C}_1^{(N)}(\gamma_1,\gamma_2)^\textnormal{c} \right ) \leq C_1 e^{-c_1 \gamma_1^2} + C_2 e^{-c_2 \gamma_2 \sqrt{N}} + C_3 \gamma_2.
\]
\end{defn} 

The most natural initialization is the uniform measure \(\mu_{N \times r}\) on \(\mathcal{S}_{N,r}\). We claim:

\begin{lem} \label{lem: invariant measure Langevin}
The uniform measure \(\mu_{N \times r}\) on \(\mathcal{S}_{N,r}\) satisfies Condition 1 and weakly satisfies Condition 0 at level \(\infty\).
\end{lem}

The proof of Lemma~\ref{lem: invariant measure Langevin} is deferred to Appendix~\ref{appendix: invariant measure}. \\

\subsection{Main results in nonasymptotic form}

We are now in the position to state our main results in nonasymptotic form under Langevin dynamics with Hamiltonian \(H\). Throughout this section, we assume that the initial correlations with the signal are positive, i.e., \(m_{ij}(\boldsymbol{X}_0)>0\) for every \(i,j \in [r]\). Equivalently, for each \(N\) the initialization measure \(\mu_N^0\) is supported on
\[
\mathcal{X}^+ \coloneqq \{\boldsymbol{X} \in \mathcal{S}_{N,r} \colon m_{ij}^{(N)}(\boldsymbol{X}) > 0 \ \forall i,j\in[r]\}.
\]
Under this assumption, we have \(r_\textnormal{c} =r\), and we may drop absolute values in all subsequent statements.

Our first result determines the sample complexity required to eﬃciently recover the leading spike.

\begin{prop}[Recovery of leading spike for \(p \geq 3\)] \label{thm: strong recovery first spike Langevin p>2}
Fix \(\beta \in (0,\infty)\), \(p \geq 3\), and integer \(n\ge 1\). Consider a sequence of initialization measures \(\mu_N^0\in \mathcal{M}_1(\mathcal{S}_{N,r})\) supported on \(\mathcal{X}^+\), and satisfying Condition \(0\) at level \(n\) (for inverse temperature \(\beta\)), and Condition \(1\). Then, for every \(\gamma_0 > 0\), \(\gamma_1 > 1 > \gamma_2 > 0\), \(\varepsilon > 0\), and \(0 < C_0 < \frac{1}{2} - \frac{1}{2}\left ( \frac{\gamma_2}{3 \gamma_1} \right)^{p-2}\), if 
\[
\lambda_1 > \frac{1+C_0}{1-C_0} \left (\frac{3 \gamma_1}{\gamma_2}\right )^{p-2} \lambda_2 \quad \textnormal{and} \quad \sqrt{M} \gtrsim \frac{(n+2) \gamma_0}{\beta p \lambda_r^2 C_0 \gamma_2^{p-1}}  N^{\frac{p-1}{2} - \frac{n}{2(n+1)}}, 
\]
then there exists \(T_0 \le k_1\frac{\gamma_2}{(n+2) \gamma_0}  N^{-\frac{1}{2(n+1)}} + k_2 N^{-\frac{p-2}{2(p-1)}}\) such that for every \(T > T_0\) and \(N\) sufficiently large,
\[
\int_{\mathcal{X}^+} \mathbb{Q}_{\boldsymbol{X}} \left (\inf_{t \in [T_0,T]} m_{11}(\boldsymbol{X}_t^\beta) \geq 1 - \varepsilon \right ) d \mu_N^0 (\boldsymbol{X})  \geq 1 -  \eta(N, n,\gamma_0, \gamma_1,\gamma_2,\varepsilon,T),
\]
with \(\mathbb{P}\)-probability at least \(1 - \exp(-K N)\), where \(\eta=\eta(N, n,\gamma_0, \gamma_1,\gamma_2,\varepsilon, T)\) is given by
\[
\begin{split}
\eta & = C_1 e^{-c_1 \gamma_0^2} + C_2 e^{-c_2 \gamma_1^2} + C_3 e^{- c_3 \gamma_2 \sqrt{N}} + C_4 \gamma_2 + K_1 e^{- \gamma_0^3 (n+2) N^{\frac{1}{2(n+1)}} / (K_1 \gamma_2)} \\
& \quad + r^2 K_2 e^{- \gamma_2 \gamma_0 (n+2)N^{\frac{1}{2(n+1)}} / K_2} + K_3 e^{- N^{\frac{p-2}{2(p-1)}} / K_3 }  + K_4 e^{- \varepsilon^2 N / (K_4T)}.
\end{split}
\]
\end{prop}

Our second main result determines the sample complexity required to exactly recover all \(r\) spikes \(\boldsymbol{v}_1, \ldots, \boldsymbol{v}_r\). We introduce the following notation. For any set \(\mathcal{E}\), we define the hitting time \(\mathcal{T}_{\mathcal{E}}\) of the set \(\mathcal{E}\) by Langevin process \((\boldsymbol{X}_t^\beta)_{t \ge 0}\) as
\[
\mathcal{T}_{\mathcal{E}} \coloneqq \inf \{t \ge 0 \colon \boldsymbol{X}_t^\beta \in \mathcal{E}\}.
\]
We then introduce the exact recovery event: for every \(\varepsilon > 0\), 
\begin{equation} \label{eq: set strong recovery p>2}
R(\varepsilon) = \left \{ \boldsymbol{X} \colon m_{ii}(\boldsymbol{X}) \geq 1 - \varepsilon  \enspace \forall i \in [r] \enspace \textnormal{and} \enspace m_{k \ell} (\boldsymbol{X}) \lesssim \log(N)^{-\frac{1}{2}}N^{-\frac{p-1}{4}} \enspace \forall k, \ell \in [r], k \neq \ell \right \}.
\end{equation}

\begin{prop}[Exact recovery of all spikes for \(p \geq 3\)] \label{thm: strong recovery all spikes Langevin p>2}
Fix \(\beta \in (0,\infty)\) and \(p \geq 3\). Consider a sequence of initialization measures \(\mu_N^0 \in \mathcal{M}_1(\mathcal{S}_{N,r})\) supported on \(\mathcal{X}^+\) and satisfying Condition \(1\). Then, for every \(\gamma_1 > 1 >\gamma_2 > 0\), \(\varepsilon > 0\), and \(C_0 \in (0, \frac{1}{2})\), there exists \(\Lambda = \Lambda(p,\beta,\{\lambda_i\}_{i=1}^r) > 0\) and \(C=C(p,\beta,\{\lambda_i\}_{i=1}^r,\varepsilon) >0\) such that, if 
\[
\lambda_i >\frac{1+C_0}{1-C_0} \left (\frac{3\gamma_1}{\gamma_2}\right )^{p-2} \lambda_{i+1} \enspace \textnormal{for every} \enspace 1 \le i \le r-1 \quad \textnormal{and} \quad \sqrt{M} \ge \frac{\Lambda}{\beta p \lambda_r^2 C_0 \gamma_2^{p-1}} N^{\frac{p-1}{2}}, 
\]
then for \(N\) sufficiently large,
\[
\int_{\mathcal{X}^+} \mathbb{Q}_{\boldsymbol{X}} \left ( \mathcal{T}_{R(\varepsilon)}  \le C N^{-1/2}\right ) d \mu_N^0 (\boldsymbol{X})  \geq 1 - \eta(N,\gamma_1, \gamma_2,\varepsilon), 
\]
with \(\mathbb{P}\)-probability at least \(1 - \exp(-K N)\), where \(\eta=\eta(N,\gamma_1, \gamma_2, \varepsilon)\) is given by
\[
\begin{split}
\eta & = C_1 e^{- c_1 \gamma_1^2} + C_2 \gamma_2 + C_3 e^{- c_3 \gamma_2 \sqrt{N}} + r^3 K_1 e^{- \gamma_2^2 \sqrt{N}/K_1} + r K_2 e^{- N^{\frac{p+1}{2(p-1)}}/K_2} + r K_3 e^{- \varepsilon^2 N^{3/2} / K_3}.
\end{split}
\]
Furthermore, the correlations \(\{m_{ij}\}_{i,j=1}^r\) follow a sequential elimination.
\end{prop}

In light of Proposition~\ref{thm: strong recovery all spikes Langevin p>2}, we observe that exact recovery of multiple spikes is feasible only after the off-diagonal correlations \(m_{ij}\) for \(i \neq j\) decay past their initial scale \(\Theta(N^{-\frac{1}{2}})\). In particular, efficient recovery is achieved once these correlations reach the smaller threshold \(\log(N)^{-\frac{1}{2}} N^{-\frac{p-1}{4}}\). \\

We now present our main results for the matrix PCA problem, i.e., the case \(p=2\). We focus first on the regime where the SNRs are separated by constants of order one; that is, we assume
\[
\lambda_i =\lambda_{i+1} (1 + \kappa_i), \quad 1 \le i \le r-1,
\]
for constants \(\kappa_i >0\) independent of \(N\). As in the case \(p \ge 3\), we identify two distinct algorithmic thresholds for the sample complexity: one for efficient recovery of the leading spike and another for exact recovery of all spikes. 
\begin{prop}[Recovery of leading spike for \(p=2\)] \label{thm: strong recovery first spike Langevin p=2}
Fix \(\beta \in (0,\infty)\), \(p = 2\), and an integer \(n \ge 1\). Consider a sequence of initialization measures \(\mu_N^0 \in \mathcal{M}_1(\mathcal{S}_{N,r})\) satisfying Condition \(0\) at level \(n\) (for inverse temperature \(\beta\)) and Condition \(1\). Then, for every \(\gamma_0 > 0\), \(\gamma_1 > 1 >\gamma_2 > 0\), and \(\varepsilon > 0\), there exists \(C_0 \in (0, \frac{1}{2})\) such that if 
\[
\sqrt{M}\gtrsim \frac{(n+2) \gamma_0}{\beta \lambda_r^2 C_0 \gamma_2} N^{\frac{1}{2(n+1)}} ,
\]
then there exists \(T_0 \le c \frac{\gamma_2}{(n+2) \gamma_0} \log(N) N^{-\frac{1}{2(n+1)}}\) such that for every \(T > T_0\) and every \(N\) sufficiently large,
\[
\int_{\mathcal{S}_{N,r}} \mathbb{Q}_{\boldsymbol{X}} \left (\inf_{t \in [T_0,T]} m_{11}(\boldsymbol{X}_t^\beta) \geq 1 - \varepsilon \right ) d \mu_N^0 (\boldsymbol{X})  \geq 1 -  \eta (N, n,\gamma_0,\gamma_1,\gamma_2,\varepsilon,T), 
\]
with \(\mathbb{P}\)-probability at least \(1 - \exp(-K  N)\), where \(\eta=\eta (N, n,\gamma_0,\gamma_1,\gamma_2,\varepsilon,T)\) is given by
\[
\begin{split}
\eta &= C_1 e^{-c_1 \gamma_0^2} + C_2 e^{-c_2 \gamma_1^2} + C_3 \gamma_2 + C_4 e^{- c_4 \gamma_2 \sqrt{N}} + K_1 e^{ -  \gamma_0^3 (n+2) N^{\frac{1}{2(n+1)}} / (K_1 \gamma_2 \log(N))} \\
& \quad + K_2 r^2 e^{-\gamma_0 \gamma_2 (n+2) N^{\frac{1}{2(n+1)}} / \log(N)} + K_3 e^{- \varepsilon^2 N/ (K_3T)}. 
\end{split}
\]
\end{prop}

Our second main result for \(p=2\) concerns the exact recovery of all spikes. Consider the following event: for every \(\varepsilon > 0\) and \(C_0 \in (0,1)\), \(R(\varepsilon, C_0)\) denote 
\begin{equation} \label{eq: set strong recovery p=2}
R(\varepsilon, C_0) \coloneqq \left \{ \boldsymbol{X} \colon m_{ii}(\boldsymbol{X}) \geq 1 - \varepsilon  \enspace \forall i \in [r] \enspace \textnormal{and} \enspace m_{k \ell}(\boldsymbol{X}) \lesssim N^{-\frac{1}{2} \left ( 1 - \frac{1-C_0}{1+C_0} \frac{\lambda_r^2}{\lambda_1^2}\right )} \enspace \forall k, \ell \in [r], k \neq \ell \right \}.
\end{equation}

\begin{prop}[Exact recovery of all spikes for \(p=2\)] \label{thm: strong recovery all spikes Langevin p=2}
Fix \(\beta \in (0,\infty)\), \(p = 2\), and an integer \(n \ge 1\). Let \(\kappa\) denote \(\kappa = \min_{1 \le i \le r-1} \kappa_i\). Consider a sequence of initialization measures \(\mu_N^0 \in \mathcal{M}_1(\mathcal{S}_{N,r})\) satisfying Condition \(0\) at level \(n\) (for inverse temperature \(\beta\)) and Condition \(1\). Then, for every \(\gamma_0\), \(\gamma_1 > 1 >\gamma_2 > 0\), there exist \(\varepsilon_0 >0\) and \(c_0 \in (0,\frac{1}{2} \wedge \frac{\kappa}{2+\kappa})\) such that for every \(\varepsilon < \varepsilon_0\), \(C_0 < c_0\), if 
\[
\sqrt{M}\gtrsim \frac{(n+2) \gamma_0}{\beta \lambda_r^2 C_0 \gamma_2} N^{\frac{1}{2} \left ( \frac{1+C_0}{1-C_0} - \frac{\lambda_r^2}{\lambda_1^2}\right )} \quad \mathrm{and} \log(N) \gtrsim  \frac{1}{1 +\kappa} \log \left (\frac{1}{\kappa} \right ), 
\]
then there exists \(T_0 \le \frac{\gamma_2}{(n+2) \gamma_0} \log(N) N^{-\frac{1}{2} \left ( \frac{1+C_0}{1-C_0} - \frac{\lambda_r^2}{\lambda_1^2}\right )}\) such that for \(N\) sufficiently large,
\[
\int_{\mathcal{S}_{N,r}} \mathbb{Q}_{\boldsymbol{X}} \left (\mathcal{T}_{R(\varepsilon,C_0)} \le T_0 \right ) d \mu_N^0  (\boldsymbol{X})  \geq 1 - \eta (N,n,\gamma_0, \gamma_1, \gamma_2,\varepsilon),
\]
with \(\mathbb{P}\)-probability at least \(1 - \exp(-K N)\), where \(\eta = \eta (N,n,\gamma_0, \gamma_1, \gamma_2, \varepsilon)\) is given by
\[
\begin{split}
\eta &= C_1 e^{-c_1 \gamma_0^2} + C_2 e^{-c_2 \gamma_1^2} + C_3 \gamma_2 + C_4 e^{- c_4 \gamma_2 \sqrt{N}} + K_1 e^{-\gamma_0^3  (n+2)N^{\frac{1}{2} \left ( \frac{1+C_0}{1-C_0} - \frac{\lambda_r^2}{\lambda_1^2}\right )}/(K_1 \gamma_2 \log(N))} \\
& \quad +K_2 e^{-\gamma_0\gamma_2 (n+2)N^{\frac{1}{2} \left ( \frac{1+C_0}{1-C_0} - \frac{\lambda_r^2}{\lambda_1^2}\right )}/(K_2 \log(N))} + K_3 e^{-\varepsilon^2 N^{1+\frac{1}{2} \left ( \frac{1+C_0}{1-C_0} - \frac{\lambda_r^2}{\lambda_1^2}\right )}/K_3}
\end{split}
\]
Furthermore, the correlations \(\{m_{ij}\}_{i,j=1}^r\) follow a sequential elimination.
\end{prop}

Our final main result in this section concerns strong recovery when \(p=2\) and all signal strengths are equal, i.e., \(\lambda_1 = \cdots = \lambda_r \equiv \lambda >0\). In this case, the random landscape \(H\) is invariant under rotations. Therefore, rather than precisely characterizing the recovery of each planted signal by analyzing the evolution of the correlations as in the previous cases, we focus here on \emph{subspace recovery}, meaning recovery of the correct subspace shared by the signal vectors \(\boldsymbol{v}_1, \ldots, \boldsymbol{v}_r\). Specifically, we can look at the distance between the orthogonal projections \(\boldsymbol{X}\boldsymbol{X}^\top\) and \(\boldsymbol{V}\boldsymbol{V}^\top\), where \(\boldsymbol{V}=[\boldsymbol{v}_1, \ldots,\boldsymbol{v}_r] \in \mathcal{S}_{N,r}\) and \(\boldsymbol{X}=[\boldsymbol{x}_1, \ldots,\boldsymbol{x}_r] \in \mathcal{S}_{N,r}\). This can be quantified by 
\[
\frac{1}{N^2} \norm{\boldsymbol{X}\boldsymbol{X}^\top-\boldsymbol{V}\boldsymbol{V}^\top}^2_{\textnormal{F}} = 2 \left ( r-\Tr \left( \boldsymbol{M}^\top \boldsymbol{M}\right) \right),
\]
where \(\boldsymbol{M}\) denotes the \emph{correlation matrix} defined by \(\boldsymbol{M} = \frac{1}{N}\boldsymbol{V}^\top \boldsymbol{X}\). To analyze this, we study the behavior of the eigenvalues \(\theta_1, \ldots, \theta_r\) of \( \boldsymbol{G} = \boldsymbol{M}^\top \boldsymbol{M}\in \R^{r \times r}\) under Langevin dynamics, as stated in the following theorem. To this end, we need to ensure that the eigenvalues of \(\boldsymbol{G}\) at initialization are on the typical scale \(\Theta (N^{-1})\).

\begin{defn}[Condition 1']\label{def: condition 1 prime}
For every \(N \in \N\) and \(\gamma_1 > \gamma_2 >0\), define
\[
\mathcal{C}_1^{'(N)}(\gamma_1,\gamma_2) = \left \{ \boldsymbol{X} \in \mathcal{S}_{N,r} \colon \frac{\gamma_2}{N} \le \theta_i (\boldsymbol{G}(\boldsymbol{X})) < \frac{\gamma_1}{N} \enspace \textnormal{for every} \, i \in [r] \right \}.
\]
We say that a sequence \((\mu_N)_{N \ge 1}\) of probability measure on \(\mathcal{M}_1(\mathcal{S}_{N,r})\) satisfies \emph{Condition \(1'\)} if there exist constants \(C_1, c_1, C_2, c_2, C_3>0\) such that, for every \(N \in \N\) and \(\gamma_1 > \gamma_2 > 0\),
\[
\mu_N \left ( \mathcal{C}_1{'(N)}(\gamma_1,\gamma_2)^\textnormal{c} \right) \le C_1 e^{-c_1 \gamma_1^2} + C_2  e^{-c_2 \gamma_2 \sqrt{N}} + C_3 \gamma_2 .
\]
\end{defn} 

Lemma~\ref{lem: concentration eigenvalues} ensures that the uniform measure \(\mu_{N \times r}\) on \(\mathcal{S}_{N,r}\) satisfies Condition \(1'\). 

Finally, the following result determines the sample complexity for subspace recovery. We write \(\theta_{\min}(\boldsymbol{X}) \coloneqq \min_{1 \le i \le r} \theta_i(\boldsymbol{X})\).

\begin{prop}[Subspace recovery for \(p=2\)] \label{thm: strong recovery isotropic Langevin p=2}
Assume \(\lambda_1 = \cdots = \lambda_r \equiv \lambda > 0\). Fix \(\beta \in (0,\infty)\), \(p = 2\), and an integer \(n \ge 1\). Consider a sequence of initialization measures \(\mu_N^0 \in \mathcal{M}_1(\mathcal{S}_{N,r})\) satisfying Condition \(0\) at level \(n\) (for inverse temperature \(\beta\)) and Condition \(1'\). Then, for every \(\gamma_0>0\) and \(\gamma_1 > 1 >\gamma_2 > 0\), there exist \(\varepsilon_0 >0\) and \(c_0 \in (0,\frac{1}{3r})\) such that for every \(\varepsilon < \varepsilon_0\) and \(C_0 < c_0\), if
\[
\sqrt{M}\gtrsim \frac{(n+2) \gamma_0^2 \gamma_1}{\beta \lambda^2 C_0 \gamma_2} N^{\delta(n,C_0)}, 
\]
then there exists \(T_0 \lesssim \frac{1+C_0^2}{(n+2) \gamma_0^2 \gamma_1} \log(N) N^{- \delta(n,C_0)}\) such that for every \(T > T_0\),
\[
\int_{\mathcal{S}_{N,r}} \mathbb{Q}_{\boldsymbol{X}} \left (\inf_{t \in [T_0,T]} \theta_{\min} ( \boldsymbol{X}_t^\beta ) \ge 1 - \varepsilon \right ) d \mu_N^0 (\boldsymbol{X}) \geq 1 - \eta (N, n,\gamma_0, \gamma_1, \gamma_2, \varepsilon,T),
\]
with \(\mathbb{P}\)-probability at least \(1 - \exp(-K N)\), where \(\delta(n,C_0) = \frac{1}{2(n+1)} \vee \frac{2 C_0 r}{1-C_0r}\) and \(\eta=\eta (N, n,\gamma_0, \gamma_1, \gamma_2, \varepsilon,T)\) is given by
\[
\begin{split}
\eta &= C_1 e^{-c_1 \gamma_0^2} + C_2 e^{-c_2 \gamma_1^2} + C_3 \gamma_2 + C_4 e^{- c_4 \gamma_2 \sqrt{N}} + K_1 e^{-\gamma_0^4\gamma_1 (n+2) N^{\delta(n,C_0)}/(K_1 \log(N))} \\
& \quad + K_2 e^{-\gamma_0^2 \gamma_2 (n+2) N^{\delta(n,C_0)}/K_2} +  K_3 e^{- \gamma_0^2 \gamma_1 (n+2) N^{1 - \frac{4C_0r}{1+C_0r} + \delta(n,C_0)} / (K_3 C_0^2 \log(N))} + K_4 e^{-N \varepsilon^2/(K_4 T)}.
\end{split}
\]
\end{prop}

Propositions~\ref{thm: strong recovery first spike Langevin p>2} and~\ref{thm: strong recovery all spikes Langevin p>2} are proved in Section~\ref{section: proof recovery Langevin p>2}. Propositions~\ref{thm: strong recovery first spike Langevin p=2} and~\ref{thm: strong recovery all spikes Langevin p=2} are proved in Section~\ref{section: proof recovery Langevin p=2}, while the proof of Proposition~\ref{thm: strong recovery isotropic Langevin p=2} is provided in Section~\ref{section: proof recovery Langevin isotropic p=2}. Our asymptotic results in Subsection~\ref{subsection: main asymptotic results} follow directly from the above statements together with the definitions of the initial conditions. We refer the reader to the proof of Theorem 1.3 in~\cite{benarous2024gradientflow} for arguments that can be adapted to the present setting of Langevin dynamics.
\section{Preliminary results} \label{preliminaries}

In this section, we present preliminary results necessary for proving the main results in Section~\ref{section: Langevin}. Specifically, we study the regularity of the Hamiltonian \(H_0 \colon \mathcal{S}_{N,r} \to \R\) and introduce the \emph{bounding flows method} from~\cite{ben2020bounding}, which plays a crucial role in deriving estimates for \(\norm{L_{0,\beta} m_{ij}^{(N)}}_\infty\). Additionally, we present the evolution equations governing the correlations \(\{m_{ij}^{(N)}\}_{1 \le i,j \le r}\) as well as the equations satisfied by the entries of \(\boldsymbol{M}^\top \boldsymbol{M}\), where \(\boldsymbol{M} = (m_{ij}^{(N)})_{1 \le i,j \le r}\). The results of this section generalize those presented in~\cite{arous2020algorithmic}. To simplify notation, we will omit explicit reference to \(N\) in \(m_{ij}^{(N)}(\boldsymbol{X})\) and instead write \(m_{ij}(\boldsymbol{X})\).

\subsection{Regularity of noise, ladder relations, and bounding flows method}

Recall that the Hamiltonian \(H_0 \colon \mathcal{S}_{N,r} \to \R\) defined by
\[
H_0 (\boldsymbol{X}) = N^{-\frac{p-1}{2}} \sum_{i=1}^r \lambda_i \langle \boldsymbol{W}, \boldsymbol{x}_i^{\otimes p}\rangle,
\]
where \(\boldsymbol{W} \in (\R^N)^{\otimes p}\) is an order-\(p\) tensor with i.i.d.\ entries \(W_{i_1, \ldots, i_p} \sim \mathcal{N}(0,1)\) and \(\mathcal{S}_{N,r}\) is the normalized Stiefel manifold. Similar to the works~\cite{ben2020bounding, arous2020algorithmic}, we work with the \(\mathcal{G}\)-norm which is motivated by the homogeneous Sobolev norm. 

\begin{defn}[\(\mathcal{G}\)-norm on \(\mathcal{S}_{N,r}\)] \label{def: G norm}
For any integer \(k\), we say that a function \(F \colon \mathcal{S}_{N,r} \to \R\) is in the space \(\mathcal{G}^k(\mathcal{S}_{N,r})\) if
\[
\norm{F}_{\mathcal{G}^k} \coloneqq \sum_{0 \leq \ell \leq k} N^{\ell/2} \norm{|\nabla^\ell F|_{\text{\textnormal{op}}}}_{L^\infty(\mathcal{S}_{N,r})} < \infty.
\]
Here, \(\nabla^\ell F\) denotes the \(\ell\)th Riemannian (covariant) derivative of \(F\), defined as a tensor field of order \(\ell\). For every \(\boldsymbol{X} \in \mathcal{S}_{N,r}\), it defines an \(\ell\)-linear map on the tangent space \(T_{\boldsymbol{X}} \mathcal{S}_{N,r}\):
\[
\nabla^\ell F(\boldsymbol{X}) \colon T_{\boldsymbol{X}} \mathcal{S}_{N,r} \times \cdots \times T_{\boldsymbol{X}} \mathcal{S}_{N,r} \to \R,
\]
and this map is defined recursively by
\[
\nabla^\ell F (\boldsymbol{X} ; \boldsymbol{U}_1, \ldots, \boldsymbol{U}_\ell) = \nabla_{\boldsymbol{U}_1} \nabla^{\ell-1} F (\boldsymbol{X} ; \boldsymbol{U}_2, \ldots, \boldsymbol{U}_\ell) - \sum_{i=2}^\ell \nabla^{\ell-1} F (\boldsymbol{X} ; \boldsymbol{U}_2, \ldots, \nabla_{\boldsymbol{U}_1} \boldsymbol{U}_i, \ldots, \boldsymbol{U}_\ell) ,
\]
for all \(\boldsymbol{U}_1, \ldots, \boldsymbol{U}_\ell \in T_{\boldsymbol{X}} \mathcal{S}_{N,r}\). The operator norm of \(\nabla^\ell F \) is given by
\[
|\nabla^\ell F|_{\textnormal{op}} (\boldsymbol{X}) = \sup_{\boldsymbol{U}_1, \ldots, \boldsymbol{U}_\ell \in T_{\boldsymbol{X}} \mathcal{S}_{N,r} \atop \norm{\boldsymbol{U}_i}_{\textnormal{F}} \le 1} |\nabla^\ell F (\boldsymbol{X} ; \boldsymbol{U}_1, \ldots, \boldsymbol{U}_\ell) |,
\]
where \(\norm{\boldsymbol{U}}_{\textnormal{F}} = \sqrt{\Tr (\boldsymbol{U}^\top \boldsymbol{U})}\) is the Frobenius norm. For further details, see~\cite[Section 10.7]{boumal2023} and the references therein.
\end{defn}

We emphasize that this definition is a generalization of the \(\mathcal{G}\)-norm introduced by~\cite{ben2020bounding} for functions on the sphere \(\mathbb{S}^{N-1}(\sqrt{N})\). We now present an important estimate for the \(\mathcal{G}\)-norm of \(H_0\).

\begin{lem}[Regularity of \(H_0\)] \label{lem: regularity H0}
For every \(n\), there exist positive constants \(C_1\) and \(C_2\), depending on \(p\) and \(n\), such that
\[
\mathbb{P}\left(\norm{H_0}_{\mathcal{G}^n} \geq C_1 \left (\sum_{i=1}^r \lambda_i \right ) N \right) \leq \exp \left (- C_2 \frac{(\sum_{i=1}^r \lambda_i)^2}{\sum_{i=1}^r \lambda_i^2} N\right ).
\]
\end{lem}

Lemma~\ref{lem: regularity H0} reduces to~\cite[Theorem 4.3]{ben2020bounding} in the special case \(r=1\). Its proof follows the same strategy as that of~\cite{ben2020bounding} and one can therefore mimic the arguments; thereby we thus left it. We next present the ladder relations, which will be useful to bound \(\norm{L_{0,\beta} m_{ij}}_\infty\), where we recall from~\eqref{eq: generator Langevin noise} that the generator \(L_{0,\beta}\) is given by \(L_{0,\beta} = \Delta- \beta \langle \nabla H_0, \hat{\nabla}\cdot \rangle \). 
 
 \begin{lem}[Ladder relations] \label{lem: ladder relations}
Let \(L\) be a linear operator acting on the space of smooth functions \(F \colon \mathcal{S}_{N,r} \to \R\), and let \(n \geq m \geq 1\) be integers. Define 
\[
\norm{L}_{\mathcal{G}^n \to \mathcal{G}^m}  = \sup_{F \in  \mathcal{G}^n(\mathcal{S}_{N,r})} \frac{\norm{LF}_{\mathcal{G}^m}}{ \norm{F}_{\mathcal{G}^n}}.
\]
Then, for every \(n \geq 1\), there exists a constant \(c(n)\) such that for every \(N, r\), and every \(G \in \mathcal{G}^n (\mathcal{S}_{N,r})\),
\begin{align}
\norm{\Delta}_{\mathcal{G}^n \to \mathcal{G}^{n-2}} &\leq r \label{eq: ladder relation 1} \\
\norm{\langle \nabla G, \nabla \cdot \rangle}_{\mathcal{G}^n \to \mathcal{G}^{n-1}} &\leq \frac{c(n)}{N} \norm{G}_{\mathcal{G}^n}.\label{eq: ladder relation 2}
\end{align}
\end{lem}
\begin{proof}
We first prove~\eqref{eq: ladder relation 1}. Since \(\Delta F = \Tr(\nabla^2 F)\) and \(\nabla \Tr(F) = \Tr(\nabla F)\), we have
\[
\left \lvert \nabla^\ell \Delta F \right \rvert _{\text{\textnormal{op}}} = \left \lvert \nabla^\ell \Tr (\nabla^2 F) \right \rvert _{\text{\textnormal{op}}} = \left \lvert \Tr (\nabla^{\ell + 2} F) \right \rvert _{\text{\textnormal{op}}} .
\]
Moreover, since \(\dim(\mathcal{S}_{N,r})= Nr- \frac{r(r+1)}{2} \leq Nr\), we have \(\left \lvert \Tr (\nabla^{\ell + 2} F) \right \rvert _{\text{\textnormal{op}}}  \leq Nr \left \lvert \nabla^{\ell + 2} F\right \rvert _{\text{\textnormal{op}}}\), thus
\[
\norm{\Delta F}_{\mathcal{G}^{n-2}} = \sum_{\ell=0}^{n-2}N^{\ell/2} \norm{|\nabla^\ell \Delta F|_{\text{\textnormal{op}}}}_\infty \leq r \sum_{\ell=0}^{n-2} N^{(\ell + 2)/2} \norm{|\nabla ^{\ell+2} \Delta F |_{\text{\textnormal{op}}}}_\infty \leq r\norm{F}_{\mathcal{G}^{n}},
\]
as desired. Next, we show~\eqref{eq: ladder relation 2}. By the general Leibniz rule and Cauchy-Schwarz, we have
\[
\begin{split}
\norm{\langle \nabla G, \nabla F \rangle}_{\mathcal{G}^{n-1}}& =\sum_{\ell=0}^{n-1} N^{\ell/2} \sum_{k=0}^\ell {\ell \choose k}  \norm{|\langle \nabla^{k+1} G, \nabla^{\ell -k +1} F \rangle |_{\text{\textnormal{op}}}}_{L^\infty}\\
&\leq \sum_{0 \leq k \leq \ell \leq n-1} N^{\ell/2} {\ell \choose k} \norm{| \nabla^{k+1} G |_{\text{\textnormal{op}}}}_\infty \norm{| \nabla^{\ell-k+1} F |_{\text{\textnormal{op}}}}_\infty\\
&\leq \frac{c(n)}{N}\norm{G}_{\mathcal{G}^n} \norm{F}_{\mathcal{G}^n},
\end{split}
\]
where the inequality in the last line follows by definition of the \(\mathcal{G}\)-norm.
\end{proof}

Using the ladder relations we can estimate \(\norm{L_{0,\beta} m_{ij}}_\infty\) for every \(1 \leq i,j \leq r\). In light of Lemma~\ref{lem: regularity H0}, for every \(n \geq 1\), there exist constants \(K = K (p,n, \{\lambda_i\}_{i=1}^r)\) and \(C = C (p,n, \{\lambda_i\}_{i=1}^r)\) such that 
\[
\norm{H_0}_{\mathcal{G}^n} \leq C N, 
\]
with \(\mathbb{P}\)-probability at least \(1 - \exp(- K N)\). Moreover, a direct computation shows \(\norm{m_{ij}}_{\mathcal{G}^n} \leq c(n)\). Therefore, by Lemma~\ref{lem: ladder relations}, there exists a constant \(\Lambda = \Lambda(p, r, \{\lambda_i\}_{i=1}^r,\beta)\) such that 
\begin{equation} \label{eq: bound norm L_0m_ij}
\norm{L_{0,\beta} m_{ij}}_\infty \leq \norm{\Delta m_{ij}}_\infty + \beta \norm{ \langle \nabla H_0, \hat{\nabla} m_{ij}\rangle}_\infty \leq r \norm{m_{ij}}_{\mathcal{G}^2} + \frac{\beta}{N} \norm{H_0}_{\mathcal{G}^1} \norm{m_{ij}}_{\mathcal{G}^1} \leq \Lambda
\end{equation}
with \(\mathbb{P}\)-probability at least \(1 - \exp(-KN)\). This estimate turns out to be suboptimal. A sharper bound is achieved using the \emph{bounding flows} method. This technique was introduced in~\cite{ben2020bounding} and later applied in~\cite{arous2020algorithmic} to provide a precise control over the evolution of functions under Langevin dynamics and gradient flows on \(\mathbb{S}^{N-1}(\sqrt{N})\). Here, we extend the method in order to obtain more accurate bounds for the evolution of functions under Langevin dynamics on \(\mathcal{S}_{N,r}\). In particular, the following result generalizes~\cite[Theorem 5.3]{arous2020algorithmic}.

\begin{lem}[Bounding flows on \(\mathcal{S}_{N,r}\)] \label{lem: bounding flows}
Let \(I_\gamma = [-\frac{\gamma}{\sqrt{N}}, \frac{\gamma}{\sqrt{N}}]\) for \(\gamma>0\), and let \(D \subset \mathcal{S}_{N,r}\) be open. Consider an Itô process \((\boldsymbol{X}_t)_{t \ge 0}\) taking values in \(D\), with infinitesimal generator \(L\), and initial condition \(\boldsymbol{X}_0 \in D\). We write \(\mathbb{Q}_{\boldsymbol{X}_0}\) for the law of \((\boldsymbol{X}_t)_{t\ge0}\) started from \(\boldsymbol{X}_0\). Let the exit time from \(D\) be
\[
\mathcal{T}_{D^\textnormal{c}} = \inf \{ t \ge 0 \colon \boldsymbol{X}_t \notin D\}.
\]
Let \(F \colon D \to \R\) be a smooth function. Assume that the following conditions are satisfied for some integer \(n \geq 1\):
\begin{enumerate}
\item[(1)] \(L\) is a differential operator of the form \(L = L_0 + \sum_{1 \leq i,j \leq r} a_{ij}(\boldsymbol{X}) A_{ij}\), where
\begin{enumerate}
\item[(a)] \(A_{ij} = \langle \nabla \psi_{ij}, \hat{\nabla} \cdot \rangle\) for some function \(\psi_{ij} \in C^\infty\) satisfying \(\norm{\psi_{ij}}_{\mathcal{G}^1} \leq c_1 N\),
\item[(b)] \(a_{ij} \in C (\mathcal{S}_{N,r})\),
\item[(c)] \(L_0 = \Delta + \langle \nabla U, \hat{\nabla} \cdot \rangle \) for some \(U \in C^{\infty} (\mathcal{S}_{N,r})\) satisfying \(\norm{U}_{\mathcal{G}^{2n}} \leq c_2(n) N\).
\end{enumerate}
\item[(2)] \(F\) is smooth with \(\norm{F}_{\mathcal{G}^{2n}} \leq c_3(n) \).
\item[(3)] There exists \(\gamma > 0\) such that \(L_0^k F(\boldsymbol{X}_0) \in I_{\gamma}\) for every \(0 \leq k \leq n-1\).
\item[(4)] For every \(i,j \in [r]\) there exist \(\epsilon \in (0,1)\) and \(T_0^{(ij)}>0\), possibly depending on \(\epsilon\), such that for every \(t \leq \mathcal{T}_{D^\text{c}} \wedge T_0^{(ij)}\), 
\[
\int_0^t |a_{ij}(\boldsymbol{X}_s)| d s \leq \epsilon |a_{ij}(\boldsymbol{X}_t)|.
\]
\end{enumerate}
Then, there exists a constant \(K_1>0\) depending only on \(c_1,c_2,c_3\), and \(\gamma\), such that, for every \(T_0 > 0\), 
\begin{equation} \label{eq: bounding flow}
| F(\boldsymbol{X}_t)| \leq K_1 \left (  \frac{\gamma}{\sqrt{N}} \sum_{k=0}^{n-1} t^k + t^n + \frac{1}{1-\epsilon} \sum_{1 \leq i,j \leq r}\int_0^t |a_{ij}(\boldsymbol{X}_s)| ds\right )
\end{equation}
for every \(t \leq \mathcal{T}_{D^\text{c}} \wedge \min_{1 \leq i,j \leq r} T_0^{(ij)} \wedge T_0\) with \(\mathbb{Q}_{\boldsymbol{X}_0}\)-probability at least \(1- K_2 \exp \left ( - \frac{\gamma^2 }{K_2 T_0}\right)\). 

If item (1)(c) is replaced by
\begin{enumerate}
\item[(c')] \(L_0 = \langle \nabla U, \hat{\nabla} \cdot\rangle\) for some \(U \in C^\infty\) such that \(\norm{U}_{\mathcal{G}^{2n}} \leq c_2N\),
\end{enumerate}
then~\eqref{eq: bounding flow} holds deterministically for every \(t \leq \mathcal{T}_{D^\text{c}} \wedge \min_{1 \leq i,j \leq r} T_0^{(ij)} \wedge T_0\).

If item (3) is replaced by
\begin{enumerate}
\item[(3')] There exist \(T_1,\gamma >0\) such that \(\boldsymbol{X}_0\) satisfies \(e^{tL_0}F(\boldsymbol{X}_0) \in I_\gamma\) for every \(t< T_1\),
\end{enumerate}
then~\eqref{eq: bounding flow} holds for every \(t \leq \mathcal{T}_{D^\text{c}} \wedge \min_{1 \leq i,j \leq r} T_0^{(ij)} \wedge T_0 \wedge T_1 \wedge 1\), with \(\mathbb{Q}_{\boldsymbol{X}}\)-probability at least \(1- K_2 \exp \left ( - \frac{\gamma^2 }{K_2 T_0}\right)\).
\end{lem}

\begin{proof}
We mimic the proof of~\cite[Theorem 5.3]{arous2020algorithmic}. We claim that the function \(F\) can be expanded as
\begin{equation} \label{eq: expansion via Ito}
\begin{split}
F(t) & = F(0) + M_t^F + \sum_{k=1}^{n-1} \int_0^t \cdots \int_0^{t_{k -1}} \left (L_0^k F(0) + M_{t_k}^{L_0^k F} \right) dt_k \cdots dt_1  \\
& \quad + \int_0^t \cdots \int_0^{t_{n -1}} L_0^n F(t_n) dt_n \cdots dt_1 \\
& \quad + \sum_{1 \leq i,j \leq r} \sum_{k=1}^n \int_0^t \cdots \int_0^{t_{k -1}} a_{ij}(t_k) A_{ij} L_0^{k-1} F(t_k) dt_k \cdots dt_1.
\end{split}
\end{equation}
The proof is by induction on \(n\). The claim is verified for \(n=1\) by Lemma~\ref{lem: Ito} since
\[
F(t) = F(0) + M_t^F + \int_0^t LF(s) ds = F(0) + M_t^F + \int_0^t L_0 F(s) ds + \sum_{1 \leq i,j \leq r} \int_0^t a_{ij}(s) A_{ij} F(s) ds.
\]
Assume that the result holds in the \(n\)th case. By definition of \(M_{t_n}^{L_0^n F}\) we find that
\[
\begin{split}
L_0^n F(t_n) &= L_0^n F(0) + M_{t_n}^{L_0^n F} + \int_0^{t_n} L L_0^n F(t_{n+1}) dt_{n+1}\\
&= L_0^n F(0) + M_{t_n}^{L_0^n F} + \int_0^{t_n} L_0^{n+1} F(t_{n+1})dt_{n+1}+ \sum_{1 \leq i,j \leq r} \int_0^{t_n} a_{ij}(t_{n+1}) A_{ij} L_0^n F(t_{n+1})dt_{n+1}.
\end{split}
\]
Therefore, for the \((n+1)\)st case we use the above equality to expand the second-to-last term as 
\[
\begin{split}
&\int_0^t \cdots \int_0^{t_{n-1}} L_0^n F(t_n) dt_n \cdots dt_1 \\
& = \int_0^t \cdots \int_0^{t_{n-1}}  \left (L_0^n F(0) + M_{t_n}^{L_0^n F} \right) dt_n \cdots dt_1 +  \int_0^t \cdots \int_0^{t_{n-1}} \int_0^{t_n} L_0^{n+1} F(t_{n+1}) dt_{n+1} dt_n \cdots dt_1\\
& \quad + \sum_{1 \leq i,j \leq r} \int_0^t \cdots \int_0^{t_{n-1}} \int_0^{t_n} a_{ij}(t_{n+1}) A_{ij} L_0^n F(t_{n+1}) dt_{n+1} dt_n \cdots dt_1.
\end{split}
\]
Combining the terms yields the desired expression by induction. 

We next bound the absolute values in~\eqref{eq: expansion via Ito} term-by-term. We start with the first line of~\eqref{eq: expansion via Ito} and claim that for every \(1 \leq k \leq n-1\),
\begin{equation} \label{eq: claim}
\norm{L_0^k F}_{\mathcal{G}^{2n-2k}} \leq (r+c(n)c_2)^k \norm{F}_{\mathcal{G}^{2n}} \leq (r+c(n)c_2)^k c_3. 
\end{equation}
The second inequality in~\eqref{eq: claim} follows by assumption (2). We next prove the first inequality in~\eqref{eq: claim} by induction on \(k\). The claim is verified for \(k=1\) by Lemma~\ref{lem: ladder relations} since
\[
\norm{L_0 F}_{\mathcal{G}^{2n-2}} \leq r \norm{F}_{\mathcal{G}^{2n}} + \frac{2^n -1}{N} \norm{U}_{\mathcal{G}^{2n}} \norm{F}_{\mathcal{G}^{2n}} \leq (r+(2^n -1)c_2) \norm{F}_{\mathcal{G}^{2n}},
\]
where we used assumption (1c). Assume that~\eqref{eq: claim} holds in the \(k\)th case. Then, in the \((k+1)\)th case we have
\[
\begin{split}
\norm{L_0^{k+1}F}_{\mathcal{G}^{2n-2(k+1)}} & \leq r \norm{L_0^k F}_{\mathcal{G}^{2n-2k}} + \frac{2^n-1}{N} \norm{U}_{\mathcal{G}^{2n-2k}} \norm{L_0^k F}_{\mathcal{G}^{2n-2k}} \\
& \leq (r+(2^n-1)c_2) \norm{L_0^k F}_{\mathcal{G}^{2n-2k}} \\
& \leq (r+(2^n-1)c_2) (r+(2^n-1)c_2)^k \norm{F}_{\mathcal{G}^{2n}},
\end{split}
\]
where the last inequality follows by induction hypothesis. This proves the claim~\eqref{eq: claim}. It then follows that
\[
\norm{|\nabla_{\mathcal{S}_{N,r}} L_0^k F|_{\textnormal{op}}}_\infty \leq N^{-1/2} \norm{L_0^k F}_{\mathcal{G}^{2n-2k}} \leq C N^{-1/2},
\]
where the first inequality follows by Definition~\ref{def: G norm} and the second inequality by~\eqref{eq: claim}. Therefore, by Lemma~\ref{lem: Doob inequality}, there exists a universal constant \(K_2 >0\) such that for every \(\gamma, T_0 > 0\) and every \(N\),
\[
\sup_{\boldsymbol{X} \in \mathcal{S}_{N,r}} \mathbb{Q}_{\boldsymbol{X}} \left (\sup_{t \in [0,T_0]} |M_t^{L_0^k F}| \leq \frac{\gamma}{\sqrt{N}} \right ) \geq 1 - K_2 \exp(-\gamma^2 / (K_2 T_0)).
\]
For every \(0 \leq k \leq n-1\), we then bound \(|L_0^k F(0)|\) by \(\gamma N^{-\frac{1}{2}}\) using assumption (3). We therefore have that the first line of~\eqref{eq: expansion via Ito} is bounded by
\[
\sum_{k=0}^{n-1} \int_0^t \cdots \int_0^{t_{k-1}} \left(\left |L_0^k F(0) \right| + \left |M_{t_k}^{L_0^k F}\right | \right) dt_k \cdots dt_1 \leq \frac{2\gamma}{\sqrt{N}}\sum_{k=0}^{n-1} \frac{t^k}{k!} \leq \frac{2\gamma}{\sqrt{N}}\sum_{k=0}^{n-1} t^k,
\]
with \(\mathbb{Q}_{\boldsymbol{X}_0}\)-probability at least \(1 - K_2\exp(-\gamma^2 / (K_2 T_0))\). Since \(|L_0^n F| = \norm{L_0^n F}_\infty \leq C(n,r,c_2,c_3)\) by~\eqref{eq: claim}, the second line of~\eqref{eq: expansion via Ito} is bounded by
\[
\int_0^t \cdots \int_0^{t_{n-1}}|L_0^n F(t_n)| dt_n \cdots dt_1 \leq C \frac{t^n}{n!} \leq C' t^n.
\]
We turn to the last line of~\eqref{eq: expansion via Ito}. According to Lemma~\ref{lem: ladder relations} and~\eqref{eq: claim}, it follows that 
\[
|A_{ij} L_0^{k-1} F| = \norm{A_{ij} L_0^{k-1} F}_\infty \leq \frac{1}{N} \norm{\psi_{ij}}_{\mathcal{G}^1} \norm{L_0^{k-1} F}_{\mathcal{G}^1} \leq \tilde{C}(n,r,c_1,c_2,c_3).
\]
Therefore, using assumption (4), we bound the last line of~\eqref{eq: expansion via Ito} by
\[
\begin{split}
\sum_{1 \leq i,j \leq r} \sum_{k=1}^n \int_0^t \cdots \int_0^{t_{k-1}} |a_{ij}(t_k)| dt_k \cdots dt_1 &\leq \sum_{1 \leq i,j \leq r} \sum_{k=1}^n \epsilon ^{k-1} \int_0^t |a_{ij}(s)| ds \leq \frac{1}{1-\epsilon} \sum_{1 \leq i,j \leq r} \int_0^t |a_{ij}(s)| ds.
\end{split}
\]
Choosing \(K = 2 \vee C' \vee \tilde{C}\) gives the desired estimate. 

To prove the result under assumption \((3')\), we first note that the third term in~\eqref{eq: expansion via Ito} can be rewritten as  
\[
\sum_{k=1}^{n-1} \int_0^t \cdots \int_0^{t_{k-1}} L_0^k F(0) dt_k \cdots d t_1 = \sum_{k=1}^{n-1} L_0^k F(0) \frac{t^k}{k!}.
\]
For every \(t \in (0,1)\), an order-$n$ Taylor expansion gives
\[
e^{tL_0}F(0) = \sum_{k=1}^{n-1}L_0^{k}F(0)\frac{t^{k}}{k!} + L_0^{n} F(\tilde{t}) \frac{t^{n}}{n!},
\]
for some $\tilde{t} \in [0,t]$. Therefore, the following estimate holds according to the ladder relations (see Lemma~\ref{lem: ladder relations}):
\[
\vert e^{tL_0}F(0)-\sum_{k=1}^{n-1}L_0^{k}F(0)\frac{t^{k}}{k!} \vert \leq  \norm{L_0^{n}F}_\infty\frac{t^{n}}{n!}.
\]
\end{proof}

\subsection{Evolution equations}

We are interested in the evolution of the correlations \(m_{ij}\) and of the matrix-valued function \(\boldsymbol{G}(\boldsymbol{X}) = \boldsymbol{M}^\top \boldsymbol{M}\) with \(\boldsymbol{M} = (m_{ij} (\boldsymbol{X}))_{1 \le i,j \le r}\) under Langevin dynamics. We start by recalling Itô's formula in \(\R^{N\times r}\).

\begin{lem}[Itô's formula in \(\R^{N\times r}\)] \label{lem: Ito}
Let \(\boldsymbol{X}_t\) solve the following \(\R^{N\times r}\)-valued SDE:
\[
d\boldsymbol{X}_t = a(\boldsymbol{X}_t) dt + b d\boldsymbol{B}_t, t >0 
\]
with initial solution \(\boldsymbol{X}_0 \in \R^{N \times r}\). Here, \((\boldsymbol{B}_t)_{t \geq 0}\) is a \(\R^{N \times r}\)-valued Brownian motion, \(a \colon \R^{N \times r} \to \R^{N \times r}\) is a measurable function, and \(b\) is a constant. Suppose that \(F\) is a \(C^2\) function from some open domain \(D \subset \R^{N \times r}\) into \(\R\). Suppose that almost surely, \(\boldsymbol{X}_t \in D\) for all \(t \geq0\). Then, the process \((F (\boldsymbol{X}_t ) )_{ t \geq 0}\) satisfies
\[
F(\boldsymbol{X}_t) = F(\boldsymbol{X}_0) + M_t^F + \int_0^t LF(\boldsymbol{X}_s) ds,
\]
where \(M_t^F = b \int_0^t \langle \hat{\nabla}F(\boldsymbol{X}_s), d\boldsymbol{B}_s \rangle \) is a local martingale such that \([M_t^F] \leq b^2 t \norm{ \hat{\nabla}F }_{L^\infty}^2\) and \(L F(\boldsymbol{X}_t)=\frac{b^2}{2} \hat{\Delta}F(\boldsymbol{X}_t) + \langle \hat{\nabla}F(\boldsymbol{X}_t), \hat{\nabla}a(\boldsymbol{X}_t) \rangle \). 
\end{lem}

For every \(1 \leq i,j \leq r\), the correlations \(m_{ij}\) are smooth functions from \(\mathcal{S}_{N,r} \subset \R^{N \times r}\) to \(\R\). Thus, according to the Itô's formula, we have
\[
m_{ij}(\boldsymbol{X}_t^\beta) = m_{ij}(\boldsymbol{X}_0) + M_t^{m_{ij}} + \int_0^t L_\beta m_{ij}(\boldsymbol{X}_s^\beta) ds,
\]
where \(M_t^{m_{ij}} = \sqrt{2} \int_0^t \langle \nabla m_{ij}(\boldsymbol{X}_s^\beta), d \boldsymbol{B}_s\rangle \) and \(L_\beta m_{ij}(\boldsymbol{X}_t^\beta) =\Delta m_{ij}(\boldsymbol{X}_t^\beta) - \beta \langle \nabla H(X_t^\beta), \hat{\nabla}m_{ij}(\boldsymbol{X}_t^\beta)\rangle\), as stated by~\eqref{eq: generator Langevin}. An explicit computation of the generator gives the following evolution equations for \(\{m_{ij}(X_t^\beta)\}_{1 \leq i,j \leq r}\). We write \(m_{ij}(t) = m_{ij}(\boldsymbol{X}_t^\beta)\) to simplify notation slightly.

\begin{lem}[Evolution equation for \(m_{ij}\)] \label{lem: evolution equation m_ij}
For every \(1 \leq i,j \leq r\), the evolution equation for \(m_{ij}\) is given by
\[
dm_{ij}(t) = L_\beta m_{ij}(t) dt + d M_t^{m_{ij}},
\]
where 
\[
L_\beta m_{ij} = L_{0,\beta} m_{ij} + \beta \sqrt{M} p \lambda_i \lambda_j m_{ij}^{p-1}  -  \beta \sqrt{M}\frac{p}{2}\sum_{1 \leq k, \ell \leq r} \lambda_k m_{kj} m_{k \ell} m_{i \ell}\left (\lambda_j m_{kj}^{p-2} + \lambda_\ell m_{k \ell}^{p-2} \right) ,
\]
and
\[
L_{0,\beta} m_{ij}  = \Delta m_{ij}  - \beta \langle \nabla H_0, \hat{\nabla}m_{ij} \rangle.
\]
\end{lem}
\begin{proof}
According to~\eqref{eq: generator Langevin}, we have 
\[
L_\beta m_{ij} = \Delta m_{ij} - \beta \langle \nabla  H, \hat{\nabla}m_{ij} \rangle = L_{0,\beta} m_{ij} + \beta  \langle \nabla  \Phi, \hat{\nabla}m_{ij} \rangle, 
\]
where \(\Phi(\boldsymbol{X}) = N \sqrt{M} \sum_{1 \le i,j \le r} \lambda_i \lambda_j m_{ij}^p(\boldsymbol{X})\). The Riemannian gradient of \(\Phi\) on the manifold \(\mathcal{S}_{N,r}\) is given by 
\[
\nabla \Phi(\boldsymbol{X}) = \hat{\nabla}\Phi(\boldsymbol{X}) - \frac{1}{2N}\boldsymbol{X} \left (\boldsymbol{X}^\top \hat{\nabla}\Phi(\boldsymbol{X}) + \hat{\nabla}\Phi(\boldsymbol{X})^\top \boldsymbol{X} \right ).
\]
An explicit computation gives that
\[
\begin{split}
\left (\nabla \Phi(\boldsymbol{X}) \right )_j & = 
\nabla_{\boldsymbol{x}_j} \Phi(\boldsymbol{X})  - \frac{1}{2N} \sum_{i=1}^r \left ( \boldsymbol{X}^\top \hat{\nabla}\Phi(\boldsymbol{X})  + \hat{\nabla}\Phi(\boldsymbol{X})^\top \boldsymbol{X} \right )_{ij} \boldsymbol{x}_i\\
& =  \sqrt{M} \sum_{k=1}^r p \lambda_k\lambda_j m_{kj}^{p-1} \boldsymbol{v}_k  - \sqrt{M} \frac{p}{2} \sum_{k=1}^r \sum_{\ell=1}^r  \lambda_k m_{kj} m_{k\ell} \left (\lambda_j  m_{kj}^{p-2} +  \lambda_\ell m_{k\ell}^{p-2} \right ) \boldsymbol{x}_\ell.
\end{split}
\]
Since \(\nabla_{\boldsymbol{x}_j} m_{ij} = \frac{\boldsymbol{v}_i}{N}\) and \(\langle \boldsymbol{v}_i, \boldsymbol{v}_j \rangle = N\delta_{ij}\), we obtain 
\[
\langle \nabla  \Phi, \hat{\nabla}m_{ij} \rangle = \sqrt{M} p \lambda_i \lambda_j m_{ij}^{p-1} - \sqrt{M}\frac{p}{2} \sum_{1 \le k,\ell \le r} \lambda_k m_{i \ell}m_{kj} m_{k\ell} \left (\lambda_j  m_{kj}^{p-2} +  \lambda_\ell m_{k\ell}^{p-2} \right ).
\]
\end{proof}

We also recall a classical estimate to bound the martingale part of the evolution of a function under Langevin dynamics, based on the Doob's maximal inequality.

\begin{lem}[Sub-Gaussian tail bound for local martingales] \label{lem: Doob inequality} 
Let \(F \colon \mathcal{S}_{N,r} \to \R\) be a smooth function such that \(\norm{F}_{\mathcal{G}^1} \leq K\) and let \(F(t) = F(\boldsymbol{X}_t^\beta)\) denote its evolution under Langevin dynamics~\eqref{eq: langevin dynamics}. Then, for every \(\varepsilon,T>0\), and every \(N\), 
\begin{equation*} 
\sup_{\boldsymbol{X}_0 \in \mathcal{S}_{N,r}} \mathbb{Q}_{\boldsymbol{X}_0} \left ( \sup_{t \in [0,T]} \left| M_t^F \right | \geq \varepsilon \right )  \leq 2\exp \left (- \frac{N  \varepsilon^2}{4 K^2 T} \right ).
\end{equation*}
\end{lem}

\begin{proof}
According to Lemma~\ref{lem: Ito}, \(M_t^F\) is a local martingale and its quadratic variation \(\left [M_t^F \right ]\) satisfies 
\[
\left [M_t^F \right ] \leq 2 t \norm{\nabla F}^2_{L^\infty} \leq \frac{2t}{N} \norm{F}_{\mathcal{G}^1}^2 \leq \frac{2K^2 t}{N} ,
\]
where the second inequality follows by definition of the \(\mathcal{G}^1\)-norm. For every \(\lambda \geq 0\), \(Z^\lambda_t \coloneqq \exp(\lambda M_t^F - \lambda^2 [M_t^F]/2)\) is a positive super-martingale such that \(Z^\lambda_0 = 1\) and \(\sup_{\boldsymbol{X}_0} \E_{\mathbb{Q}_{\boldsymbol{X}_0}} \left [ Z^\lambda_t\right ] \leq 1\) for all \(t \ge 0\). Therefore, for every \(t, \lambda, \varepsilon \geq 0\), we have 
\[
\begin{split}
\sup_{\boldsymbol{X}_0} \mathbb{Q}_{\boldsymbol{X}_0} \left ( \sup_{t \in [0,T]} M_t^F \geq \varepsilon \right ) & \leq  \sup_{\boldsymbol{X}_0} \mathbb{Q}_{\boldsymbol{X}_0} \left ( \sup_{t \in [0,T]} Z_t^\lambda \geq e^{\lambda \varepsilon - \frac{\lambda^2}{2}\frac{2K^2T}{N}} \right ) \\
& \leq e^{\lambda^2\frac{ K^2T}{N}- \lambda \varepsilon}  \sup_{\boldsymbol{X}_0} \E_{\mathbb{Q}_{\boldsymbol{X}_0}} \left [ Z^\lambda_0 \right ]  = e^{\lambda^2 \frac{K^2T}{N}- \lambda \varepsilon} ,
\end{split}
\]
where we used the maximal inequality for the super-martingale \(Z^\lambda\) for the last inequality. Optimizing over \(\lambda\ge0\) by choosing \(\lambda = \varepsilon N/(2K^2 T)\) gives
\[
\sup_{\boldsymbol{X}_0} \mathbb{Q}_{\boldsymbol{X}_0} \left ( \sup_{t \in [0,T]} M_t^F \geq \varepsilon \right )  \leq \exp \left (-\frac{N  \varepsilon^2}{4 K^2 T} \right ).
\]
Applying the same argument to \(-M_t^F\), we obtain
\[
\sup_{\boldsymbol{X}_0} \mathbb{Q}_{\boldsymbol{X}_0} \left ( \sup_{t \in [0,T]} (-M_t^F) \geq \varepsilon \right )  \leq \exp \left (-\frac{N  \varepsilon^2}{4 K^2 T} \right ),
\]
yielding the desired result by union bound.
\end{proof}

In particular, since \(\norm{m_{ij}}_{\mathcal{G}^n} \leq c(n)\), and Lemma~\ref{lem: Doob inequality} yields that there exists a universal constant \(K>0\) such that, for every \(\gamma,T>0\), and every \(N\),
\begin{equation} \label{eq: Doob inequality}
\sup_{\boldsymbol{X}} \mathbb{Q}_{\boldsymbol{X}} \left ( \sup_{t \in [0,T]} \left| M_t^{m_{ij}} \right | \geq \frac{\gamma}{\sqrt{N}} \right )  \leq 2 \exp \left (- \frac{\gamma^2}{K T} \right ).
\end{equation}

In addition to the evolution of the correlations \(m_{ij}\) under Langevin dynamics, we are also interested in the evolution of the matrix-valued function \(\boldsymbol{G} \colon \mathcal{S}_{N,r} \to \R^{r \times r}\) defined by
\begin{equation} \label{eq: matrix G}
\boldsymbol{G}(\boldsymbol{X}) = \boldsymbol{M}^\top\boldsymbol{M} = \frac{1}{N^2} \boldsymbol{X}^\top \boldsymbol{V} \boldsymbol{V}^\top \boldsymbol{X},
\end{equation}
where each entry \(G_{ij}\) is a smooth function from \(\mathcal{S}_{N,r}\) to \(\R\) such that \(G_{ij}(\boldsymbol{X}) = \sum_{k=1}^r m_{ki}(\boldsymbol{X}) m_{kj}(\boldsymbol{X})\). Note that \(G_{ij}=G_{ji}\) for every \(i,j\in [r]\), that is, \(\boldsymbol{G}(\boldsymbol{X})\) is a symmetric matrix.

\begin{lem}[Evolution equation for \(G_{ij}\)] \label{lem: evolution equation for G_ij}
For every \(1 \le i,j \le r\), the process \(G_{ij}(t) \coloneqq G_{ij}(\boldsymbol{X}_t^\beta)\) evolves according to
\[
d G_{ij}(t) = L_\beta G_{ij}(t) dt + dM_t^{G_{ij}},
\]
where \(M_t^{G_{ij}} = \sqrt{2} \int_0^t \langle \nabla G_{ij}(s), d \boldsymbol{B}_s \rangle\) is the martingale part of the evolution, and the generator \(L_\beta G_{ij}(t)\) is given by \(L_\beta G_{ij}(t) = L_{0,\beta} G_{ij}(t) + \hat{L}_\beta G_{ij}(t) \) and is of the form
\begin{equation} \label{eq: noise generator G_ij}
L_{0,\beta} G_{ij} = \sum_{k=1}^r \left ( m_{ki} L_{0,\beta}m_{kj} + m_{kj} L_{0,\beta} m_{ki}\right) + \frac{2}{N} \left (r \delta_{ij} - G_{ij} \right),
\end{equation}
and 
\begin{equation} \label{eq: population generator G_ij}
\hat{L}_\beta G_{ij} = 4 \beta \sqrt{M} \lambda^2 (G_{ij} - (\boldsymbol{G}^2)_{ij}).
\end{equation}
Moreover, the quadratic variation of the martingale \(M_t^{G_{ij}}\) satisfies
\begin{equation} \label{eq: quadratic variation G_ij}
[M_t^{G_{ij}}] \leq  \frac{2r}{N} \int_0^t \norm{\boldsymbol{M}(s)}_{\textnormal{op}}^2 ds.
\end{equation}
\end{lem}

\begin{proof}
For every \(i,j \in [r]\), the summary statistic \(G_{ij} \colon \mathcal{S}_{N,r} \to \R\) is a smooth function, thus by Itô's Lemma~\ref{lem: Ito} we have
\[
G_{ij}(\boldsymbol{X}_t) = G_{ij}(\boldsymbol{X}_0) + M_t^{G_{ij}} + \int_0^t L_\beta G_{ij}(\boldsymbol{X}_s) ds,
\]
where \(M_t^{G_{ij}} = \sqrt{2} \int_0^t \langle \nabla G_{ij}(\boldsymbol{X}_s), d \boldsymbol{B}_s \rangle\) is a local martingale and \(L_\beta G_{ij}(\boldsymbol{X})\) is given by \(L_\beta G_{ij}(\boldsymbol{X}) = L_{0,\beta}G_{ij}(\boldsymbol{X}) + \beta \langle \nabla \Phi(\boldsymbol{X}), \hat{\nabla} G_{ij}(\boldsymbol{X}) \rangle\) with \(\Phi(\boldsymbol{X}) = H_0(\boldsymbol{X}) -H(\boldsymbol{X})\). By an explicit computation as done in the proof of Lemma~\ref{lem: evolution equation m_ij}, we then find that the population generator \(\hat{L}_\beta = \beta \langle \nabla \Phi(\boldsymbol{X}), \hat{\nabla}\cdot)\) satisfies
\[
\hat{L}_\beta G_{ij} = 4 \beta \sqrt{M} \lambda^2 (G_{ij} - (\boldsymbol{G}^2)_{ij}).
\]
Moreover, according to~\cite[Proposition 1.3.1]{hsu2008brief} the quadratic variation \([M_t^{G_{ij}}]\) of \(M_t^{G_{ij}}\) satisfies
\[
[M_t^{G_{ij}}] = \int_0^t \norm{\nabla G_{ij}(s)}_{\textnormal{F}}^2 ds \leq \int_0^t \norm{\hat{\nabla}G_{ij}(s)}_{\textnormal{F}}^2 ds ,
\]
where we recall that \(\nabla\) denotes the Euclidean gradient. Since \(\nabla_{\boldsymbol{x}_k} m_{ij}(\boldsymbol{X}) = \frac{\boldsymbol{v}_i}{N} \delta_{kj}\), we have 
\[
\norm{\hat{\nabla}G_{ij}}_{\textnormal{F}}^2 = \Tr((\hat{\nabla}G_{ij})^\top (\hat{\nabla}G_{ij})) = \frac{2}{N} \sum_{k=1}^r (m_{ki}^2 + m_{kj}^2) = \frac{2}{N} (G_{ii} + G_{jj}) \leq  \frac{2}{N} \Tr(\boldsymbol{G}).
\]
In particular, we find that 
\[
[M_t^{G_{ij}}] \leq \frac{2}{N} \int_0^t \Tr(\boldsymbol{G}(s)) ds = \frac{2}{N} \int_0^t \norm{\boldsymbol{M}(s)}_{\textnormal{F}}^2 ds\leq  \frac{2r}{N} \int_0^t \norm{\boldsymbol{M}(s)}_{\textnormal{op}}^2 ds,
\]
as desired. 
\end{proof}

We readily obtain the following corollary.

\begin{cor}[Evolution equation for \(\boldsymbol{G}\)] \label{cor: evolution equation for G}
The evolution equation for \(\boldsymbol{G}\) is given by
\[
\boldsymbol{G}(t) = \boldsymbol{G}(0) + M_t^{\boldsymbol{G}} + \int_0^t \boldsymbol{L}_\beta \boldsymbol{G}(s) ds,
\]
where \(M_t^{\boldsymbol{G}} = (M_t^{G_{ij}})_{ij} \in \R^{r \times r}\) and \(\boldsymbol{L}_\beta \boldsymbol{G}(t) =(L_\beta G_{ij}(t))_{ij} \in \R^{r \times r}\). In particular, \(\boldsymbol{L}_\beta \boldsymbol{G}\) is of the form \(\boldsymbol{L}_\beta \boldsymbol{G} = \boldsymbol{L}_{0,\beta} \boldsymbol{G} + \hat{\boldsymbol{L}}_\beta \boldsymbol{G}\), where 
\begin{equation}\label{eq: noise generator G}
\boldsymbol{L}_{0,\beta} \boldsymbol{G} = \left (\boldsymbol{L}_{0,\beta} \boldsymbol{M}\right)^\top \boldsymbol{M} + \boldsymbol{M}^\top \left (\boldsymbol{L}_{0,\beta} \boldsymbol{M}\right) + \frac{2}{N} \left (r \boldsymbol{I}_r - \boldsymbol{G} \right),
\end{equation}
and
\begin{equation}\label{eq: population generator G}
\hat{\boldsymbol{L}}_\beta \boldsymbol{G} = 4 \beta \sqrt{M} \lambda^2 (\boldsymbol{G} - \boldsymbol{G}^2).
\end{equation}
\end{cor}

Similarly to the sub-Gaussian tail bound for the martingale part of the evolution of the correlations \(m_{ij}\) given by~\eqref{eq: Doob inequality} obtained by the Doob's maximal inequality (see Lemma~\ref{lem: Doob inequality}), we have the following result.

\begin{lem} \label{lem: Doob max inequality for operator norm M}
For every \(\gamma >0\) and \(T>0\), suppose there exists a positive, increasing, deterministic function \(f \colon [0,T] \to \R_+\) such that for every \(\boldsymbol{X}_0 \in \mathcal{S}_{N,r}\), the process \((\boldsymbol{M}(t))_{t\in[0,T]}\) satisfies
\[
\int_0^t \norm{\boldsymbol{M}(s)}^2_{\textnormal{op}} ds \le f(t),
\]
for every \(t \in [0,T]\), \(\mathbb{Q}_{\boldsymbol{X}_0}\)-a.s. Then,
\[
\sup_{\boldsymbol{X}_0 \in \mathcal{S}_{N,r}} \mathbb{Q}_{\boldsymbol{X}_0} \left (\sup_{t \in [0,T]} \norm{M_t^{\boldsymbol{G}}}_{\textnormal{op}}  \geq \frac{\gamma}{\sqrt{N}}f(T)^{1/2} \right ) \leq r(r+1) e^{-\gamma^2/(4r^3)}.
\]
\end{lem}

\begin{proof}
We first note that \(\norm{M_t^{\boldsymbol{G}}}_{\textnormal{op}} \leq r \max_{1 \le i,j \le r} |M_t^{G_{ij}}|\), where \(M_t^{G_{ij}}\) is a local martingale given by Lemma~\ref{lem: evolution equation for G_ij} such that its quadratic variation satisfies~\eqref{eq: quadratic variation G_ij}, i.e., for every \(i,j \in [r]\)
\[
[M_t^{G_{ij}}] \leq \frac{2r}{N} \int_0^t \norm{\boldsymbol{M}(s)}_{\textnormal{op}}^2 ds \leq \frac{2r}{N} f(t).
\]
For every \(\lambda \ge 0\), we now introduce the exponential martingale \(Z^\lambda_t = \exp(\lambda M_t^{G_{ij}} - \lambda^2 [M_t^{G_{ij}}]/2)\). The same reasoning used in the proof of Lemma~\ref{lem: Doob inequality} gives that
\[
\sup_{\boldsymbol{X}_0 \in \mathcal{S}_{N,r}}  \mathbb{Q}_{\boldsymbol{X}_0} \left ( \sup_{t \in [0,T]} | M_t^{G_{ij}} | \geq \varepsilon \right )  \leq 2 \exp \left (- \frac{N \varepsilon^2}{4rf(T)} \right ),
\]
for every \(i,j \in [r]\). From this, we then have that 
\[
\begin{split}
\sup_{\boldsymbol{X}_0 \in \mathcal{S}_{N,r}} \mathbb{Q}_{\boldsymbol{X}_0} \left ( \sup_{t \in [0,T]} \norm{M_t^{\boldsymbol{G}}}_{\textnormal{op}} \geq \varepsilon \right) & \leq \sup_{\boldsymbol{X}_0 \in \mathcal{S}_{N,r}}  \mathbb{Q}_{\boldsymbol{X}_0} \left ( \sup_{t \in [0,T]} \max_{1 \leq i \le j \leq r} | M_t^{G_{ij}} | \geq \frac{\varepsilon}{r} \right)\\
& \leq \frac{r(r+1)}{2} \sup_{\boldsymbol{X}_0 \in \mathcal{S}_{N,r}}  \mathbb{Q}_{\boldsymbol{X}_0} \left ( \sup_{t \in [0,T]} | M_t^{G_{11}} | \geq \frac{\varepsilon}{r} \right) \\
& \leq r(r+1) \exp \left (- \frac{N \varepsilon^2}{4r^3 f(T)} \right ).
\end{split}
\]
Choosing \(\varepsilon = \frac{\gamma}{\sqrt{N}}  f(T)^{1/2}\) gives the desired result, since \(f\) is an increasing function by assumption.
\end{proof}

\subsection{Comparison inequalities}

We first report Lemma 5.1 of~\cite{arous2020algorithmic} that provides simple comparison inequalities for functions. 

\begin{lem}[Bounds on functions] \label{lem: Gronwall}
Let \(\gamma >0\) with \(\gamma \neq 1\), \(c >0\), and \(f \in C_{\text{loc}}([0,T))\) with \(f(0)>0\). 
\begin{itemize}
\item[(a)] Suppose that there exists \(T\) such that \(f\) satisfies the integral inequality
\begin{equation} \label{eq: gronwall}
f(t) \geq a + \int_0^t c f^\gamma(s) ds,
\end{equation}
for every \(t \leq T\) and some \(a >0\). Then, for \(t \geq 0\) satisfying \((\gamma-1)ca^{\gamma-1}t < 1\), we have
\[
f(t) \geq a \left ( 1 - (\gamma-1) c a^{\gamma-1} t\right )^{-\frac{1}{\gamma-1}}.
\]
\item[(b)] If the integral inequality~\eqref{eq: gronwall} holds in reverse, i.e., if \(f(t) \leq a + \int_0^t c f^\gamma(s) ds\), then the corresponding upper bound holds.
\item[(c)] If \(\gamma >1\), then \(T \leq t_{\ast}\), where \(t_{\ast} = \left ( (\gamma-1)ca^{\gamma-1} \right )^{-1}\) is called the blow-up time.
\item[(d)] If~\eqref{eq: gronwall} holds with \(\gamma = 1\), then the Grönwall's inequality gives \(f(t) \geq a\exp(ct)\).
\end{itemize}
\end{lem}

Lemma~\ref{lem: Gronwall} will be useful in providing bounds on the correlation functions \(m_{ij}^{(N)}\) for both \(p \ge 3\) and \(p=2\). For subspace recovery when \(p=2\), the matrix-valued function \(\boldsymbol{G} \colon [0,\infty) \to \R^{r\times r}\) defined by~\eqref{eq: matrix G} is not differentiable and we need an integral comparison inequality, as stated in the following lemma. By definition, \(\boldsymbol{G}\) takes values in the cone of symmetric positive semidefinite matrices. Denote by \(\theta_1(t), \ldots, \theta_r(t)\) the eigenvalues of \(\boldsymbol{G}(t)\), and set
\[
\theta_{\min}(t) \coloneqq \min_{1 \le i \le r} \theta_i(t), 
\qquad 
\theta_{\max}(t) \coloneqq \max_{1 \le i \le r} \theta_i(t).
\]

\begin{lem}  \label{lem: integral inequality matrix case}
Fix \(a >0\) and define \(\boldsymbol{A} \colon [0,\infty) \to \R^{r \times r}\) as the unique \(C^1\) solution of
\[
\begin{cases}
\frac{d}{dt} \boldsymbol{A}(t) = a \boldsymbol{G}(t) (\boldsymbol{I}_r - \boldsymbol{G}(t)),\\
\boldsymbol{A}(0) = \boldsymbol{G}(0).
\end{cases},
\]
Denote by \(\mu_1(t),\dots,\mu_r(t)\) the eigenvalues of \(\boldsymbol{A}(t)\). Then, for every \(i \in [r]\) and every \(t \geq 0\), 
\begin{equation} \label{eq: integral inequality eigenvalues 1}
\theta_i (0) + a \int_0^t ( \theta_{\min}(s) - \theta_{\max}^2(s)) ds \leq \mu_i \left(t\right) \leq \theta_i (0) + a \int_0^t (\theta_{\max}(s) - \theta_{\min}^2(s)) ds.
\end{equation}
Moreover, if for some \(t \geq 0\), \(\theta_{\max}(s) \le \frac{1}{2}\) for all \(s \in [0,t]\), then for every \(i \in [r]\),
\begin{equation} \label{eq: integral inequality eigenvalues 2}
\theta_i (0) + a \int_0^t ( \theta_{\min}(s) - \theta_{\min}^2(s) ) ds \leq \mu_i \left(\boldsymbol{A}(t)\right) \leq \theta_i (0) + a \int_0^t (\theta_{\max}(s) - \theta_{\max}^2(s) )ds.
\end{equation}
\end{lem}

\begin{proof}
Since \(\boldsymbol{G}(t)\) is symmetric for every \(t\) and \(\boldsymbol{A}(0) = \boldsymbol{G}(0)\) is symmetric, the ODE implies that \(\boldsymbol{A}(t)\) is symmetric for every \(t\) and admits the integral representation
\[
\boldsymbol{A}(t) = \boldsymbol{G}(0) + a \int_0^t \boldsymbol{G}(s)\left(\boldsymbol{I}_r - \boldsymbol{G}(s)\right) ds.
\]
Fix \(s\ge 0\). Let \(\theta_1(s),\ldots,\theta_r(s)\) be the eigenvalues of \(\boldsymbol{G}(s)\) and let \(\boldsymbol{u}_1(s),\dots,\boldsymbol{u}_r(s)\) be an orthonormal eigenbasis. Then
\[
\boldsymbol{G}(s) = \sum_{i=1}^r \theta_i(s) \boldsymbol{u}_i(s)\boldsymbol{u}_i(s)^\top,
\qquad
\boldsymbol{G}(s)^2 = \sum_{i=1}^r \theta_i(s)^2 \boldsymbol{u}_i(s)\boldsymbol{u}_i(s)^\top,
\]
so that
\[
\boldsymbol{G}(s)\left(\boldsymbol{I}_r - \boldsymbol{G}(s)\right) = \sum_{i=1}^r \left (\theta_i(s) - \theta_i(s)^2\right) \boldsymbol{u}_i(s)\boldsymbol{u}_i(s)^\top.
\]
Moreover, we have \(\theta_{\min}(s) \le \theta_i(s) \le \theta_{\max}(s)\) for every \(i \in [r]\).

Therefore, for every \(i \in [r]\), 
\[
\theta_i (s) - \theta_i^2 (s) \ge \theta_{\min}(s)( 1-\theta_{\max}(s))  \ge \theta_{\min}(s) - \theta_{\max}(s)^2,
\]
and similarly
\[
\theta_i (s) - \theta_i^2(s) \le \theta_{\max}(s)(1-\theta_{\min}(s))  \le \theta_{\max}(s) - \theta_{\min}(s)^2.
\]
Combining both inequalities, for every \(i \in [r]\), we have
\begin{equation} \label{eq: eig comparison}
\theta_{\min}(s) - \theta_{\max}(s)^2 \le \theta_i(s) - \theta_i^2(s) \le \theta_{\max}(s) - \theta_{\min}(s)^2.
\end{equation}
This implies that
\[
\left (\theta_{\min}(s) - \theta_{\max}(s)^2\right)\boldsymbol{I}_r \preceq \boldsymbol{G}(s) - \boldsymbol{G}(s)^2 \preceq \left(\theta_{\max}(s) - \theta_{\min}(s)^2\right)\boldsymbol{I}_r,
\]
where for any symmetric matrices \(\boldsymbol{A}\) and \(\boldsymbol{B}\), we write \(\boldsymbol{A} \preceq \boldsymbol{B}\) if \(\boldsymbol{B} - \boldsymbol{A}\) is positive semi-definite. Integrating over \(s\in[0,t]\) and multiplying by \(a>0\) yields
\[
a \int_0^t \left(\theta_{\min}(s) - \theta_{\max}(s)^2\right)  ds \boldsymbol{I}_r \preceq a \int_0^t \left(\boldsymbol{G}(s) - \boldsymbol{G}(s)^2\right) ds \preceq
a \int_0^t \left(\theta_{\max}(s) - \theta_{\min}(s)^2\right) ds \boldsymbol{I}_r.
\]
Adding \(\boldsymbol{G}(0)\) to all three sides gives
\[
\boldsymbol{G}(0) + a \int_0^t \left(\theta_{\min}(s) - \theta_{\max}(s)^2\right) ds \boldsymbol{I}_r \preceq \boldsymbol{A}(t) \preceq \boldsymbol{G}(0) + a \int_0^t \left(\theta_{\max}(s) - \theta_{\min}(s)^2\right) ds \boldsymbol{I}_r.
\]
If \(\boldsymbol{A}\preceq \boldsymbol{B}\) for symmetric matrices, then \(\lambda_i(\boldsymbol{A})\le \lambda_i(\boldsymbol{B})\) for every \(i\), by the Courant-Fischer principle. Since the eigenvalues of \(\boldsymbol{G}(0) + c\boldsymbol{I}_r\) are \(\theta_i(0) + c\), applying this eigenvalue monotonicity gives
\[
\theta_i(0) + a \int_0^t \left(\theta_{\min}(s) - \theta_{\max}(s)^2\right) ds \le \mu_i\left(t\right) \le \theta_i(0) + a \int_0^t \left(\theta_{\max}(s) - \theta_{\min}(s)^2\right) ds,
\]
which is~\eqref{eq: integral inequality eigenvalues 1}. We now refine the bounds when \(\theta_{\max}(s)\le 1/2\). Assume that \(\theta_{\max}(s) \le 1/2\) for all \(s\in[0,t]\). On \([0,1/2]\), the function \(x\mapsto x-x^2\) is increasing, so for each \(s\in[0,t]\) and all \(i\),~\eqref{eq: eig comparison} becomes
\[
\theta_{\min}(s) - \theta_{\min}(s)^2 \le \theta_i(s) - \theta_i(s)^2 \le \theta_{\max}(s) - \theta_{\max}(s)^2.
\]
Repeating the argument above, we obtain the second inequality~\eqref{eq: integral inequality eigenvalues 2}. This completes the proof.
\end{proof}

\section{Proofs for \(p \geq 3\)} \label{section: proof recovery Langevin p>2}

This section is devoted to the proof of Propositions~\ref{thm: strong recovery first spike Langevin p>2} and~\ref{thm: strong recovery all spikes Langevin p>2}. To simplify notation, we write the correlation functions as \(m_{ij} (\boldsymbol{X})\) instead of \(m_{ij}^{(N)}(\boldsymbol{X})\), and define the time-dependent quantities \(m_{ij}(t) \coloneqq m_{ij} (\boldsymbol{X}_t^\beta)\). Throughout this section, we assume that the initial correlations with the signal are positive, i.e., \(m_{ij}(\boldsymbol{X}_0)>0\) for every \(i,j \in [r]\). Equivalently, for every \(N\) the initialization measure \(\mu_N^0\) is supported on
\[
\mathcal{X}^+ \coloneqq \{\boldsymbol{X} \in \mathcal{S}_{N,r} \colon m_{ij}^{(N)}(\boldsymbol{X}) > 0 \ \forall i,j\in[r]\}.
\]

\subsection{Recovery of leading spike} 

We begin by proving Proposition~\ref{thm: strong recovery first spike Langevin p>2}. As a first step, we derive sample-complexity bounds and identify a sufficient separation between the top two SNRs that guarantees weak recovery of the leading spike. To that end, we define the weak-recovery event
\[
E \coloneqq \{\boldsymbol{X} \colon m_{11} (\boldsymbol{X}) > c_1 N^{-\frac{p-2}{2(p-1)}} \enspace \mathrm{and} \enspace m_{ij}(\boldsymbol{X})  \le c_2 N^{-1/2} \enspace \forall \, (i,j) \neq (1,1)\},
\]
for some constants \(c_1,c_2>0\). The event \(E\) corresponds to microscopic but nontrivial correlation with the first spike, while all other correlations remain of order \(N^{-1/2}\).

\begin{lem} \label{lem: weak recovery first spike Langevin p>2}
Let \(\beta \in (0,\infty)\) and \(p \geq 3\). Consider a sequence of initialization measures \(\mu_N^0 \in \mathcal{M}_1(\mathcal{S}_{N,r})\) supported on \(\mathcal{X}^+\). Fix parameters \(n \geq 1, \gamma_0> 0, \gamma_1 > 1 > \gamma_2 > 0\), and choose \(0 < C_0 < \frac{1}{2} -\frac{1}{2} \left ( \frac{\gamma_2}{3 \gamma_1} \right)^{p-2}\). Then, there exist constants \(C,K, K_1, K_2 >0\) (depending only on the fixed parameters) such that, if 
\[
\lambda_1 >\frac{1+C_0}{1-C_0} \left(\frac{3 \gamma_1}{\gamma_2}\right)^{p-2} \lambda_2 \quad \mathrm{and} \quad \sqrt{M} \ge C \frac{(n+2) \gamma_0}{\beta p \lambda_r^2 C_0 \gamma_2^{p-1}} N^{\frac{p-1}{2}-\frac{n}{2(n+1)}}, 
\]
then for \(N\) sufficiently large,
\[
\begin{split}
& \int_{\mathcal{X}^+} \mathbb{Q}_{\boldsymbol{X}} \left ( \mathcal{T}_E \gtrsim \frac{\gamma_2}{{(n+2)\gamma_0 N^{\frac{1}{2(n+1)}}}}  \right )  \mathbf{1}_{\{\mathcal{C}_{0,\beta}^{(N)} (n,\gamma_0) \cap \mathcal{C}_1^{(N)} (\gamma_1,\gamma_2)\}} (\boldsymbol{X}) d \mu_N^0 (\boldsymbol{X}) \\
& \leq K_1 \exp \left (- \frac{\gamma_0^3 (n+2)}{\gamma_2 K_1} N^{\frac{1}{2(n+1)}}\right)  + K_2 r^2 \exp \left (- \frac{\gamma_2 \gamma_0 (n+2)}{K_2} N^{\frac{1}{2(n+1)} } \right),
\end{split}
\]
with \(\mathbb{P}\)-probability at least \(1 - \exp(-K N)\). Here, \(\gtrsim\) denotes a constant depending only on \(p\).
\end{lem}

The strategy for proving weak recovery of the first spike is as follows. The separation assumption on \(\lambda_1\) ensures that the correlation \(m_{11}\) grows faster than all other correlations. As a consequence, \(m_{11}\) is the first correlation to reach the microscopic threshold \(\tilde{\varepsilon}_N = \tilde{\gamma} N^{-\frac{1}{2}}\), for some sufficiently large constant \(\tilde{\gamma}\) of order \(1\). Thus, even if \(m_{11}(0) < \max_{i,j} m_{ij}(0)\), the strength of \(\lambda_1\) ensures that \(m_{11}\) exceeds \(\tilde{\varepsilon}_N\) before any other correlation. We further show that \(m_{11}\) continues to grow and is also the first correlation to exceed the critical threshold \(c_1N^{-\frac{p-2}{2(p-1)}}\), whereas the remaining correlations stay of order \(N^{-1/2}\).

We next show that weak recovery implies strong recovery.

\begin{lem} \label{lem: weak implies strong recovery Langevin p>2}
Let \(\beta \in (0,\infty)\) and \(p \geq 3\). For every \(\varepsilon > 0\), there exist constants \(C = C(p,\beta, \{\lambda_i\}_{i=1}^r)>0\) and \(K, K_1,K_2>0\) such that, if \(\sqrt{M} \geq \frac{C}{\beta p \lambda_r^2} N^{\frac{p-2}{2}}\), then there exists \(T_1 \le c N^{-\frac{p-2}{2(p-1)}}\) for a sufficiently large constant $c=c(p,\beta,\{\lambda_i\})$, such that for all \(T \geq T_1\) and all \(N\) sufficiently large,
\[
\inf_{\boldsymbol{X} \in E} \mathbb{Q}_{\boldsymbol{X}} \left (\inf_{t \in [T_1,T]} m_{11}(\boldsymbol{X}_t^\beta) \geq 1-\varepsilon \right ) \geq 1 - K_1 \exp \left ( - \frac{ N^{\frac{p-2}{2(p-1)}}}{K_1} \right ) - K_2  \exp \left ( - \frac{ \varepsilon^2 N}{K_2T} \right ),
\]
with \(\mathbb{P}\)-probability at least \(1 - \exp(-K N)\).
\end{lem}

Having Lemmas~\ref{lem: weak recovery first spike Langevin p>2} and~\ref{lem: weak implies strong recovery Langevin p>2} at hand, Proposition~\ref{thm: strong recovery first spike Langevin p>2} follows by the strong Markov property. 

\begin{proof}[\textbf{Proof of Proposition~\ref{thm: strong recovery first spike Langevin p>2}}]
Set $T_0 \coloneqq k_1 \frac{\gamma_2}{(n+2)\gamma_0}N^{-\frac{1}{2(n+1)}}$ and let $T_1 \le k_2 N^{-\frac{p-2}{2(p-1)}}$ be as in Lemma~\ref{lem: weak implies strong recovery Langevin p>2}. Integrating over the initial condition \(\boldsymbol{X} \sim \mu_N^0\) and conditioning on the event \(\{\mathcal{T}_E \le T_0\}\) gives for all \(T \ge T_0+T_1\),
\[
\begin{split}
& \int_{\mathcal{S}_{N,r}} \mathbb{Q}_{\boldsymbol{X}} \left (\inf_{t \in [T_0+T_1,T]} m_{11}(\boldsymbol{X}_t^\beta) \geq 1 - \varepsilon \right ) d \mu_N^0(\boldsymbol{X}) \\
&\geq \int_{\mathcal{S}_{N,r}}  \mathbb{Q}_{\boldsymbol{X}} \left (\inf_{t \in [T_0+T_1,T]} m_{11}(\boldsymbol{X}_t^\beta) \geq 1-\varepsilon  \, \Big | \, \mathcal{T}_E \le T_0 \right )  \mathbb{Q}_{\boldsymbol{X}} \left (\mathcal{T}_E \leq  T_0  \right )d \mu_N^0(\boldsymbol{X}).
\end{split}
\]
On the event \(\{\mathcal T_E \le T_0\}\), we have \(\boldsymbol{X}_{\mathcal{T}_E}^\beta\in E\) by definition of \(\mathcal{T}_E\). Moreover, on \(\{\mathcal T_E \le T_0\}\) one has \([T_0+T_1,T]\subseteq[\mathcal T_E+T_1,T]\), and therefore
\[
\mathbb{Q}_{\boldsymbol{X}} \left (\inf_{t \in [T_0+T_1,T]} m_{11}(\boldsymbol{X}_t^\beta) \geq 1-\varepsilon  \, \Big | \, \mathcal{T}_E \le T_0 \right )
\ge
\mathbb{Q}_{\boldsymbol{X}} \left (\inf_{t \in [\mathcal T_E+T_1,T]} m_{11}(\boldsymbol{X}_t^\beta) \geq 1-\varepsilon  \, \Big | \, \mathcal{T}_E \le T_0 \right ).
\]
Applying the strong Markov property at \(\mathcal T_E\) yields
\[
\begin{split}
\mathbb{Q}_{\boldsymbol{X}} \left (\inf_{t \in [\mathcal{T}_E+T_1,T]} m_{11}(\boldsymbol{X}_t^\beta) \geq 1-\varepsilon  \, \Big | \, \mathcal{T}_E \le T_0 \right )  \ge \inf_{\boldsymbol{X} \in E} \mathbb{Q}_{\boldsymbol{X}} \left (\inf_{t \in [T_1,T]} m_{11}(\boldsymbol{X}_t^\beta) \geq 1-\varepsilon \right ),
\end{split}
\]
which implies that
\[
\begin{split}
& \int_{\mathcal{S}_{N,r}} \mathbb{Q}_{\boldsymbol{X}} \left (\inf_{t \in [T_0+T_1,T]} m_{11}(\boldsymbol{X}_t^\beta) \geq 1 - \varepsilon \right ) d \mu_N^0(\boldsymbol{X}) \\
&\geq \left [\inf_{\boldsymbol{X} \in E} \mathbb{Q}_{\boldsymbol{X}} \left (\inf_{t \in [T_1,T]} m_{11}(\boldsymbol{X}_t^\beta) \geq 1-\varepsilon \right )  \right] \int_{\mathcal{S}_{N,r}} \mathbb{Q}_{\boldsymbol{X}} \left (\mathcal{T}_E \leq T_0 \right )d \mu_N^0(\boldsymbol{X}).
\end{split}
\]
We estimate the first term according to Lemma~\ref{lem: weak implies strong recovery Langevin p>2}. For the integral we use Lemma~\ref{lem: weak recovery first spike Langevin p>2} as follows:
\[
\begin{split}
\int_{\mathcal{S}_{N,r}} \mathbb{Q}_{\boldsymbol{X}} \left (\mathcal{T}_E > T_0  \right )d \mu_N^0(\boldsymbol{X}) & \leq \mu_N^0 (\mathcal{C}_{0,\beta}^{(N)}(n,\gamma_0)^\textnormal{c}) + \mu_N^0 ( \mathcal{C}_1 (\gamma_1,\gamma_2)^\textnormal{c} ) \\
& \quad + \int_{\mathcal{S}_{N,r}} \mathbb{Q}_{\boldsymbol{X}} \left (\mathcal{T}_E > T_0 \right ) \boldsymbol{1}_{\{\mathcal{C}_{0,\beta}^{(N)}(n,\gamma_0) \cap \mathcal{C}_1^{(N)}(\gamma_1,\gamma_2)\}} d \mu_N^0(\boldsymbol{X}).
\end{split}
\]
Combining Lemma~\ref{lem: weak recovery first spike Langevin p>2} with Definitions~\ref{def: condition 0} and~\ref{def: condition 1} gives the estimate for the integral. This establishes strong recovery of the first spike for the Langevin dynamics when \(p \ge 3\).
\end{proof}

It remains to prove the two intermediate results, namely Lemmas~\ref{lem: weak recovery first spike Langevin p>2} and~\ref{lem: weak implies strong recovery Langevin p>2}. We begin with the proof of Lemma~\ref{lem: weak recovery first spike Langevin p>2}. In what follows, for every \(a > 0\) and \(i,j \in [r]\), we denote by \(\mathcal{T}_a^{(ij)}\) the hitting time
\[
\mathcal{T}_a^{(ij)} \coloneqq \inf \{t \ge 0 \colon m_{ij} (t) \ge a \}. 
\]

\begin{proof}[\textbf{Proof of Lemma~\ref{lem: weak recovery first spike Langevin p>2}}]
Let \(\mathcal{A} = \mathcal{A}(n, \gamma_0, \gamma_1, \gamma_2)\) denote the event 
\[
\mathcal{A}(n,\gamma_0,\gamma_1, \gamma_2) = \left \{\boldsymbol{X}_0 \sim \mu_N^0 \colon \boldsymbol{X}_0 \in \mathcal{C}_{0,\beta}^{(N)}(n,\gamma_0) \cap \mathcal{C}_1^{(N)}(\gamma_1,\gamma_2) \right\},
\]
where the events \(\mathcal{C}_{0,\beta}^{(N)}(n,\gamma_0)\) and \(\mathcal{C}_1^{(N)}(\gamma_1,\gamma_2)\) are given by Definitions~\ref{def: condition 0} and~\ref{def: condition 1}, respectively. On the event \(\mathcal{C}_1^{(N)} (\gamma_1,\gamma_2)\), for every \(i,j \in [r]\) there exists \(\gamma_{ij} \in (\gamma_2,\gamma_1)\) such that 
\[
m_{ij}(0) = \gamma_{ij} N^{-\frac{1}{2}}. 
\]
For some \(T_0^{(ij)} >0\) to be chosen later, we define the event \(\mathcal{A}^{(ij)}= \mathcal{A}^{(ij)}(n,\gamma_0, \gamma_1,\gamma_2, T_0^{(ij)})\) by
\[
\mathcal{A}^{(ij)}(n,\gamma_0, \gamma_1,\gamma_2, T_0^{(ij)})=  \mathcal{A}(n,\gamma_0,\gamma_1, \gamma_2) \cap \left \{\sup_{t \in [0,T_0^{(ij)}]} |M_t^{m_{ij}}| \leq  \frac{\gamma_2}{2\sqrt{N}}  \right \},
\]
where we recall that according to Lemma~\ref{lem: Doob inequality} and~\eqref{eq: Doob inequality}, there exists an absolute constant \(K_1 > 0\) such that 
\begin{equation*}
\sup_{\boldsymbol{X}} \mathbb{Q}_{\boldsymbol{X}} \left ( \sup_{t \in [0,T_0^{(ij)}]} |M_t^{m_{ij}}| \geq \frac{\gamma_2}{2\sqrt{N}} \right ) \leq K_1 \exp \left (-\frac{\gamma_2^2}{4 K_1 T_0^{(ij)}} \right).
\end{equation*}
Moreover, for every \(i,j \in [r]\), we let \(\mathcal{T}_{L_{0,\beta}}^{(ij)}\) denote the hitting time of the set
\[
\left \{\boldsymbol{X} \colon |L_{0,\beta} m_{ij}(\boldsymbol{X})| > C_0 \beta \sqrt{M} p\lambda_i \lambda_j m_{ij}^{p-1}(\boldsymbol{X}) \right \},
\]
where \(C_0 \in (0,\frac{1}{2})\) is a constant independent of \(N\). Note that on the event \(\mathcal{A}\)---and in particular on \(\mathcal{C}_{0,\beta}^{(N)}(n,\gamma_0)\)---we have
\[
|L_{0,\beta} m_{ij}(\boldsymbol{X}_0)| \leq \frac{\gamma_0}{ \sqrt{N}} \leq C_0 \beta \sqrt{M} p \lambda_i \lambda_j \left ( \frac{\gamma_2}{\sqrt{N}} \right )^{p-1}  \leq C_0 \beta \sqrt{M} p\lambda_i \lambda_j m_{ij}^{p-1}(0),
\]
provided that \(\sqrt{M} \geq \frac{\gamma_0}{C_0 \beta p \lambda_i \lambda_j \gamma_2^{p-1}}N^{\frac{p-2}{2}}\), which holds by assumption. Therefore, by continuity of the process \(\boldsymbol{X}_t^\beta\), \(\mathcal{T}_{L_{0,\beta}}^{(ij)} >0\) on the event \(\mathcal{A}\). We also define the hitting time \(\mathcal{T}_{L_{0,\beta}}\) of the set
\[
\left \{\boldsymbol{X} \colon \sup_{1 \leq k, \ell \leq r} \{|L_{0,\beta} m_{k \ell}(\boldsymbol{X})| \}> C_0 \beta \sqrt{M} p\lambda_1^2 m_{11}^{p-1}(\boldsymbol{X}) \right \}.
\]
It follows again by continuity of the process \(\boldsymbol{X}_t^\beta\) that \(\mathcal{T}_{L_{0,\beta}}>0\), and by construction we have \(\mathcal{T}_{L_{0,\beta}} \leq \mathcal{T}_{L_{0,\beta}}^{(11)}\).

We now fix \(i,j \in [r]\) and work on the event \(\mathcal{A}^{(ij)}\). We introduce the threshold \(\varepsilon_N \coloneqq c N^{-\frac{p-2}{2(p-1)}}\) for some constant \(c>0\), and consider the generator expression from Lemma~\ref{lem: evolution equation m_ij}:
\[
L_\beta m_{ij} = L_{0,\beta} m_{ij} + \beta \sqrt{M} p \lambda_i \lambda_j m_{ij}^{p-1} -  \beta \sqrt{M}  \frac{p}{2} \sum_{1 \leq k,\ell \leq r} \lambda_k m_{i \ell}m_{kj} m_{k \ell} (\lambda_j m_{kj}^{p-2} + \lambda_\ell m_{k \ell}^{p-2}).
\]
As a consequence, for every \(t \leq \mathcal{T}_{L_{0,\beta}}^{(ij)} \wedge \mathcal{T}_{L_{0,\beta}} \wedge \min_{1 \leq k, \ell \leq r }\mathcal{T}_{\varepsilon_N}^{(k \ell)}\), we obtain the comparison bounds
\[
(1-C_0) \beta \sqrt{M} p \lambda_i \lambda_j m_{ij}^{p-1}(t) \leq L_\beta m_{ij}(t) \leq (1+ C_0) \beta \sqrt{M} p \lambda_i \lambda_j m_{ij}^{p-1}(t),
\]
provided that
\[
N^{p-3} \ge \left ( \frac{r^2 \lambda_1^2 c^{p+1}}{\lambda_r^2 \gamma_2^{p-2}} \right)^{2(p-1)}.
\]
(When \(p=3\), it suffices to choose \(c\) sufficiently small, independently of \(N\).) Since the evolution equation for \(m_{ij}\) under Langevin dynamics satisfies 
\[
m_{ij}(t) = m_{ij}(0) + M_t^{m_{ij}} + \int_0^t L_\beta m_{ij}(s) ds, 
\]
we obtain the integral inequality
\begin{equation} \label{eq: integral inequality}
\frac{\gamma_{ij}}{2\sqrt{N}} +  (1-C_0) \beta \sqrt{M} p \lambda_i \lambda_j \int_0^t m_{ij}^{p-1}(s) ds
\leq m_{ij}(t) \leq \frac{3\gamma_{ij}}{2\sqrt{N}} +  (1+ C_0)  \beta \sqrt{M} p \lambda_i \lambda_j \int_0^t m_{ij}^{p-1}(s) ds,
\end{equation}
for every \(t \leq \mathcal{T}_{L_{0,\beta}}^{(ij)} \wedge \mathcal{T}_{L_{0,\beta}} \wedge \min_{1 \leq k, \ell \leq r }\mathcal{T}_{\varepsilon_N}^{(k \ell)} \wedge T_0^{(ij)}\). Applying items (a) and (b) of Lemma~\ref{lem: Gronwall}, we have the comparison inequality
\begin{equation} \label{eq: comparison inequality}
\ell_{ij}(t) \leq m_{ij}(t) \leq u_{ij}(t),
\end{equation}
for all \(t\) in the same interval, where the functions \(\ell_{ij}\) and \(u_{ij}\) are given by
\begin{equation} \label{eq: function ell}
\ell_{ij}(t) = \frac{\gamma_{ij}}{2\sqrt{N}} \left ( 1 -(1 - C_0)\beta \sqrt{M} p(p-2)  \lambda_i \lambda_j \left ( \frac{\gamma_{ij}}{2\sqrt{N}} \right )^{p-2} t \right )^{-\frac{1}{p-2}},
\end{equation}
and 
\begin{equation} \label{eq: function u}
u_{ij}(t) = \frac{3\gamma_{ij}}{2\sqrt{N}} \left (1 - (1 + C_0)\beta \sqrt{M} p(p-2)  \lambda_i \lambda_j \ \left ( \frac{3\gamma_{ij}}{2\sqrt{N}} \right )^{p-2} t \right )^{-\frac{1}{p-2}},
\end{equation}
respectively. We now define \(T_{\ell, \varepsilon_N}^{(ij)}\) as the time at which the lower bound \(\ell_{ij}(t)\) reaches the threshold \(\varepsilon_N\), i.e., 
\begin{equation} \label{eq: T lower varepsilon N}
T_{\ell, \varepsilon_N}^{(ij)} = \frac{1 -\left (\frac{\gamma_{ij}}{2} \right)^{p-2} N^{-\frac{p-2}{2(p-1)}}}{(1 - C_0)\beta \sqrt{M} p(p-2)  \lambda_i \lambda_j \left( \frac{\gamma_{ij}}{2\sqrt{N}} \right)^{p-2} }.
\end{equation}
Similarly, we define \(T_{u, \varepsilon_N}^{(ij)}\) by the condition \(u_{ij}(T_{u,\varepsilon_N}^{(ij)}) = \varepsilon_N\), i.e.,
\begin{equation} \label{eq: T upper varepsilon N}
T_{u,\varepsilon_N}^{(ij)} = \frac{1 - \left(\frac{3\gamma_{ij}}{2} \right)^{p-2}N^{-\frac{p-2}{2(p-1)}}}{(1 + C_0)\beta \sqrt{M} p(p-2)  \lambda_i \lambda_j \left( \frac{3\gamma_{ij}}{2\sqrt{N}}\right)^{p-2} }.
\end{equation}
On the event \(\mathcal{A}^{(ij)}\), we therefore have \(T_{u,\varepsilon_N}^{(ij)}  \leq \mathcal{T}_{\varepsilon_N}^{(ij)} \leq T_{\ell, \varepsilon_N}^{(ij)}\). We choose \(T_0^{(ij)} = T_{\ell, \varepsilon_N}^{(ij)} >0\). By assumption, we have 
\begin{equation} \label{eq: lambda_0}
\lambda_1 = \lambda_0 \frac{1+C_0}{1-C_0} \left ( \frac{3 \gamma_1}{\gamma_2} \right )^{p-2} \lambda_2,
\end{equation}
for some constant \(\lambda_0>1\). Then, for every \((i,j) \neq (1,1)\),
\[
\begin{split}
T_{\ell, \varepsilon_N}^{(11)} & = \frac{1 -\left (\frac{\gamma_{11}}{2} \right)^{p-2} N^{-\frac{p-2}{2(p-1)}}}{(1 - C_0)\beta \sqrt{M} p(p-2)  \lambda_1^2 \left( \frac{\gamma_{11}}{2\sqrt{N}} \right)^{p-2} }\\
& \le \frac{1 - \left(\frac{\gamma_2}{2} \right)^{p-2} N^{-\frac{p-2}{2(p-1)}}}{\lambda_0 (1 + C_0) \beta \sqrt{M}p(p-2) \lambda_i \lambda_j \left(\frac{\gamma_{ij}}{2\sqrt{N}} \frac{\gamma_{11}}{\gamma_{ij}}  \right)^{p-2}  \left(\frac{3\gamma_1}{\gamma_2} \right)^{p-2} } \\
& \le \frac{1 - \left(\frac{3\gamma_{ij}}{2} \right)^{p-2}N^{-\frac{p-2}{2(p-1)}}}{(1 + C_0)\beta \sqrt{M} p(p-2)  \lambda_i \lambda_j \left( \frac{3\gamma_{ij}}{2\sqrt{N}}\right)^{p-2} } = T_{u, \varepsilon_N}^{(ij)},
\end{split}
\]
provided \(N \ge \left (  \frac{\lambda_0 (3 \gamma_1)^{p-2} - \gamma_2^{p-2}}{2^{p-2} (\lambda_0-1)}\right)^{\frac{2(p-1)}{p-2}}\). Therefore, on the event \(\cap_{1 \le k, \ell \le r} \mathcal{A}^{(k \ell)}\), we obtain
\[
\mathcal{T}_{\varepsilon_N}^{(11)} = \min_{1 \le k, \ell \le r} \mathcal{T}_{\varepsilon_N}^{(k \ell)}, 
\]
as soon as we can show that 
\begin{equation} \label{goal1}
\mathcal{T}_{\varepsilon_N}^{(11)} \leq \min_{1 \le k, \ell \le r} \mathcal{T}_{L_{0, \beta}}^{(k \ell)} \wedge \mathcal{T}_{L_{0,\beta}}.
\end{equation}
Our goal is thus to prove~\eqref{goal1}.\\

To achieve this, we seek an estimate for \(L_{0,\beta} m_{ij}\) for every \(i,j \in [r]\). To this end, we apply Lemma~\ref{lem: bounding flows} to the function \(F_{ij}(\boldsymbol{X}) = L_{0,\beta} m_{ij} (\boldsymbol{X})\). We see that if we let \(\psi_{k \ell}(\boldsymbol{X}) = \langle \boldsymbol{v}_k, \boldsymbol{x}_\ell \rangle\), \(a_{k \ell}(\boldsymbol{X}) = \beta p \sqrt{M} \lambda_k \lambda_\ell m_{k \ell}^{p-1}(\boldsymbol{X})\) and \(U = H_0\), then condition (1) is satisfied with \(\mathbb{P}\)-probability at least \(1 - \exp(-K N)\) for every \(n \geq 1\) according to Lemma~\ref{lem: regularity H0}. The function \(F_{ij}\) is smooth and for every \(n \geq 1\) satisfies \(\norm{F_{ij}}_{\mathcal{G}^{2n}} \leq \Lambda\) with \(\mathbb{P}\)-probability at least \(1 - \exp(-K N)\) according to~\eqref{eq: bound norm L_0m_ij}, thus condition (2) is verified. Condition (3) follows by assumption on the initial data, i.e., the event \(\mathcal{C}_{0,\beta}^{(N)}(n,\gamma_0)\). It therefore remains to verify condition (4). Fix \(i,j \in [r]\). Using the lower bound from the integral inequality~\eqref{eq: integral inequality}, on the event \(\mathcal{A}^{(ij)}\), we have
\begin{equation} \label{eq: bound a_ij}
\int_0^t |a_{ij}(s)| ds \leq \frac{1}{1 - C_0} \left (m_{ij}(t)-\frac{\gamma_{ij}}{2\sqrt{N}}\right) \leq \frac{1}{1-C_0} m_{ij}(t),
\end{equation}
for every \(t \leq \mathcal{T}_{L_{0,\beta}}^{(ij)} \wedge \mathcal{T}_{L_{0,\beta}} \wedge \min_{1 \leq k, \ell \leq r} \mathcal{T}_{\varepsilon_N}^{(k \ell)} \wedge T_0^{(11)}\). We observe that at time \(t=0\), for every \(\xi > 0\) we have 
\[
\xi \beta \sqrt{M} p \lambda_i \lambda_j \left (\ell_{ij}(0)\right)^{p-1} = \xi \beta \sqrt{M} p \lambda_i \lambda_j \left ( \frac{\gamma_{ij}}{2\sqrt{N}} \right)^{p-1} \geq C \xi (n+2) \gamma_0 N^{-\frac{n}{2(n+1)}} \geq \ell_{ij}(0),
\]
where we used the assumption \(\sqrt{M} \geq C \frac{(n+2) \gamma_0}{\beta p \lambda_r^2 C_0 \gamma_2^{p-1}}  N^{\frac{p-1}{2} - \frac{n}{2(n+1)}}\). Now, from~\eqref{eq: comparison inequality} we know that \(m_{ij}\) is lower bounded by \(\ell_{ij}\) over the time interval of interest. Since \(\ell_{ij}(t)\) is increasing and satisfies the above inequality at \(t=0\), it follows that
\[
m_{ij}(t) \leq \xi \beta \sqrt{M} p \lambda_i \lambda_j m_{ij}^{p-1}(t),
\]
and therefore, combining with~\eqref{eq: bound a_ij}, we obtain
\[
\int_0^t |a_{ij}(s)| ds \leq \frac{1}{1- C_0} m_{ij}(t) \leq \frac{\xi}{1-C_0} \beta \sqrt{M} p \lambda_i \lambda_j m_{ij}^{p-1}(t) ,
\]
for every \(t \leq \mathcal{T}_{L_{0,\beta}}^{(ij)} \wedge \mathcal{T}_{L_{0,\beta}}  \wedge \min_{1 \leq k, \ell \leq r} \mathcal{T}_{\varepsilon_N}^{(k \ell)} \wedge T_0^{(11)}\). Choosing \(\xi = (1-C_0)/2\) yields condition (4) with \(\epsilon = 1/2\). Thus, by Lemma~\ref{lem: bounding flows}: there exists a constant \(K_1 > 0\) such that on the event \(\cap_{1 \le k,\ell \le r} \mathcal{A}^{(k \ell)}\),
\begin{equation} \label{eq: bounding flow langevin p>3}
|L_{0,\beta}m_{ij}(t)| \leq K_1 \left(\frac{\gamma_0}{\sqrt{N}} \sum_{k=0}^{n-1} t^k  + t^n + 2 \sum_{1 \leq k, \ell \leq r} \int_0^t |a_{k \ell}(s)| ds \right),
\end{equation}
for every \(t \leq \min_{1 \le k,\ell \le r} \mathcal{T}_{L_{0,\beta}}^{(k \ell)} \wedge \mathcal{T}_{L_{0,\beta}} \wedge \min_{1 \le k,\ell \le r} \mathcal{T}_{\varepsilon_N}^{(k \ell)} \wedge T_0^{(11)}\), with \(\mathbb{Q}_{\boldsymbol{X}}\)-probability at least \(1 - K_2 \exp \left (- \gamma_0^2 /(K_2 T_0^{(11)}) \right )\) and with \(\mathbb{P}\)-probability at least \(1 - \exp(-K N)\). 

Given the estimate~\eqref{eq: bounding flow langevin p>3}, in the remainder of the proof we show that \(m_{11}\) is the first correlation to reach \(\varepsilon_N\) and that \(\mathcal{T}_{\varepsilon_N}^{(11)} \leq \mathcal{T}_{L_{0,\beta}}\). To this end, we first introduce an intermediate threshold \(\tilde{\varepsilon}_N = \tilde{\gamma} N^{-\frac{1}{2}}\) for a sufficiently large constant \(\tilde{\gamma} > \gamma_1\) of order \(1\). Our first step is to show that 
\[
\mathcal{T}_{\tilde{\varepsilon}_N}^{(11)} \leq \min_{1 \leq k, \ell \leq r} \mathcal{T}_{L_{0,\beta}}^{(k \ell)} \wedge \mathcal{T}_{L_{0,\beta}}.
\]
In particular, we choose \(\tilde \gamma\) so that
\begin{equation} \label{eq: tilde gamma}
\lambda_0 \geq \frac{(2 \tilde{\gamma})^{p-2}}{(2 \tilde{\gamma})^{p-2} - (3 \gamma_1)^{p-2}}, \quad \textnormal{so that} \enspace \tilde{\gamma} \ge \frac{3}{2} \left( \frac{\lambda_0}{\lambda_0-1}\right)^{\frac{1}{p-2}} \gamma_1,
\end{equation} 
where \(\lambda_0\) is given in~\eqref{eq: lambda_0}. Since the integral inequality~\eqref{eq: integral inequality} and comparison inequality~\eqref{eq: comparison inequality} remain valid for \(t \leq \mathcal{T}_{L_{0,\beta}}^{(ij)} \wedge \mathcal{T}_{L_{0,\beta}} \wedge \min_{1 \leq k, \ell \leq r }\mathcal{T}_{\tilde{\varepsilon}_N}^{(k \ell)} \wedge T_0^{(ij)}\), we can define \(T_{\ell, \tilde{\varepsilon}_N}^{(ij)}\) and \(T_{u, \tilde{\varepsilon}_N}^{(ij)}\) similarly to~\eqref{eq: T lower varepsilon N} and~\eqref{eq: T upper varepsilon N}, and we see that for every \( (i,j) \neq (1,1)\),
\[
\begin{split}
T_{\ell, \tilde{\varepsilon}_N}^{(11)} & = \frac{1 -\left (\frac{\gamma_{11}}{2 \tilde{\gamma}} \right)^{p-2}}{(1 - C_0)\beta \sqrt{M} p(p-2)  \lambda_1^2 \left( \frac{\gamma_{11}}{2\sqrt{N}} \right)^{p-2} } \\
& \leq \frac{1}{\lambda_0 (1 + C_0)\beta \sqrt{M} p(p-2)  \lambda_i \lambda_j \left( \frac{3\gamma_{ij}}{2\sqrt{N}} \right)^{p-2} } \\
&\leq \frac{1 -\left (\frac{3\gamma_{ij}}{2\tilde{\gamma}} \right)^{p-2}}{(1+ C_0)\beta \sqrt{M} p(p-2)  \lambda_i \lambda_j \left( \frac{3\gamma_{ij}}{2\sqrt{N}} \right)^{p-2} }  = T_{u, \tilde{\varepsilon}_N}^{(ij)},
\end{split}
\]
where we used~\eqref{eq: lambda_0} and~\eqref{eq: tilde gamma}. To conclude this step, we will show that, on the event \(\cap_{1 \leq k, \ell \leq r} \mathcal{A}^{(k \ell)}\), 
\[
\sup_{1 \le i, j \le r} |L_{0,\beta} m_{ij} (t)| \le C_0 \beta \sqrt{M}  p \inf_{1 \le i, j \le r} \lambda_i \lambda_j m_{ij}^{p-1}(t).
\]
Since the estimate~\eqref{eq: bounding flow langevin p>3} holds for every \(t \leq \mathcal{T}_{L_{0,\beta}}^{(ij)} \wedge \mathcal{T}_{L_{0,\beta}} \wedge \min_{1 \leq k, \ell \leq r }\mathcal{T}_{\tilde{\varepsilon}_N}^{(k \ell)} \wedge T_0^{(ij)}\), a sufficient condition for this is to show that each term on the right-hand side of~\eqref{eq: bounding flow langevin p>3} is bounded above by \( \frac{C_0 \beta \sqrt{M} p}{n+2} \inf_{1 \le k, \ell \le r} \lambda_k \lambda_\ell m_{k \ell}^{p-1}(t)\) for all  \(t \leq \min_{1 \le k,\ell \le r} \mathcal{T}_{L_{0,\beta}}^{(k \ell)} \wedge \mathcal{T}_{L_{0,\beta}} \wedge \min_{1 \leq k, \ell \leq r }\mathcal{T}_{\tilde{\varepsilon}_N}^{(k \ell)} \wedge T_0^{(11)}\). We verify this term by term:
\begin{itemize}
\item[(i)] For every \(1 \le i,j \le r\) and \(t \leq \mathcal{T}_{L_{0,\beta}}^{(ij)}  \wedge \mathcal{T}_{L_{0,\beta}} \wedge \min_{1 \leq k, \ell \leq r} \mathcal{T}_{\tilde{\varepsilon}_N}^{(k \ell)} \wedge T_0^{(11)}\), the lower bound in~\eqref{eq: comparison inequality} implies
\[
\frac{C_0 \beta \sqrt{M} p \lambda_i \lambda_j}{n+2} m_{ij}^{p-1}(t) \geq \frac{C_0  \beta \sqrt{M}  p \lambda_i \lambda_j}{n+2} \ell_{ij}^{p-1}(t) \geq \frac{C_0  \beta \sqrt{M}  p \lambda_i \lambda_j}{n+2} \ell_{ij}^{p-1}(0).
\]
Thus, for every \(0 \leq k \leq n-1\), using the assumption on \(\sqrt{M}\), we have
\[
\frac{C_0 \beta \sqrt{M} p \lambda_i \lambda_j}{n+2} \left (\frac{\gamma_{ij}}{2 \sqrt{N}} \right)^{p-1} \geq C  \gamma_0 N^{-\frac{n}{2(n+1)}} \geq K_1 \frac{\gamma_0}{\sqrt{N}} t^k,
\]
provided \(t^k = \mathcal{O}(1)\), which certainly holds for all \(k \leq n-1\) and \(t \leq T_0^{(11)}\) since \(T_0^{(11)} < 1\). 

\item[(ii)] A sufficient condition to control the second term is given by \(F(t) \leq G(t)\), where \(F(t) = K_1 t^n\) and \(G(t)= \frac{C_0 \beta \sqrt{M} p \lambda_i \lambda_j}{n+2} \ell_{ij}^{p-1}(t)\). By an easy computation, we have for any \(k \leq n\),
\[
F^{(k)}(t) = K_1 n(n-1) \cdots (n-k+1)t^{n-k}
\]
and
\[
G^{(k)}(t) = \frac{C_0 \beta \sqrt{M} p \lambda_i \lambda_j \prod_{i=1}^k \left ( \frac{p-1}{p-2} + (i-1) \right)}{n+2}\left(\frac{\gamma_{ij}}{2 \sqrt{N}}\right)^{p-1} \left(\frac{1}{t_\ast^{(ij)}}\right)^k \left(1-\frac{t}{t_\ast^{(ij)}}\right)^{- \left(\frac{p-1}{p-2} + k\right)},
\]
where \(t_\ast^{(ij)}\) denotes the blow-up time of \(\ell_{ij}\) which is given by
\[
\frac{1}{t_\ast^{(ij)}} \coloneqq(1 - C_0) \beta \sqrt{M} p(p-2)  \lambda_i \lambda_j \left ( \frac{\gamma_{ij}}{2\sqrt{N}} \right )^{p-2}.
\]
For \(k \leq n-1\), it holds that \(G^{(k)}(0) \geq 0 = F^{(k)}(0)\). For \(k=n\), we obtain the lower bound
\[
\begin{split}
G^{(n)}(t) & \ge \frac{(\sqrt{M} \beta p \lambda_i \lambda_j)^{n+1} C_0 (1-C_0)^n}{n+2} \left ( \frac{\gamma_2}{2 \sqrt{N}}\right )^{p-1+n(p-2)} \left(1-\frac{t}{t_\ast^{(ij)}}\right)^{- \left(\frac{p-1}{p-2} + n\right)} \\
& \ge\frac{C^{n+1} (1-C_0)^n (n+2)^n \gamma_0^{n+1}}{C_0^n \gamma_2^n 2^{p-1+n(p-2)}}  \left(1-\frac{t}{t_\ast^{(ij)}}\right)^{- \left(\frac{p-1}{p-2} + n\right)}\\
& \geq K_1 n! = F^{(n)}(t),
\end{split}
\]
provided \(t/t_\ast^{(ij)}\) is of order \(1\) over the considered time interval, which certainly holds since \(T_0^{(11)} \leq t_\ast^{(11)} < t_\ast^{(ij)}\).

\item[(iii)] We control the last term as follows. According to the integral inequality~\eqref{eq: integral inequality}, on the event \(\cap_{1 \leq k, \ell \le r} \mathcal{A}^{(k \ell)}\), we have
\[
2 \sum_{1 \le k ,\ell \le r} \int_0^t |a_{k \ell}(s)|ds \leq \frac{2r^2}{1-C_0} \max_{1 \le k, \ell \le r}\{m_{k \ell}(t)\} \leq \frac{2 r^2}{1-C_0} \tilde{\varepsilon}_N =  \frac{2 r^2}{1-C_0}  \frac{\tilde \gamma}{\sqrt{N}},
\]
for \(t \leq \min_{1 \le k, \ell \le r} \mathcal{T}_{L_{0,\beta}}^{(k \ell)} \wedge \mathcal{T}_{L_{0,\beta}}  \wedge \min_{1 \leq k, \ell \leq r} \mathcal{T}_{\tilde{\varepsilon}_N}^{(k \ell)} \wedge T_0^{(11)}\). From the lower bound in~\eqref{eq: comparison inequality}, we also have 
\[
\frac{C_0 \beta \sqrt{M} p \lambda_i \lambda_j}{n+2} m_{ij}^{p-1}(t) \geq \frac{C_0 \beta \sqrt{M} p \lambda_i \lambda_j}{n+2} \ell_{ij}^{p-1}(t) \geq \frac{C_0 \beta \sqrt{M} p \lambda_i \lambda_j}{n+2} \ell_{ij}^{p-1}(0),
\]
for \(t \leq \min_{1 \le k, \ell \le r} \mathcal{T}_{L_{0,\beta}}^{(k \ell)}  \wedge \mathcal{T}_{L_{0,\beta}} \wedge \min_{1 \le k, \ell \le r} \mathcal{T}_{\tilde{\varepsilon}_N}^{(k \ell)} \wedge T_0^{(11)}\). Using the assumption on \(\sqrt{M}\), it follows that
\[
\frac{C_0 \beta \sqrt{M} p \lambda_i \lambda_j}{n+2} \left (\frac{\gamma_{ij}}{2 \sqrt{N}} \right)^{p-1} \ge \frac{C \gamma_0}{2^{p-1}N^{\frac{n}{2(n+1)}}}  \geq K_1 \frac{2r^2}{1-C_0} \frac{\tilde \gamma}{\sqrt{N}},
\]
where the last inequality holds provided \(N \ge \left ( \frac{K_1 2^p r^2 \tilde{\gamma}}{C \gamma_0 (1-C_0)} \right)^{2(n+1)}\). Thus, the integral term is also bounded appropriately. 
\end{itemize}
On the event \(\cap_{1 \leq k,\ell \leq r} \mathcal{A}^{(k \ell)}\), we therefore have that 
\(\min_{1 \leq k, \ell \leq r} \mathcal{T}_{\tilde{\varepsilon}_N}^{(k \ell)} \leq \min_{1 \le k, \ell \le r} \mathcal{T}_{L_{0,\beta}}^{(k \ell)}  \wedge \mathcal{T}_{L_{0,\beta}}\), which implies that 
\[
\mathcal{T}_{\tilde{\varepsilon}_N}^{(11)} = \min_{1 \le k, \ell \le r} \mathcal{T}_{\tilde{\varepsilon}_N}^{(k \ell)},
\]
with \(\mathbb{Q}_{\boldsymbol{X}}\)-probability at least \(1 - K_2 \exp \left (- \gamma_0^2/ (K_2 T_0^{(11)}) \right )\) and with \(\mathbb{P}\)-probability at least \(1 - \exp(-K N)\). \\

We now show that \(m_{11}\) remains the dominant correlation and, in particular, reaches the threshold \(\varepsilon_N\) before any other correlation. From Lemma~\ref{lem: evolution equation m_ij} we see that at time \(t = \mathcal{T}_{\tilde{\varepsilon}_N}^{(11)}\),
\[
L_\beta m_{11}(t) \geq (1-C_0) \beta \sqrt{M} p\lambda_1^2m_{11}^{p-1}(t) = (1-C_0) \beta \sqrt{M} p\lambda_1^2 \left ( \frac{\tilde{\gamma}}{\sqrt{N}}\right)^{p-1}
\]
and for every \((i,j) \neq (1,1)\),
\[
\begin{split}
L_\beta m_{ij}(t)  & \leq C_0 \beta \sqrt{M} p\lambda_1^2m_{11}^{p-1}(t) + \beta \sqrt{M} p \lambda_i \lambda_j m_{ij}^{p-1}(t) \\
& = C_0 \beta \sqrt{M} p\lambda_1^2 \left ( \frac{\tilde{\gamma}}{\sqrt{N}}\right)^{p-1} + \beta \sqrt{M} p \lambda_i \lambda_j m_{ij}^{p-1}(t) .
\end{split}
\]
For every \((i,j) \neq (1,1)\), we upper bound
\[
m_{ij}(t) \le u_{ij} (T_{\ell, \tilde{\varepsilon}_N}^{(11)}) \le \frac{3 \gamma_{ij}}{2 \sqrt{N}} \left (1 - \frac{1}{\lambda_0} \right)^{- \frac{1}{p-2}},
\]
so that
\[
L_\beta m_{ij}(t) \leq  \beta \sqrt{M} p \lambda_1^2 \left (\frac{\tilde{\gamma}}{\sqrt{N}} \right)^{p-1} \left ( C_0 + \frac{1}{\lambda_0} \frac{1-C_0}{1+C_0} \left ( \frac{\gamma_2}{3 \gamma_1} \right)^{p-2} \right) ,
\]
where we used~\eqref{eq: lambda_0} and~\eqref{eq: tilde gamma}. Since \(\lambda_0>1\) and \(\frac{1-C-0}{1+C_0} < 1\), we see that for every \(C_0 < \frac{1}{2}- \frac{1}{2}\left ( \frac{\gamma_2}{3 \gamma_1} \right)^{p-2}\), 
\[
C_0 + \frac{1}{\lambda_0} \frac{1-C_0}{1+C_0} \left ( \frac{\gamma_2}{3 \gamma_1} \right)^{p-2}  \le 1-C_0,
\]
leading to
\[
L_\beta m_{ij}(t) \leq L_\beta m_{11}(t).
\]
As a consequence, on the event \(\cap_{1 \leq k,\ell \leq r} \mathcal{A}^{(k \ell)}\), since \(m_{11}(\mathcal{T}_{\tilde{\varepsilon}_N}^{(11)}) > m_{ij}(\mathcal{T}_{\tilde{\varepsilon}_N}^{(11)})\) and \(L_\beta m_{11}(\mathcal{T}_{\tilde{\varepsilon}_N}^{(11)}) \geq L_\beta m_{ij}(\mathcal{T}_{\tilde{\varepsilon}_N}^{(11)})\) for every \((i,j) \neq (1,1)\), we obtain that \(m_{11}(t)>m_{ij}(t)\) for all \(\mathcal{T}_{\tilde{\varepsilon}_N}^{(11)} \leq t \leq \min_{1 \leq k,\ell \leq r} \mathcal{T}_{L_{0,\beta}}^{(k\ell)} \wedge \mathcal{T}_{L_{0,\beta}} \wedge \min_{1 \leq k,\ell \leq r} \mathcal{T}_{\varepsilon_N}^{(k\ell)} \wedge T_0^{(11)}\), ensuring that
\[
\mathcal{T}_{\varepsilon_N}^{(11)} = \min_{1 \leq k, \ell \leq r} \mathcal{T}_{\varepsilon_N}^{(k \ell)},
\]
with \(\mathbb{Q}_{\boldsymbol{X}}\)-probability at least \(1 - K_2 \exp \left (- \gamma_0^2/ (K_2 T_0^{(11)}) \right )\) and with \(\mathbb{P}\)-probability at least \(1 - \exp(-K N)\).\\ 

Finally, it remains to show that on the event \(\cap_{1 \leq k,\ell \leq r} \mathcal{A}^{(k \ell)}\), \(\mathcal{T}_{\varepsilon_N}^{(11)} \leq \mathcal{T}_{L_{0,\beta}}\) with high \(\mathbb{Q}_{\boldsymbol{X}}\)- and \(\mathbb{P}\)-probability. We recall that the estimate~\eqref{eq: bounding flow langevin p>3} holds for \(t \le \min_{1 \leq k, \ell \leq r} \mathcal{T}_{L_{0,\beta}}^{(k\ell)} \wedge \mathcal{T}_{L_{0,\beta}} \wedge \mathcal{T}_{\varepsilon_N}^{(11)} \wedge T_0^{(11)}\) with \(\mathbb{Q}_{\boldsymbol{X}}\)-probability at least \(1 - K_2 \exp \left (- \gamma_0^2 / (K_2T_0^{(11)}) \right )\) and with \(\mathbb{P}\)-probability at least \(1 - \exp(-K N)\). Using similar computations as before, condition (4) of Lemma~\ref{lem: bounding flows} is satisfied since
\[
\int_0^t |a_{ij}(s)| ds \le \frac{1}{2} a_{11}(t),
\]
for every \(\mathcal{T}_{\tilde{\varepsilon}_N}^{(11)} \leq t \leq  \min_{1 \le k,\ell \le r} \mathcal{T}_{L_{0,\beta}}^{(k \ell)}  \wedge \mathcal{T}_{L_{0,\beta}} \wedge \mathcal{T}_{\varepsilon_N}^{(11)} \wedge T_0^{(11)}\). This implies that, on the event \(\cap_{1 \leq k, \ell \leq r} \mathcal{A}^{(k\ell)}\),
\begin{equation} \label{eq: bounding flow langevin p>3 2}
|L_{0,\beta}m_{ij}(t)| \leq K_1 \left(\frac{\gamma_0}{\sqrt{N}} \sum_{k=0}^{n-1} t^k +t^n + 2r^2 \int_0^t |a_{11}(s)| ds \right),
\end{equation}
for every \(\mathcal{T}_{\tilde{\varepsilon}_N}^{(11)} \leq t \leq  \min_{1 \le k,\ell \le r} \mathcal{T}_{L_{0,\beta}}^{(k \ell)}  \wedge \mathcal{T}_{L_{0,\beta}} \wedge \mathcal{T}_{\varepsilon_N}^{(11)} \wedge T_0^{(11)}\), with \(\mathbb{Q}_{\boldsymbol{X}}\)-probability at least \(1 - K_2 \exp \left (- \gamma_0^2 / (K_2T_0^{(11)}) \right )\) and with \(\mathbb{P}\)-probability at least \(1 - \exp(-K N)\). In the same way as before, by the assumption on \(\sqrt{M}\), each term in the right-hand side of~\eqref{eq: bounding flow langevin p>3 2} is bounded above by \( \frac{C_0 \beta \sqrt{M} p \lambda_1^2}{n+2} m_{11}^{p-1}(t)\) for every \(\mathcal{T}_{\tilde{\varepsilon}_N}^{(11)} \leq t \leq  \min_{1 \le k,\ell \le r} \mathcal{T}_{L_{0,\beta}}^{(k \ell)}  \wedge \mathcal{T}_{L_{0,\beta}} \wedge \mathcal{T}_{\varepsilon_N}^{(11)} \wedge T_0^{(11)}\). This ensures 
\[
\mathcal{T}_{\varepsilon_N}^{(11)}  \le \mathcal{T}_{L_{0,\beta}},
\]
with high \(\mathbb{Q}_{\boldsymbol{X}}\)- and \(\mathbb{P}\)-probability. 

According to~\eqref{eq: function u}, on the event \(\cap_{1 \leq k,\ell \leq r} \mathcal{A}^{(k \ell)}\), we have for every \((i,j) \neq (1,1)\), 
\[
\begin{split}
m_{ij}(\mathcal{T}_{\varepsilon_N}^{(11)} ) \le u_{ij}(T_{\ell, \varepsilon_N}^{(11)}) & = \frac{3\gamma_{ij}}{2\sqrt{N}} \left (1 - \frac{1 + C_0}{1-C_0} \frac{\lambda_i \lambda_j}{\lambda_1^2}  \left ( \frac{3\gamma_{ij}}{\gamma_{11}} \right )^{p-2} \left ( 1 -\left (\frac{\gamma_{11}}{2} \right)^{p-2} N^{-\frac{p-2}{2(p-1)}}\right) \right )^{-\frac{1}{p-2}}\\
& \le \frac{3\gamma_{ij}}{2\sqrt{N}} \left ( \frac{\lambda_0}{ \lambda_0-1} \right )^{-\frac{1}{p-2}}\\
& \le \frac{\tilde{\gamma}}{\sqrt{N}},
\end{split}
\]
where the last inequality follows from~\eqref{eq: tilde gamma}. Thus, when \(m_{11}\) reaches \(\varepsilon_N\), all the other correlations are at most \(\tilde{\gamma}N^{-1/2}\). This shows that \(\mathcal{T}_E = \mathcal{T}_{\varepsilon_N}^{(11)} \). Therefore, on the event \(\mathcal{C}_{0,\beta}^{(N)}(n,\gamma_0) \cap \mathcal{C}_1^{(N)}(\gamma_1,\gamma_2)\), 
\[
\mathcal{T}_E = \mathcal{T}_{\varepsilon_N}^{(11)} \le T_0^{(11)} \lesssim \frac{\gamma_2}{(n+2) \gamma_0 N^{\frac{1}{2(n+1)}}},
\]
with \(\mathbb{Q}_{\boldsymbol{X}}\)-probability at least 
\[
1  - K_2 r^2\exp \left (- \gamma_2 \gamma_0(n+2)N^{\frac{1}{2(n+1)}} /(4K_2)\right ) -K_2 \exp \left (- \gamma_0^3(n+2)N^{\frac{1}{2(n+1)}} / (K_2 \gamma_2)\right ),
\]
and with \(\mathbb{P}\)-probability at least \(1 - \exp(-K  N)\). This completes the proof of Lemma~\ref{lem: weak recovery first spike Langevin p>2}.
\end{proof}

We finally provide the proof of the second intermediate result, namely Lemma~\ref{lem: weak implies strong recovery Langevin p>2}.

\begin{proof}[\textbf{Proof of Lemma~\ref{lem: weak implies strong recovery Langevin p>2}}]
Set \(\varepsilon_N = c_1 N^{-\frac{p-2}{2(p-1)}} \) for a constant \(c_1>0\) and recall that the weak-recovery event is
\[
E = \{\boldsymbol{X} \colon m_{11} (\boldsymbol{X}) > \varepsilon_N \enspace \mathrm{and} \enspace m_{ij}(\boldsymbol{X})  \le c_2 N^{-1/2} \enspace \forall \, (i,j) \neq (1,1)\},
\]
for a constant \(c_2>0\). Fix \(\varepsilon > 0\) and define the stopping time
\[
\tau \coloneqq \mathcal{T}_{\varepsilon_N/2}^{(11,-)} \wedge \mathcal{T}_{\varepsilon}^{(11,+)} \wedge \min_{(k,\ell) \neq (1,1)} \mathcal{T}_{c \varepsilon_N}^{(k\ell,+)} ,
\]
for a constant \(c>0\), where for any \(a >0\),
\[
\begin{split}
\mathcal{T}_a^{(ij,+)} & \coloneqq  \inf \left \{t \ge 0 \colon m_{ij}(t) \ge a \right \},\\ 
\mathcal{T}_a^{(ij,-)} & \coloneqq  \inf \left \{t \ge 0 \colon m_{ij}(t) \le a \right \}.
\end{split}
\]
We analyze the dynamics on the interval \([0,\tau]\).

By Lemma~\ref{lem: evolution equation m_ij}, on the interval \([0,\tau]\),
\[
\begin{split}
L_\beta m_{11}(t) & \geq - \norm{L_{0,\beta} m_{11}}_\infty + \beta \sqrt{M} p \lambda_1^2 m_{11}^{p-1}(t) (1-m_{11}^2(t) - r^2 m_{12}^2(t)) \\
& \geq - \norm{L_{0,\beta} m_{11}}_\infty + (1-\varepsilon^2 -r^2 c^2 \varepsilon_N^2) \beta \sqrt{M} p \lambda_1^2 m_{11}^{p-1}(t).
\end{split}
\]
Note that \(1-\varepsilon^2 -r^2 c^2\varepsilon_N^2 \ge 1 - \varepsilon^2 - \frac{C_0}{2} > 0\) for a choice of \(C_0 \in (0, 2(1-\varepsilon^2))\) and provided \(N > \left ( \frac{2r^2 c^2c_1^2}{C_0}\right)^{\frac{p-1}{p-2}}\). Furthermore, according to~\eqref{eq: bound norm L_0m_ij}, we have
\[
\norm{L_{0,\beta} m_{11}}_\infty \leq \Lambda \leq  \frac{C_0}{2} \beta \sqrt{M} p \lambda_1^2 \left(\frac{\varepsilon_N}{2}\right)^{p-1} \le \frac{C_0}{2} \beta \sqrt{M} p \lambda_1^2 m_{11}^{p-1}(t)
\]
with \(\mathbb{P}\)-probability at least \(1 - \exp(-K N)\), provided \(\sqrt{M} \ge \frac{2^p \Lambda}{C_0 \beta p \lambda_1^2 c_1^{p-1}} N^{\frac{p-2}{2}}\), which certainly holds by assumption. Combining the two estimates gives, on \([0,\tau]\),
\[
L_\beta m_{11}(t) \geq (1-\varepsilon^2 - C_0) \beta \sqrt{M} p \lambda_1^2 m_{11}^{p-1}(t) > 0.
\]
Applying Lemma~\ref{lem: Gronwall} with initial condition \(m_{11}(0) \ge \varepsilon_N\), we obtain that on \([0,\tau]\),
\[
m_{11}(t) \ge \ell_{11}(t),
\]
where
\begin{equation} \label{eq m11}
\ell_{11}(t) \coloneqq \frac{c_1 N^{-\frac{p-2}{2(p-1)}}}{2} \left (1 - (1-\varepsilon^2 - C_0) \beta \sqrt{M} p (p-2) \lambda_1^2 \left ( \frac{c_1 N^{-\frac{p-2}{2(p-1)}} }{2}\right)^{p-2} t \right )^{-\frac{1}{p-2}},
\end{equation}
with \(\mathbb{Q}_{\boldsymbol{X}}\)-probability at least \(1 - K_1 \exp(- N^{\frac{1}{p-1}} / (16 K_1 \tau))\) and with \(\mathbb{P}\)-probability at least \(1 - \exp(-K N)\). Thus, by Lemma~\ref{lem: Gronwall}, \(m_{11}(t)\) admits a strictly increasing lower bound \(\ell_{11}(t)\) on \([0,\tau]\) with \(\ell_{11}(0) \ge \varepsilon_N/2\), yielding
\[
\mathcal{T}_{\varepsilon}^{(11,+)} \wedge \min_{(k,\ell) \neq (1,1)} \mathcal{T}_{c\varepsilon_N}^{(k\ell,+)} \le \mathcal{T}_{\varepsilon_N/2}^{(11,-)}. 
\]
Define \(t_\ast\) by \(\ell_{11}(t_\ast)= \varepsilon\). From~\eqref{eq m11}, a direct computation yields
\[
t_\ast = \frac{1 - \left (\frac{c_1}{2 \varepsilon} N^{-\frac{p-2}{2(p-1)}}\right )^{p-2}}{(1-\varepsilon^2 - C_0) \beta \sqrt{M} p (p-2) \lambda_1^2 \left (\frac{c_1}{2} N^{-\frac{p-2}{2(p-1)}}\right )^{p-2}} \lesssim N^{-\frac{p-2}{2(p-1)}} ,
\]
where we used the assumption on \(\sqrt{M}\) and we denote by \(\lesssim\) a constant depending only on \(p\) and \(\varepsilon\). To conclude, it suffices to show that \(m_{ij}(t_\ast) \le c \varepsilon_N\) for all \((i,j) \neq (1,1)\). By Lemma~\ref{lem: evolution equation m_ij} and~\eqref{eq: bound norm L_0m_ij}, we have the crude bound
\[
L_\beta m_{ij}(t) \le \norm{L_{0,\beta} m_{ij}}_\infty +\beta \sqrt{M} p \lambda_i \lambda_j m_{ij}^{p-1}(t)  \le \Lambda+\beta \sqrt{M} p \lambda_i \lambda_j m_{ij}^{p-1}(t) \le \tilde \Lambda ,
\]
for every \(t \le \mathcal{T}_{c \varepsilon_N}^{(ij,+)}\), with \(\mathbb{P}\)-probability at least \(1 - \exp(-K N)\). Here, \(\tilde \Lambda\) denotes a constant depending on \(\beta, p, \{\lambda_i\}_{i=1}^r\). Applying Lemma \ref{lem: Gronwall} with \(m_{ij}(0) \le c_2 N^{-1/2}\) gives
\[
m_{ij}(t) \le \frac{3c_2}{2\sqrt{N}} + \tilde \Lambda t,
\]
up to time \(\mathcal{T}_{c \varepsilon_N}^{(ij,+)}\), with \(\mathbb{Q}_{\boldsymbol{X}}\)-probability at least \(1 - K_2 \exp(- c_2^2/ (4 K_2 \mathcal{T}_{c\varepsilon_N}^{(ij,+)}))\) and with \(\mathbb{P}\)-probability at least \(1 - \exp(-K N)\). Since \(N^{-1/2} = o(\varepsilon_N)\) for \(p \ge 3\), we see that at time \(t_\ast \sim N^{-\frac{p-2}{2(p-1)}}\), \(m_{ij}(t_\ast) = \mathcal{O}(\varepsilon_N)\) for any \((i,j)\neq(1,1)\). Thus all non-leading correlations reach at most the scale \(\varepsilon_N\) until \(m_{11}(t)\) reaches \(\varepsilon\). Hence, \(t_\ast\le \min_{(k,\ell) \neq (1,1)} \mathcal{T}_{c \varepsilon_N}^{(k \ell,+)}\) and, since \(\mathcal{T}_{\varepsilon}^{(11,+)} \le t_\ast\), we have
\begin{equation} \label{eq: from varepsilon_N to epsilon}
\mathbb{Q}_{\boldsymbol{X}} \left ( \mathcal{T}_\varepsilon^{(11,+)} \lesssim N^{-\frac{p-2}{2(p-1)}} \right ) \ge 1 - K_3 \exp \left ( - \frac{ N^{\frac{p-2}{2(p-1)}}}{K_3} \right ),
\end{equation}
with \(\mathbb{P}\)-probability at least \(1 - \exp(-K N)\). 

Similarly, we control the time needed for \(m_{11}\) to go from \(\varepsilon\) to \(1-\varepsilon\). Define the stopping time
\[
\tau' \coloneqq \mathcal{T}_{\varepsilon/2}^{(11,-)} \wedge \mathcal{T}_{1-\varepsilon}^{(11,+)}  \wedge \min_{(k,\ell) \neq (1,1)} \mathcal{T}_{c' \varepsilon_N}^{(k\ell,+)} ,
\]
where \(c'>0\) is a constant of order \(1\). On the time interval \([\mathcal{T}_{\varepsilon}^{(11,+)},\tau']\) we have
\[
\frac{\varepsilon}{2} \le m_{11}(t) \le 1-\varepsilon \qquad\text{and}\qquad
m_{1j}(t)\le c' \varepsilon_N, \quad j\neq 1.
\]
By Lemma~\ref{lem: evolution equation m_ij} and~\eqref{eq: bound norm L_0m_ij}, for \(t \in [\mathcal{T}_{\varepsilon}^{(11,+)},\tau']\),
\[
\begin{split}
L_\beta m_{11}(t) &\ge - \norm{L_{0,\beta} m_{11}}_\infty  + \beta \sqrt{M} p \lambda_1^2 m_{11}^{p-1}(t) \left(1 - m_{11}^2(t) - r^2 m_{12}^2(t)\right) \\
& \ge  - \Lambda + \beta \sqrt{M} p \lambda_1^2 \left ( \frac{\varepsilon}{2} \right)^{p-1} \left(1 - (1-\varepsilon)^2 - r^2 c'^2 \varepsilon_N^2\right)\\
& \ge - \Lambda + \beta \sqrt{M} p \lambda_1^2 2^{-p} \varepsilon^{p-1} (2\varepsilon-\varepsilon^2)\\
& \ge c_\varepsilon \beta \sqrt{M} ,
\end{split}
\]
for some constant \(c_\varepsilon>0\) depending only on \(\varepsilon, p,\lambda_1\), provided \(N\) is large enough so that \(1 - (1-\varepsilon)^2 - r^2 c'^2 \varepsilon_N^2 \ge \frac{2\varepsilon-\varepsilon^2}{2} > 0\), with \(\mathbb{P}\)-probability at least \(1-\exp(-KN)\). Applying Lemma~\ref{lem: Gronwall} on the interval \([\mathcal{T}_{\varepsilon}^{(11,+)},\tau']\) with initial condition \(m_{11}(\mathcal{T}_{\varepsilon}^{(11,+)}) \ge \varepsilon\), we obtain 
\[
m_{11}(t) \ge \frac{\varepsilon}{2} + c_\varepsilon \beta \sqrt{M} (t - \mathcal{T}_{\varepsilon}^{(11,+)}),
\]
with \(\mathbb{Q}_{\boldsymbol{X}}\)-probability at least \(1 - K_4 \exp(- \varepsilon^2N/(16\tau'K_4))\) and with \(\mathbb{P}\)-probability at least \(1 - \exp(- K N)\).
Let
\[
\Delta t_\varepsilon  \coloneqq \frac{1-3\varepsilon/2}{c_\varepsilon \beta \sqrt{M}}.
\]
Then for all \(t \in [\mathcal{T}_{\varepsilon}^{(11,+)}, \mathcal{T}_{\varepsilon}^{(11,+)} + \Delta t_\varepsilon]\) we have
\[
m_{11}(t)  \ge \frac{\varepsilon}{2} + c_\varepsilon \beta \sqrt{M}\, (t - \mathcal{T}_{\varepsilon}^{(11,+)}) \ge  \frac{\varepsilon}{2} + c_\varepsilon \beta \sqrt{M}\, \Delta t_\varepsilon
= 1-\varepsilon.
\]
As before, it remains to control the non-leading correlations. Lemma~\ref{lem: evolution equation m_ij} and~\eqref{eq: bound norm L_0m_ij} with initial time \(\mathcal{T}_\varepsilon^{(11,+)}\), we obtain
\[
m_{ij}(t) \le \frac{c\varepsilon_N}{2} + \tilde \Lambda (t - \mathcal{T}_\varepsilon^{(11,+)}),
\]
for every \(\mathcal{T}_\varepsilon^{(11,+)} \le t \le \mathcal{T}_{c'\varepsilon_N}^{(ij,+)}\), with \(\mathbb{Q}_{\boldsymbol{X}}\)-probability at least \(1 - K_5 \exp(- c^2 c_1^2 N^{\frac{1}{p-1}}/(16 \mathcal{T}_{c'\varepsilon_N}^{(ij,+)}K_5))\). Since
\[
\Delta t_\varepsilon  \coloneqq \frac{1-3\varepsilon/2}{c_\varepsilon \beta \sqrt{M}} \lesssim N^{-\frac{p-2}{2}}= o(\varepsilon_N),
\]
we deduce that, for \(N\) large enough and all \((i,j)\neq (1,1)\),
\[
m_{ij}(t) \le C_1 \varepsilon_N, 
\]
for evey \(\mathcal{T}_\varepsilon^{(11,+)} \le t \le \mathcal{T}_\varepsilon^{(11,+)}  + \Delta t_\varepsilon\), with \(\mathbb{Q}_{\boldsymbol{X}}\)-probability at least \(1 - K_5 \exp(- c^2 c_1^2 N^{\frac{1}{p-1}+\frac{p-2}{2}}/K_5)\). In particular, \(\mathcal{T}_{1-\varepsilon}^{(11,+)} \le \min_{(k,\ell)\neq(1,1)}\mathcal{T}_{c'\varepsilon_N}^{(k\ell,+)}\), and it follows that
\[
\mathcal{T}_{1-\varepsilon}^{(11,+)} = \mathcal{T}_{\varepsilon}^{(11,+)}  + \left( \mathcal{T}_{1-\varepsilon}^{(11,+)} - \mathcal{T}_{\varepsilon}^{(11,+)} \right) \le \mathcal{T}_{\varepsilon}^{(11,+)} + \Delta t_\varepsilon \lesssim N^{-\frac{p-2}{2(p-1)}}+N^{-\frac{p-2}{2}} \lesssim N^{-\frac{p-2}{2(p-1)}} ,
\]
with \(\mathbb{Q}_{\boldsymbol{X}}\)-probability at least \(1 - K_6\exp(-N^{\frac{p-2}{2(p-1)}}/K_7)\ -K_7 \exp(- \varepsilon^2 N^{p/2}/K_7)\) and with \(\mathbb{P}\)-probability at least \(1 - \exp(- K N)\).
\end{proof}

\subsection{Recovery of all spikes} \label{subsection: recovery of all spikes p>2}

We now prove Proposition~\ref{thm: strong recovery all spikes Langevin p>2} on the exact recovery of all \(r\) spikes. The argument proceeds iteratively: at each step we show that one additional spike direction is recovered, while all correlations involving the remaining coordinates decay to a sufficiently small scale. Fix \(\varepsilon > 0\). For every \(i \in [r]\), define the strong-recovery event for spike \(\boldsymbol{v}_i\) by
\begin{equation} \label{eq: strong recovery v_i}
R_i(\varepsilon) \coloneqq \left \{ \boldsymbol{X} \in\mathcal{S}_{N,r} \colon m_{ii}(\boldsymbol{X}) \geq 1-\varepsilon \enspace \textnormal{and} \enspace \max_{j\neq i} \{ m_{ij}(\boldsymbol X) \vee m_{ji}(\boldsymbol X) \} \lesssim \log(N)^{-\frac{1}{2}}N^{-\frac{p-1}{4}} \right \}.
\end{equation}
Then, for \(1 \le k \le r\) the events 
\[
\begin{split}
E_k (\varepsilon) & \coloneqq \cap_{1 \leq i \leq k} R_i(\varepsilon) \cap   \left \{ \boldsymbol{X} \colon m_{ij}(\boldsymbol{X}) \in \Theta(N^{-\frac{1}{2}}) \enspace \forall \, i,j \ge k+1\right \}.
\end{split}
\]
In particular, \(E_r(\varepsilon) = \cap_{i=1}^r R_i(\varepsilon) \) coincides with the event \(R(\varepsilon)\) defined in~\eqref{eq: set strong recovery p>2}. The structure of these events captures the iterative nature of the proof: once \(m_{ii}\) reaches the macroscopic level \(1-\varepsilon\), all cross-correlations \(m_{ij}\) and \(m_{ji}\) for \(j \neq i\) decay below the scale \(\log(N)^{-\frac{1}{2}}N^{-\frac{p-1}{4}}\). This decay is essential for ensuring the recovery of the next correlation \(m_{i+1,i+1}\).

Our first step is to show that the event \(E_1 (\varepsilon)\) can be reached from a random initialization satisfying Condition \(1\), provided that \(M\) scales at least as \(N^{p-1}\). From Proposition~\ref{thm: strong recovery first spike Langevin p>2} and Lemma~\ref{lem: weak recovery first spike Langevin p>2}, this implies that the lower complexity threshold \(M \gtrsim N^{p-2}\) suffices for recovering the first spike but does not ensure stability of the remaining directions.

\begin{lem} \label{lem: E_1}
Let \(\beta \in (0,\infty)\) and \(p \geq 3\). Consider a sequence of initialization measures \(\mu_N^0 \in \mathcal{M}_1(\mathcal{S}_{N,r})\) supported on \(\mathcal{X}^+\). Fix  \(\gamma_1 > 1 >\gamma_2 > 0\) and \(\varepsilon > 0\). Then, there exist constants \(C = C(p, \beta, \{\lambda_i\}_{i=1}^r)>0\), \(C_0 = C_0(\varepsilon) \in (0,\frac{1}{2})\), \(K>0\), and \(K_1,K_2,K_3>0\) (depending only on the fixed parameters) such that, if 
\[
\lambda_1 > \frac{1+C_0}{1-C_0} \left (\frac{3 \gamma_1}{\gamma_2}\right )^{p-2} \lambda_2 \quad \textnormal{and} \quad \sqrt{M} \geq \frac{C}{\beta p \lambda_r^2 C_0 \gamma_2^{p-1}} N^{\frac{p-1}{2}}, 
\]
then, for \(N\) sufficiently large,
\[
\int_{\mathcal{X}^+} \mathbb{Q}_{\boldsymbol{X}} \left (\mathcal{T}_{E_1 (\varepsilon)} \gtrsim \frac{1}{\sqrt{N}} \right ) \boldsymbol{1}_{\mathcal{C}_1^{(N)}(\gamma_1,\gamma_2)}(\boldsymbol{X}) \mathrm{d} \mu_N^0 (\boldsymbol{X}) \le \eta_1 (N, \varepsilon, \gamma_1, \gamma_2) ,
\]
with \(\mathbb{P}\)-probability at least \(1 - \exp(-K N)\), where 
\[
\eta_1 (N, \varepsilon, \gamma_1, \gamma_2) = K_1 r^2 e^{-\gamma_2^2 \sqrt{N}/ K_1} + K_2 e^{-N^{\frac{p+1}{2(p-1)}}/ K_2} + K_3 e^{- \varepsilon^2 N^{3/2}/ K_3}.
\]
Here, \(\gtrsim\) hides a constant that depends on \(p\) and \(\varepsilon\).
\end{lem}

Once the event \(E_1 (\varepsilon)\) has been reached, reaching \(E_2 (\varepsilon)\) is straightforward, and the same holds for subsequent steps. Assuming that the \((k-1)\)st event \(E_{k-1}(\varepsilon)\) has been attained, we now show that the process reaches \(E_k (\varepsilon)\) with high probability.

\begin{lem} \label{lem: E_2}
Let \(\beta \in (0,\infty)\) and \(p \geq 3\). Fix \(\gamma_1 > 1 >\gamma_2 > 0\), \(\varepsilon > 0\), and \(2 \le k \le r\). Then, there exist constants \(C = C(p, \beta, \{\lambda_i\}_{i=1}^r)>0\), \(C_0 = C_0(\varepsilon) \in (0,\frac{1}{2})\), \(K>0\), and \(K_1,K_2,K_3>0\) (depending only on the fixed parameters) such that, if 
\[
\lambda_{k-1} > \frac{1+C_0}{1-C_0} \left (\frac{3 \gamma_1}{\gamma_2}\right )^{p-2} \lambda_k \quad \textnormal{and} \quad \sqrt{M} \geq \frac{C}{\beta p \lambda_r^2 C_0 \gamma_2^{p-1}} N^{\frac{p-1}{2}}, 
\]
then there exists \(T_k \leq C_k N^{-1/2}\) for some constant \(C_k>0\) such that, for \(N\) sufficiently large,
\[
\inf_{\boldsymbol{X} \in E_{k-1}(\varepsilon)} \mathbb{Q}_{\boldsymbol{X}} \left ( \mathcal{T}_{E_k (\varepsilon)} \le T_k  \right ) \ge 1 - \eta_k (N, \varepsilon, \gamma_1,\gamma_2),
\]
with \(\mathbb{P}\)-probability at least \(1 - \exp(-K N)\), where
\[
\eta_k  (N, \varepsilon, \gamma_1,\gamma_2) = (r-(k-1))^2 K_1 e^{ -\gamma_2^2 \sqrt{N} / K_1} + K_2 e^{- N^{\frac{p+1}{2(p-1)}}/ K_2} + K_3 e^{- \varepsilon^2 N^{3/2}/ K_3}.
\]
\end{lem}

Having Lemmas~\ref{lem: E_1} and~\ref{lem: E_2} at hand, we now are in the position to provide the proof of Proposition~\ref{thm: strong recovery all spikes Langevin p>2}. 

\begin{proof}[\textbf{Proof of Proposition~\ref{thm: strong recovery all spikes Langevin p>2}}]
For every \(k \in [r]\), denote the hitting times \(\tau_k \coloneqq \mathcal{T}_{E_k(\varepsilon)}\), and set \(\tau_0 \coloneqq 0\). Recall the times \(T_1, \ldots, T_r\) from Lemmas~\ref{lem: E_1} and~\ref{lem: E_2}, and set \(T_0 \coloneqq \sum_{k=1}^r T_k\). By Lemma~\ref{lem: E_1}, for \(N\) large enough and with \(\mathbb{P}\)-probability at least \(1-\exp(-K N)\),
\[
\int_{\mathcal{X}^+} \mathbb{Q}_{\boldsymbol{X}} \left ( \tau_1 \gtrsim \frac{1}{\sqrt{N}} \right) \mathbf{1}_{C_1^{(N)} (\gamma_1,\gamma_2)}(\boldsymbol{X}) d \mu_N^0 (\boldsymbol{X}) \le \eta_1 (N,\varepsilon, \gamma_1, \gamma_2).
\]
Choosing \(T_1\) of order \(N^{-1/2}\), so that \(\{\tau_1 > T_1\} \subset \{\tau_1 \gtrsim N^{-1/2}\}\) for \(N\) large enough. Then, Lemma~\ref{lem: E_1} implies that
\begin{equation} \label{eq:tau1-bound}
\int_{\mathcal{X}^+} \mathbb{Q}_{\boldsymbol{X}} \left( \tau_1 \le T_1 \right) \mathbf{1}_{\mathcal{C}_1^{(N)}(\gamma_1,\gamma_2)} (\boldsymbol{X}) d\mu_N^0(\boldsymbol{X}) \ge 1 - \eta_1(N,\varepsilon,\gamma_1,\gamma_2).
\end{equation}
Fix \(2 \le k \le r\). On the event \(\{\tau_{k-1} < \infty\}\), the process \((\boldsymbol{X}_{\tau_{k-1}+t}^\beta)_{t \ge 0}\) is a Markov process started from some point in \(E_{k-1}(\varepsilon)\). By the strong Markov property applied at \(\tau_{k-1}\), we have
\[
\mathbb{Q}_{\boldsymbol{X}} \left ( \tau_k - \tau_{k-1} \le T_k, \tau_{k-1} < \infty \, | \, \mathcal{F}_{\tau_{k-1}} \right ) = \mathbf{1}_{\{\tau_{k-1} < \infty\}} \mathbb{Q}_{\boldsymbol{X}_{\tau_{k-1}}} \left ( \mathcal{T}_{E_k (\varepsilon)} \le T_k\right),
\]
where \(\mathcal{F}_{\tau_{k-1}} = \sigma (\boldsymbol{X}_t^\beta \colon t \le \tau_{k-1})\) is the natural filtration. Since \(\boldsymbol{X}_{\tau_{k-1}} \in E_{k-1}(\varepsilon)\) on \(\{\tau_{k-1} < \infty\}\), Lemma~\ref{lem: E_2} yields, for \(N\) large enough and with \(\mathbb{P}\)-probability at least \(1-\exp(-K N)\),
\[
\mathbb{Q}_{\boldsymbol{X}_{\tau_{k-1}}} \left( \mathcal{T}_{E_k(\varepsilon)} \le T_k \right) \ge 1 - \eta_k(N,\varepsilon, \gamma_1,\gamma_2).
\]
In particular,
\[
\mathbb{Q}_{\boldsymbol{X}} \left ( \tau_k - \tau_{k-1} \le T_k\, | \, \mathcal{F}_{\tau_{k-1}} \right ) \ge \mathbb{Q}_{\boldsymbol{X}} \left ( \tau_k - \tau_{k-1} \le T_k, \tau_{k-1} < \infty \, | \, \mathcal{F}_{\tau_{k-1}} \right ) \ge  \mathbf{1}_{\{\tau_{k-1} < \infty\}} (1 - \eta_k(N,\varepsilon, \gamma_1,\gamma_2)).
\]
Letting \(A_k \coloneqq \{\tau_k - \tau_{k-1} \le T_k\}\) and \(B_{k-1} \coloneqq \bigcap_{i=1}^{k-1} A_i\) and taking the conditional expectation with respect to \(B_{k-1}\), we obtain
\[
\begin{split}
\mathbb{Q}_{\boldsymbol{X}} \left ( A_k \, | \,B_{k-1} \right)  & = \E_{\boldsymbol{X}} \left [ \mathbb{Q}_{\boldsymbol{X}} \left ( A_k \, | \, \mathcal{F}_{\tau_{k-1}} \right)  \, | \, B_{k-1}  \right ]\\
& \ge (1 - \eta_k(N,\varepsilon, \gamma_1,\gamma_2)) \E_{\boldsymbol{X}} \left [ \mathbf{1}_{\{\tau_{k-1} < \infty \}} \, | \, B_{k-1}  \right ]\\
& = 1- \eta_k (N, \varepsilon, \gamma_1,\gamma_2),
\end{split}
\]
since \(B_{k-1} \subset \{\tau_{k-1}  < \infty\}\). Since \(B_r \subset \{\tau_r \le T_0\}\), we have
\[
\mathbb{Q}_{\boldsymbol{X}} \left ( \tau_r \le T_0 \right) \ge \mathbb{Q}_{\boldsymbol{X}} \left ( B_r \right)  = \mathbb{Q}_{\boldsymbol{X}} (A_1) \prod_{k=2}^r \mathbb{Q}_{\boldsymbol{X}} \left ( A_k \, | \,B_{k-1} \right) \ge \mathbb{Q}_{\boldsymbol{X}} (\tau_1 \le T_1) \prod_{k=2}^r (1- \eta_k (N,\varepsilon, \gamma_1,\gamma_2)).
\]
Finally, we integrate this inequality with respect to the initialization measure \(\mu_N^0\), restricted to the good initial condition set \(\mathcal{C}_1^{(N)}(\gamma_1,\gamma_2)\):
\[
\begin{split}
& \int_{\mathcal{X}^+} \mathbb{Q}_{\boldsymbol{X}} \left( \tau_r \le T_0 \right) \boldsymbol{1}_{\mathcal{C}_1^{(N)}(\gamma_1,\gamma_2)} (\boldsymbol{X}) \mathrm{d}\mu_N^0(\boldsymbol{X}) \\
&\quad \ge
\left( \prod_{k=2}^r (1 - \eta_k(N,\varepsilon, \gamma_1,\gamma_2)) \right) \int_{\mathcal{X}^+} \mathbb{Q}_{\boldsymbol{X}}\!\left( \tau_1 \le T_1 \right) \boldsymbol{1}_{\mathcal{C}_1^{(N)}(\gamma_1,\gamma_2)} (\boldsymbol{X}) \mathrm{d}\mu_N^0(\boldsymbol{X}) \\
& \quad \ge \left( 1 - \eta_1(N,\varepsilon,\gamma_1,\gamma_2) \right) \prod_{k=2}^r \left( 1 - \eta_k(N,\varepsilon, \gamma_1,\gamma_2) \right),
\end{split}
\]
where the last inequality follows from~\eqref{eq:tau1-bound}. Finally, we have
\[
\begin{split}
\int_{\mathcal{X}^+} \mathbb{Q}_{\boldsymbol{X}} \left( \tau_r \le T_0 \right) d \mu_N^0(\boldsymbol{X}) & \ge \int_{\mathcal{X}^+} \mathbb{Q}_{\boldsymbol{X}} \left( \tau_r \le T_0 \right) \boldsymbol{1}_{\{\mathcal{C}_1^{(N)}(\gamma_1,\gamma_2)\}}(\boldsymbol{X}) \mu_N^0(\boldsymbol{X}) - \mu_N^0 \left( \mathcal{C}_1^{(N)}(\gamma_1,\gamma_2)^{\mathrm{c}} \right) \\
& \ge  \left( 1 - \eta_1(N, \varepsilon,\gamma_1,\gamma_2) \right) \prod_{k=2}^r \left( 1 - \eta_k(N,\varepsilon, \gamma_1,\gamma_2) \right) - \mu_N^0 \left( \mathcal{C}_1^{(N)}(\gamma_1,\gamma_2)^{\mathrm{c}} \right).
\end{split}
\]
We can bound \(\mu_N^0 (\mathcal{C}_1^{(N)}(\gamma_1,\gamma_2)^\textnormal{c})\) according to Definition~\ref{def: condition 1}. Since \(\tau_r = \mathcal{T}_{R(\varepsilon)}\), we obtain the desired result.
\end{proof}

We therefore need to prove the two intermediate results, namely Lemmas~\ref{lem: E_1} and~\ref{lem: E_2}. 
\begin{proof}[\textbf{Proof of Lemma~\ref{lem: E_1}}]
The proof of this lemma is divided into four steps, which we briefly describe. We first show that \(m_{11}\) is the earliest correlation to reach the microscopic threshold \(\varepsilon_N \coloneqq c N^{-\frac{p-2}{2(p-1)}}\) for some constant \(c>0\), while the other correlations remain of order \(\Theta(N^{-\frac{1}{2}})\). As proved in Lemma~\ref{lem: weak implies strong recovery Langevin p>2}, reaching this threshold already guarantees strong recovery of the spike \(\boldsymbol{v}_1\). Next, we see that, once \(m_{11}\) crosses the threshold \(N^{-\frac{p-2}{2p}}\), the correlations \(m_{i1}\) and \(m_{1i}\) for \(i \neq 1\) begin to decay from the order \(\Theta(N^{-\frac{1}{2}})\), and their dynamics are well approximated by \(\dot{m}_{i1} \approx - \lambda_1^2 m_{11}^p m_{i1}\) and \(\dot{m}_{1i} \approx - \lambda_1^2 m_{11}^p m_{1i}\), respectively. In particular, we show that these correlations drop to \(N^{-\frac{p-1}{4}}\log(N)^{-\frac{1}{2}}\) within a microscopic time. Finally, we study the evolution of the remaining correlations \(m_{ij}\) for \(i,j \neq 1\) as \(m_{11}\) crosses \(\varepsilon_N\) which is approximately \(\dot{m}_{ij} \approx \lambda_i \lambda_j m_{ij}^{p-1} - r^2 \lambda_1^2 m_{11}^{p-1} m_{i1} m_{1j}\). We show that although these terms may experience a temporary decrease, the maximal decay is bounded by a constant, so that \(m_{ij}\) remain on the scale \(\Theta(N^{-\frac{1}{2}})\). 

Throughout, for every \(a >0\) and \(i,j \in [r]\), we define the hitting times \(\mathcal{T}_a^{(ij,+)}\) and \(\mathcal{T}_a^{(ij,-)}\) by
\[
\begin{split}
\mathcal{T}_a^{(ij,+)} & = \inf \left \{ t \ge 0 \colon m_{ij}(t) \ge a \right \},\\
\mathcal{T}_a^{(ij,-)} & = \inf \left \{ t \ge 0 \colon m_{ij}(t) \le a \right \}.
\end{split}
\]

\textbf{Step 1: Evolution of the correlations until the first correlation reaches \(\varepsilon_N\).} On the initial event \(\mathcal{C}_1^{(N)}(\gamma_1,\gamma_2)\) given by Definition~\ref{def: condition 1}, for every \(i,j \in [r]\) there exists \(\gamma_{ij} \in (\gamma_2,\gamma_1)\) such that \(m_{ij}(0) = \gamma_{ij}N^{-\frac{1}{2}}\). For some \(T_0^{(ij)} >0\) to be chosen later, we then define the event \(\mathcal{A}^{(ij)}= \mathcal{A}^{(ij)}(\gamma_1, \gamma_2, T_0^{(ij)})\) by
\[
\mathcal{A}^{(ij)}(\gamma_1,\gamma_2,T_0^{(ij)})=  \mathcal{C}_1^{(N)}(\gamma_1,\gamma_2) \cap \left \{ \sup_{t \in [0,T_0^{(ij)}]} |M_t^{m_{ij}}| \leq \frac{\gamma_2}{2\sqrt{N}} \right \},
\]
where we recall that according to~\eqref{eq: Doob inequality}, there exists a constant \(K_2 > 0\) such that 
\[
\sup_{\boldsymbol{X}} \mathbb{Q}_{\boldsymbol{X}} \left ( \sup_{t \in [0,T_0^{(ij)}]} |M_t^{m_{ij}}| \geq \frac{\gamma_2}{2 \sqrt{N}} \right ) \leq K_2 \exp \left (-\frac{\gamma_2^2}{4 K_2 T_0^{(ij)}} \right).
\]
In the following, we fix \(i,j \in [r]\) and place ourselves on the event \(\mathcal{A}^{(ij)}\). Let \(\varepsilon_N = cN^{-\frac{p-2}{2(p-1)}}\) for some constant \(c>0\). Given the generator expansion by Lemma~\ref{lem: evolution equation m_ij}, i.e., 
\[
L_\beta m_{ij} = L_{0,\beta} m_{ij} + \beta \sqrt{M} p \lambda_i \lambda_j m_{ij}^{p-1} - \beta \sqrt{M} \frac{p}{2} \sum_{1 \leq k,\ell \leq r} \lambda_k m_{i \ell}m_{kj} m_{k \ell} (\lambda_j m_{kj}^{p-2} + \lambda_\ell m_{k \ell}^{p-2}),
\]
we see that
\[
- \norm{L_{0,\beta} m_{ij}}_\infty + \beta \sqrt{M} p \lambda_i \lambda_j m_{ij}^{p-1}(t) \leq L_\beta m_{ij}(t) \leq \norm{L_{0,\beta} m_{ij}}_\infty + \beta \sqrt{M}  p \lambda_i \lambda_j m_{ij}^{p-1}(t),
\]
for \(t \leq \min_{1 \leq k, \ell \leq r} \mathcal{T}_{\varepsilon_N}^{(k \ell,+)}\), provided that
\[
N^{p-3} \ge \left ( \frac{r^2 \lambda_1^2 c^{p+1}}{\lambda_r^2 \gamma_2^{p-2}} \right)^{2(p-1)}.
\]
(When \(p=3\), it suffices to choose \(c\) sufficiently small, independently of \(N\).) According to Lemma~\ref{lem: regularity H0} and especially to~\eqref{eq: bound norm L_0m_ij}, there exists a constant \(\Lambda = \Lambda(\beta, p, \{\lambda_i\}_{i=1}^r)\) and \(K>0\) such that \(\norm{L_{0,\beta} m_{ij}}_\infty \leq \Lambda \) with \(\mathbb{P}\)-probability at least \(1 - \exp(-K N)\). We then observe that for \(t \leq \min_{1 \leq k, \ell \leq r} \mathcal{T}_{\varepsilon_N}^{(k \ell,+)}\),
\[
C_0 \beta \sqrt{M} p \lambda_i \lambda_j m_{ij}^{p-1}(t) \geq C \frac{\Lambda}{\gamma_2^{p-1}} N^{\frac{p-1}{2}} m_{ij}^{p-1}(t) \geq \Lambda,
\]
for some constant \(C_0 \in (0,\frac{1}{2})\), where we used the facts that \(\sqrt{M} \geq C \frac{\Lambda}{\beta p \lambda_r^2 C_0 \gamma_2^{p-1}}N^{\frac{p-1}{2}}\) and that \(m_{ij}(t) \geq \gamma_2N^{-\frac{1}{2}}\). On the event \(\mathcal{A}^{(ij)}\), we therefore obtain the integral inequality
\[
\frac{\gamma_{ij}}{2\sqrt{N}} + (1 - C_0) \beta \sqrt{M} p \lambda_i \lambda_j \int_0^t m_{ij}^{p-1}(s) ds \leq m_{ij}(t) \leq \frac{3\gamma_{ij}}{2\sqrt{N}} + (1+C_0) \beta \sqrt{M} p \lambda_i \lambda_j \int_0^t m_{ij}^{p-1}(s) ds,
\]
for all \(t \leq \min_{1 \leq k, \ell \leq r} \mathcal{T}_{\varepsilon_N}^{(k \ell,+)} \wedge T_0^{(ij)}\), with \(\mathbb{P}\)-probability at least \(1 - \exp(-K N)\). Applying Lemma~\ref{lem: Gronwall}, we obtain the comparison inequality 
\[
\ell_{ij}(t) \leq m_{ij}(t) \leq u_{ij}(t),
\]
for all \(t \leq \min_{1 \leq k, \ell \leq r} \mathcal{T}_{\varepsilon_N}^{(k \ell,+)} \wedge T_0^{(ij)}\), where \(\ell_{ij}\) and \(u_{ij}\) are given by
\[
\ell_{ij}(t) = \frac{\gamma_{ij}}{2\sqrt{N}} \left ( 1- (1-C_0) \beta \sqrt{M} p(p-2) \lambda_i \lambda_j \left ( \frac{\gamma_{ij}}{2\sqrt{N}}\right )^{p-2} t \right )^{-\frac{1}{p-2}}
\]
and
\[
u_{ij}(t) = \frac{3 \gamma_{ij}}{2 \sqrt{N}} \left ( 1-  (1+C_0) \beta \sqrt{M} p(p-2) \lambda_i \lambda_j\left ( \frac{3 \gamma_{ij}}{2 \sqrt{N}}\right )^{p-2} t \right )^{-\frac{1}{p-2}},
\]
respectively. We define \(T_{\ell,\varepsilon_N}^{(ij)}\) as the hitting time of \(\varepsilon_N\) for the lower comparison function, i.e., \(\ell_{ij}(T_{\ell,\varepsilon_N}^{(ij)}) =\varepsilon_N \), which yields
\[
T_{\ell,\varepsilon_N}^{(ij)} = \frac{1 - \left ( \gamma_{ij}/2\right )^{p-2} N^{-\frac{p-2}{2(p-1)}}}{ (1-C_0) \beta \sqrt{M} p(p-2) \lambda_i \lambda_j \left (\frac{ \gamma_{ij}}{2 \sqrt{N}}\right )^{p-2}}.
\]
Similarly, let \(T_{u,\varepsilon_N}^{(ij)}\) be defined by \(u_{ij}(T_{u,\varepsilon_N}^{(ij)}) = \varepsilon_N\), that is,
\[ 
T_{u,\varepsilon_N}^{(ij)} = \frac{1 - \left (3\gamma_{ij}/2\right )^{p-2} N^{-\frac{p-2}{2(p-1)}}}{ (1+C_0) \beta \sqrt{M} p(p-2) \lambda_i \lambda_j \left (\frac{3\gamma_{ij}}{2 \sqrt{N}}\right )^{p-2}}.
\]
On the event \(\mathcal{A}^{(ij)}\), the comparison implies \(T_{u,\varepsilon_N}^{(ij)} \leq \mathcal{T}_{\varepsilon_N}^{(ij,+)} \leq T_{\ell,\varepsilon_N}^{(ij)}\). Next, by assumption, for every \(2 \le i \le r\),
\[
\lambda_1 = \lambda_0 \frac{1 + C_0}{1 - C_0}\left(\frac{ 3\gamma_1}{\gamma_2} \right)^{p-2} \lambda_i,
\]
for some constant \(\lambda_0 > 1\). This implies that, for every \((i,j) \neq (1,1)\),
\[
T_{\ell,\varepsilon_N}^{(11)}  \leq \frac{1 - \left ( \frac{\gamma_{11}}{2}\right )^{p-2} N^{-\frac{p-2}{2(p-1)}}}{\lambda_0 (1+C_0) \beta \sqrt{M} p(p-2) \lambda_i \lambda_j \left (\frac{ \gamma_{ij}}{2\sqrt{N}}\right )^{p-2} \left (\frac{ 3 \gamma_1}{\gamma_2}\right )^{p-2}} \leq T_{u,\varepsilon_N}^{(ij)},
\]
provided that \(N > \left ( \frac{\lambda_0}{\lambda_0 - 1}\right )^{\frac{2(p-1)}{p-2}} \left ( \frac{3}{2} \gamma_1\right )^{2(p-1)}\). As a consequence, on the event \(\cap_{1 \leq k, \ell \leq r} \mathcal{A}^{(k \ell)}\), 
\[
\mathcal{T}_{\varepsilon_N}^{(11,+)} = \min_{1 \leq k, \ell \leq r} \mathcal{T}_{\varepsilon_N}^{(k \ell,+)}
\]
with \(\mathbb{P}\)-probability at least \(1-\exp(-KN)\). That is, \(m_{11}\) is the first correlation that reaches the threshold \(\varepsilon_N\). In particular, on this event, 
\[
\mathcal{T}_{\varepsilon_N}^{(11,+)} \leq T_0^{(11)} \leq \frac{2^{p-2} \gamma_2}{C (p-2)\sqrt{N}} ,
\]
with \(\mathbb{P}\)-probability at least \(1 -\exp(-KN)\). Finally, since \(\mathcal{T}_{\varepsilon_N}^{(11,+)} \leq T_{\ell, \varepsilon_N}^{(11)}\) and \(u_{ij}\) is monotone increasing, we may estimate, on the event \(\mathcal{A}^{(11)} \cap \mathcal{A}^{(ij)}\), 
\[
m_{ij}(\mathcal{T}_{\varepsilon_N}^{(11,+)}) \le u_{ij}(\mathcal{T}_{\varepsilon_N}^{(11,+)}) \leq u_{ij}(T_{\ell, \varepsilon_N}^{(11)}).
\]
Substituting \(T_{\ell, \varepsilon_N}^{(11)}\) into the expression for \(u_{ij}\) yields 
\begin{equation} \label{eq: upper bound m_ij at T_epsilon}
\begin{split}
u_{ij}(T_{\ell, \varepsilon_N}^{(11)}) & = \frac{3 \gamma_{ij}}{2 \sqrt{N}} \left ( 1 -  \frac{\lambda_i \lambda_j}{\lambda_1^2} \frac{1+C_0}{1-C_0}\left (\frac{3 \gamma_{ij}}{\gamma_{11}} \right )^{p-2} \left (1 - \left(\gamma_{11} /2\right)^{p-2} N^{-\frac{p-2}{2(p-1)}}\right)  \right )^{-\frac{1}{p-2}}\\
& \leq  \frac{3\gamma_{ij}}{2\sqrt{N}} \left ( \frac{\lambda_0}{\lambda_0-1}\right )^{\frac{1}{p-2}}.
\end{split}
\end{equation}
Therefore, on the event \(\cap_{1 \leq k, \ell \leq r} \mathcal{A}^{(k \ell)}\), we conclude that for every \((i,j) \neq (1,1)\),
\[
m_{ij}(\mathcal{T}_{\varepsilon_N}^{(11,+)}) = \gamma_{ij}' N^{-\frac{1}{2}},
\]
for some constant \(\gamma_{ij}'>0\), with \(\mathbb{P}\)-probability at least \(1-\exp(-KN)\). \newline

\textbf{Step 2: Recovery of the first spike.} Next, we study the evolution of \(m_{11}(t)\) for \(t \geq \mathcal{T}_{\varepsilon_N}^{(11,+)}\). Fix \(\varepsilon \in (0,1)\) and set \(\gamma_N \coloneqq \gamma N^{- \frac{1}{2}}\) for some sufficiently large constant \(\gamma >0\) independent of \(N\). We introduce the stopping time
\[
\widehat{\mathcal{T}} \coloneqq \mathcal{T}_\varepsilon^{(11,+)} \wedge \mathcal{T}_{\varepsilon_N/2}^{(11,-)} \wedge \min_{(k,\ell) \neq (1,1)} \mathcal{T}_{\gamma_N}^{(k \ell,+)}.
\]
By the generator expansion given in Lemma~\ref{lem: evolution equation m_ij}, for every \(\mathcal{T}_{\varepsilon_N}^{(11,+)} \le t \le \widehat{\mathcal{T}} \), we have  
\[
\begin{split}
L_\beta m_{11}(t) & \geq - \norm{L_{0,\beta} m_{11}}_\infty + \beta \sqrt{M} p \lambda_1^2 m_{11}^{p-1}(t) (1 - m_{11}^2(t) -r^2 m_{12}(t)) \\
& \geq - \norm{L_{0,\beta} m_{11}}_\infty + (1 - \varepsilon^2 - r^2 \gamma_N^2) \beta \sqrt{M} p \lambda_1^2 m_{11}^{p-1}(t) \\
& \ge - \norm{L_{0,\beta} m_{11}}_\infty + \left (1 - \varepsilon^2 - \frac{C_0}{2} \right) \beta \sqrt{M} p \lambda_1^2 m_{11}^{p-1}(t) ,
\end{split}
\]
for some constant \(0 < C_0 < 2(1-\varepsilon^2)\), provided \(N \ge \frac{2r^2 \gamma^2}{C_0}\). Moreover, with \(\mathbb{P}\)-probability at least \(1 - \exp(-K N)\), we have \(\norm{L_{0,\beta} m_{11}}_\infty \leq \Lambda\), so that
\[
\norm{L_{0,\beta} m_{11}}_\infty \leq \Lambda \le \frac{C_0}{2} \beta \sqrt{M} p \lambda_1^2 m_{11}^{p-1}(t) ,
\]
where we used that \(\sqrt{M} \geq C \frac{\Lambda}{\beta p \lambda_r^2 C_0 \gamma_2^{p-1}}N^{\frac{p-1}{2}}\) for some constant \(C>0\). It therefore follows that \(L_\beta m_{11}\) is bounded below by
\[
L_\beta m_{11}(t) \geq (1- \varepsilon^2 -C_0)  \beta \sqrt{M} p \lambda_1^2 m_{11}^{p-1}(t) >0,
\]
for \(\mathcal{T}_{\varepsilon_N}^{(11,+)} \le t \le \widehat{\mathcal{T}}\), with \(\mathbb{P}\)-probability at least \(1-\exp(-KN)\). Applying Lemma~\ref{lem: Gronwall} on \([\mathcal{T}_{\varepsilon_N}^{(11,+)},  \widehat{\mathcal{T}}]\) gives
\[
m_{11}(t) \geq \frac{\varepsilon_N}{2} \left ( 1- (1- \varepsilon^2-C_0) \beta \sqrt{M} p (p-2) \lambda_1^2 \left( \frac{\varepsilon_N}{2}\right )^{p-2} (t-\mathcal{T}_{\varepsilon_N}^{(11,+)}) \right )^{-\frac{1}{p-2}}
\]
for every \(\mathcal{T}_{\varepsilon_N}^{(11,+)} \le t \le  \widehat{\mathcal{T}}\), with \(\mathbb{Q}_{\boldsymbol{X}}\)-probability at least \(1 - K_1 \exp (-N^{\frac{1}{p-1}}/(4K_1  \widehat{\mathcal{T}}))\) and with \(\mathbb{P}\)-probability at least \(1-\exp(-KN)\). In particular, the lower bound is monotone increasing and equals \(\varepsilon_N/2\) at \(t = \mathcal{T}_{\varepsilon_N}^{(11,+)}\). Hence, \(m_{11}(t)\) cannot return below \(\varepsilon_N/2\) before reaching \(\varepsilon\). It follows that \(\mathcal{T}_{\varepsilon_N/2}^{(11,-)} > \mathcal{T}_\varepsilon^{(11,+)}\) and \(\widehat{\mathcal{T}}=\mathcal{T}_\varepsilon^{(11,+)} \wedge \min_{(k,\ell) \neq (1,1)} \mathcal{T}_{\gamma_N}^{(k \ell,+)}\). A straightforward computation then yields
\begin{equation} \label{eq: time difference}
 \widehat{\mathcal{T}} - \mathcal{T}_{\varepsilon_N}^{(11,+)} \leq \frac{1 - \left ( \frac{1}{ 2 \varepsilon N^{\frac{p-2}{2(p-1)}}}\right )^{p-2} }{(1-\varepsilon^2-C_0) \beta \sqrt{M} p(p-2) \lambda_1^2 \left ( \frac{N^{-\frac{p-2}{2(p-1)}}}{2}\right )^{p-2}} \le \frac{2^{p-2} C_0 \gamma_2^{p-1}}{C (p-2) (1-\varepsilon^2-C_0)N^{\frac{2p-3}{2(p-1)}}} ,
\end{equation}
with \(\mathbb{Q}_{\boldsymbol{X}}\)-probability at least \(1 - K_1 \exp (-N^{\frac{2p-1}{2(p-1)}}/(4K_1) )\) and with \(\mathbb{P}\)-probability at least \(1-\exp(-KN)\), where the second inequality in~\eqref{eq: time difference} follows from the assumption on \(\sqrt{M}\).  

Define 
\[
\mathcal{E}_N \coloneqq \left \{\mathcal{T}_\varepsilon^{(11,+)} \le  \min_{(k,\ell) \ne (1,1)} \mathcal{T}_{\gamma_N}^{(k \ell,+)} \right \}.
\]
Then, on \(\mathcal{E}_N\), we have \( \widehat{\mathcal{T}} = \mathcal{T}_\varepsilon^{(11,+)}\), and thus the first spike is weakly recovered at level \(\varepsilon\) before any competing correlation reaches the scale \(\gamma_N\). Combining the estimate for \(\mathcal{T}_{\varepsilon_N}^{(11,+)}\) obtained in Step~1 with~\eqref{eq: time difference}, we obtain that, on \(\mathcal{E}_N \cap\cap_{1 \leq k, \ell \leq r} \mathcal{A}^{(k \ell)} \),
\[
\mathcal{T}_\varepsilon^{(11,+)} = \mathcal{T}_{\varepsilon_N}^{(11,+)} + \left ( \mathcal{T}_\varepsilon^{(11,+)} -\mathcal{T}_{\varepsilon_N}^{(11,+)} \right) \le C_\varepsilon \frac{2^{p-2} \gamma_2}{C(p-2) \sqrt{N}},
\]
for all \(N\) sufficiently large, where \(C_\varepsilon >0\) is a constant independent of \(N\). This holds with \(\mathbb{Q}_{\boldsymbol{X}}\)-probability at least \(1 - K_1 \exp (-N^{\frac{2p-1}{2(p-1)}}/(4K_1) )\) and with \(\mathbb{P}\)-probability at least \(1-\exp(-KN)\). In Steps 3 and 4 we prove that \(\mathcal{E}_N \to 1\) in probability to \(1\) as \(N \to \infty\). Moreover, we show that after time \(\mathcal{T}_\varepsilon^{(11,+)}\) the mixed correlations \(m_{1i}(t)\) and \(m_{i1}(t)\) for \(i \neq 1\) decay below the prescribed threshold, while the remaining correlations \(m_{ij}(t)\) with \(i,j\neq 1\) remain of order \(N^{-1/2}\). In Step 5, we further prove that, once \(m_{11}\) reaches \(\varepsilon\), it reaches \(1-\varepsilon\), and that the corresponding recovery event \(E_1(\varepsilon)\) is attained within time \(\mathcal{O}(N^{-1/2})\) with high probability.
\newline

\textbf{Step 3: Evolution of \(m_{i1}(t)\) and \(m_{1i}(t)\) for \(i \neq 1\) for \(t \ge \mathcal{T}_{\varepsilon_N}^{(11,+)}\).} 
From Step 1, as \(m_{11}\) crosses the threshold \(\varepsilon_N = N^{-\frac{p-2}{2(p-1)}}\), on the event \(\cap_{1 \leq k, \ell \leq r} \mathcal{A}^{(k \ell)}\), we have \(m_{ij}(\mathcal{T}_{\varepsilon_N}^{(11,+)}) = \gamma_{ij}' N^{-\frac{1}{2}}\) for some constant \(\gamma_{ij}'>0\), with \(\mathbb{P}\)-probability at least \(1-\exp(-KN)\). According to the generator expansion given by Lemma~\ref{lem: evolution equation m_ij}, for \(i \neq 1\) we obtain the crude upper bounds
\[
\begin{split}
L_\beta m_{i1}(t) & \leq \norm{L_{0,\beta} m_{i1}}_\infty + \beta \sqrt{M} p \lambda_1 \lambda_i m_{i1}^{p-1}(t),\\
L_\beta m_{1i}(t) & \leq \norm{L_{0,\beta} m_{1i}}_\infty+ \beta \sqrt{M} p \lambda_1 \lambda_i m_{1i}^{p-1}(t).
\end{split}
\]
Since \(\norm{L_{0,\beta} m_{i1}}_\infty \le \Lambda\) with \(\mathbb{P}\)-probability at least \(1-\exp(-KN)\), we have
\[
\begin{split}
L_\beta m_{i1}(t) & \le \Lambda +  \beta \sqrt{M} p \lambda_1 \lambda_i m_{i1}^{p-1}(t),\\
L_\beta m_{1i}(t) & \le \Lambda +  \beta \sqrt{M} p \lambda_1 \lambda_i m_{1i}^{p-1}(t).
\end{split}
\]
with \(\mathbb{P}\)-probability at least \(1-\exp(-KN)\). On the event \(\cap_{1 \leq k, \ell \leq r} \mathcal{A}^{(k \ell)}\), it follows for \(t \in [\mathcal T_{\varepsilon_N}^{(11,+)}, \widehat{\mathcal T}]\) that
\[
\begin{split}
m_{1i}(t) & \le \frac{3\gamma_{i1}'}{2 \sqrt{N}} + \Lambda (t - \mathcal T_{\varepsilon_N}^{(11,+)}) +  \beta \sqrt{M} p \lambda_1 \lambda_i \int_{\mathcal T_{\varepsilon_N}^{(11,+)}}^t m_{1i}^{p-1}(s) ds\\
& \le \frac{3\gamma_{i1}'}{2 \sqrt{N}} + \Lambda (\widehat{\mathcal{T}} - \mathcal T_{\varepsilon_N}^{(11,+)}) +  \beta \sqrt{M} p \lambda_1 \lambda_i \int_{\mathcal T_{\varepsilon_N}^{(11,+)}}^t m_{1i}^{p-1}(s) ds\\
& \le \frac{C_1}{\sqrt{N}}+  \beta \sqrt{M} p \lambda_1 \lambda_i \int_{\mathcal T_{\varepsilon_N}^{(11,+)}}^t m_{1i}^{p-1}(s) ds,
\end{split}
\]
where the last inequality follows from~\eqref{eq: time difference} and \(C_1 >0\) is a constant independent of \(N\), with \(\mathbb{P}\)-probability at least \(1-\exp(-KN)\). Applying Lemma~\ref{lem: Gronwall} and repeating the comparison argument from Step~1, we obtain, on the event \(\cap_{1 \leq k, \ell \leq r} \mathcal{A}^{(k \ell)}\),
\[
\sup_{\mathcal{T}_{\varepsilon_N}^{(11,+)}   \le t \le \widehat{\mathcal{T}}} \{ m_{i1}(t),m_{1i}(t)\} \le C_1 N^{-1/2},
\]
for some constant \(C_1>0\), with \(\mathbb{P}\)-probability at least \(1-\exp(-KN)\). Choosing \(\gamma> C_1\) in \(\gamma_N = \gamma N^{-1/2}\), we have
\[
m_{i1}(t),m_{1i}(t) < \gamma_N,
\]
for all \( t\in[\mathcal{T}_{\varepsilon_N}^{(11,+)},\mathcal{T}_\varepsilon^{(11,+)} \wedge \min_{(k,\ell) \neq (1,1)} \mathcal{T}_{\gamma_N}^{(k\ell,+)}]\), and by continuity this implies
\[
\mathcal T_{\gamma_N}^{(i1,+)} > \mathcal T_\varepsilon^{(11,+)},
\qquad
\mathcal T_{\gamma_N}^{(1i,+)} > \mathcal T_\varepsilon^{(11,+)},
\quad \forall\,i\neq 1,
\]
with \(\mathbb{P}\)-probability at least \(1-\exp(-KN)\).

We now study the evolution of \(m_{i1}\) and \(m_{1i}\) as \(t \ge \mathcal{T}_\varepsilon^{(11,+)}\). We note by Lemma~\ref{lem: evolution equation m_ij} that actually, as \(m_{11}\) exceeds \(N^{-\frac{p-2}{2p}}\), there exists a constant \(c_1>0\) such that
\[
\begin{split}
L_\beta m_{i1}(t) & \leq \norm{L_{0,\beta} m_{i1}}_\infty +  \beta \sqrt{M} p \lambda_1 \lambda_i m_{i1}^{p-1}(t) - \beta \sqrt{M} p \lambda_1^2 m_{11}^p(t) m_{i1}(t) \\
& \leq \norm{L_{0,\beta} m_{i1}}_\infty  - c_1 \beta \sqrt{M} p \lambda_1^2 m_{11}^p(t) m_{i1}(t) .
\end{split}
\] 
The same holds for the correlations \(m_{1i}\) for \(i \neq 1\). We therefore introduce the stopping time \(\mathcal{T}_{\textnormal{low}}\) given by 
\begin{equation} \label{eq: T lower}
\mathcal{T}_{\textnormal{low}} = \inf \left \{t \ge \mathcal{T}_{\varepsilon_N}^{(11,+)} \colon \max \left \{ \max_{i \neq 1} m_{i1}(t),  \max_{i \neq 1} m_{1i}(t) \right \} \lesssim \log(N)^{-1/2} N^{-\frac{p-1}{4}} \right \}.
\end{equation}
Note that \(\mathcal{T}_{\textnormal{low}}  >  \mathcal{T}_{\varepsilon_N}^{(11,+)}\). We then note that if \(\mathcal{T}_{N^{-\frac{p-2}{2p}}}^{(11,+)}  \leq \mathcal{T}_{\textnormal{low}} \leq \mathcal{T}_{\varepsilon}^{(11,+)}\), then there is nothing to prove since we have already shown in Step 2 that \(\mathcal{T}_{\varepsilon}^{(11,+)} \lesssim N^{-\frac{1}{2}}\) with high \(\mathbb{Q}_{\boldsymbol{X}}\)- and \(\mathbb{P}\)-probability. It therefore remains to consider the case \(\mathcal{T}_{\varepsilon}^{(11,+)} < \mathcal{T}_{\textnormal{low}}\). To this end, fix \(i \neq 1\) and look at \(L_\beta m_{1i}\) which is bounded for \(\mathcal{T}_\varepsilon^{(11,+)} \leq t \leq \mathcal{T}_{\textnormal{low}}\) by
\[
L_\beta m_{i1}(t) \leq \Lambda - c_1 \beta \sqrt{M} p \lambda_1^2 m_{11}^p(t) m_{i1}(t) \leq - \frac{c_1}{2} \beta \sqrt{M} p\lambda_1^2 \varepsilon^p m_{i1}(t),
\] 
with \(\mathbb{P}\)-probability at least \(1 - \exp(-K N)\) provided \(\sqrt{M} \geq \frac{2 \Lambda \sqrt{\log(N)} N^{\frac{p-1}{4}}}{c \beta p \lambda_1^2 \varepsilon^p}\), which certainly holds by assumption. By item (d) of Lemma~\ref{lem: Gronwall}, it then follows that, on the event \(\cap_{1 \le k, \ell \le r} \mathcal{A}^{(k \ell)}\),
\[
m_{i1}(t) \leq \frac{3C_1}{2\sqrt{N}} - \frac{c_1}{2}\beta \sqrt{M} p\lambda_1^2 \varepsilon^p \int_{\mathcal{T}_\varepsilon^{(11)}}^t m_{i1}(s) ds  \leq \frac{3C_1}{2 \sqrt{N}}  \exp \left(- \frac{c_1}{2} \beta \sqrt{M}  p \lambda_1^2  \varepsilon^p  (t-\mathcal{T}_\varepsilon^{(11)}) \right),
\]
for \(\mathcal{T}_\varepsilon^{(11,+)}  \leq t \leq \mathcal{T}_{\textnormal{low}}\), with \(\mathbb{P}\)-probability at least \(1-\exp(-KN)\). Since \(m_{i1}(t)\) is upper bounded by a decreasing function for every \(\mathcal{T}_\varepsilon^{(11,+)}  \leq t \leq \mathcal{T}_{\textnormal{low}}\), it then follows that \(m_{i1}(t)\) is decreasing for \(t \ge \mathcal{T}_\varepsilon^{(11,+)}\) and will never reach \(\gamma_N\). In particular, we have
\begin{equation} \label{eq: delta T}
\mathcal{T}_{\textnormal{low}} - \mathcal{T}_\varepsilon^{(11,+)} \leq \frac{2}{c_1 \beta \sqrt{M}p\lambda_1^2  \varepsilon^p} \log( \frac{3C_1}{2}\sqrt{\log(N)} N^{\frac{p-3}{4}}) \lesssim \log(N) N^{-\frac{p-1}{2}},
\end{equation}
where \(\lesssim\) hides a constant depending on \(p, \varepsilon\). This ensures that, on \(\mathcal{E}_N \cap\cap_{1 \leq k, \ell \leq r} \mathcal{A}^{(k \ell)} \),
\[
\mathcal{T}_{\textnormal{low}} =\mathcal{T}_\varepsilon^{(11,+)} + (\mathcal{T}_{\textnormal{low}} - \mathcal{T}_\varepsilon^{(11,+)})\lesssim  \frac{\gamma_2}{\sqrt{N}},
\]
for all \(N\) sufficiently large, with \(\mathbb{Q}_{\boldsymbol{X}}\)-probability at least \(1 - K_1 \exp (-N^{\frac{2p-1}{2(p-1)}}/(4K_1) )\) and with \(\mathbb{P}\)-probability at least \(1-\exp(-KN)\).
\newline 

\textbf{Step 4: Evolution of \(m_{ij}(t)\) for \(i, j \neq 1\) as \(t \geq \mathcal{T}_{\varepsilon_N}^{(11,+)}\).} It remains to control the correlations \(m_{ij}(t)\) for \(i, j \neq 1\) as \(t \geq \mathcal{T}_{\varepsilon_N}^{(11)}\). We proceed in two stages: first up to \(\widehat{\mathcal{T}}\), and then after \(m_{11}\) reaches the macroscopic threshold \(\varepsilon\).

As in Step 3, for \(\mathcal{T}_{\varepsilon_N}^{(11,+)} \le t \le \widehat{\mathcal{T}}\), Lemma \ref{lem: evolution equation m_ij} yields the crude upper bound
\[
L_\beta m_{ij}(t) \le \norm{L_{0,\beta}m_{ij}}_\infty + \beta \sqrt{M} p \lambda_i \lambda_j m_{ij}^{p-1}(t).
\]
On the event \(\cap_{1 \le k, \ell \le r} \mathcal{A}^{(k \ell)}\), we therefore have
\[
m_{ij}(t) \le \frac{C_2}{\sqrt{N}} + \beta \sqrt{M} p \lambda_i \lambda_j \int_{\mathcal{T}_{\varepsilon_N}^{(11,+)}}^t m_{ij}^{p-1}(s) ds,
\]
with \(\mathbb{P}\)-probability at least \(1- \exp(-KN)\), where \(C_2>0\) is independent of \(N\). By Lemma \ref{lem: Gronwall} (comparison argument as in Step 1), it follows that
\[
\sup_{\mathcal{T}_{\varepsilon_N}^{(11,+)} \le t \le \widehat{\mathcal{T}}} m_{ij}(t) < C_2 N^{-1/2}.
\]
Choosing \(\gamma>C_2\) in \(\gamma_N = \gamma N^{-1/2}\), we conclude that  
\[
\mathcal{T}_{\gamma_N}^{(ij,+)} > \mathcal{T}_\varepsilon^{(11,+)}, \quad \forall \, i,j \neq 1,
\]
with \(\mathbb{P}\)-probability at least \(1- \exp(-KN)\). In particular we deduce that, on the event \(\cap_{1 \le k, \ell \le r} \mathcal{A}^{(k \ell)}\),
\[
\min_{(k,\ell) \neq (1,1)} \mathcal{T}_{\gamma_N}^{(k \ell,+)} > \mathcal{T}_\varepsilon^{(11,+)},
\]
with \(\mathbb{P}\)-probability at least \(1- \exp(-KN)\). 

We now fix \(i,j \neq 1\) and study the evolution of \(m_{ij}\) for \(t \geq\mathcal{T}_\varepsilon^{(11,+)}\). According to Lemma~\ref{lem: evolution equation m_ij} the generator \(L_\beta m_{ij}\) is bounded above by
\[
L_\beta m_{ij}(t) \le \lVert L_{0,\beta} m_{ij} \rVert_\infty + \beta \sqrt{M} p \lambda_i \lambda_j m_{ij}^{p-1}(t) - \frac{1}{2} \beta \sqrt{M} p \lambda_1^2 m_{11}^{p-1}(t) m_{i1}(t) m_{1i}(t).
\]
We note that actually as \(m_{11}(t) \ge N^{-\frac{p-3}{2(p-1)}}\), the term \(\lambda_1^2 m_{11}^{p-1}(t) m_{i1}(t) m_{1j}(t)\) may be larger than \(\lambda_i \lambda_j m_{ij}^{p-1}(t)\) and may lead to a decrease in \(m_{ij}\). Recalling the stopping time \(\mathcal{T}_{\textnormal{low}}\) introduced by~\eqref{eq: T lower}, we see that if \(\mathcal{T}_{N^{-\frac{p-2}{2p}}}^{(11,+)} \leq \mathcal{T}_{\textnormal{low}} \leq \mathcal{T}_{N^{-\frac{p-3}{2(p-1)}}}^{(11,+)}\), then we still have the estimate
\begin{equation}  \label{eq: bound generator increase Lm_{22}}
-\norm{L_{0,\beta} m_{ij}}_\infty + \beta \sqrt{M} p \lambda_i \lambda_j m_{ij}^{p-1}(t) \leq L_\beta m_{ij}(t) \leq \norm{L_{0,\beta} m_{ij}}_\infty + \beta \sqrt{M} p \lambda_i \lambda_j m_{ij}^{p-1}(t),
\end{equation}
for every \(\mathcal{T}_{N^{-\frac{p-3}{2(p-1)}}}^{(11,+)} \leq t \leq \mathcal{T}_\varepsilon^{(11,+)} \wedge \mathcal{T}_{\gamma_N}^{(ij,+)}\), and the evolution \(m_{ij}\) keeps increasing but, as seen above, it will stay on the order \(N^{-1/2}\). Otherwise, if \(\mathcal{T}_{\textnormal{low}} > \mathcal{T}_{N^{-\frac{p-3}{2(p-1)}}}^{(11,+)}\), then \(L_\beta m_{ij}(t)\) is bounded according to
\begin{equation}  \label{eq: bound generator decrease Lm_{ij}}
\begin{split}
L_\beta m_{ij}(t) & \geq - \norm{L_{0,\beta} m_{ij}}_\infty - r^2 \beta \sqrt{M} p \lambda_1^2 m_{11}^{p-1}(t) m_{i1}^{p-1}(t) m_{1j}^{p-1}(t) \\
L_\beta m_{ij}(t) &\leq \norm{L_{0,\beta} m_{ij}}_\infty - \frac{1}{2}\beta \sqrt{M} p \lambda_1^2 m_{11}^{p-1}(t) m_{i1}^{p-1}(t) m_{1j}^{p-1}(t),
\end{split}
\end{equation}
for \(\mathcal{T}_{N^{-\frac{p-3}{2(p-1)}}}^{(11,+)} \leq t \leq \mathcal{T}_{\textnormal{low}} \wedge \mathcal{T}_\varepsilon^{(11,+)} \wedge  \mathcal{T}_{\gamma_N}^{(ij,+)}\), thus the evolution of \(m_{ij}\) on this time interval is decreasing. We therefore need to quantify the decrease of \(m_{ij}\) between \(\mathcal{T}_{N^{-\frac{p-3}{2(p-1)}}}^{(11,+)}\) and \(\mathcal{T}_{\textnormal{low}}\). We may assume without loss of generality that \(\mathcal{T}_\varepsilon^{(11,+)} < \mathcal{T}_{\textnormal{low}}\) in order to have the maximal decrease of \(m_{ij}\). It is then sufficient to show that the decrease of \(m_{ij}\) between \(\mathcal{T}_{N^{-\frac{p-3}{2(p-1)}}}^{(11,+)}\) and \(\mathcal{T}_\varepsilon^{(11,+)}\) is by a constant, thus ensuring that as \(m_{11}\) exceeds \(\varepsilon\) the correlations \(m_{ij}\) are still on a scale \(\Theta(N^{-\frac{1}{2}})\). We claim that we have shown that \(m_{ij}(\mathcal{T}_\varepsilon^{(11,+)}) = \gamma_{ij}'' N^{-\frac{1}{2}}\) for some constant \(\gamma_{ij}'' >0\). Then, we see from~\eqref{eq: bound generator decrease Lm_{ij}} that
\[
L_\beta m_{ij}(t) \geq - \norm{L_{0,\beta} m_{ij}}_\infty - r^2 \beta \sqrt{M} p \lambda_1^2 m_{11}^{p-1}(t) m_{i1}(t) m_{1j}(t) \geq - c_2 r^2 \beta \sqrt{M} p \lambda_1^2 (1-\varepsilon)^{p-1} N^{-1},
\]
for every \(\mathcal{T}_\varepsilon^{(11,+)} \leq t \leq \mathcal{T}_{\textnormal{low}} \wedge \mathcal{T}_{1-\varepsilon}^{(11,+)}\), with \(\mathbb{P}\)-probability at least \(1 - \exp(-K N)\). We therefore obtain that
\[
m_{ij}(t) \geq m_{ij}(\mathcal{T}_\varepsilon^{(11,+)}) + M_t^{m_{ij}} - c_2 r^2 \beta \sqrt{M} p \lambda_1^2 (1-\varepsilon)^{p-1} N^{-1} (t - \mathcal{T}_\varepsilon^{(11,+)})
\]
for \(\mathcal{T}_\varepsilon^{(11,+)} \leq t \leq \mathcal{T}_{\textnormal{low}} \wedge \mathcal{T}_{1-\varepsilon}^{(11,+)}\), yielding 
\[
\begin{split}
m_{ij}(\mathcal{T}_{\textnormal{low}}) & \geq \frac{\gamma_{ij}''}{2 \sqrt{N}} - c_2 r^2 \beta \sqrt{M} p \lambda_1^2 (1-\varepsilon)^{p-1} N^{-1} \left ( \mathcal{T}_{\textnormal{low}} - \mathcal{T}_\varepsilon^{(11)}\right) \\
&\geq \frac{\gamma_{ij}''}{2 \sqrt{N}}- c_3 (1-\varepsilon)^{p-1} \frac{\log(N)}{\sqrt{N} \sqrt{N}} \approx \frac{C_3}{\sqrt{N}} ,
\end{split}
\]
on the event \(\cap_{1 \le k, \ell \le r}\mathcal{A}^{(k \ell)}\), with \(\mathbb{P}\)-probability at least \(1 - \exp(-K  N)\), where we used the bound given by~\eqref{eq: delta T}.

It therefore remains to prove the claim that \(m_{ij} (\mathcal{T}_\varepsilon^{(11,+)}) =\gamma_{ij}'' N^{-\frac{1}{2}}\). We first note that there is nothing to prove for \(p=3\) since \(m_{ij}\) may decrease once \(m_{11}\) is macroscopic since \(m_{ij}\) start decreasing as \(m_{11}\) exceeds \(N^{-\frac{p-3}{2(p-1)}} \approx \varepsilon\). We therefore consider \(p \geq 4\). In particular, we observe that for every \(\nu \in (0, \frac{p-3}{2(p-1)}]\) and every \(\delta < \nu\), the time difference \(\mathcal{T}_{N^{-\nu + \delta}}^{(11,+)} - \mathcal{T}_{N^{-\nu}}^{(11,+)}\) is bounded above by
\[
\begin{split}
\mathcal{T}_{N^{-\nu + \delta}}^{(11,+)} - \mathcal{T}_{N^{-\nu}}^{(11,+)} & \leq 2 \left ( T_{\ell ,N^{-\nu+\delta}}^{(11,+)} - T_{\ell,N^{-\nu}}^{(11,+)} \right )= \frac{2}{ (1-C_0) \beta \sqrt{M} p (p-2) \lambda_1^2 } N^{\nu(p-2)} \left (1 - N^{-\delta(p-2)}\right),
\end{split}
\]
where we used~\eqref{eq: time difference}. We will use the above estimate to provide a bound on the correlation \(m_{ij}(t)\) for \(\mathcal{T}_{N^{-\nu}}^{(11,+)} \le t \le \mathcal{T}_{N^{-\nu + \delta}}^{(11,+)}\). According to~\eqref{eq: bound generator decrease Lm_{ij}}, for every \(\mathcal{T}_{N^{-\nu}}^{(11,+)} \leq t \leq \mathcal{T}_{N^{-\nu+\delta}}^{(11,+)}\), we bound \(L_\beta m_{ij}\) below by
\[
\begin{split}
L_\beta m_{ij}(t) & \geq - \norm{L_{0,\beta} m_{ij}}_\infty - r^2 \beta \sqrt{M} p \lambda_1^2 m_{11}^{p-1}(t) m_{i1}(t) m_{1j}(t)\\
& \geq - \Lambda - r^2 \beta \sqrt{M}p  \lambda_1^2 N^{- (\nu-\delta) (p-1) -1}\\
& \geq -2 r^2 \beta \sqrt{M}p  \lambda_1^2 N^{- (\nu-\delta) (p-1) -1},
\end{split}
\]
with \(\mathbb{P}\)-probability at least \(1-\exp(-KN)\), provided \(\sqrt{M} \geq \frac{\Lambda}{\beta p r^2\lambda_1^2} N^{(\nu-\delta) (p-1)+1}\) which certainly holds since by assumption \((\nu-\delta) (p-1)+1 \leq \frac{p-1}{2}\). We therefore obtain that
\[
m_{ij}(t) \geq m_{ij}(\mathcal{T}_{N^{-\nu}}^{(11,+)}) + M_t^{m_{ij}} - r^2 \beta \sqrt{M} p \lambda_1^2 N^{- (\nu-\delta) (p-1) -1} \left (t - \mathcal{T}_{N^{-\nu}}^{(11)} \right ),
\]
for every \(\mathcal{T}_{N^{-\nu}}^{(11,+)} \leq t \leq \mathcal{T}_{N^{-\nu+\delta}}^{(11,+)}\), yielding 
\begin{equation} \label{eq: maximal decrease of m_{22}}
\begin{split}
m_{ij}( \mathcal{T}_{N^{-\nu+\delta}}^{(11,+)} ) & \geq \frac{1}{2} m_{ij}(\mathcal{T}_{N^{-\nu}}^{(11,+)})  - r^2 \beta \sqrt{M} p \lambda_1^2 N^{- (\nu-\delta) (p-1) -1} \left ( \mathcal{T}_{N^{-\nu+\delta}}^{(11,+)} - \mathcal{T}_{N^{-\nu}}^{(11,+)}\right )\\
& \geq \frac{1}{2} m_{ij}(\mathcal{T}_{N^{-\nu}}^{(11,+)})-C \frac{N^{\delta(p-1)} -N^\delta}{N N^\nu}.
\end{split}
\end{equation}
We see that if \(\nu = \frac{p-3}{2(p-1)}\) and \(\delta < \frac{\nu}{p-1}\), then 
\[
m_{ij}(\mathcal{T}_{N^{-\nu+\delta}}^{(11,+)}) \geq \frac{\gamma_{ij}'}{2\sqrt{N}} - \frac{C}{\sqrt{N}} \frac{N^{\delta(p-1)}-N^\delta}{N^{\nu + \frac{1}{2}}} \geq \frac{c}{\sqrt{N}}.
\]
In particular, we can divide the interval \([\mathcal{T}_{N^{-\frac{(p-3)}{2(p-1)}}}^{(11,+)}, \mathcal{T}_{\varepsilon}^{(11,+)}]\) into a finite number \(n(\delta)\) of small intervals and by iterating the argument above until \(N^{-\nu + \delta} \approx \varepsilon\) we can show that at each step the decrease of \(m_{ij}\) is at most by a constant. This ensures that at \(t = \mathcal{T}_\varepsilon^{(11,+)}\) the correlation \(m_{ij}(t)  \in \Theta(N^{-\frac{1}{2}})\). This completes Step 4. 

As a consequence, on the high-probability event from Steps 3-4, we have 
\[
\mathcal{T}_{\varepsilon}^{(11,+)} \le \min_{(k,\ell) \neq (1,1)} \mathcal{T}_{\gamma_N}^{(k\ell,+)},
\]
so that \(\mathcal{E}_N \to 1\) in probability as \(N \to \infty\).\newline

\textbf{Step 5: Strong recovery of the first spike.} 
We now show strong recovery of \(\boldsymbol{v}_1\). We note from Lemma~\ref{lem: evolution equation m_ij} that \(L_\beta m_{11}\) satisfies
\[
L_\beta m_{11}(t) \geq - \norm{L_{0,\beta}m_{11}}_\infty + 2^{-p+1} \beta \sqrt{M} p \lambda_1^2 \varepsilon^{p-1} (2\varepsilon-\varepsilon^2),
\]
for all \(\mathcal{T}_\varepsilon^{(11,+)} \leq t \leq \mathcal{T}_{\varepsilon/2}^{(11,-)} \wedge \mathcal{T}_{1-\varepsilon}^{(11,+)}\). It therefore follows that 
\[
m_{11}(t) \geq \frac{\varepsilon}{2} + \frac{c}{2^{p-1}} \beta \sqrt{M} p \lambda_1^2 \varepsilon^{p-1} (2\varepsilon-\varepsilon^2) (t - \mathcal{T}_\varepsilon^{(11)}),
\]
for every \(\mathcal{T}_\varepsilon^{(11,+)} \leq t \leq \mathcal{T}_{\varepsilon/2}^{(11,-)} \wedge \mathcal{T}_{1-\varepsilon}^{(11,+)}\), with \(\mathbb{Q}_{\boldsymbol{X}}\)-probability at least \(1 - K_2 \exp (- \varepsilon^2 N/(4K_2( \mathcal{T}_{\varepsilon/2}^{(11,-)} \wedge \mathcal{T}_{1-\varepsilon}^{(11,+)})))\) and with \(\mathbb{P}\)-probability at least \(1-\exp(-KN)\). As a consequence, we have \(\mathcal{T}_{1 - \varepsilon}^{(11,+)} < \mathcal{T}_{ \varepsilon/2}^{(11,-)}\) and we find that 
\[
\mathcal{T}_{1 - \varepsilon}^{(11,+)} - \mathcal{T}_\varepsilon^{(11,+)} \leq \frac{(1- 3\varepsilon/2) 2^{p-1}}{c \beta \sqrt{M} p \lambda_1^2 \varepsilon^{p-1}} \lesssim \frac{1}{N^{\frac{p-1}{2}}} < \frac{1}{\sqrt{N}},
\]
with \(\mathbb{Q}_{\boldsymbol{X}}\)-probability at least \(1 - K_2 \exp (-\varepsilon^2 N/(4K_2 \mathcal{T}_{1-\varepsilon}^{(11,+)}))\) and with \(\mathbb{P}\)-probability at least \(1-\exp(-KN)\). In particular, we see that on the event \(\cap_{1 \le k, \ell \le r} \mathcal{A}^{(k \ell)}\), 
\[
\mathcal{T}_{1-\varepsilon}^{(11)} = \mathcal{T}_{\varepsilon}^{(11)}  + (\mathcal{T}_{1-\varepsilon}^{(11)} - \mathcal{T}_{\varepsilon}^{(11)}) \lesssim \frac{1}{\sqrt{N}},
\]
with \(\mathbb{Q}_{\boldsymbol{X}}\)-probability at least \(1 - K_2 \exp (-N^{\frac{p+1}{2(p-1)}}/(4K_2)) - K_2 \exp (- \varepsilon^2 N/(4K_2 \mathcal{T}_{1-\varepsilon}^{(11)}))\) and with \(\mathbb{P}\)-probability at least \(1-\exp(-KN)\). \newline

\textbf{Step 6: Conclusion.} 
According to Steps 2 and 3, on the event \(\cap_{1 \le k, \ell \le r} \mathcal{A}^{(k \ell)}\),
\[
\mathcal{T}_{E_1(\varepsilon)} = \mathcal{T}_{1-\varepsilon}^{(11)} \vee \mathcal{T}_{\textnormal{low}} \lesssim \frac{1}{\sqrt{N}},
\]
with \(\mathbb{Q}_{\boldsymbol{X}}\)-probability at least \(1 -K_2 \exp(- N^{\frac{p+1}{2(p-1)}} / (4K_2)) - K_2 \exp(-\varepsilon^2 N^{3/2} /(4K_2))\) and with \(\mathbb{P}\)-probability at least \(1 - \exp(-K N)\). We then choose \(T_0^{(11)} = T_{\ell, \varepsilon_N}^{(11)}\) and \(T_0^{(ij)} = \mathcal{T}_{\textnormal{low}}\) for every \((i,j) \neq (1,1)\). In this way, we have on the initial event \(\mathcal{C}_1^{(N)}(\gamma_1,\gamma_2)\),
\[
\mathcal{T}_{E_1(\varepsilon)} \lesssim \frac{1}{\sqrt{N}},
\]
with \(\mathbb{Q}_{\boldsymbol{X}}\)-probability at least 
\begin{equation}1 - r^2 K_2 \exp(-\gamma_2^2 \sqrt{N}/(4K_2)) - K_2 \exp(- N^{\frac{p+1}{2(p-1)}} / (4K_2)) - K_2 \exp(-\varepsilon^2 N^{3/2} /(4K_2)),
\end{equation}
and with \(\mathbb{P}\)-probability at least \(1 - \exp(-K N)\), thus completing the proof.
\end{proof}

It remains to prove Lemma~\ref{lem: E_2}.

\begin{proof}[\textbf{Proof of Lemma~\ref{lem: E_2}}]
We prove the statement for \(k=2\) since the proof will be identical for the other cases. Fix \(\varepsilon \in (0,1)\) and assume \(\boldsymbol{X}_0 \in E_1(\varepsilon)\). In particular, \(m_{11}(0) \ge 1- \varepsilon\), \(\max_{i \neq 1} \{m_{1i} (0), m_{i1}(0)\} \lesssim \log(N)^{-1/2} N^{-\frac{p-1}{4}}\), and \(m_{ij}(0) = \gamma_{ij} N^{-1/2}\) with \(\gamma_{ij} = \mathcal{O}(1)\) for \(i,j \neq 1\). We first show that the evolution of the correlations \(m_{11}\) and \(m_{1i}, m_{i1}\) for \(i \neq 1\) are stable for all \(t \ge 0\). This can indeed be easily see from the generator expansion given by Lemma~\ref{lem: evolution equation m_ij}. In particular, we have
\[
L_\beta m_{11}(t) \ge - \norm{L_{0,\beta} m_{11}}_\infty + \beta \sqrt{M} p \lambda_1^2 m_{11}^{p-1}(t)(1-m_{11}^2(t)),
\]
for every \(t \ge 0\). Arguing as in Step 5 of Lemma~\ref{lem: E_1}, \( \norm{L_{0,\beta} m_{11}}_\infty \le \Lambda\) with \(\mathbb{P}\)-probability at least \(1- \exp(-KN)\), and we obtain that \(m_{11}(t) \ge 1-2\varepsilon\) for all \(t \in [0,T_2]\), where \(T_2 \asymp N^{-1/2}\) is the time horizon specified below. Similarly, Lemma~\ref{lem: evolution equation m_ij} yields for \(i \neq 1\),
\[
L_\beta m_{1i}(t) \leq \norm{L_{0,\beta} m_{1i}}_\infty - c\beta \sqrt{M} p \lambda_1^2 m_{11}^p(t) m_{1i}(t)
\]
for all \(t \ge 0\). Since \(m_{11}(t) \ge 1-2 \varepsilon\) on \([0,T_2]\), the drift is negative and arguing as in Step 3 of the proof of Lemma~\ref{lem: E_1}, we can show that
\[
\max_{i \neq 1} \{m_{1i} (0), m_{i1}(0)\} \lesssim \log(N)^{-1/2} N^{-\frac{p-1}{4}},
\]
for all \(t \in [0,T_2]\), with high \(\mathbb{Q}_{\boldsymbol{X}}\)- and \(\mathbb{P}\)-probability.
 
We therefore consider the evolution of the correlations \(m_{ij}\) for \(i, j \neq 1\). Let \(\varepsilon_N = c N^{-\frac{p-2}{2(p-1)}}\) for some constant \(c>0\). By the generator expansion from Lemma~\ref{lem: evolution equation m_ij}, i.e., 
\[
L_\beta m_{ij} = L_{0,\beta} m_{ij} + \beta \sqrt{M} p \lambda_i \lambda_j m_{ij}^{p-1} - \beta \sqrt{M} \frac{p}{2} \sum_{1 \leq k,\ell \leq r} \lambda_k m_{i \ell}m_{kj} m_{k \ell} (\lambda_j m_{kj}^{p-2} + \lambda_\ell m_{k \ell}^{p-2}),
\]
we see that for every \(i,j \neq 1\), 
\[
-\norm{L_{0,\beta} m_{ij}}_\infty +\beta \sqrt{M} p \lambda_i \lambda_j m_{ij}^{p-1}(t) \le L_\beta m_{ij}(t) \leq \norm{L_{0,\beta} m_{ij}}_\infty + \beta \sqrt{M} p \lambda_i \lambda_j m_{ij}^{p-1}(t),
\]
for all \(t \le \min_{2 \le k, \ell \le r} \mathcal{T}_{\varepsilon_N}^{(k \ell)}\). Indeed, the terms associated with \(m_{11}\) in the generator expansion are also accompanied by \(m_{i1}\) and \(m_{1i}\) which make that globally they are small compared to the term \(\beta \sqrt{M} p \lambda_i \lambda_j m_{ij}^{p-1}\), for \(N\) sufficiently large. We can therefore proceed exactly as done in the proof of Lemma~\ref{lem: E_1} by mimicking all the steps. This shows that, for \(N\) sufficiently large,
\[
\begin{split}
& \inf_{\boldsymbol{X}_0 \in E_1(\varepsilon)} \mathbb{Q}_{\boldsymbol{X}_0} \left ( \mathcal{T}_{E_2(\varepsilon)} \lesssim N^{-1/2}  \right ) \\
& \geq 1 - (r-1)^2 K_2 \exp (-\gamma_2^2 \sqrt{N} / (4K_2)) - K_2 \exp(- N^{\frac{p+1}{2(p-1)}}/(4K_2)) - K_2 \exp(-\varepsilon^2 N^{3/2}/(4K_2)),
\end{split}
\]
with \(\mathbb{P}\)-probability at least \(1 - \exp(-K N)\).
\end{proof}

\section{Proofs for \(p = 2\) and distinct SNRs} \label{section: proof recovery Langevin p=2}

This subsection is devoted to the proofs of Propositions~\ref{thm: strong recovery first spike Langevin p=2} and~\ref{thm: strong recovery all spikes Langevin p=2}. Throughout this section, we also assume that the SNRs satisfy
\[
\lambda_i = \lambda_{i+1} (1 + \kappa_i), \quad  1 \le i \le r-1, 
\]
where the constants \(\kappa_i > 0\) are independent of \(N\).

\subsection{Recovery of leading spike} 

The proof of Proposition~\ref{thm: strong recovery first spike Langevin p=2} follows a similar structure to the proof of Proposition~\ref{thm: strong recovery first spike Langevin p>2}. We begin to state the result on weak recovery of the first spike. We recall that, for every \(a >0\) and \(i,j \in [r]\), \(\mathcal{T}_a^{(ij)}\) denotes the hitting time
\[
\mathcal{T}_a^{(ij)} = \inf \{t \ge 0 \colon m_{ij} (t) \ge a\}. 
\]

\begin{lem}  \label{lem: weak recovery first spike Langevin p=2}
Let \(\beta \in (0,\infty)\) and \(p = 2\). Consider a sequence of initialization measures \(\mu_N^0 \in \mathcal{M}_1(\mathcal{S}_{N,r})\). Fix parameters \(n \geq 1, \gamma_0> 0, \gamma_1 > 1 > \gamma_2 > 0\). Then, there exist \(\varepsilon_0 > 0\) and \(c_0 \in (0, \frac{1}{2} \wedge \frac{\kappa_1}{2 + \kappa_1})\) such that for every \(\varepsilon < \varepsilon_0\), \(C_0 < c_0\), \(\sqrt{M} \gtrsim \frac{(n+2) \gamma_0}{\beta \lambda_r^2 C_0 \gamma_2} N^{\frac{1}{2(n+1)}}\), and \(N\) sufficiently large (in particular \(N \ge \varepsilon^{-2}\)),
\[
\begin{split}
&\int_{\mathcal{S}_{N,r}} \mathbb{Q}_{\boldsymbol{X}} \left (\mathcal{T}_\varepsilon^{(11)} \gtrsim \frac{\gamma_2 \log(\varepsilon \sqrt{N})}{(n+2)\gamma_0 N^{\frac{1}{2(n+1)}}} \right ) \boldsymbol{1}_{\{\mathcal{C}_{0,\beta}^{(N)}(n,\gamma_0) \cap \mathcal{C}_1^{(N)}(\gamma_1,\gamma_2)\}} (\boldsymbol{X})d\mu_N^0(\boldsymbol{X}) \\
& \leq K_1 e^{- \gamma_0^3  (n+2) N^{\frac{1}{2(n+1)}} / (K_1 \gamma_2 \log(N))} + r^2 K_2 e^{- \gamma_2 \gamma_0 (n+2) N^{\frac{1}{2(n+1)}} / (K_2 \log(N))},
\end{split}
\]
with \(\mathbb{P}\)-probability at least \(1 - \exp(-K N)\). Here, \(\gtrsim\) denotes a constant depending only on \(p\).
\end{lem}

Strong recovery of the first spike follows straightforwardly from the weak recovery result, as stated in the following lemma.

\begin{lem} \label{lem: weak implies strong recovery Langevin p=2}
Let \(\beta \in (0,\infty)\) and \(p = 2\). Then, for every \(\varepsilon > 0\) and \(n \ge 1\), there exists \(T_1 \le c  \log(N) N^{-\frac{1}{2(n+1)}}\) and \(c>0\) such that for all \(T \geq T_1\) and all \(N\) sufficiently large,
\[
\begin{split}
\inf_{\{\boldsymbol{X} \colon m_{11}(\boldsymbol{X}) \geq \varepsilon\}} \mathbb{Q}_{\boldsymbol{X}} \left (\inf_{t \in [T_1,T]} m_{11}(\boldsymbol{X}_t^\beta) \geq 1-\varepsilon \right ) & \geq 1 - 2 K_1 e^{- \varepsilon^2 N / (K_2 T)},
\end{split}
\]
with \(\mathbb{P}\)-probability at least \(1 - \exp(-K N)\).
\end{lem}

The proof of Lemma~\ref{lem: weak implies strong recovery Langevin p=2} follows the same strategy as the proof of Lemma~\ref{lem: weak implies strong recovery Langevin p>2} and is therefore omitted. Proposition~\ref{thm: strong recovery first spike Langevin p=2} is established in the same manner as Proposition~\ref{thm: strong recovery first spike Langevin p>2}, by combining Lemmas~\ref{lem: weak recovery first spike Langevin p=2} and~\ref{lem: weak implies strong recovery Langevin p=2} through the strong Markov property. We now turn to Lemma~\ref{lem: weak recovery first spike Langevin p=2}, whose proof closely parallels that of Lemma~\ref{lem: weak recovery first spike Langevin p>2}.

\begin{proof}[\textbf{Proof of Lemma~\ref{lem: weak recovery first spike Langevin p=2}}]
We let \(\mathcal{A} = \mathcal{A}(n, \gamma_0, \gamma_1, \gamma_2)\) denote the event 
\[
\mathcal{A}(n,\gamma_0,\gamma_1, \gamma_2) = \left \{\boldsymbol{X}_0 \sim \mu_N^0 \colon \boldsymbol{X}_0 \in \mathcal{C}_{0,\beta}^{(N)}(n,\gamma_0) \cap \mathcal{C}_1^{(N)}(\gamma_1,\gamma_2) \right\},
\]
where \(\mathcal{C}_{0,\beta}^{(N)}(n,\gamma_0)\) and \(\mathcal{C}_1^{(N)}(\gamma_1,\gamma_2)\) are given in Definitions~\ref{def: condition 0} and~\ref{def: condition 1}, respectively. We note that on \(\mathcal{C}_1^{(N)}(\gamma_1,\gamma_2)\), for every \(i,j \in [r]\) there exists \(\gamma_{ij} \in (\gamma_2,\gamma_1)\) such that \(m_{ij}(0) = \gamma_{ij} N^{-\frac{1}{2}}\). For some \(T_0 >0\) to be chosen later, we then define the event \(\mathcal{A}^{(ij)}= \mathcal{A}^{(ij)}(n,\gamma_0, \gamma_1,\gamma_2, T_0)\) by
\[
\mathcal{A}^{(ij)}(n,\gamma_0, \gamma_1,\gamma_2, T_0)=  \mathcal{A}(n,\gamma_0,\gamma_1, \gamma_2) \cap \left \{\sup_{t \in [0,T_0]} |M_t^{m_{ij}}| \leq  \frac{\gamma_2}{2\sqrt{N}}  \right \},
\]
where we recall that according to Lemma~\ref{lem: Doob inequality} and~\eqref{eq: Doob inequality}, there exists a constant \(K_2 > 0\) such that 
\begin{equation*}
\sup_{\boldsymbol{X}} \mathbb{Q}_{\boldsymbol{X}} \left ( \sup_{t \in [0,T_0^{(ij)}]} |M_t^{m_{ij}}| \geq \frac{\gamma_2}{2\sqrt{N}} \right ) \leq K_2 \exp \left (-\frac{\gamma_2^2}{4 K_2 T_0^{(ij)}} \right).
\end{equation*}
Moreover, for every \(i,j \in [r]\), we let \(\mathcal{T}_{L_{0,\beta}}^{(ij)}\) denote the hitting time of the set
\[
\left \{\boldsymbol{X} \colon |L_{0,\beta} m_{ij}(\boldsymbol{X})| > 2 C_0 \beta \sqrt{M}\lambda_i \lambda_j m_{ij}(\boldsymbol{X}) \right \},
\]
where \(C_0 \in (0,\frac{1}{2})\) is a constant which does not depend on \(N\). We note that on \(\mathcal{C}_{0,\beta}^{(N)}(n,\gamma_0)\), 
\[
|L_{0,\beta} m_{ij}(\boldsymbol{X}_0)| \leq \frac{\gamma_0}{ \sqrt{N}} \leq 2 C_0 \beta \sqrt{M} \lambda_i \lambda_j \frac{\gamma_2}{\sqrt{N}},
\]
provided \(\sqrt{M} \geq \frac{\gamma_0}{2C_0 \beta\lambda_i \lambda_j \gamma_2}\), which certainly holds by assumption. Therefore, on the event \(\mathcal{C}_{0,\beta}^{(N)}(n,\gamma_0)\), we have 
\[
|L_{0,\beta} m_{ij}(\boldsymbol{X}_0)| \leq 2 C_0 \beta \sqrt{M} \lambda_i \lambda_j m_{ij}(\boldsymbol{X}_0),
\]
and by continuity of the process \(\boldsymbol{X}_t\), \(\mathcal{T}_{L_{0,\beta}}^{(ij)} >0\). We also introduce the hitting time \(\mathcal{T}_{L_{0,\beta}}\) of the set
\[
\left \{\boldsymbol{X} \colon \sup_{1 \leq k, \ell \leq r}|L_{0,\beta} m_{k \ell}(\boldsymbol{X})| > 2 C_0 \beta \sqrt{M} \lambda_1^2 m_{11}(\boldsymbol{X}) \right \}.
\]
We also have that \(\mathcal{T}_{L_{0,\beta}}>0\) and we note that \(\mathcal{T}_{L_{0,\beta}} \leq \mathcal{T}_{L_{0,\beta}}^{(11)}\).

In the following, we fix \(i,j \in [r]\) and place ourselves on the event \(\mathcal{A}^{(ij)}\). We introduce the microscopic threshold \(\tilde{\varepsilon}_N = \tilde{\gamma} N^{-1/2}\), where \(\tilde{\gamma} > \gamma_1\) is a sufficiently large constant to be determined later. Since \(\tilde{\gamma} > \gamma_1\), it follows that \(\min_{1 \le i,j \le r} \mathcal{T}_{\tilde{\varepsilon}_N}^{(ij)} >0\). From Lemma~\ref{lem: evolution equation m_ij}, we have
\[
L_\beta m_{ij}(t) = L_{0,\beta} m_{ij}(t) + 2 \beta \sqrt{M} \lambda_i \lambda_j m_{ij}(t) -  \beta \sqrt{M}  \sum_{1 \leq k,\ell \leq r} \lambda_k (\lambda_j +\lambda_\ell) m_{i \ell}(t) m_{kj}(t) m_{k \ell}(t).
\]
As a consequence, for every \(t \leq \mathcal{T}_{L_{0,\beta}}^{(ij)} \wedge \mathcal{T}_{L_{0,\beta}} \wedge \min_{1 \leq k, \ell \leq r }\mathcal{T}_{\tilde{\varepsilon}_N}^{(k \ell)}\), we obtain the comparison bounds
\[
2(1-C_0) \beta \sqrt{M} \lambda_i \lambda_j m_{ij}(t) \leq L_\beta m_{ij}(t) \leq 2(1+ C_0) \beta \sqrt{M} \lambda_i \lambda_j m_{ij}(t),
\]
provided that \(N \gtrsim \frac{ r^2 \lambda_1^2 \tilde{\gamma}^3}{C_0\lambda_r^2 \gamma_2}\). According to the evolution equation for \(m_{ij}\) under Langevin dynamics, namely 
\[
m_{ij}(t) = m_{ij}(0) + M_t^{m_{ij}} + \int_0^t L_\beta m_{ij}(s) ds,
\]
we obtain the integral inequality
\begin{equation} \label{eq: integral inequality p=2}
\frac{\gamma_{ij}}{2\sqrt{N}} + 2 (1-C_0) \beta \sqrt{M} \lambda_i \lambda_j \int_0^t m_{ij}(s) ds
\leq m_{ij}(t) \leq \frac{3\gamma_{ij}}{2\sqrt{N}} + 2 (1+ C_0)  \beta \sqrt{M} \lambda_i \lambda_j \int_0^t m_{ij}(s) ds,
\end{equation}
which holds for \(t \leq \mathcal{T}_{L_{0,\beta}}^{(ij)} \wedge \mathcal{T}_{L_{0,\beta}} \wedge \min_{1 \leq k, \ell \leq r }\mathcal{T}_{\tilde{\varepsilon}_N}^{(k \ell)} \wedge T_0\). Grönwall's inequality (see item (d) of Lemma~\ref{lem: Gronwall}) then yields
\begin{equation} \label{eq: comparison inequality p=2}
\ell_{ij}(t) \leq m_{ij}(t) \leq u_{ij}(t),
\end{equation}
for all \(t\) in the same time interval, where the functions \(\ell_{ij}\) and \(u_{ij}\) are given by 
\[
\ell_{ij}(t) = \frac{\gamma_{ij}}{2\sqrt{N}} \exp \left(2(1-C_0) \beta \sqrt{M}\lambda_i \lambda_j t \right),
\]
and
\[
u_{ij}(t) = \frac{3\gamma_{ij}}{2\sqrt{N}} \exp \left(2(1 + C_0) \beta \sqrt{M}\lambda_i \lambda_j t \right),
\]
respectively. We now define \(T_{\ell, \tilde{\varepsilon}_N}^{(ij)}\) by \(\ell_{ij}(T_{\ell, \tilde{\varepsilon}_N}^{(ij)}) = \tilde{\varepsilon}_N\), i.e.,
\begin{equation} \label{eq: T lower bound p=2} 
T_{\ell, \tilde{\varepsilon}_N}^{(ij)} = \frac{\log(\tilde{\gamma}) - \log(\gamma_{ij}/2)}{2 (1-C_0) \beta \sqrt{M}\lambda_i\lambda_j}.
\end{equation}
Similarly, we let \(T_{u, \tilde{\varepsilon}_N}^{(ij)}\) satisfy \(u_{ij}(T_{u, \tilde{\varepsilon}_N}^{(ij)}) = \tilde{\varepsilon}_N\), so that
\begin{equation}  \label{eq: T upper bound p=2} 
T_{u,\tilde{\varepsilon}_N}^{(ij)} = \frac{\log(\tilde{\gamma})- \log (3 \gamma_{ij}/2)}{2 (1+C_0) \beta \sqrt{M} \lambda_i\lambda_j}.
\end{equation}
On the event \(\mathcal{A}^{(ij)}\), \(T_{u, \tilde{\varepsilon}_N}^{(ij)} \leq \mathcal{T}_{\tilde{\varepsilon}_N}^{(ij)} \leq T_{\ell,\tilde{\varepsilon}_N}^{(ij)}\). 

Our goal is to show that \(\min_{1 \le i,j \le r} \mathcal{T}_{\tilde{\varepsilon}_N}^{(ij)} \le \mathcal{T}_{L_{0,\beta}}\) and that \(\min_{1 \le i,j \le r} \mathcal{T}_{\tilde{\varepsilon}_N}^{(ij)} = \mathcal{T}_{\tilde{\varepsilon}_N}^{(11)}\), noting that \(\mathcal{T}_{L_{0,\beta}} \le \mathcal{T}_{L_{0,\beta}}^{(11)}\). Choose \(\tilde{\gamma}>0\) such that
\begin{equation} \label{eq: tilde gamma p=2}
\log(\tilde{\gamma}) \geq \frac{\kappa_1 +1 }{\kappa_1 -2} \log \left ( \frac{3 \gamma_1}{2} \right ).
\end{equation}
Then, for every \((i,j) \neq (1,1)\), we have  
\[
\begin{split}
T_{\ell, \tilde{\varepsilon}_N}^{(11)} & \leq \frac{\log(\tilde{\gamma}) - \log(\gamma_2/2)}{2(1-C_0) \beta \sqrt{M} \lambda_i \lambda_j \prod_{k=1}^{i-1} (1+\kappa_k) \prod_{\ell=1}^{j-1} (1+\kappa_\ell)} \\
& \leq \frac{\log(\tilde{\gamma})}{2(1-C_0) \beta \sqrt{M} \lambda_i \lambda_j (1 + \kappa_1)}\\
& \leq \frac{\log(\tilde{\gamma}) - \log(3 \gamma_1/2)}{2(1+C_0) \beta \sqrt{M} \lambda_i \lambda_j} \leq T_{u, \tilde{\varepsilon}_N}^{(ij)},
\end{split}
\]
where for the first inequality we used the fact that by definition \(\lambda_1 = \lambda_i (1+\kappa_1) \cdots (1+\kappa_{i-1})\) and that \(\lambda_1^2 \geq \lambda_i \lambda_j (1+\kappa_1)^2 \geq \lambda_i \lambda_j (1+\kappa_1)\) for every \(i \geq 2\), and for the last inequality we used the fact that by~\eqref{eq: tilde gamma p=2},
\[
\log(\tilde{\gamma}) \geq \frac{(1-C_0) (1+\kappa_1)}{(1-C_0) (1+\kappa_1) - (1+C_0)} \log \left ( \frac{3 \gamma_1}{2} \right ).
\]
We need to choose \(C_0 < \frac{1}{2} \wedge \frac{\kappa_1}{2 + \kappa_1}\) in order to ensure that \((1-C_0)(1+\kappa_1) - (1+C_0) >0\). In particular, it follows that
\[
\mathcal{T}_{\tilde{\varepsilon}_N}^{(11)} = \min_{1 \le i,j \le r} \mathcal{T}_{\tilde{\varepsilon}_N}^{(ij)} ,
\]
as soon as we can show
\[
\min_{1 \le i,j \le r}\mathcal{T}_{\tilde{\varepsilon}_N}^{(ij)} \le \min_{1 \le i,j \le r}\mathcal{T}_{L_{0,\beta}}^{(ij)} \wedge \mathcal{T}_{L_{0,\beta}} \wedge T_0.
\]

As done in the proof of Lemma~\ref{lem: weak recovery first spike Langevin p>2} for \(p \geq 3\), to get an estimate for \(L_{0,\beta} m_{ij}\) for every \(i,j \in [r]\), we wish to apply Lemma~\ref{lem: bounding flows} to the function \(F_{ij}(\boldsymbol{X}) = L_{0,\beta} m_{ij}(\boldsymbol{X})\). Conditions (1)-(3) are easily verified using the same arguments of the proof of Lemma~\ref{lem: weak recovery first spike Langevin p>2}. To see condition (4), we note that for every \(i,j \in [r]\), on the event \(\mathcal{A}^{(ij)}\),
\begin{equation} \label{eq: bound a_ij p=2}
\int_0^t |a_{ij}(s)| ds \leq \frac{1}{1 - C_0} \left (m_{ij}(t)-\frac{\gamma_{ij}}{2\sqrt{N}}\right) \leq \frac{1}{1-C_0} m_{ij}(t),
\end{equation}
for every \(t \leq \mathcal{T}_{L_{0,\beta}}^{(ij)} \wedge \mathcal{T}_{L_{0,\beta}} \wedge \min_{1 \leq k, \ell \leq r} \mathcal{T}_{\tilde{\varepsilon}_N}^{(k \ell)} \), where we used the lower bound in the integral inequality~\eqref{eq: integral inequality p=2} and recall that \(a_{ij}(t)=2\beta \sqrt{M}\lambda_i \lambda_jm_{ij}(t)\). We then observe that at time \(t=0\), for every \(\xi > 0\) we have 
\[
\xi 2 \beta \sqrt{M} \lambda_i \lambda_j \ell_{ij}(0) = \xi  \beta \sqrt{M} \lambda_i \lambda_j \frac{\gamma_{ij}}{\sqrt{N}} \geq C \xi (n+2) \gamma_0 N^{-\frac{n}{2(n+1)}} \geq \ell_{ij}(0),
\]
where we used that \(\sqrt{M} \geq C \frac{n+2}{\beta \lambda_r^2 C_0 (1-C_0) \gamma_2} \gamma_0 N^{ \frac{1}{2(n+1)}}\) for some constant \(C >0\). Since by~\eqref{eq: comparison inequality} the function \(m_{ij}(t)\) is lower bounded by \(\ell_{ij}(t)\) for every \(t \leq \mathcal{T}_{L_{0,\beta}}^{(ij)} \wedge \mathcal{T}_{L_{0,\beta}} \wedge \min_{1 \leq k, \ell \leq r} \mathcal{T}_{\tilde{\varepsilon}_N}^{(k \ell)}\) and since \(\ell_{ij}(t)\) is an increasing function satisfying the above inequality, we therefore obtain that 
\[
m_{ij}(t) \leq 2 \xi \beta \sqrt{M} \lambda_i \lambda_j m_{ij}(t),
\]
so that from~\eqref{eq: bound a_ij p=2} it follows that
\[
\int_0^t |a_{ij}(s)| ds \leq \frac{1}{1- C_0} m_{ij}(t) \leq \frac{\xi}{1-C_0} 2 \beta \sqrt{M} \lambda_i \lambda_j m_{ij}(t) ,
\]
for \(t \leq \mathcal{T}_{L_{0,\beta}}^{(ij)} \wedge \mathcal{T}_{L_{0,\beta}}  \wedge \min_{1 \leq k, \ell \leq r} \mathcal{T}_{\tilde{\varepsilon}_N}^{(k \ell)}\). Choosing \(\xi = (1-C_0)/2\) yields condition (4) with \(\varepsilon = 1/2\). From Lemma~\ref{lem: bounding flows} it follows that there exists \(K > 0\) such that on the event \(\cap_{1 \leq k, \ell \leq r} \mathcal{A}^{(k \ell)}\), for every \(t \leq \min_{1 \leq k, \ell \leq r} \mathcal{T}_{L_{0,\beta}}^{(k \ell)} \wedge \mathcal{T}_{L_{0,\beta}} \wedge \min_{1 \leq k, \ell \leq r}\mathcal{T}_{\tilde{\varepsilon}_N}^{(k \ell)} \), it holds that
\begin{equation} \label{eq: bound L0m_ij p=2}
|L_{0,\beta} m_{ij}(t)| \leq K \left(\frac{\gamma_0}{\sqrt{N}} \sum_{k=0}^{n-1}t^k + t^n + 2 \sum_{1 \leq k ,\ell \leq r} \int_0^t |a_{k \ell}(s)| ds\right),
\end{equation}
with \(\mathbb{Q}_{\boldsymbol{X}}\)-probability at least \(1 - K_2 \exp (- \gamma_0^2 / ( K_2 T_0))\) and with \(\mathbb{P}\)-probability at least \(1 - \exp(-K N)\). In order to deduce that, on the event \(\cap_{1 \leq k,\ell \leq r} \mathcal{A}^{(k \ell)}\), \(\mathcal{T}_{\tilde{\varepsilon}_N}^{(11)} = \min_{1 \leq k, \ell \leq r} \mathcal{T}_{\tilde{\varepsilon}_N}^{(k \ell)}\), we need to show that \(\min_{1 \leq k, \ell \leq r} \mathcal{T}_{\tilde{\varepsilon}_N}^{(k \ell)} \leq \min_{1 \leq k, \ell \leq r} \mathcal{T}_{L_{0,\beta}}^{(k \ell)} \wedge \mathcal{T}_{L_{0,\beta}}\). It therefore suffices to show that for every \(i,j \in [r]\), each term in the sum~\eqref{eq: bound L0m_ij p=2} is upper bounded by \(\frac{2 C_0 \beta \sqrt{M} \lambda_i \lambda_j}{n+2} m_{ij}(t)\) for every \(t \leq \min_{1 \leq k, \ell \leq r} \mathcal{T}_{L_{0,\beta}}^{(k \ell)} \wedge \mathcal{T}_{L_{0,\beta}} \wedge \min_{1 \leq k, \ell \leq r} \mathcal{T}_{\tilde{\varepsilon}_N}^{(k \ell)} \). 
\begin{itemize}
\item[(i)] We start by observing that on the event \(\cap_{1 \leq k, \ell \leq r} \mathcal{A}^{(k \ell)}\),  
\[
\frac{2 C_0 \beta \sqrt{M} \lambda_i \lambda_j}{n+2} m_{ij}(t) \geq \frac{2 C_0 \beta \sqrt{M} \lambda_i \lambda_j}{n+2} \ell_{ij}(t) \gtrsim \gamma_0 N^{-\frac{n}{2(n+1)}} \exp((n+2) \gamma_0 N^{\frac{1}{2(n+1)}}t),
\]
for \(t \leq \min_{1 \leq k, \ell \leq r} \mathcal{T}_{L_{0,\beta}}^{(k \ell)} \wedge \mathcal{T}_{L_{0,\beta}} \wedge \min_{1 \leq k, \ell \leq r}\mathcal{T}_{\tilde{\varepsilon}_N}^{(k \ell)}\), where we used the assumption \(\sqrt{M} \gtrsim \frac{(n+2)\gamma_0}{\beta \lambda_r^2 C_0(1-C_0) \gamma_2} N^{\frac{1}{2(n+1)}}\). We then see that for every \(0 \leq k \leq n-1\),
\[
\exp((n+2) \gamma_0 N^{\frac{1}{2(n+1)}}t) \geq \frac{(n+2)^k \gamma_0^k N^{\frac{k}{2(n+1)}}}{k!} t^k \gtrsim t^k
\]
for \(t \leq T_0< \log(\varepsilon \sqrt{N})/((n+2) \gamma_0 N^{\frac{1}{2(n+1)}})\). We therefore have that for every \(0 \leq k \leq n-1\),
\[
\frac{2 C_0 \beta \sqrt{M} \lambda_i \lambda_j}{n+2} m_{ij}(t) \gtrsim K \frac{\gamma_0}{\sqrt{N}} t^k.
\]
\item[(ii)] We next observe that a sufficient condition to control the second term is given by \(F(t) = Kt^n \leq \frac{2C_0 \beta \sqrt{M} \lambda_i \lambda_j}{n+2} \ell_{ij}(t) = G(t)\). An explicit computation shows that for every \(k \leq n\), 
\[
F^{(k)}(t) = K n (n-1) \cdots (n-k+1)t^{n-k}
\]
and 
\[
G^{(k)}(t) = \frac{C_0 \beta \sqrt{M} \lambda_i \lambda_j}{n+2} \frac{\gamma_{ij}}{\sqrt{N}} \left (2(1-C_0) \beta \sqrt{M} \lambda_i \lambda_j \right )^k \exp( 2 (1-C_0) \beta \sqrt{M} \lambda_i \lambda_j t).
\]
For \(k \leq n-1\), it holds that \(F^{(k)}(0) = 0 \leq G^{(k)}(0)\), and for \(k=n\) we have
\[
\begin{split}
G^{(n)}(t) & = \frac{\left (\beta \sqrt{M} \lambda_i \lambda_j \right)^{n+1} \gamma_{ij} C_0 2^n (1-C_0)^n}{(n+2) \sqrt{N}}  \exp( 2 (1-C_0) \beta \sqrt{M} \lambda_i \lambda_j t)\\
& \gtrsim \frac{2^n (n+2)^n \gamma_0^{n+1}}{\gamma_2^n C_0^n (1-C_0)} \exp((n+2) \gamma_0 N^{\frac{1}{2(n+1)}}t)\\
& \geq Kn! = F^{(n)}(t),
\end{split}
\]
which gives the bound for the second term.
\item[(iii)] To bound the last term, we note from~\eqref{eq: integral inequality p=2} that on the event \(\cap_{1 \leq k, \ell \leq r} \mathcal{A}^{(k \ell)}\),
\[
2\sum_{1 \leq k, \ell \leq r} \int_0^t \vert a_{k \ell}(s) \vert ds \leq \frac{2 r^2}{1-C_0} \max_{1 \leq k, \ell \leq r} \{m_{k \ell}(t)\} \leq \frac{2 r^2}{1-C_0} \tilde{\varepsilon}_N =  \frac{2 r^2}{1-C_0} \frac{\tilde{\gamma}}{\sqrt{N}},
\]
for \(t \leq \min_{1 \leq k, \ell \leq r} \mathcal{T}_{L_{0,\beta}}^{(k \ell)} \wedge \mathcal{T}_{L_{0,\beta}} \wedge \min_{1 \leq k, \ell \leq r}  \mathcal{T}_{\tilde{\varepsilon}_N}^{(k \ell)} \wedge T_0\). Then, since by the assumption on \(\sqrt{M}\) we have
\[
\frac{C_0 \beta \sqrt{M} \lambda_i \lambda_j}{n+2} \frac{\gamma_{ij}}{\sqrt{N}} \gtrsim \frac{\gamma_0}{1-C_0} N^{-\frac{n}{2(n+1)}} \ge K \frac{2r^2}{1-C_0} \frac{\tilde{\gamma}}{\sqrt{N}},
\]
it follows that 
\[
\frac{2C_0 \beta \sqrt{M} \lambda_i \lambda_j}{n+2} m_{ij}(t) \geq  \frac{2C_0 \beta \sqrt{M} \lambda_i \lambda_j}{n+2} \ell_{ij}(0) \gtrsim K \frac{2r^2}{1-C_0} \frac{\tilde{\gamma}}{\sqrt{N}},
\]
for \(t \leq \min_{1 \leq k, \ell \leq r} \mathcal{T}_{L_{0,\beta}}^{(k \ell)} \wedge \mathcal{T}_{L_{0,\beta}} \wedge \min_{1 \leq k, \ell \leq r}  \mathcal{T}_{\tilde{\varepsilon}_N}^{(k \ell)}\). This shows the last bound.
\end{itemize}
We therefore have that, on the event \(\cap_{1 \leq k, \ell \leq r} \mathcal{A}^{(k \ell)}\), 
\(\min_{1 \leq k, \ell \leq r}\mathcal{T}_{\tilde{\varepsilon}_N}^{(k \ell)} \leq \min_{1 \leq k, \ell \leq r} \mathcal{T}_{L_{0,\beta}}^{(k \ell)} \wedge \mathcal{T}_{L_{0,\beta}}\) so that it follows 
\[
\mathcal{T}_{\tilde{\varepsilon}_N}^{(11)} = \min_{1 \leq k, \ell \leq r} \mathcal{T}_{\tilde{\varepsilon}_N}^{(k \ell)},
\]
with \(\mathbb{Q}_{\boldsymbol{X}}\)-probability at least \(1 - K_2 \exp (- \gamma_0^2 / ( K_2 T_0))\) and with \(\mathbb{P}\)-probability at least \(1 - \exp(-K N)\).

Now we show that \(m_{11}\) remains dominant until it reaches \(\varepsilon\). Define the hitting time 
\[
\tau \coloneqq \inf \left \{t \ge \mathcal{T}_{\tilde{\varepsilon}_N}^{(11)} \colon m_{11}(t) \le \max_{(i,j)\neq (1,1)} m_{ij}(t) \right\},
\]
so that \(m_{ij}(t) \le m_{11}(t)\) for all \((i,j) \neq (1,1)\) on \([\mathcal{T}_{\tilde{\varepsilon}_N}^{(11)} , \tau)\). Choose \(\varepsilon_0\) such that \(Cr^2 \varepsilon_0^2 \le C_0\). Then, for \(\varepsilon \le \varepsilon_0\), according to the generator expansion \(L_\beta m_{ij}\) given by Lemma~\ref{lem: evolution equation m_ij}, we have that
\[
2(1-2C_0) \beta \sqrt{M} \lambda_1^2 m_{11}(t) \le L_\beta m_{11}(t) \le 2(1+C_0) \beta \sqrt{M} \lambda_1^2 m_{11}(t),
\]
for all \(t \le \mathcal{T}_{L_{0,\beta}} \wedge \tau \wedge \mathcal{T}_\varepsilon^{(11)}\). This implies that the comparison inequality~\eqref{eq: comparison inequality p=2} holds for \(m_{11}\) over the time interval
\[
\mathcal{T}_{\tilde{\varepsilon}_N}^{(11)} \leq t \leq \tau \wedge \mathcal{T}_\varepsilon^{(11)} \wedge \mathcal{T}_{L_{0,\beta}}  \wedge T_0.
\]
In particular, letting \(T_{\ell, \varepsilon}^{(11)}\) be defined by \(\ell_{11}(T_{\ell, \varepsilon}^{(11)}) = \varepsilon\), we have \(\mathcal{T}_\varepsilon^{(11)} \le T_{\ell, \varepsilon}^{(11)}\). For \((i,j) \neq (1,1)\), we always have the upper drift bound 
\[
L_\beta m_{ij}(t) \le 2 (1+C_0) \beta \sqrt{M} \lambda_i \lambda_j m_{ij}(t),
\]
for all \(t \le  \mathcal{T}_{L_{0,\beta}}^{(ij)}\), hence \(m_{ij}(t) \le u_{ij}(t)\) for all \(t \le  \mathcal{T}_{L_{0,\beta}}^{(ij)} \wedge T_0\). Choosing \(T_0 = T_{\ell, \varepsilon}^{(11)}\) and evaluating at \(t=T_{\ell,\varepsilon}^{(11)}\) gives
\[
m_{ij} (T_{\ell,\varepsilon}^{(11)}) \le u_{ij} (T_{\ell,\varepsilon}^{(11)}) = \frac{3 \gamma_{ij}}{2 \sqrt{N}} \left (\frac{4 \varepsilon^2 N}{\gamma_{11}^2} \right)^{\frac{1}{2} \frac{1+C_0}{1-C_0} \frac{\lambda_i \lambda_j}{\lambda_1^2}} \lesssim N^{-\frac{1}{2} \left ( 1- \frac{1+C_0}{1-C_0} \frac{\lambda_i \lambda_j}{\lambda_1^2}\right)}.
\]
For \(N\) sufficiently large, we have \(m_{ij} (T_{\ell,\varepsilon}^{(11)}) < \varepsilon\) for all \((i,j) \neq (1,1)\). Hence, \(\tau > T_{\ell,\varepsilon}^{(11)}\ge \mathcal{T}_\varepsilon^{(11)}\), so \(m_{11}\) remains strictly dominant up to \(\mathcal{T}_\varepsilon^{(11)}\), and therefore
\[
\mathcal{T}_\varepsilon^{(11)} =\min_{1 \le k, \ell \le r} \mathcal{T}_\varepsilon^{(k\ell)}.
\]

It therefore remains to show that \(\mathcal{T}_\varepsilon^{(11)} \leq \mathcal{T}_{L_{0,\beta}}^{(11)}\) with high \(\mathbb{Q}_{\boldsymbol{X}}\)- and \(\mathbb{P}\)-probability. We want to use the estimate~\eqref{eq: bound L0m_ij p=2}. Since each \(a_{k \ell}(t) \leq a_{11}(t)\) for \(\mathcal{T}_{\tilde{\varepsilon}_N}^{(11)} \leq t \leq \min_{1 \leq k, \ell \leq r} \mathcal{T}_{L_{0,\beta}}^{(k \ell)} \wedge \mathcal{T}_{L_{0,\beta}} \wedge \mathcal{T}_\varepsilon^{(11)} \), it follows from~\eqref{eq: bound L0m_ij p=2} that on the event \(\cap_{1 \leq k, \ell \leq r} \mathcal{A}^{(k \ell)}\), for every \(\mathcal{T}_{\tilde{\varepsilon}_N}^{(11)} \leq t \leq \min_{1 \leq k, \ell \leq r} \mathcal{T}_{L_{0,\beta}}^{(k \ell)} \wedge \mathcal{T}_{L_{0,\beta}} \wedge \mathcal{T}_\varepsilon^{(11)} \),
\begin{equation} \label{eq: bounding flow langevin p=2 2}
|L_{0,\beta} m_{ij}(t)| \leq K \left(\frac{\gamma_0}{\sqrt{N}} \sum_{k=0}^{n-1} t^k +t^n + 2r^2 \int_0^t |a_{11}(s)| ds \right),
\end{equation}
with \(\mathbb{Q}_{\boldsymbol{X}}\)-probability at least \(1 - K_2 \exp \left (- \gamma_0^2 / (K_2  T_0) \right )\) and with \(\mathbb{P}\)-probability at least \(1 - \exp(-K N)\). In the same way as before, we can show that each term in the sum~\eqref{eq: bounding flow langevin p=2 2} is upper bounded by \(\frac{2 C_0 \beta \sqrt{M} \lambda_1^2}{n+2} m_{11}(t)\) for \(\mathcal{T}_{\tilde{\varepsilon}_N}^{(11)} \leq t \leq  \min_{1 \leq k, \ell \leq r} \mathcal{T}_{L_{0,\beta}}^{(k\ell)} \wedge \mathcal{T}_{L_{0,\beta}} \wedge \mathcal{T}_\varepsilon^{(11)} \), ensuring that \(\mathcal{T}_\varepsilon^{(11)} \le \mathcal{T}_{L_{0,\beta}}\). It then follows that on the event \(\mathcal{C}_{0,\beta}^{(N)}(n,\gamma_0) \cap \mathcal{C}_1^{(N)}(\gamma_1,\gamma_2)\) we have
\[
\mathcal{T}_\varepsilon^{(11)} \le T_{\ell, \varepsilon}^{(11)} \lesssim \frac{\gamma_2 \log(\varepsilon\sqrt{N})}{(n+2) \gamma_0 N^{\frac{1}{2(n+1)}}} 
\]
with \(\mathbb{Q}_{\boldsymbol{X}}\)-probability at least 
\[
1 - K_2 \exp \left ( - \frac{\gamma_0^3 (n+2)N^{\frac{1}{2(n+1)}}}{K_2 \gamma_2 \log(N)}\right ) - K_2 r^2 \exp \left (- \frac{\gamma_2 \gamma_0 (n+2)N^{\frac{1}{2(n+1)}}}{4K_2 \log(N)} \right ),
\]
and \(\mathbb{P}\)-probability at least \(1 - \exp(-K N)\).
\end{proof}

\subsection{Recovery of multiple spikes} 

In the following, we focus on the proof of exact recovery of all spikes for \(p=2\) under a suitable separation of the SNRs. Throughout we assume \(\lambda_i = (1+\kappa_i)\lambda_{i+1}\) with \(\kappa_i>0\) for \(1 \le i \le r-1\). Moreover, we set \(\kappa \coloneqq \min_{1 \le i \le r-1} \kappa_i\). Following a strategy similar to that used in the proof of Proposition~\ref{thm: strong recovery all spikes Langevin p>2} in the previous section, we proceed through several steps, each focusing on establishing the weak recovery of one signal direction. 

Let \(C_0 \in \left ( 0, \frac{\kappa}{2+\kappa} \right)\). For every \(1 \leq k \leq r\) and every \(k \le i,j \le r\) with \((i,j) \neq (k,k)\), define
\[
\delta_{ij}^{(k)} (C_0) \coloneqq 1 - \frac{1+C_0}{1-C_0} \frac{\lambda_i \lambda_j}{\lambda_k^2} \quad \textnormal{and} \quad \xi_{ij}^{(k)} (C_0) \coloneqq 1 - \frac{1-C_0}{1+C_0} \frac{\lambda_i \lambda_j}{\lambda_k^2}.
\]
By construction, \(0 < \delta_{ij}^{(k)} (C_0) < \xi_{ij}^{(k)} (C_0) < 1\), and both are symmetric in \((i,j)\). To lighten notation, we sometimes omit the dependence on \(C_0\). For \(\varepsilon >0, C_0 \in \left ( 0, \frac{\kappa}{2+\kappa} \right)\), and \(i \in [r]\), define the weak and strong recovery regions as
\[
\begin{split}
W_i(\varepsilon,C_0) & \coloneqq \left \{ \boldsymbol{X} \colon m_{ii}(\boldsymbol{X}) \geq \varepsilon \enspace \textnormal{and} \enspace \max_{j>i} \{ m_{ij}(\boldsymbol{X}) \vee m_{ji}(\boldsymbol{X})\} \lesssim N^{-\frac{\delta_{ij}^{(i)}}{2}} \right \},\\
S_i(\varepsilon,C_0) & \coloneqq \left \{ \boldsymbol{X} \colon m_{ii}(\boldsymbol{X}) \geq 1 - \varepsilon \enspace \textnormal{and} \enspace \max_{j>i} \{m_{ij}(\boldsymbol{X}) \vee m_{ji}(\boldsymbol{X}) \} \lesssim N^{-\frac{\xi_{rr}^{(1)}}{2}} \right \}.
\end{split}
\]
For \(k \in [r]\) define the the intermediate ``recovering'' events as
\[
E_k (\varepsilon,C_0) \coloneqq \left ( \cap_{i=1}^{k-1} S_i(\varepsilon,C_0) \right) \cap W_k (\varepsilon,C_0) \cap \left \{\boldsymbol{X} \colon N^{-\frac{\xi_{ij}^{(k)}}{2}} \lesssim m_{ij}(\boldsymbol{X}) \lesssim N^{-\frac{\delta_{ij}^{(k)}}{2}} \enspace \forall \,  i,j \ge k+1  \right \}.
\]
Finally, the recovery event for \(p=2\) given in~\eqref{eq: set strong recovery p=2} corresponds to \(R(\varepsilon, C_0) = \cap_{1 \le i \le r} S_i (\varepsilon,C_0) \). 

The first step consists in showing that, starting from any initialization satisfying Conditions \(0\) and \(1\), Langevin dynamics reaches \(E_1(\varepsilon,C_0)\) on a time scale of order \(\log(\varepsilon \sqrt{N})N^{-\delta(C_0) /2}\) provided \(\sqrt{M} \gtrsim N^{\delta(C_0)/2}\), where \(\delta (C_0)>0\) depends on the ratio between the signal sizes as follows:
\[
\delta (C_0) \coloneqq \frac{1+C_0}{1-C_0} - \frac{\lambda_r^2}{\lambda_1^2} .
\]

\begin{lem} \label{lem: weak recovery first spike Langevin p=2 2}
Let \(\beta \in (0,\infty)\) and \(p=2\). Consider a sequence of initialization measures \(\mu_N^0 \in \mathcal{M}_1(\mathcal{S}_{N,r})\). Fix parameters \(n \geq 1, \gamma_0>0\) and \(\gamma_1 > 1 > \gamma_2 > 0\). Then there exist \(\varepsilon_0 > 0\), \(c_0 \in (0, \frac{1}{2} \wedge \frac{\kappa_1}{2 + \kappa_1})\), and constants \(C,K,K_1,K_2>0\) such that for every \(\varepsilon < \varepsilon_0\) and every \(C_0 < c_0\), if
\[
\sqrt{M} \geq C\frac{(n+2) \gamma_0}{\beta C_0 \lambda_r^2 \gamma_2} N^{\delta(C_0) /2}
\]
and if \(N\) large enough so that
\[
\log(\varepsilon \sqrt{N}) \geq \log(3 \gamma_1/2) + \log(3 \gamma_1/\gamma_2) \frac{1+C_0}{(1-C_0)(1+\kappa_1) -(1+C_0)},
\]
then 
\[
\begin{split}
& \int_{\mathcal{S}_{N, r}}  \mathbb{Q}_{\boldsymbol{X}}\left(\mathcal{T}_{E_1(\varepsilon,C_0)} \ge \frac{1}{C} \frac{ \gamma_2 \log(\varepsilon \sqrt{N}) N^{-\delta(C_0)/2}}{(n+2) \gamma_0 \gamma_1} \right) \boldsymbol{1}_{ \left \{\mathcal{C}_{0,\beta}^{(N)}(n,\gamma_0) \cap \mathcal{C}_1^{(N)}(\gamma_1,\gamma_2) \right \}}(\boldsymbol{X}) d \mu_N^0(\boldsymbol{X}) \\
& \le K_1 \exp \left ( - \frac{\gamma_0^3 (n+2) N^{\delta(C_0)/2}}{K_1 \gamma_2 \log(N)} \right) + r^2 K_2 \exp \left ( - \frac{\gamma_2 \gamma_0 (n+2) N^{\delta(C_0)/2}}{K_2 \log(N)} \right),
\end{split}
\]
with \(\mathbb{P}\)-probability \(1-\exp(- K N)\).
\end{lem}

Having the first step at hand, we next show that \(\boldsymbol{x}_1\) achieves strong recovery with \(\boldsymbol{v}_1\) and that \(\boldsymbol{x}_2\) achieves weak recovery with \(\boldsymbol{v}_2\), i.e., we show that from the event \(E_1 (\varepsilon, C_0)\) we obtain the event \(E_2 (\varepsilon, C_0)\). In general, having the event \(E_k (\varepsilon, C_0)\) with \(1 \le k \le r-1\) at hand, we can reach the event \(E_{k+1} (\varepsilon, C_0)\), as stated by the following lemma. 

\begin{lem} \label{lem: weak recovery k spike Langevin p=2} 
Let \(\beta \in (0,\infty)\) and \(p=2\). Fix \(\gamma_1 > 1 > \gamma_2 > 0\) and \(2 \le k \le r\). Then there exist \(\varepsilon_0 > 0\), \(c_0 \in (0, \frac{1}{2} \wedge \frac{\kappa}{2+\kappa})\), and constants \(\Lambda = \Lambda(p,\beta, \{\lambda_i\}_{i=1}^r)\), \(C, K, K_1, K_2>0\) such that for every \(\varepsilon < \varepsilon_0\), \(C_0 < c_0\), if
\[
\sqrt{M} \ge \frac{\Lambda}{\beta C_0 \lambda_r^2 \gamma_2} N^{\delta(C_0)/2}, 
\]
and 
\[
\log(N) \ge  C\frac{1}{(1 + \kappa_k)^2} \log(\frac{1}{\kappa_k}), 
\]
then there exists \(T_k \lesssim \gamma_2 \log(\varepsilon N^{\xi_{kk}^{(k-1)/2}}) N^{-\delta(C_0)/2}\) such that for all \(N\) large enough, 
\[
\inf_{\boldsymbol{X} \in E_{k-1} (\varepsilon,C_0)} \mathbb{Q}_{\boldsymbol{X}} \left( \mathcal{T}_{E_k (\varepsilon,C_0)} \le T_k \right) \ge 1 - K_1 e^{-\varepsilon^2 N^{1+\delta(C_0)/2}/K_1} ,
\]
with \(\mathbb{P}\)-probability at least \(1 - \exp\left(- K N \right)\).
\end{lem}

The last phase is to show that strong recovery of the last spike follows straightforwardly from \(E_r(\varepsilon,C_0)\).

\begin{lem} \label{lem: strong recovery r spike Langevin p=2}
Let \(\beta \in (0,\infty)\) and \(p=2\). Fix \(\gamma_1 > 1 > \gamma_2 > 0\) and \(2 \le k \le r\). Then there exist \(\varepsilon_0 > 0\), \(c_0 \in (0, \frac{1}{2} \wedge \frac{\kappa}{2+\kappa})\), and constants \(\Lambda = \Lambda(p,\beta, \{\lambda_i\}_{i=1}^r)\), \(C, K, K_1, K_2>0\) such that for every \(\varepsilon < \varepsilon_0\), \(C_0 < c_0\), if
\[
\sqrt{M} \ge \frac{\Lambda}{\beta C_0 \lambda_r^2 \gamma_2} N^{\delta(C_0)/2}, 
\]
and 
\[
\log(N) \ge  C\frac{1}{(1 + \kappa_k)^2} \log(\frac{1}{\kappa_k}), 
\]
then there exists \(T_r \lesssim \gamma_2 \log(\varepsilon N^{\xi_{rr}^{(r-1)/2}}) N^{-\delta(C_0)/2} \) such that for all \(N\) large enough, 
\[
\inf_{\boldsymbol{X} \in E_r(\varepsilon,C_0)} \mathbb{Q}_{\boldsymbol{X}} \left(  \mathcal{T}_{R(\varepsilon, C_0)} \le T_r \right) \geq 1 - K_1 e^{-\varepsilon^2 N^{1+\delta(C_0)/2}/K_1},
\]
with \(\mathbb{P}\)-probability at least \(1 - \exp(-K N)\).
\end{lem}

Having Lemmas~\ref{lem: weak recovery first spike Langevin p=2 2},~\ref{lem: weak recovery k spike Langevin p=2}, and~\ref{lem: strong recovery r spike Langevin p=2} at hand, we deduce Proposition~\ref{thm: strong recovery all spikes Langevin p=2} using the strong Markov property, as done for the proof of  Proposition~\ref{thm: strong recovery all spikes Langevin p>2} in Subsection~\ref{subsection: recovery of all spikes p>2}. In the remainder of this subsection we prove Lemma~\ref{lem: weak recovery first spike Langevin p=2 2} and Lemma~\ref{lem: weak recovery k spike Langevin p=2} for \(k=2\). The proof for general \(k\ge3\) follows by the same induction scheme and is omitted.

\begin{proof}[\textbf{Proof of Lemma~\ref{lem: weak recovery first spike Langevin p=2 2}}] 
Fix \(n \ge 1\), \(\gamma_0>0\), and \(\gamma_1 >1> \gamma_2 >0\). Let \(\mathcal{A} = \mathcal{A}(n,\gamma_0,\gamma_1,\gamma_2)\) denote the initial event 
\[
\mathcal{A}(n,\gamma_0,\gamma_1,\gamma_2) = \left \{\boldsymbol{X}_0\sim \mu_N^0 \colon \boldsymbol{X}_0 \in \mathcal{C}_{0,\beta}^{(N)}(n,\gamma_0) \cap \mathcal{C}_1^{(N)}(\gamma_1,\gamma_2) \right \}.
\]
On \(\mathcal{C}_1^{(N)}(\gamma_1,\gamma_2)\), for every \(i,j \in [r]\) there exists \(\gamma_{ij} \in (\gamma_2,\gamma_1)\) such that \(m_{ij}(0) = \gamma_{ij} N^{-\frac{1}{2}}\). Let \(T_0 >0\) be a deterministic time horizon (to be chosen later), and define the event \(\mathcal{A}^{(ij)}= \mathcal{A}^{(ij)}(n,\gamma_0, \gamma_1,\gamma_2, T_0)\) by
\[
\mathcal{A}^{(ij)}(n,\gamma_0,\gamma_1,\gamma_2, T_0)=  \mathcal{A}(n,\gamma_0,\gamma_1, \gamma_2) \cap \left \{\sup_{t \in [0,T_0]} |M_t^{m_{ij}}| \leq  \frac{\gamma_2}{2\sqrt{N}}  \right \},
\]
where we recall that according to Lemma~\ref{lem: Doob inequality} and~\eqref{eq: Doob inequality}, there exists a constant \(K_2 > 0\) such that 
\[
\sup_{\boldsymbol{X}} \mathbb{Q}_{\boldsymbol{X}} \left ( \sup_{t \in [0,T_0]} |M_t^{m_{ij}}| \geq \frac{\gamma_2}{2\sqrt{N}} \right ) \leq K_2 \exp \left (-\frac{\gamma_2^2}{4 K_2 T_0} \right).
\]
Moreover, for every \(i,j \in [r]\) we let \(\mathcal{T}_{L_{0,\beta}}^{(ij)}\) denote the hitting time of the set
\[
\left \{\boldsymbol{X} \colon |L_{0,\beta} m_{ij}(\boldsymbol{X})| >  C_0 \beta \sqrt{M} \lambda_i \lambda_j m_{ij}(\boldsymbol{X}) \right \},
\]
for some order 1 constant \(C_0 \in (0, 1/2)\). We note that on \(\mathcal{C}_{0,\beta}^{(N)}(n,\gamma_0)\), 
\[
|L_{0,\beta} m_{ij}(\boldsymbol{X}(0))| \leq \frac{\gamma_0}{\sqrt{N}} \leq  C_0 \beta \sqrt{M} \lambda_i \lambda_j \frac{\gamma_2}{\sqrt{N}},
\]
provided \(\sqrt{M} \geq \frac{\gamma_0}{C_0 \beta \lambda_i \lambda_j \gamma_2}\), which certainly holds by assumption. Therefore, on the initial event \(\mathcal{A}\), we have \(|L_{0,\beta} m_{ij}(0)|\leq 2 C_0 \beta \sqrt{M} \lambda_i \lambda_j m_{ij}(0)\), thus by continuity of the process \(\boldsymbol{X}_t\) it follows that \(\mathcal{T}_{L_{0,\beta}}^{(ij)} > 0\) for every \(i,j \in [r]\).

In the following, we fix \(i,j \in [r]\) and place ourselves on the event \(\mathcal{A}^{(ij)}\). Recalling the generator expansion given by Lemma~\ref{lem: evolution equation m_ij}, i.e.,
\[
L_\beta m_{ij}(t) = L_{0,\beta} m_{ij}(t) + 2 \beta \sqrt{M} \lambda_i \lambda_j m_{ij}(t) -  \beta \sqrt{M}  \sum_{1 \leq k,\ell \leq r} \lambda_k (\lambda_j +\lambda_\ell) m_{i \ell}(t) m_{kj}(t) m_{k \ell}(t),
\]
we have for \(t \le \min_{1 \leq k, \ell \leq r }\mathcal{T}_{\varepsilon}^{(k \ell)}\), 
\[
\left | \sum_{1 \leq k,\ell \leq r} \lambda_k (\lambda_j +\lambda_\ell) m_{i \ell}(t) m_{kj}(t) m_{k \ell}(t) \right| \le 2 r^2 \lambda_1^2 \varepsilon^2 m_{ij}(t).
\]
Choose \(\varepsilon_0>0\) such that for all \(\varepsilon < \varepsilon_0\),
\[
r^2 \lambda_1^2 \varepsilon^2 \le \frac{C_0}{2} \lambda_r^2.
\]
This implies that, for every \(t \leq \mathcal{T}_{L_{0,\beta}}^{(ij)} \wedge \min_{1 \leq k, \ell \leq r }\mathcal{T}_{\varepsilon}^{(k \ell)}\),
\[
2(1-C_0) \beta \sqrt{M} \lambda_i \lambda_j m_{ij}(t) \leq L_\beta m_{ij}(t) \leq  2(1+C_0)\beta \sqrt{M} \lambda_i \lambda_j m_{ij}(t),
\]
We then obtain the integral inequality
\begin{equation} \label{eq: integral inequality p=2 2}
\frac{\gamma_{ij}}{2\sqrt{N}} + 2 (1-C_0) \beta \sqrt{M} \lambda_i \lambda_j \int_0^t m_{ij}(s) ds \leq m_{ij}(t) \leq \frac{3\gamma_{ij}}{2\sqrt{N}} + 2 (1+ C_0)  \beta \sqrt{M} \lambda_i \lambda_j \int_0^t m_{ij}(s) ds,
\end{equation}
for all \(t \leq \mathcal{T}_{L_{0,\beta}}^{(ij)}  \wedge \min_{1 \leq k, \ell \leq r }\mathcal{T}_{\varepsilon}^{(k \ell)} \wedge T_0\). Furthermore, by the Grönwall's inequality (see item (d) of Lemma~\ref{lem: Gronwall}) we have
\begin{equation} \label{eq: comparison inequality p=2 2}
\frac{\gamma_{ij}}{2\sqrt{N}} \exp \left(2(1-C_0) \beta \sqrt{M}\lambda_i \lambda_j t \right) = \ell_{ij}(t) \leq m_{ij}(t) \leq u_{ij}(t) =  \frac{3\gamma_{ij}}{2\sqrt{N}} \exp \left(2(1 + C_0) \beta \sqrt{M}\lambda_i \lambda_j t \right),
\end{equation}
for every \(t \leq \mathcal{T}_{L_{0,\beta}}^{(ij)} \wedge \min_{1 \leq k, \ell \leq r }\mathcal{T}_{\varepsilon}^{(k \ell)} \wedge T_0\). We then define \(T_{\ell, \varepsilon}^{(ij)}\) by \(\ell_{ij}(T_{\ell, \varepsilon}^{(ij)}) = \varepsilon\), i.e.,
\begin{equation} \label{eq: T lower bound p=2 2} 
T_{\ell, \varepsilon}^{(ij)} = \frac{\log(\varepsilon \sqrt{N}) - \log(\gamma_{ij}/2)}{2 (1-C_0) \beta \sqrt{M}\lambda_i\lambda_j}.
\end{equation}
Similarly, we let \(T_{u, \varepsilon}^{(ij)}\) satisfy \(u_{ij}(T_{u, \varepsilon}^{(ij)}) = \varepsilon\), so that
\begin{equation}  \label{eq: T upper bound p=2 2} 
T_{u, \varepsilon}^{(ij)} = \frac{\log(\varepsilon \sqrt{N})- \log (3 \gamma_{ij}/2)}{2 (1+C_0) \beta \sqrt{M} \lambda_i\lambda_j}.
\end{equation}
We note that on the event \(\mathcal{A}^{(ij)}\), \(T_{u, \varepsilon}^{(ij)} \leq \mathcal{T}_{\varepsilon}^{(ij)} \leq T_{\ell, \varepsilon}^{(ij)}\). Moreover, we have 
\[
T_{\ell, \varepsilon}^{(11)} \leq \frac{\log(\varepsilon \sqrt{N}) - \log(\gamma_2/2)}{2(1-C_0)\beta \sqrt{M}\lambda_i \lambda_j (1+\kappa_1)} \leq \frac{\log(\varepsilon \sqrt{N}) - \log(3\gamma_1/2)}{2(1+C_0)\beta \sqrt{M}\lambda_i \lambda_j} \leq T_{u,\varepsilon}^{(ij)},
\]
where we used the fact that by assumption \( \log(\varepsilon \sqrt{N}) \geq \log(3 \gamma_1/2) + \log(3 \gamma_1/\gamma_2) \frac{1+C_0}{(1-C_0)(1+\kappa_1) -(1+C_0)}\). We choose \(T_0 \coloneqq T_{\ell, \varepsilon}^{(11)}\).

In order to deduce that \(\mathcal{T}_\varepsilon^{(11)} = \min_{1 \leq k, \ell \leq r} \mathcal{T}_\varepsilon^{(k\ell)}\), we have to show that on the event \(\cap_{1 \leq k, \ell \le r} \mathcal{A}^{(k \ell)}\) it holds that \(\min_{1 \leq k, \ell \leq r} \mathcal{T}_\varepsilon^{(k \ell)} \leq \min_{1 \leq k, \ell \leq r} \mathcal{T}_{L_{0,\beta}}^{(k \ell)}\) with high \(\mathbb{Q}_{\boldsymbol{X}}\)- and \(\mathbb{P}\)-probability. To this end, we wish to apply Lemma~\ref{lem: bounding flows} to the function \(F_{ij}(\boldsymbol{X}) = L_{0,\beta} m_{ij}(\boldsymbol{X})\). We refer the reader to the proof of Lemma~\ref{lem: weak recovery first spike Langevin p=2} for a proof of conditions (1)-(4). We therefore have that there exists a constant \(K>0\) such that on the event \(\cap_{1 \le k, \ell \le r} \mathcal{A}^{(k \ell)}\),
\begin{equation} \label{eq: bound L0m_ij p=2 2}
|L_{0,\beta} m_{ij}(t)| \leq K \left(\frac{\gamma_0}{\sqrt{N}} \sum_{k=0}^{n-1}t^k + t^n + 2 \sum_{1 \leq k ,\ell \leq r} \int_0^t |a_{k \ell}(s)| ds\right),
\end{equation}
for \(t \leq \min_{1 \leq k, \ell \leq r} \mathcal{T}_{L_{0,\beta}}^{(k \ell)}  \wedge \min_{1 \leq k, \ell \leq r}\mathcal{T}_{\varepsilon}^{(k \ell)} \wedge T_0\), with \(\mathbb{Q}_{\boldsymbol{X}}\)-probability at least \(1 - K_2 \exp(- \gamma_0^2 /( K_2 T_0))\) and with \(\mathbb{P}\)-probability at least \(1 - \exp(-K N)\). Our goal is to show that, on the event \(\cap_{1 \leq k,\ell \leq r} \mathcal{A}^{(k \ell)}\), \(\min_{1 \leq k, \ell \leq r} \mathcal{T}_\varepsilon^{(k \ell)} \leq \min_{1 \leq k, \ell \leq r} \mathcal{T}_{L_{0,\beta}}^{(k \ell)}\) so that we have \(\mathcal{T}_\varepsilon^{(11)} = \min_{1 \le k, \ell \leq r} \mathcal{T}_\varepsilon^{(k \ell)}\). It therefore suffices to show that for every \(i,j \in [r]\), each term in the sum~\eqref{eq: bound L0m_ij p=2 2} is upper bounded by \(\frac{C_0 \beta \sqrt{M} \lambda_i \lambda_j}{n+2} m_{ij}(t)\) for every \(t \leq \min_{1 \leq k, \ell \leq r} \mathcal{T}_{L_{0,\beta}}^{(k \ell)} \wedge \min_{1 \leq k, \ell \leq r} \mathcal{T}_\varepsilon^{(k \ell)} \wedge T_0\). 
\begin{itemize}
\item[(i)] We start by observing that on the event \(\cap_{1 \leq k, \ell \leq r} \mathcal{A}^{(k \ell)}\),  
\[
\frac{C_0 \beta \sqrt{M} \lambda_i \lambda_j}{n+2} m_{ij}(t) \geq \frac{C_0 \beta \sqrt{M} \lambda_i \lambda_j}{n+2} \ell_{ij}(t) = \frac{C_0 \beta \sqrt{M} \lambda_i \lambda_j}{n+2} \frac{\gamma_{ij}}{2\sqrt{N}} \exp(2(1-C_0) \beta \sqrt{M} \lambda_i \lambda_j t),
\]
for \(t \leq \min_{1 \leq k, \ell \leq r} \mathcal{T}_{L_{0,\beta}}^{(k \ell)} \wedge \min_{1 \leq k, \ell \leq r}\mathcal{T}_\varepsilon^{(k \ell)} \wedge T_0\). We then see that for every \(0 \leq k \leq n-1\),
\[
\exp(2(1-C_0) \beta \sqrt{M} \lambda_i \lambda_j t) \geq \frac{(2(1-C_0) \beta \sqrt{M} \lambda_i \lambda_j)^k}{k!} t^k.
\]
Therefore, a sufficient condition for the first estimate to hold is given by 
\[
\frac{K \gamma_0}{\sqrt{N}}t^k \leq \frac{C_0 \beta \sqrt{M} \lambda_i \lambda_j}{n+2} \frac{\gamma_{ij}}{2\sqrt{N}}\frac{(2(1-C_0) \beta \sqrt{M} \lambda_i \lambda_j)^k}{k!} t^k,
\]
which implies that
\[
\sqrt{M} \geq \frac{1}{\beta \lambda_i \lambda_j} \left (\frac{K \gamma_0 (n+2) k!}{2^{k-1} C_0 (1-C_0)^k \gamma_{ij}} \right )^{\frac{1}{k+1}} 
\]
for every \(0 \leq k \leq n-1\), which certainly holds by assumption.

\item[(ii)] We next observe that a sufficient condition to control the second term is given by \(F(t) = Kt^n \leq \frac{C_0 \beta \sqrt{M} \lambda_i \lambda_j}{n+2} \ell_{ij}(t) = G(t)\). An explicit computation shows that for every \(k \leq n\), 
\[
F^{(k)}(t) = K n (n-1) \cdots (n-k+1)t^{n-k}
\]
and 
\[
G^{(k)}(t) = \frac{C_0 \beta \sqrt{M} \lambda_i \lambda_j}{n+2} \frac{\gamma_{ij}}{2\sqrt{N}} \left (2(1-C_0) \beta \sqrt{M} \lambda_i \lambda_j \right )^k \exp( 2 (1-C_0) \beta \sqrt{M} \lambda_i \lambda_j t).
\]
For \(k \leq n-1\), it holds that \(F^{(k)}(0) = 0 \leq G^{(k)}(0)\), and for \(k=n\) we have
\[
\begin{split}
G^{(n)}(t) & = \frac{\left (\beta \sqrt{M} \lambda_i \lambda_j \right)^{n+1} \gamma_{ij} C_0 2^{n-1} (1-C_0)^n}{(n+2) \sqrt{N}}  \exp( 2 (1-C_0) \beta \sqrt{M} \lambda_i \lambda_j t)\\
& \gtrsim \frac{2^{n-1} (n+2)^n \gamma_0^{n+1}}{\gamma_2^n C_0^n (1-C_0)} \exp((n+2) \gamma_0 N^{\frac{1}{2(n+1)}}t)\\
& \geq Kn! = F^{(n)}(t),
\end{split}
\]
which gives the bound for the second term.
\item[(iii)] To bound the last term, we note from~\eqref{eq: integral inequality p=2 2} that on the event \(\cap_{1 \leq k, \ell \leq r} \mathcal{A}^{(k \ell)}\),
\[
2\sum_{1 \leq k, \ell \leq r} \int_0^t \vert a_{k \ell}(s) \vert ds \leq \frac{2 r^2}{1-C_0} \max_{1 \leq k, \ell \leq r} m_{k \ell}(t) \leq \frac{2 r^2}{1-C_0} \max_{1 \leq k, \ell \leq r} u_{k \ell}(t),
\]
for \(t \leq \min_{1 \leq k, \ell \leq r} \mathcal{T}_{L_{0,\beta}}^{(k \ell)} \wedge  \min_{1 \leq k, \ell \leq r}  \mathcal{T}_\varepsilon^{(k \ell)} \wedge T_0\). Therefore a sufficient condition for the last estimate is given by 
\[
\frac{2 r^2}{1-C_0} \max_{1 \leq k, \ell \leq r} u_{k \ell}(t) \leq \frac{C_0 \beta \sqrt{M} \lambda_i \lambda_j}{n+2} \ell_{ij}(t),
\]
which implies that 
\[
\sqrt{M} \geq \frac{6r^2 (n+2)}{C_0 (1-C_0) \beta \lambda_i \lambda_j \gamma_{ij}} \max_{1 \le k, \ell \le r} \{\gamma_{k \ell} \exp(2\beta \sqrt{M} \left ( (1+C_0) \lambda_k \lambda_\ell - (1-C_0) \lambda_i \lambda_j \right ) t)\},
\]
for \(t \leq \min_{1 \leq k, \ell \leq r} \mathcal{T}_{L_{0,\beta}}^{(k \ell)} \wedge \min_{1 \leq k, \ell \leq r}  \mathcal{T}_\varepsilon^{(k \ell)} \wedge T_0\). Since the right-hand side is increasing, the maximum value of this function is attained at \(T_0\). We therefore obtain that
\[
\sqrt{M} \geq \frac{6r^2 (n+2) \gamma_1}{C_0 (1-C_0) \beta \lambda_i \lambda_j \gamma_{ij}} \left ( \frac{2 \varepsilon \sqrt{N}}{\gamma_{11}}\right )^{\frac{1+C_0}{1-C_0} - \frac{\lambda_i \lambda_j}{\lambda_1^2}},
\]
which holds by assumption.
\end{itemize}
We therefore have that, on the event \(\cap_{1 \leq k, \ell \leq r} \mathcal{A}^{(k \ell)}\), 
\(\min_{1 \leq k, \ell \leq r}\mathcal{T}_\varepsilon^{(k \ell)} \leq \min_{1 \leq k, \ell \leq r} \mathcal{T}_{L_{0,\beta}}^{(k \ell)}\), so that  
\[
\min_{1 \leq k, \ell \leq r}\mathcal{T}_\varepsilon^{(k \ell)} = \mathcal{T}_\varepsilon^{(11)} \lesssim \frac{\gamma_2 \log(\varepsilon \sqrt{N}) N^{-\delta(C_0)/2}}{(n+2) \gamma_0 \gamma_1},
\]
with \(\mathbb{Q}_{\boldsymbol{X}}\)-probability at least \(1 - K_2 \exp (- \gamma_0^3 \gamma_1 (n+2) N^{\delta(C_0)/2}/ ( K_2 \gamma_2 \log(N)))\) and with \(\mathbb{P}\)-probability at least \(1 - \exp(-K N)\). 

Furthermore, by the comparison estimate~\eqref{eq: comparison inequality p=2 2}, on the event \(\cap_{1 \leq k, \ell \leq r} \mathcal{A}^{(k\ell)}\), we have for every \((i,j) \neq (1,1)\),
\[
m_{ij}(\mathcal T^{(11)}_{\varepsilon})\le u_{ij}(\mathcal T^{(11)}_{\varepsilon}) \le u_{ij}(T^{(11)}_{\ell,\varepsilon}),
\]
where we used \(\mathcal T^{(11)}_{\varepsilon}\le T^{(11)}_{\ell,\varepsilon}\) and that \(u_{ij}\) is increasing. Using~\eqref{eq: T lower bound p=2 2} yields
\[
u_{ij}(T^{(11)}_{\ell,\varepsilon}) =\frac{3\gamma_{ij}}{2\sqrt N}\exp\!\left(2(1+C_0)\beta\sqrt M\,\lambda_i\lambda_j T^{(11)}_{\ell,\varepsilon}\right) = C_1^{(ij)}\,N^{-\delta_{ij}^{(1)}(C_0)/2},
\]
where \(C_1^{(ij)} \coloneqq \frac{3}{2}\gamma_{ij} \left (\frac{2 \varepsilon}{\gamma_{11}} \right )^{\frac{1+C_0}{1-C_0} \frac{\lambda_i \lambda_j}{\lambda_1^2}}\). Similarly, on the event \(\cap_{1 \leq k, \ell \leq r} \mathcal{A}^{(k\ell)}\), we have for every \((i,j) \neq (1,1)\),
\[
m_{ij}(\mathcal T^{(11)}_{\varepsilon}) \ge \ell_{ij}(\mathcal T^{(11)}_{\varepsilon}) \ge \ell_{ij}(T^{(11)}_{u,\varepsilon}),
\]
since \(\mathcal{T}_\varepsilon^{(11)} \ge T_{u,\varepsilon}^{(11)}\) and \(\ell_{ij}\) is increasing. Using~\eqref{eq: T upper bound p=2 2} gives
\[
\ell_{ij}(T^{(11)}_{u,\varepsilon})
=\frac{\gamma_{ij}}{2\sqrt N}\exp\!\left(2(1-C_0)\beta\sqrt M\,\lambda_i\lambda_j T^{(11)}_{u,\varepsilon}\right)
= C_2^{(ij)}\,N^{-\xi_{ij}^{(1)}(C_0)/2},
\]
where \(C_2^{(ij)} \coloneqq \frac{\gamma_{ij}}{2} \left (\frac{2 \varepsilon}{3 \gamma_{11}} \right )^{\frac{1-C_0}{1+C_0} \frac{\lambda_i \lambda_j}{\lambda_1^2}}\). Since \(\gamma_{ij} \in (\gamma_2, \gamma_1)\) and \(\lambda_i \lambda_j \le \lambda_1^2\), the constants \(C_1^{(ij)}\) and \(C_2^{(ij)}\) are bounded above and below by positive constants depending only on \(\varepsilon, \gamma_1, \gamma_2, C_0\), and \(\{\lambda_i\}_{i=1}^r\). Consequently, at time \(\mathcal T^{(11)}_{\varepsilon}\) we have \(m_{11}\) reaches \(\varepsilon\) and the other correlations satisfy the required polynomial bounds, so that \(\boldsymbol{X}^\beta_{\mathcal T^{(11)}_{\varepsilon}}\in E_1(\varepsilon,C_0)\). This completes the proof of Lemma~\ref{lem: weak recovery first spike Langevin p=2 2}.
\end{proof}

\begin{proof}[\textbf{Proof of Lemma~\ref{lem: weak recovery k spike Langevin p=2}}]
We show Lemma~\ref{lem: weak recovery k spike Langevin p=2} for \(k=2\), that is, we prove the one-step implication \(E_1(\varepsilon,C_0)\to E_2(\varepsilon,C_0)\). Fix \(\varepsilon\in(0,\varepsilon_0)\) and \(C_0\in(0,c_0)\), and assume \(\boldsymbol{X}_0 \in E_1(\varepsilon,C_0)\). 

Throughout, for every \(a>0\) and \(i,j\in [r]\), we define the hitting times \(\mathcal{T}_a^{(ij,+)}\) and \(\mathcal{T}_a^{(ij,-)}\) by
\[
\begin{split}
\mathcal{T}_a^{(ij,+)} & = \inf \{t \ge 0 \colon m_{ij}(t) \ge a\},\\
\mathcal{T}_a^{(ij,-)} & = \inf \{t \ge 0 \colon m_{ij}(t) \le a\}.
\end{split}
\]
Fix an intermediate threshold \(\varepsilon'\in(\varepsilon,1)\) to be chosen later. Fix also a small \(\delta>0\) and define \(\delta_N^{(ij)} \coloneqq N^{- \frac{\delta_{ij}^{(1)} -\delta}{2}}\).

\textbf{Step 1: Growth of \(m_{11}(t)\) until it reaches \(\varepsilon'\).}
Recall the generator expansion given in Lemma~\ref{lem: evolution equation m_ij} for \(p=2\):
\[
L_\beta m_{11}(t) = L_{0,\beta}m_{11}(t)+2\beta\sqrt M\,\lambda_1^2 m_{11}(t)
-\beta\sqrt M\sum_{k,\ell}\lambda_k(\lambda_1+\lambda_\ell)m_{k1}(t)m_{1\ell}(t)m_{k\ell}(t).
\]
For every \(t \leq \mathcal{T}_{\varepsilon'}^{(11,+)} \wedge \mathcal{T}_{\varepsilon/2}^{(11,-)} \wedge \min_{2 \le i,j \le r} \left \{ \mathcal{T}_{\delta_N^{(1j)}}^{(1j,+)} \wedge \mathcal{T}_{\delta_N^{(i1)}}^{(i1,+)} \wedge \mathcal{T}_{\delta_N^{(ij)}}^{(ij,+)} \right \}\), 
\[
\begin{split}
L_\beta m_{11} (t) & \geq - \norm{L_{0,\beta} m_{11}}_\infty + 2 \beta \sqrt{M} \lambda_1^2 m_{11}(t) \left ( 1 - m_{11}^2(t) - (r^2-1) \max_{2 \le i,j \le r} \{m_{i1}(t) m_{1j}(t)\} \right) \\
& \geq - \norm{L_{0,\beta} m_{11}}_\infty + 2 \beta \sqrt{M} \lambda_1^2 m_{11}(t) \left ( 1 - m_{11}^2(t) - (r^2-1) N^{- (\delta_{12}-\delta)} \right) \\
&  \geq - \norm{L_{0,\beta} m_{11}}_\infty +  \beta \sqrt{M} \lambda_1^2 m_{11}(t) \left ( 1 - m_{11}^2(t) \right),
\end{split}
\]
 where the last inequality holds provided that \(N > \left (\frac{2(r^2-1)}{1 - (\varepsilon')^2} \right )^{\frac{1}{\delta_{12}^{(1)} -\delta}}\). Therefore, we have 
\[
L_\beta m_{11}(t) \geq - \Lambda + \beta \sqrt{M} \lambda_1^2 m_{11}(t) (1-m_{11}^2(t)) \geq \frac{1}{2} \beta \sqrt{M} \lambda_1^2 m_{11}(t) (1 - m_{11}^2(t)) > 0,
\]
with \(\mathbb{P}\)-probability at least \(1 - \exp(-K N)\), provided that \(\sqrt{M} \geq \frac{2 \Lambda}{\beta \lambda_1^2 m_{11}(t) (1 - m_{11}^2(t))}\), which certainly holds by assumption. Recalling that
\[
m_{11}(t) = m_{11}(0) + M_t^{m_{11}} + \int_0^t L_\beta m_{11}(s) ds,
\]
we obtain the integral inequality given by
\begin{equation} \label{eq: lower bound integral inequality p=2}
m_{11}(t) \geq \frac{\varepsilon}{2} + \frac{1}{2}\beta \sqrt{M} \lambda_1^2 \int_0^t m_{11}(s)(1-m_{11}^2(s))ds,
\end{equation}
for every \(t \leq \mathcal{T}_{\varepsilon'}^{(11,+)} \wedge \mathcal{T}_{\varepsilon/2}^{(11,-)} \wedge \min_{2 \le i,j \le r} \left \{ \mathcal{T}_{\delta_N^{(1j)}}^{(1j,+)} \wedge \mathcal{T}_{\delta_N^{(i1)}}^{(i1,+)} \wedge \mathcal{T}_{\delta_N^{(ij)}}^{(ij,+)} \right \} \wedge T_0\), with \(\mathbb{Q}_{\boldsymbol{X}}\)-probability at least \(1- K_1 \exp \left(- \varepsilon^2 N / (4K_1T_0) \right)\) and with \(\mathbb{P}\)-probability at least \(1 - \exp(-K N)\). Since the function \(x \mapsto x (1-x^2)\) is locally Lipschitz on \((\varepsilon,\infty)\) for every \(\varepsilon >0\), the equation
\[
\begin{cases}
\dot f =  a f(1-f^2)\\
f(0) = b
\end{cases},
\]
admits as unique solution \(f(t) = e^{at} \left ( e^{2at}+\frac{1}{b^2}-1\right)^{-\frac{1}{2}}\). It then follows from~\eqref{eq: lower bound integral inequality p=2} that 
\[
m_{11}(t) \geq \frac{\exp(\beta \sqrt{M}\lambda_1^2 t/2)}{\sqrt{ \exp(\beta \sqrt{M} \lambda_1^2 t) +\frac{4}{\varepsilon^2}-1}},
\]
for every \(t \leq \mathcal{T}_{\varepsilon'}^{(11,+)} \wedge \mathcal{T}_{\varepsilon/2}^{(11,-)} \wedge \min_{2 \le i,j \le r} \left \{ \mathcal{T}_{\delta_N^{(1j)}}^{(1j,+)} \wedge \mathcal{T}_{\delta_N^{(i1)}}^{(i1,+)} \wedge \mathcal{T}_{\delta_N^{(ij)}}^{(ij,+)} \right \} \wedge T_0\). We deduce that \(\mathcal{T}_{\varepsilon'}^{(11,+)} \leq \mathcal{T}_{\varepsilon/2}^{(11,-)}\). In particular, we have
\begin{equation} \label{eq: T varepsilon'}
\mathcal{T}_{\varepsilon'}^{(11,+)} \wedge \min_{2 \le i,j \le r} \left \{ \mathcal{T}_{\delta_N^{(1j)}}^{(1j,+)} \wedge \mathcal{T}_{\delta_N^{(i1)}}^{(i1,+)} \wedge \mathcal{T}_{\delta_N^{(ij)}}^{(ij,+)} \right \} \wedge T_0  \leq \frac{1}{\beta \sqrt{M} \lambda_1^2} \log \left(\frac{(\varepsilon')^2}{1-(\varepsilon')^2} \frac{4 - \varepsilon^2}{\varepsilon^2}\right),
\end{equation}
with \(\mathbb{Q}_{\boldsymbol{X}}\)-probability at least \(1- K_1 \exp \left(- \varepsilon^2 N / (4 K_1 T_0) \right)\) and with \(\mathbb{P}\)-probability at least \(1 - \exp(-K N)\).\\

\textbf{Step 2: Evolution of \(m_{i1}(t)\), \(m_{1j}(t)\) and \(m_{ij}(t)\) for every \(t \leq \mathcal{T}_{\varepsilon'}^{(11,+)}\).}
We first observe that for \(i,j \neq 1\),
\[
\begin{split}
L_\beta m_{i1}(t) & \leq \norm{L_{0,\beta} m_{i1}}_\infty + 2 \beta \sqrt{M} \lambda_1 \lambda_i m_{i1}(t) - \beta \sqrt{M} \sum_{1 \leq k, \ell \leq r} \lambda_k (\lambda_1 + \lambda_\ell) m_{k1}(t) m_{i \ell}(t) m_{k \ell}(t) \\
& \leq \Lambda + 2 \beta \sqrt{M} \lambda_1 \left (\lambda_i - \lambda_1 m_{11}^2(t) \right ) m_{i1} (t),
\end{split}
\]
and similarly,
\[
\begin{split}
L_\beta m_{1j}(t) & \leq \norm{L_{0,\beta} m_{1j}}_\infty + 2 \beta \sqrt{M} \lambda_1 \lambda_j m_{1j}(t) - \beta \sqrt{M} \sum_{1 \leq k, \ell \leq r} \lambda_k (\lambda_j + \lambda_\ell) m_{kj}(t) m_{1\ell}(t) m_{k \ell}(t) \\
& \leq \Lambda + \beta \sqrt{M} \lambda_1 \left (2\lambda_j - (\lambda_1+\lambda_j) m_{11}^2(t) \right ) m_{1j} (t),
\end{split}
\]
for every \(t \leq \mathcal{T}_{\varepsilon'}^{(11,+)} \wedge \min_{2 \le i,j \le r} \left \{ \mathcal{T}_{\delta_N^{(1j)}}^{(1j,+)} \wedge \mathcal{T}_{\delta_N^{(i1)}}^{(i1,+)} \wedge \mathcal{T}_{\delta_N^{(ij)}}^{(ij,+)} \right \}\), with \(\mathbb{P}\)-probability at least \(1 - \exp(-K N)\). In particular, we note that the correlation \(m_{i1}\) starts decreasing as soon as  
\[
m_{11}^2 > \frac{\lambda_i}{\lambda_1} = \frac{1}{\prod_{\ell=1}^{i-1} (1 + \kappa_\ell)}, 
\]
so that as \(m_{11}^2\) exceeds \(\frac{1}{1+\kappa_1}\), all correlations \(m_{21},\ldots, m_{r1}\) are decreasing. In the same way, we see that the correlation \(m_{1j}\) starts decreasing as soon as 
\[
m_{11}^2 > \frac{2 \lambda_j}{\lambda_1 + \lambda_j} = \frac{2}{1 + \prod_{\ell=1}^{j-1} (1+\kappa_\ell)}, 
\]
so that the condition \(m_{11}^2 > \frac{2}{2+\kappa_1}\) ensures that \(m_{12},\ldots, m_{1r}\) are decreasing. Since \(\frac{1}{1 + \kappa_1} < \frac{2}{2 + \kappa_1}\), we deduce that, as soon as \(m_{11}^2 = \frac{2}{2+\kappa_1} + \omega \) for \(0 < \omega < \frac{\kappa_1}{2 + \kappa_1}\), all correlations \(m_{i1}\) and \(m_{1j}\) are decreasing. Without loss of generality, we may assume that \(\omega = \frac{1}{2} \frac{\kappa_1}{2+\kappa_1}\), and we set
\[
(\varepsilon')^2 \coloneqq  \frac{4 + \kappa_1}{2(2+\kappa_1)}.
\]
According to~\eqref{eq: T varepsilon'}, we have
\begin{equation} \label{upper bound time p=2}
\begin{split}
\mathcal{T}_{\varepsilon'}^{(11,+)} \wedge \min_{2 \le i,j \le r} \left \{ \mathcal{T}_{\delta_N^{(1j)}}^{(1j,+)} \wedge \mathcal{T}_{\delta_N^{(i1)}}^{(i1,+)} \wedge \mathcal{T}_{\delta_N^{(ij)}}^{(ij,+)} \right \} \wedge T_0 & \leq \frac{1}{\beta \sqrt{M} \lambda_1^2} \log \left ( \frac{4+\kappa_1}{\kappa_1} \frac{4 - \varepsilon^2}{\varepsilon^2} \right ) ,
\end{split}
\end{equation}
with \(\mathbb{Q}_{\boldsymbol{X}}\)-probability at least \(1- K_2 \exp \left(- \varepsilon^2 N / (4 K_2 T_0) \right)\) and with \(\mathbb{P}\)-probability at least \(1 - \exp(-K N)\). For the correlations \(m_{1i}\) and \(m_{i1}\) to change scales during this interval, it requires a time of order \(\log(N)\), thus they are still upper bounded in a scale \(N^{-\frac{\delta_{1i}^{(1)}}{2}}\), in particular,
\[
\mathcal{T}_{\varepsilon'}^{(11,+)} \wedge \min_{2 \le i,j \le r}  \mathcal{T}_{\delta_N^{(ij)}}^{(ij,+)} \wedge T_0  \le \min_{2 \le i,j \le r} \{\mathcal{T}_{\delta_N^{(1j)}}^{(1j,+)} \wedge \mathcal{T}_{\delta_N^{(i1)}}^{(i1,+)}\}.
\] 

It remains to consider the evolution of the correlations \(m_{ij}\) for \(2 \le i,j \le r\) during \([0,  \mathcal{T}_{\varepsilon'}^{(11,+)}]\). We aim to show that during this interval the functions \(m_{ij}\) remain increasing. We first observe that for \(2 \le i,j \le r\),
\[
\begin{split}
L_\beta m_{ij}(t) & \ge - \norm{L_{0,\beta}m_{ij}}_\infty + 2 \beta \sqrt{M} \lambda_i \lambda_j m_{ij}(t) - \beta \sqrt{M} \sum_{1 \le k, \ell \leq r} \lambda_k (\lambda_j + \lambda_\ell) m_{kj}(t) m_{i\ell}(t) m_{k \ell}(t) \\ 
& \ge - \Lambda + 2 \beta \sqrt{M} \lambda_i \lambda_j m_{ij}(t) - 2r^2 \beta \sqrt{M} \lambda_1^2 m_{1j}(t) m_{i1}(t) m_{11}(t),
\end{split}
\]
for every \(t \le \mathcal{T}_{\varepsilon'}^{(11,+)} \wedge \min_{2 \le i,j \le r}  \mathcal{T}_{\delta_N^{(ij)}}^{(ij,+)} \wedge T_0 \), with \(\mathbb{P}\)-probability at least \(1 - \exp(-K  N)\). Let \(T_1^{(ij)}\) denote the hitting time for the set
\[
\left \{ \boldsymbol{X} \colon 2 r^2 \lambda_1^2 m_{11}(\boldsymbol{X}) m_{i1}(\boldsymbol{X}) m_{1j}(\boldsymbol{X}) > C_0 \lambda_i \lambda_j m_{ij}(\boldsymbol{X}) \right \}.
\]
Then, for every \(t \le \mathcal{T}_{\varepsilon'}^{(11,+)} \wedge \min_{2 \le i,j \le r}  \mathcal{T}_{\delta_N^{(ij)}}^{(ij,+)} \wedge T_0 \wedge T_1^{(ij)}\),
\[
\begin{split}
L_\beta m_{ij}(t) & \ge - \Lambda + 2 \beta \sqrt{M} \lambda_i \lambda_j m_{ij}(t) - 2r^2 \beta \sqrt{M} \lambda_1^2 m_{1j}(t) m_{i1}(t) m_{11}(t)\\
& \ge 2(1-C_0) \beta \sqrt{M} \lambda_i \lambda_j m_{ij}(t),
\end{split}
\]
with \(\mathbb{P}\)-probability at least \(1 - \exp(-K  N)\), provided \(\sqrt{M} \geq \frac{\Lambda}{C_0 \beta \lambda_i \lambda_j m_{ij}(t)}\), which certainly holds by assumption. From this, for every \(t \le \mathcal{T}_{\varepsilon'}^{(11,+)} \wedge \min_{2 \le i,j \le r}  \mathcal{T}_{\delta_N^{(ij)}}^{(ij,+)} \wedge T_0 \wedge T_1^{(ij)}\), we obtain
\[
\begin{split}
m_{ij}(t) & \geq  m_{ij}(0) + M_t^{m_{ij}} + 2 (1-C_0) \beta \sqrt{M} \lambda_i \lambda_j \int_0^t m_{ij}(s) ds \\
& \geq \frac{N^{-\frac{\xi_{ij}^{(1)}}{2}}}{2} +2(1-C_0) \beta \sqrt{M} \lambda_i \lambda_j \int_0^t m_{ij}(s) ds \\
& \ge \frac{N^{-\frac{\xi_{ij}^{(1)}}{2}} }{2} \exp \left ( 2(1-C_0) \beta \sqrt{M} \lambda_i \lambda_j t \right ),
\end{split}
\]
with \(\mathbb{Q}_{\boldsymbol{X}}\)-probability at least \(1 - K_2 \exp(- N^{\frac{1-C_0}{1+C_0} \frac{\lambda_i \lambda_j}{\lambda_1^2}} / (4 K_2 T_0))\) and with \(\mathbb{P}\)-probability at least \(1 - \exp(-K N)\), where the last inequality follows from item (d) of Lemma~\ref{lem: Gronwall}. We denote by \(\ell_{ij}\) the function given by
\[
\ell_{ij}(t) = \frac{N^{-\frac{\xi_{ij}^{(1)}}{2}} }{2} \exp \left ( 2(1-C_0) \beta \sqrt{M} \lambda_i \lambda_j t \right ).
\]
Furthermore, we always have the brutal upper bound on the generator \(L_\beta m_{ij}\), i.e., for every \(i,j \in [r]\) and \(t \geq 0\) it holds that
\[
L_\beta m_{ij}(t) \leq \Lambda + 2 \beta \sqrt{M} \lambda_i \lambda_j m_{ij}(t) \leq 2 (1+C_0)\beta \sqrt{M} \lambda_i \lambda_j m_{ij}(t),
\]
with \(\mathbb{P}\)-probability at least \(1 - \exp(-K N)\). Therefore, we find that, on the same time interval,
\[
m_{ij}(t) \leq \frac{m_{ij}(0)}{2} + 2(1+C_0) \beta \sqrt{M} \lambda_i \lambda_j \int_0^t m_{ij}(s) ds \leq \frac{N^{-\delta_{ij}^{(1)}/2}}{2} \exp \left (2(1+C_0) \beta \sqrt{M} \lambda_i \lambda_j t \right),
\]
with \(\mathbb{Q}_{\boldsymbol{X}}\)-probability at least \(1 - K_2 \exp(- N^{\frac{1+C_0}{1-C_0} \frac{\lambda_i \lambda_j}{\lambda_1^2}}/ (4K_2 T_0))\) and with \(\mathbb{P}\)-probability at least \(1 - \exp(-K N)\). We denote by \(u_{ij}\) the function given by  
\[
u_{ij}(t) =  \frac{N^{-\delta_{ij}^{(1)}/2}}{2} \exp \left (2(1+C_0) \beta \sqrt{M} \lambda_i \lambda_j t \right).
\]

A sufficient condition to show that \(\mathcal{T}_{\varepsilon'}^{(11,+)} \wedge T_0 \leq \min_{2 \le k, \ell \le r} T_1^{(k\ell)}\) is therefore given by \(\mathcal{T}_{\varepsilon'}^{(11,+)} \wedge T_0 \leq \min_{2 \le k, \ell \le r} \tilde{T}_1^{(k\ell)}\), where \(\tilde{T}_1^{(ij)}\) denotes the hitting time for the set
\[
\left \{ \boldsymbol{X} \colon 2 r^2 \lambda_1^2 u_{11}(\boldsymbol{X}) u_{i1}(\boldsymbol{X}) u_{1j}(\boldsymbol{X})  > C_0 \lambda_i \lambda_j \ell_{ij}(\boldsymbol{X}) \right \}.
\]
The inequality defining the above event can be written as
\[
2 r^2 \lambda_1^2 \varepsilon' \frac{N^{-\delta_{1i}^{(1)}/2}}{2} \frac{N^{-\delta_{1j}^{(1)}/2}}{2} e^{2(1+C_0) \beta \sqrt{M} \lambda_1 (\lambda_i + \lambda_j) t} > C_0 \lambda_i \lambda_j \frac{N^{-\xi_{ij}^{(1)}/2}}{2} e^{2(1-C_0)\beta \sqrt{M} \lambda_i \lambda_j t}, 
\]
yielding an explicit expression for \( \tilde{T}_1^{(ij)}\), i.e., 
\[
\tilde{T}_1^{(ij)} = \frac{\frac{1}{2} \left ( \delta_{1i}^{(1)} + \delta_{1j}^{(1)} - \xi_{ij}^{(1)} \right ) \log(N) + \log(C_0 \lambda_i \lambda_j /(r^2 \varepsilon' \lambda_1^2)) }{2 \beta \sqrt{M} \left ( (1+C_0) \lambda_1 (\lambda_i + \lambda_j) - (1-C_0) \lambda_i \lambda_j\right )}.
\]
We note that the term \(\delta_{1i}^{(1)} + \delta_{1j}^{(1)} - \xi_{ij}^{(1)}\), which is given by
\[
\delta_{1i}^{(1)} + \delta_{1j}^{(1)} - \xi_{ij}^{(1)}  = 1 +\frac{1-C_0}{1+C_0} \frac{\lambda_i \lambda_j}{\lambda_1^2} - \frac{1+C_0}{1-C_0} \frac{\lambda_i + \lambda_j}{\lambda_1},
\]
is positive, provided \(C_0\) sufficiently small. In particular, if \(\kappa = \min_{1 \le i \le r-1} \kappa_i\) is sufficiently large then we can easily found \(C_0 < \frac{1}{2} \wedge \frac{\kappa}{1 + \kappa}\) such that \(\delta_{1i}^{(1)} + \delta_{1j}^{(1)} - \xi_{ij}^{(1)}>0\). Otherwise, we need to take \(C_0\) sufficiently close to zero. Therefore, we can find \(c_0 = c_0(\kappa)\) such that for every \(C_0 < c_0\) the quantity \(\delta_{1i}^{(1)} + \delta_{1j}^{(1)} - \xi_{ij}^{(1)}\) is in \((0,1)\). In addition, we see that \(\tilde{T}_1^{(22)} = \min_{2 \leq k, \ell \leq r} \tilde{T}_1^{(k \ell)}\). A sufficient condition for \(\mathcal{T}_{\varepsilon'}^{(11,+)} \wedge T_0 \leq \min_{2 \le k, \ell \le r} \tilde{T}_1^{(k\ell)}\) is given by
\[
\log(N) \geq \frac{4}{2 \delta_{12}^{(1)} - \xi_{22}^{(1)}} \frac{1}{(1+\kappa_1)^2} \left (2(1+C_0) (1+\kappa_1) - (1-C_0) \right)  \log \left ( \frac{4+\kappa_1}{\kappa_1} \frac{4 - \varepsilon^2}{\varepsilon^2} \right ),
\]
which holds by assumption:
\[
\log(N) \gtrsim  \frac{4}{1+\kappa_1}  \log \left ( \frac{4+\kappa_1}{\kappa_1} \frac{4 - \varepsilon^2}{\varepsilon^2} \right ).
\]
Therefore, choosing \(T_0\) to be the upper bound given in~\eqref{upper bound time p=2}, we have that \(\mathcal{T}_{\varepsilon'}^{(11,+)} \leq \tilde{T}_1^{(ij)} \leq T_1^{(ij)}\) for all \(2 \le i,j \le r\), with \(\mathbb{Q}_{\boldsymbol{X}}\)-probability at least 
\[
1 - (r-1)^2 K_2 \exp \left ( -N^{\frac{1-C_0}{1+C_0} \frac{\lambda_2^2}{\lambda_1^2}} / (4K_2 T_0) \right ) - 2(r-1)K_2 \exp \left ( -N^{\frac{1+C_0}{1-C_0} \frac{\lambda_2}{\lambda_1}} / (4 K_2 T_0) \right )
\]
and with \(\mathbb{P}\)-probability at least \(1 - \exp(-KN)\). We deduce that \(m_{ij}\) does not become decreasing during \([0,\mathcal{T}_{\varepsilon'}^{(11,+)}]\) and remains at least of order \(N^{-\frac{\xi_{ij}^{(1)}}{2}}\).\\

\textbf{Step 3: Evolution of all \(m_{ij}(t)\) for \(\mathcal{T}_{\varepsilon'}^{(11,+)} \leq t \leq \mathcal{T}_{E_2(\varepsilon,C_0)}\).}
The last step focuses on the evolution of the correlations \(m_{ij}\) during the time interval \(\mathcal{T}_{\varepsilon'}^{(11,+)} \leq t \leq \mathcal{T}_{E_2 (\varepsilon,C_0)}\), where we remind that \((\varepsilon')^2 = \frac{4+\kappa_1}{2(2+\kappa_1)}\).

We start with the correlation \(m_{11}(t)\) and note that for every \(\mathcal{T}_{\varepsilon'}^{(11,+)} \leq t \leq \mathcal{T}_{\varepsilon'/2}^{(11,-)} \wedge \mathcal{T}_{1-\varepsilon}^{(11,+)}\), it holds that 
\[
L_\beta m_{11}(t) \geq - \norm{ L_{0,\beta} m_{11}}_\infty + 2 \beta \sqrt{M} \lambda_1^2 m_{11}(t) \left (1 - r^2 m_{11}^2(t)\right ).
\]
We can then proceed using Lemma~\ref{lem: weak implies strong recovery Langevin p=2} to show that \(m_{11}\) reaches \(1 - \varepsilon\) with \(\mathbb{Q}_{\boldsymbol{X}}\)-probability at least \(1 - K_2 \exp(- N\varepsilon^2 /(K_2 \mathcal{T}_{1-\varepsilon}^{(11,+)}))\) and with \(\mathbb{P}\)-probability at least \(1 - \exp(-K N)\).

We next focus on \(m_{1i}(t)\) and \(m_{i1}(t)\), and observe that for every \(\mathcal{T}_{\varepsilon'}^{(11,+)} \leq t \leq \mathcal{T}_{E_2(\varepsilon,C_0)}\), it holds that 
\[
L_\beta m_{i1}(t) \leq \norm{L_{0,\beta} m_{i1}}_\infty - 2 \beta \sqrt{M} \lambda_1^2 \frac{\kappa_1}{2+\kappa_1} m_{i1}(t) \leq -\beta \sqrt{M} \lambda_1^2 \frac{\kappa_1}{2+\kappa_1} m_{i1}(t),
\]
with \(\mathbb{P}\)-probability at least \(1 - \exp(-K  N)\), provided \(\sqrt{M} \geq \frac{\Lambda (2+\kappa_1)}{\beta \lambda_1^2 \kappa_1 m_{i1}(t)}\), which certainly holds since \(\sqrt{M} \gtrsim N^{\frac{\xi_{rr}^{(1)}}{2}}\). By Grönwall's inequality it then follows that
\[
\begin{split}
m_{i1}(t) & \leq \frac{1}{2} m_{i1}(\mathcal{T}_{\varepsilon'}^{(11,+)}) - \frac{\kappa_1}{2+\kappa_1} \beta \sqrt{M} \lambda_1^2 \int_{\mathcal{T}_{\varepsilon'}^{(11,+)}}^t m_{i1}(s)ds \\
& \leq \frac{1}{2} m_{i1}(\mathcal{T}_{\varepsilon'}^{(11,+)}) e^{- \frac{\kappa_1}{2+\kappa_1} \beta \sqrt{M} \lambda_1^2 ( t - \mathcal{T}_{\varepsilon'}^{(11,+)})}\\
& \leq \frac{N^{-\frac{\delta_{1i}^{(1)}}{2}}}{2} e^{- \frac{\kappa_1}{2+\kappa_1} \beta \sqrt{M} \lambda_1^2 ( t - \mathcal{T}_{\varepsilon'}^{(11,+)})},
\end{split}
\]
with \(\mathbb{Q}_{\boldsymbol{X}}\)-probability at least \(1 - K_3 \exp \left ( -N^{\frac{1+C_0}{1-C_0} \frac{\lambda_2}{\lambda_1}} / (4K_3 \mathcal{T}_{E_2(\varepsilon,C_0)}) \right )\) and with \(\mathbb{P}\)-probability at least \(1 - \exp(-K  N)\). This shows that the correlations \(m_{i1}\) descend to a value of order \(N^{-\frac{\xi_{rr}^{(1)}}{2}}\) in a time of order \(\frac{(\xi_{rr}^{(1)} - \delta_{i1}^{(1)})(2+\kappa_1)}{2\kappa_1 \beta \sqrt{M}\lambda_1^2} \log(N)\). 

We finally look at the correlations \(m_{ij}\) for \(2 \leq i,j \leq r\) and show that \(m_{22}\) is the second correlation to become macroscopic. We see that
\[
\begin{split}
L_\beta m_{ij}(t)  &\geq - \Lambda + 2 \beta \sqrt{M} \lambda_i \lambda_j m_{ij}(t) - \beta \sqrt{M} \sum_{1 \le k, \ell \le r} \lambda_k (\lambda_j + \lambda_\ell) m_{kj}(t) m_{i \ell}(t) m_{k \ell}(t) \\
& \geq - \Lambda + 2 \beta \sqrt{M} \lambda_i \lambda_j m_{ij}(t) - 2 r^2 \beta \sqrt{M} \lambda_1^2 m_{1j}(t) m_{i1}(t) m_{11}(t) \\
& \geq - \Lambda + 2 \beta \sqrt{M} \left ( \lambda_i \lambda_j m_{ij}(t) - r^2 \lambda_1^2 (1-\varepsilon) N^{-\delta_{rr}}\right ) \\
& \geq 2(1-C_0) \beta \sqrt{M} \lambda_i \lambda_j m_{ij}(t).
\end{split}
\]
By Grönwall's inequality (i.e., Lemma~\ref{lem: Gronwall}) we have
\[
\frac{N^{-\xi_{ij}^{(1)}/2}}{2} \exp(2(1-C_0) \beta \sqrt{M} \lambda_i \lambda_j t) \le m_{ij}(t) \leq \frac{N^{-\delta_{ij}^{(1)}/2}}{2} \exp(2(1+C_0) \beta \sqrt{M} \lambda_i \lambda_j t),
\]
with high \(\mathbb{Q}_{\boldsymbol{X}}\)- and \(\mathbb{P}\)-probability. It then follows that 
\[
\mathcal{T}_{\varepsilon}^{(22,+)} \leq \frac{\log(2 \varepsilon N^{\xi_{22}^{(1)} /2})}{2(1-C_0) \beta \sqrt{M} \lambda_2^2} .
\]
Moreover, we see that for every \(i,j \neq 1, (i,j)\neq (2,2)\),
\[
m_{ij}(\mathcal{T}_{\varepsilon}^{(22,+)}) \leq \frac{N^{-\delta_{ij}^{(1)}/2}}{2} \left( 2 \varepsilon N^{\xi_{22}^{(1)}/2} \right)^{\frac{1+C_0}{1-C_0} \frac{\lambda_i \lambda_j}{\lambda_2^2}} \lesssim N^{-\frac{1}{2} \delta_{ij}^{(2)}},
\]
and similarly
\[
m_{ij}(\mathcal{T}_{\varepsilon}^{(22,+)}) \geq \frac{N^{-\xi_{ij}^{(1)}/2}}{2} \left( 2 \varepsilon N^{\delta_{22}^{(1)}/2} \right)^{\frac{1-C_0}{1+C_0} \frac{\lambda_i \lambda_j}{\lambda_2^2}} \gtrsim N^{-\frac{1}{2} \xi_{ij}^{(2)}}.
\]
In particular, at time \(\mathcal T^{(22,+)}_{\varepsilon}\) we have \(m_{22}\) reaches \(\varepsilon\), \(m_{11}\) reaches \(1-\varepsilon\), and the other correlations satisfy the required polynomial bounds, so that \(\boldsymbol{X}^\beta_{\mathcal T^{(22,+)}_{\varepsilon}}\in E_2(\varepsilon,C_0)\). This completes the proof of Lemma~\ref{lem: weak recovery k spike Langevin p=2} for \(k=2\).  
\end{proof}

\section{Proofs for $p=2$ and equal SNRs} \label{section: proof recovery Langevin isotropic p=2}

This section is devoted to the proof of Proposition~\ref{thm: strong recovery isotropic Langevin p=2} for \(p=2\) and equal SNRs \(\lambda_1 = \cdots = \lambda_r \equiv \lambda >0\). We first prove the weak recovery of the largest eigenvalue of \(\boldsymbol{G}\). We introduce the weak-recovery event as follows: for every \(\varepsilon>0\) and \(C_0 \in (0, \frac{1}{r})\), 
\[
E (\varepsilon,C_0) = \left\{\boldsymbol{X} \colon \theta_{\max} (\boldsymbol{X}) \geq \varepsilon \enspace \textnormal{and} \enspace \theta_{\min} (\boldsymbol{X}) \gtrsim  N^{-\frac{2C_0r}{1+C_0r}} \right\},
\]
where \(\theta_i (\boldsymbol{X})\) denotes the \(i\)th eigenvalue of \(\boldsymbol{G} (\boldsymbol{X})\), \(\theta_{\max} (\boldsymbol{X})= \max_{1 \le i \le r} \theta_i (\boldsymbol{X})\) its largest eigenvalue, and \(\theta_{\min} (\boldsymbol{X})= \min_{1 \le i \le r} \theta_i (\boldsymbol{X})\) its smallest eigenvalue. Recall that \(\mathcal{T}_{E(\varepsilon,C_0)}\) denotes the hitting time of \(E(\varepsilon,C_0)\). 

\begin{lem}  \label{lem: weak recovery largest eig Langevin}
Let \(\beta \in (0,\infty)\) and consider a sequence of initialization measures \(\mu_N^0 \in \mathcal{M}_1(\mathcal{S}_{N,r})\). Then, for every \(n \geq 1\), \(\gamma_0 >0\), \(\gamma_1 > 1 > \gamma_2 > 0\), there exists \(\varepsilon_0>0\) and \(c_0 \in (0, \frac{1}{3r})\) such that for every \(\varepsilon < \varepsilon_0\), \(C_0 < c_0\), if 
\[
\sqrt{M} \gtrsim \frac{(n+2) \gamma_0^2 \gamma_1}{\beta C_0\lambda^2 \gamma_2}N^{\frac{1}{2(n+1)} \vee \frac{2C_0r}{1-C_0r}},
\] 
then, for \(N\) sufficiently large,
\[
\begin{split}
& \int_{\mathcal{S}_{N,r}} \mathbb{Q}_{\boldsymbol{X}} \left(\mathcal{T}_{E(\varepsilon,C_0)} \gtrsim \frac{\log(N) N^{- \left (\frac{1}{2(n+1)} \vee \frac{2C_0r}{1-C_0r}\right)}}{(n+2) \gamma_0^2 \gamma_1 }  \right) \boldsymbol{1}_{\{\mathcal{C}_{0,\beta}^{(N)}(n,\gamma_0) \cap \mathcal{C}_1^{'(N)}(\gamma_1,\gamma_2)\}} (\boldsymbol{X}) d\mu_N^0 (\boldsymbol{X})\\
& \lesssim K_1 \exp \left ( - \frac{\gamma_1 \gamma_0^4 (n+2) N^{\frac{1}{2(n+1)} \vee \frac{2C_0r}{1-C_0r}}}{ K_1 \log(N)} \right) + K_2 \exp \left ( -\frac{\gamma_0^2 \gamma_2 (n+2) N^{\frac{1}{2(n+1)} \vee \frac{2C_0r}{1-C_0r}}}{K_2} \right), 
\end{split}
\]
with \(\mathbb{P}\)-probability at least \(1 - \exp\left(- K N \right)\).
\end{lem}

\begin{proof}
We let \(\mathcal{A}_0 = \mathcal{A}_0(n,\gamma_0,\gamma_1,\gamma_2)\) denote the initial event given by
\[
\mathcal{A}_0(n,\gamma_0,\gamma_1,\gamma_2) = \left \{\boldsymbol{X}_0 \sim \mu_N^0 \colon \boldsymbol{X}_0 \in \mathcal{C}_{0,\beta}^{(N)}(n,\gamma_0) \cap \mathcal{C}_1^{'(N)}(\gamma_1,\gamma_2) \right \} 
\]
We note that on \(\mathcal{C}_1^{'(N)}(\gamma_1,\gamma_2) \), for every \(i \in [r]\) there exists \(\gamma_{ii} \in (\gamma_2,\gamma_1)\) such that \(\theta_i(0) = \gamma_{ii} N^{-1}\). Moreover, for \(\gamma>0\) and \(T_0 >0\) to be chosen later, we also define the event \(\mathcal{A} = \mathcal{A}(n,\gamma_0,\gamma_1,\gamma_2,\gamma, T_0)\) by
\[
\mathcal{A} = \mathcal{A}_0(n,\gamma_0,\gamma_1,\gamma_2) \cap \left \{ \sup_{t \in [0,T_0]} \norm{M_t^{\boldsymbol{G}}}_{\textnormal{op}} \leq \frac{\gamma}{\sqrt{N}} f(T_0)^{1/2} \right \},
\]
for some positive, monotone increasing function \(f\) to be determined later such that \(f(t) \ge \int_0^t \norm{\boldsymbol{M}(s)}^2_{\textnormal{op}}\) for every \(t \in [0,T_0]\). We then introduce the hitting time \(\mathcal{T}_{\boldsymbol{G}}\) for the set 
\[
\left\{t \colon  \norm{M_t^{\boldsymbol{G}}}_{\textnormal{op}} > \min_{1 \le i \le r} 2 C_0 r \beta \sqrt{M} \lambda^2 \int_0^t \theta_i(s) ds \right\},
\]
for some \(C_0 \in (0, \frac{1}{2r})\). We note that since \(\norm{M_0^{\boldsymbol{G}}}_{\textnormal{op}}=0\), 
we have \(\mathcal{T}_{\boldsymbol{G}} > 0\). We also introduce the hitting time \(\mathcal{T}_{\boldsymbol{L}_{0,\beta}}\) for the set 
\[
\left\{\boldsymbol{X} \colon \norm{\boldsymbol{L}_{0,\beta} \boldsymbol{G}(\boldsymbol{X})}_{\textnormal{op}} > 2 C_0 \beta \sqrt{M} \lambda^2 \theta_{\min}(\boldsymbol{X}) \right\}.
\]
According to~\eqref{eq: noise generator G} we have the bound
\begin{equation} \label{eq: bound norm noise generator G}
\norm{\boldsymbol{L}_{0,\beta} \boldsymbol{G}}_{\textnormal{op}} \leq 2 \norm{\boldsymbol{M}}_{\textnormal{op}} \norm{L_{0,\beta }\boldsymbol{M}}_{\textnormal{op}} + \frac{2r}{N} + \frac{2}{N} \norm{\boldsymbol{G}}_{\textnormal{op}},
\end{equation}
so that under the event \(\mathcal{A}\), 
\[
\begin{split}
\norm{\boldsymbol{L}_{0,\beta} \boldsymbol{G} (\boldsymbol{X}_0)}_{\textnormal{op}} & \leq 2 r \max_{1 \le i,j \le r} |m_{ij}(\boldsymbol{X}_0)| \max_{1 \le i,j \le r} |L_{0,\beta} m_{ij}(\boldsymbol{X}_0)| + \frac{2r}{N} + \frac{2r}{N} \sum_{k=1}^r \max_{1 \le i,j \leq r} |m_{ki}(\boldsymbol{X}_0) m_{kj}(\boldsymbol{X}_0) | \\
& \leq \frac{2r^2 \gamma_0^2}{N}  + \frac{2r}{N} + \frac{2r^2 \gamma_0^2}{N} \\
&\leq 2 C_0 \beta \sqrt{M} \lambda^2 \frac{\gamma_2}{N},
\end{split}
\]
provided \(\sqrt{M} \gtrsim \frac{r \gamma_0^2}{C_0 \beta \lambda^2 \gamma_2}\) which certainly holds by assumption. This implies that \(\mathcal{T}_{\boldsymbol{L}_{0,\beta}} >0\).

For every \(\varepsilon > 0\) and \(i \in [r]\), we then let \(\mathcal{T}_\varepsilon^{(i)}\) denote the hitting time for the set \(\{\boldsymbol{X} \colon \theta_i(\boldsymbol{X}) \geq \varepsilon\}\). We recall the evolution equation for \(\boldsymbol{G}\) under Langevin dynamics given by Corollary~\ref{cor: evolution equation for G}, i.e., 
\[
\boldsymbol{G}(t) = \boldsymbol{G}(0) + M_t^{\boldsymbol{G}} + \int_0^t \boldsymbol{L}_{0,\beta} \boldsymbol{G}(s) ds + \int_0^t \hat{\boldsymbol{L}}_{\beta} \boldsymbol{G}(s) ds ,
\]
where \(\hat{\boldsymbol{L}}_{\beta} \boldsymbol{G}\) satisfies \(\hat{\boldsymbol{L}}_{\beta} \boldsymbol{G} = 4 \beta \sqrt{M} \lambda^2 \boldsymbol{G} (\boldsymbol{I}_r - \boldsymbol{G})\). For all \(t \ge 0\) and \(i \in [r]\), we obtain by Weyl's inequality that
\[
\left | \theta_i \left (t \right ) - \mu_i \left (\boldsymbol{G}(0) + 4 \beta \sqrt{M} \lambda^2 \int_0^t  \boldsymbol{G}(s) (\boldsymbol{I}_r - \boldsymbol{G}(s)) ds \right)\right | \leq \norm{M_t^{\boldsymbol{G}} + \int_0^t \boldsymbol{L}_{0,\beta} \boldsymbol{G}(s)}_{\textnormal{op}} ds,
\]
where \(\mu_i(\boldsymbol{A})\) stands for the \(i\)th eigenvalue of \(\boldsymbol{A}\). Moreover, by the triangle inequality and the fact that \(\norm{\int_0^t \boldsymbol{L}_{0,\beta} \boldsymbol{G}(s) ds}_{\textnormal{op}} \leq r \int_0^t \norm{\boldsymbol{L}_{0,\beta} \boldsymbol{G}(s)}_{\textnormal{op}} ds\) it then follows that
\[
\left | \theta_i \left (t \right ) - \mu_i \left (\boldsymbol{G}(0) + 4 \beta \sqrt{M} \lambda^2 \int_0^t  \boldsymbol{G}(s) (\boldsymbol{I}_r - \boldsymbol{G}(s)) ds \right)\right | \leq \norm{M_t^{\boldsymbol{G}}}_{\textnormal{op}} + r \int_0^t \norm{\boldsymbol{L}_{0,\beta} \boldsymbol{G}(s)}_{\textnormal{op}} ds.
\]
According to Lemma~\ref{lem: integral inequality matrix case}, we then have that 
\[
\begin{split}
\theta_i(t) & \leq  \theta_i(0) + 4 \beta \sqrt{M} \lambda^2 \int_0^t \left ( \theta_{\max} (s) -  \theta_{\min}^2 (s) \right) ds + \norm{M_t^{\boldsymbol{G}}}_{\textnormal{op}} + r \int_0^t \norm{ \boldsymbol{L}_{0,\beta} \boldsymbol{G}(s)}_{\textnormal{op}} ds,\\
\theta_i(t) & \geq  \theta_i(0) + 4 \beta \sqrt{M} \lambda^2 \int_0^t \left ( \theta_{\min}(s) - \theta_{\max}^2 (s) \right) ds -  \norm{M_t^{\boldsymbol{G}}}_{\textnormal{op}} - r \int_0^t \norm{ \boldsymbol{L}_{0,\beta} \boldsymbol{G}(s)}_{\textnormal{op}} ds,
\end{split}
\]
for all \(t \ge 0\) and \(i \in [r]\). Then, on the event \(\mathcal{A}\), for every \(i \in [r]\) we have
\begin{equation} \label{eq: integral inequality subspace}
\frac{\gamma_{ii}}{N} + 4 (1 - C_0 r) \beta \sqrt{M} \lambda^2 \int_0^t \theta_{\min}(s) ds  \leq \theta_i(t) \leq  \frac{\gamma_{ii}}{N} + 4 (1 + C_0r) \beta \sqrt{M} \lambda^2 \int_0^t \theta_{\max}(s) ds ,
\end{equation}
for all \(t \le \mathcal{T}_{\boldsymbol{L}_{0,\beta}} \wedge \mathcal{T}_{\boldsymbol{G}} \wedge \min_{1 \le i \le r} \mathcal{T}_\varepsilon^{(i)}\). Given the integral inequality~\eqref{eq: integral inequality subspace} we then obtain from Lemma~\ref{lem: Gronwall} the following comparison inequality, 
\begin{equation} \label{eq: comparison inequality subspace}
\ell (t) \leq \theta_{\min} (t) \leq \theta_i(t) \leq \theta_{\max} (t) \leq  u(t),  
\end{equation}
for every \(i \in [r]\) and every \(t \le \mathcal{T}_{\boldsymbol{L}_{0,\beta}} \wedge \mathcal{T}_{\boldsymbol{G}} \wedge \min_{1 \le i \le r} \mathcal{T}_\varepsilon^{(i)}\), where the functions \(\ell\) and \(u\) are given by
\[
\begin{split}
\ell(t) & = \frac{\gamma_2}{N} \exp (4 (1-C_0r) \beta \sqrt{M} \lambda^2 t), \\
u(t) & = \frac{\gamma_1}{N} \exp(4 (1 + C_0 r) \beta \sqrt{M} \lambda^2 t).
\end{split}
\]
We then let \(T_{\ell, \varepsilon}\) denote the time such that \(\ell(T_{\ell, \varepsilon}) = \varepsilon\), i.e., 
\[
T_{\ell,\varepsilon} = \frac{\log(N)- \log(\gamma_2 /\varepsilon)}{4 \beta \sqrt{M} \lambda^2 (1 - C_0r)}. 
\]
Similarly, we let \(T_{u, \varepsilon}\) be such that \(u(T_{u, \varepsilon}) = \varepsilon\), i.e., 
\[
T_{u,\varepsilon} = \frac{ \log(N) - \log(\gamma_1/\varepsilon)}{4 \beta \sqrt{M} \lambda^2 (1+ C_0 r)}. 
\]
We note that on the event \(\mathcal{A}\), \(T_{u, \varepsilon} \leq \mathcal{T}_{\varepsilon}^{(i)} \leq T_{\ell, \varepsilon}\) for every \(i \in [r]\). Moreover, since the functions \(\ell\) and \(u\) are monotone increasing, we can bound \(\theta_i (\min_{1\le j \le r} \mathcal{T}_{\varepsilon}^{(j)})\) for every \(i \in [r]\) by
\[
\gamma_2 \left ( \frac{\varepsilon}{\gamma_1}\right )^{\frac{1-C_0 r}{1+C_0 r}} N^{- \frac{2 C_0 r}{1+C_0 r}}=\ell \left ( T_{u, \varepsilon} \right ) \le \theta_i \left (\min_{1\le j \le r} \mathcal{T}_{\varepsilon}^{(j)} \right ) \le u \left (T_{\ell, \varepsilon} \right ) =  \gamma_1  \left ( \frac{\varepsilon}{\gamma_2}\right )^{\frac{1+C_0 r}{1-C_0 r}} N^{\frac{2C_0r}{1-C_0r}}.
\]
In the following, we choose \(T_0 = T_{\ell,\varepsilon}\). 

Now, we wish to show that on the event \(\mathcal{A}\), \(\mathcal{T}_{\boldsymbol{G}} \wedge \mathcal{T}_{\boldsymbol{L}_{0,\beta}} \geq \min_{1 \le i \le r} \mathcal{T}_\varepsilon^{(i)}\). We first show that on the event \(\mathcal{A}\), \(\mathcal{T}_{\boldsymbol{L}_{0,\beta}} \wedge \min_{1 \le i \le r} \mathcal{T}_\varepsilon^{(i)} \wedge T_0 \le \mathcal{T}_{\boldsymbol{G}}\). We observe that 
\[
\norm{\boldsymbol{M}}^2_{\textnormal{op}} = \norm{\boldsymbol{M}^\top \boldsymbol{M}}_{\textnormal{op}} = \theta_{\max} , 
\]
so that from~\eqref{eq: comparison inequality subspace} we have
\begin{equation*} \label{eq: bound op norm M}
\int_0^t \norm{\boldsymbol{M}(s)}_{\textnormal{op}}^2 ds \leq \frac{\gamma_1}{N} \int_0^t e^{4 (1+C_0r) \beta \sqrt{M} \lambda^2 s} ds = \frac{\gamma_1}{N}\frac{e^{4 (1+C_0r) \beta \sqrt{M} \lambda^2 t} -1}{4(1+C_0r) \beta \sqrt{M} \lambda^2},
\end{equation*}
for every \(t \le \mathcal{T}_{\boldsymbol{L}_{0,\beta}} \wedge \mathcal{T}_{\boldsymbol{G}} \wedge \min_{1 \le i \le r} \mathcal{T}_\varepsilon^{(i)}\). In particular, the function \(f(t) = \frac{\gamma_1}{N}\frac{e^{4 (1+C_0r) \beta \sqrt{M} \lambda^2 t}-1}{4(1+C_0r) \beta \sqrt{M} \lambda^2} \) is non-negative, monotone increasing for all \(t >0\), and bounds \(\int_0^t \norm{\boldsymbol{M}(s)}_{\textnormal{op}}^2 ds\) by above for every \(t \leq \mathcal{T}_{\boldsymbol{L}_{0,\beta}} \wedge \mathcal{T}_{\boldsymbol{G}} \wedge \min_{1 \le i \le r} \mathcal{T}_\varepsilon^{(i)}\). In order to prove that on the event \(\mathcal{A}\), \(\mathcal{T}_{\boldsymbol{L}_{0,\beta}} \wedge \min_{1 \le i \le r} \mathcal{T}_\varepsilon^{(i)} \wedge T_0 \le \mathcal{T}_{\boldsymbol{G}}\), it suffices to show that there is \(\gamma >0\) such that 
\begin{equation} \label{eq: suff cond matrix p=2}
\frac{\gamma}{\sqrt{N}} \sqrt{f(t)} \leq \frac{\gamma_2 C_0 r}{2(1-C_0r)N}\left (e^{4(1-C_0r) \beta \sqrt{M} \lambda^2 t} - 1 \right ),
\end{equation}
for every \(t \le \mathcal{T}_{\boldsymbol{L}_{0,\beta}} \wedge \mathcal{T}_{\boldsymbol{G}} \wedge \min_{1 \le i \le r} \mathcal{T}_\varepsilon^{(i)} \wedge T_0\). Indeed, according to the comparison inequality~\eqref{eq: comparison inequality subspace}, the right-hand side of~\eqref{eq: suff cond matrix p=2} is a lower bound for the desired quantity, i.e.,
\[
\begin{split}
\frac{\gamma_2 C_0 r}{2(1-C_0r)N}\left (e^{4(1-C_0r) \beta \sqrt{M} \lambda^2 t} - 1 \right ) &= 2 C_0 r \beta \sqrt{M} \lambda^2 \frac{\gamma_2}{N} \int_0^t e^{4(1-C_0r) \beta \sqrt{M} \lambda^2 s} ds \\
&\leq 2 C_0 r \beta \sqrt{M} \lambda^2  \int_0^t \theta_{\min}(s)ds,
\end{split}
\]
for every \(t \le \mathcal{T}_{\boldsymbol{L}_{0,\beta}} \wedge \mathcal{T}_{\boldsymbol{G}} \wedge \min_{1 \le i \le r} \mathcal{T}_\varepsilon^{(i)}\). It is easily noticed that~\eqref{eq: suff cond matrix p=2} is verified at \(t=0\) for every \(\gamma >0\). Moreover, the inequality~\eqref{eq: suff cond matrix p=2} holds at \(t=T_0\), provided \(C_0r \leq 1/3\) and 
\[
\gamma^2 \leq \frac{\gamma_2^2 C_0^2 r^2 \beta \sqrt{M} \lambda^2}{\gamma_1}.
\] 
Since both sides of~\eqref{eq: suff cond matrix p=2} are increasing functions and the inequality is satisfied at \(t=0\) and at \(t=T_0\), the estimate~\eqref{eq: suff cond matrix p=2} holds for every \(t \le \mathcal{T}_{\boldsymbol{L}_{0,\beta}} \wedge \mathcal{T}_{\boldsymbol{G}} \wedge \min_{1 \le i \le r} \mathcal{T}_\varepsilon^{(i)} \wedge T_0\). This implies that \(\mathcal{T}_{\boldsymbol{L}_{0,\beta}} \wedge \min_{1 \le i \le r} \mathcal{T}_\varepsilon^{(i)} \wedge T_0 \leq \mathcal{T}_{\boldsymbol{G}}\), provided \(C_0r \leq 1/3\) and \(\gamma^2 \leq \frac{\gamma_2^2 C_0^2 r^2 \beta \sqrt{M} \lambda^2}{\gamma_1}\).

Next, we show that \(\mathcal{T}_{\boldsymbol{L}_{0, \beta}} \geq \min_{1 \le i \le r} \mathcal{T}_\varepsilon^{(i)} \wedge T_0\). To this end, we wish to apply the bounding flow method in order to estimate the operator norm \(\norm{\boldsymbol{L}_{0,\beta} \boldsymbol{G}}_{\textnormal{op}}\). We first recall from~\eqref{eq: bound norm noise generator G} that
\[
\norm{\boldsymbol{L}_{0,\beta} \boldsymbol{G}}_{\textnormal{op}} \leq 2 \norm{\boldsymbol{M}}_{\textnormal{op}} \norm{\boldsymbol{L}_{0,\beta} \boldsymbol{M}}_{\textnormal{op}} + \frac{2r}{N} +  \frac{2}{N} \norm{\boldsymbol{G}}_{\textnormal{op}}.
\]
Since \(\norm{\boldsymbol{M}}_{\textnormal{op}}^2 =  \theta_{\max}(\boldsymbol{G})\), we can bound \(\norm{\boldsymbol{M}}_{\textnormal{op}}\) via~\eqref{eq: comparison inequality subspace}. Moreover, we will use the bound \(\norm{\boldsymbol{G}}_{\textnormal{op}} \leq r^2 \max_{1 \le i, j, k \le r} |m_{ki} m_{kj}| \leq r^2\) for the last term. It remains to bound \(\norm{\boldsymbol{L}_{0,\beta} \boldsymbol{M}}_{\textnormal{op}}\) and for this purpose we will use Lemma~\ref{lem: bounding flows} with a different control on the last term. We direct the reader to the proof of Lemma~\ref{lem: weak recovery first spike Langevin p=2} for more details on the application of Lemma~\ref{lem: bounding flows}. We now consider condition (4) of Lemma~\ref{lem: bounding flows}. Since we do not have comparison inequalities for the \(m_{ij}\), in order to verify item (4) of Lemma~\ref{lem: bounding flows} we need to relate the control of the drift part of the expansion in Lemma~\ref{lem: bounding flows} to the quantity \(\norm{\boldsymbol{M} }_{\textnormal{op}}\). Recall the expression for the third term in the expansion~\eqref{eq: expansion via Ito} which is bounded by
\[
\begin{split}
\sum_{1 \leq i,j \leq r}\sum_{k=1}^n \int_0^t \cdots \int_0 ^{t_{k-1}}\vert a_{ij}(t_{k}) \vert dt_k \cdots  dt_1 & = 2 \beta \sqrt{M} \lambda^2 \sum_{k=1}^n \int_0^t \cdots \int_0 ^{t_{k-1}} \norm{\boldsymbol{M}(t_k)}_{\ell_1} dt_k \cdots dt_1\\
& \leq 2 \beta \sqrt{M} \lambda^2 r^{3/2} \sum_{k=1}^n \int_0^t \cdots \int_0 ^{t_{k-1}} \norm{\boldsymbol{M}(t_k)}_{\textnormal{op}} dt_k \cdots dt_1,
\end{split}
\]
where we used that \(a_{ij}(s) = 2 \beta \sqrt{M} \lambda^2 m_{ij}(s)\) and we used the equivalence of norms \(\norm{\boldsymbol{M}}_{\ell_1} \leq r^{3/2} \norm{\boldsymbol{M}}_{\textnormal{op}}\) (see e.g.~\cite{horn2012matrix}). Our goal is therefore to prove that 
\begin{equation} \label{eq: cond 4 matrix case}
\int_0^t \norm{\boldsymbol{M}(s)}_{\textnormal{op}} ds \leq \epsilon \norm{\boldsymbol{M}(t)}_{\textnormal{op}} ,
\end{equation}
for some \(\epsilon \in (0,1)\), ensuring that the third term in the expansion~\eqref{eq: expansion via Ito} is bounded by
\[
\begin{split}
2 \beta \sqrt{M} \lambda^2 r^{3/2} \sum_{k=1}^n \int_0^t \cdots \int_0 ^{t_{k-1}} \norm{\boldsymbol{M}(t_k)}_{\textnormal{op}} dt_k \cdots dt_1 & \leq 2 \beta \sqrt{M} \lambda^2 r^{3/2} \sum_{k=1}^n \epsilon^{k-1} \int_0^t \norm{\boldsymbol{M}(s)}_{\textnormal{op}} ds \\
& \leq 2 \beta \sqrt{M} \lambda^2 r^{3/2} \frac{1}{1-\epsilon}\int_0^t \norm{\boldsymbol{M}(s)}_{\textnormal{op}} ds.
\end{split}
\]
A sufficient condition for~\eqref{eq: cond 4 matrix case} to hold is given by 
\[
\sqrt{\frac{\gamma_1}{N}} \frac{e^{2 (1 + C_0r) \beta \sqrt{M} \lambda^2 t}-1}{2(1+C_0r) \beta \sqrt{M} \lambda^2} \leq \epsilon \sqrt{\frac{\gamma_2}{N}} e^{2(1-C_0r) \beta \sqrt{M} \lambda^2 t},
\]
for every \(t \le \mathcal{T}_{\boldsymbol{L}_{0,\beta}} \wedge \min_{1 \le i \le r} \mathcal{T}_\varepsilon^{(i)} \wedge T_0\), where we used~\eqref{eq: comparison inequality subspace}. The above inequality is easily verified at \(t=0\) and holds at \(t=T_0\) with \(\epsilon=1/2\), provided \(\sqrt{M} \gtrsim  \frac{\sqrt{\gamma_1}}{\beta \lambda^2 \sqrt{\gamma_2}} N^{\frac{C_0r}{1-C_0r}}\), which holds by assumption. We therefore have according to Lemma~\ref{lem: bounding flows} that on the event \(\mathcal{A}\),
\[
\norm{\boldsymbol{L}_{0,\beta} \boldsymbol{M}(t)}_{\textnormal{op}} \leq r \max_{1 \le i,j \leq r} |L_{0,\beta} m_{ij}(t)| \leq K r \left ( \frac{\gamma_0}{\sqrt{N}} \sum_{k=1}^{n-1} t^k + t^n + 4 \beta \sqrt{M} \lambda^2 r^{3/2} \int_0^t \norm{\boldsymbol{M}(s)}_{\textnormal{op}} ds \right ),
\]
for every \(t \le \mathcal{T}_{\boldsymbol{L}_{0,\beta}} \wedge \min_{1 \le i \le r} \mathcal{T}_\varepsilon^{(i)} \wedge T_0\), with \(\mathbb{Q}_{\boldsymbol{X}}\)-probability at least \(1 - K_2 \exp(-\gamma_0^2/(K_2 T_0))\) and with \(\mathbb{P}\)-probability at least \(1 -\exp(-KN)\). From this we obtain that, on the event \(\mathcal{A}\),
\begin{equation} \label{eq: bounding flow matrix case}
\norm{\boldsymbol{L}_{0,\beta} \boldsymbol{G}(t)}_{\textnormal{op}} \le 2K r \norm{\boldsymbol{M}(t)}_{\textnormal{op}} \left ( \frac{\gamma_0}{\sqrt{N}} \sum_{k=1}^{n-1} t^k + t^n + 4 \beta \sqrt{M} \lambda^2 r^{3/2} \int_0^t \norm{\boldsymbol{M}(s)}_{\textnormal{op}} ds \right ) + \frac{2r}{N} + \frac{2r^2}{N},
\end{equation}
for every \(t \le \mathcal{T}_{\boldsymbol{L}_{0,\beta}} \wedge \min_{1 \le i \le r} \mathcal{T}_\varepsilon^{(i)} \wedge T_0\), with \(\mathbb{Q}_{\boldsymbol{X}}\)-probability at least \(1 - K_2 \exp(-\gamma_0^2/(K_2 T_0))\) and with \(\mathbb{P}\)-probability at least \(1 -\exp(-KN)\). In order to deduce that \( \mathcal{T}_{\boldsymbol{L}_{0,\beta}} \geq \min_{1 \le i \leq r} \mathcal{T}_\varepsilon^{(i)} \wedge T_0\), it suffices to show that each term in the right-hand-side of~\eqref{eq: bounding flow matrix case} is bounded above by \(\frac{2 C_0 \beta \sqrt{M}\lambda^2}{n+2} \theta_{\min}(t)\) for all \(t \leq \mathcal{T}_{\boldsymbol{L}_{0,\beta}} \wedge \min_{1 \le i \le r} \mathcal{T}_\varepsilon^{(i)} \wedge T_0\). The reasoning is similar to the one used for the proof of Lemma~\ref{lem: weak recovery first spike Langevin p=2}.
\begin{enumerate}
\item[(i)] We first see that according to the lower bound in~\eqref{eq: comparison inequality subspace}, we have
\[
\frac{2  C_0 \beta \sqrt{M}\lambda^2}{n+2} \theta_{\min} (t) \geq \frac{2  C_0 \beta \sqrt{M}\lambda^2}{n+2} \frac{\gamma_2}{N} \exp (4(1-C_0r) \beta \sqrt{M} \lambda^2 t),
\]
for \(t \leq \mathcal{T}_{\boldsymbol{L}_{0,\beta}} \wedge \min_{1 \le i \le r} \mathcal{T}_\varepsilon^{(i)} \wedge T_0\). Similarly, according to the upper bound in~\eqref{eq: comparison inequality subspace}, we have
\[
\norm{\boldsymbol{M}(t)}_{\textnormal{op}} = \sqrt{\theta_{\max}(t)} \leq \frac{\sqrt{\gamma_1}}{\sqrt{N}} \exp (2(1+C_0r) \beta \sqrt{M} \lambda^2 t),
\]
for every \(t \leq \mathcal{T}_{\boldsymbol{L}_{0,\beta}} \wedge \min_{1 \le i \le r} \mathcal{T}_\varepsilon^{(i)} \wedge T_0\). Therefore, a sufficient condition for the first term is given by
\[
2Kr \frac{\gamma_0}{\sqrt{N}} \frac{\sqrt{\gamma_1}}{\sqrt{N}}t^k  e^{2(1+C_0r) \beta \sqrt{M} \lambda^2 t} + \frac{2r(r+1)}{N} \leq \frac{2 C_0 \beta \sqrt{M}\lambda^2}{n+2} \frac{\gamma_2}{N}  e^{4(1-C_0r) \beta \sqrt{M} \lambda^2 t},
\]
for every \(0 \leq k \leq n-1\). We observe that the inequality holds at \(t=0\). For \(t>0\) it is equivalent to verify that
\[
\frac{C_0 \beta \sqrt{M}\lambda^2 \gamma_2}{n+2} e^{2(1-3C_0r) \beta \sqrt{M} \lambda^2 t} \geq C r \gamma_0 \sqrt{\gamma_1} t^k,
\]
for some constant \(C\) which may depend on \(r\). We then see that for every \(0 \leq k \leq n-1\),
\[
\begin{split}
\frac{C_0 \beta \sqrt{M}\lambda^2 \gamma_2}{n+2} e^{2(1-3C_0r) \beta \sqrt{M} \lambda^2 t} & \gtrsim \gamma_0 \gamma_1 r N^{\frac{1}{2(n+1)} \vee \frac{2C_0r}{1-C_0r}} e^{2(1-3C_0r) (n+2) \gamma_0 \gamma_1 r t}\\
& \gtrsim  (n+2)^k \frac{t^k}{k!} \gamma_0^{k+1} \gamma_1^{k+1} r^{k+1} N^{\frac{1}{2(n+1)} \vee \frac{2C_0r}{1-C_0r}} \\
& \gtrsim r \gamma_0 \sqrt{\gamma_1} t^k,
\end{split}
\]
where the first inequality follows by assumption on \(\sqrt{M}\). This ensures that the first term is controlled as desired for every \(t \leq \mathcal{T}_{\boldsymbol{L}_{0,\beta}} \wedge \min_{1 \le i \le r} \mathcal{T}_\varepsilon^{(i)} \wedge T_0\) and every \(0 \leq k \leq n-1\).

\item[(ii)] Similar to item (i), a sufficient condition for the second term is given by
\[
2Kr \frac{\sqrt{\gamma_1}}{\sqrt{N}} t^n  e^{2( 1 + C_0 r) \beta \sqrt{M} \lambda^2 t} + \frac{2r(r+1)}{N} \leq \frac{2 C_0 \beta \sqrt{M} \lambda^2}{n+2}\frac{\gamma_2}{N} e^{4(1-C_0 r) \beta \sqrt{M} \lambda^2 t},
\] 
for all \(t \leq \mathcal{T}_{\boldsymbol{L}_{0,\beta}}  \wedge \min_{1 \le i \le r} \mathcal{T}_\varepsilon^{(i)} \wedge T_0\). The bound holds at \(t=0\) and for \(t>0\) it is equivalent to verify that 
\[
\frac{C_0 \beta \sqrt{M} \lambda^2}{n+2} \frac{\gamma_2}{\sqrt{N}} e^{2(1 -3C_0 r) \beta \sqrt{M} \lambda^2 t} \geq C r \sqrt{\gamma_1} t^n ,
\] 
for some constant \(C\) which may depend on \(r\). Proceeding as done in the proof of Lemma~\ref{lem: weak recovery first spike Langevin p=2} to verify item (ii), gives that the control holds provided
\[
\sqrt{M} \gtrsim \frac{(r \sqrt{\gamma_1}(n+2))^{\frac{1}{n+1}}}{\beta \lambda^2 C_0^{\frac{1}{n+1}} \gamma_2^{\frac{1}{n+1}} } N^{\frac{1}{2(n+1)}},
\]
which holds by assumption. 

\item[(iii)] Finally, we wish to show that 
\[
8 K r^{5/2} \beta \sqrt{M} \lambda^2 \norm{\boldsymbol{M}(t)}_{\textnormal{op}} \int_0^t \norm{\boldsymbol{M}(s)}_{\textnormal{op}} ds  + \frac{2r(r+1)}{N} \leq \frac{2 C_0 \beta \sqrt{M} \lambda^2}{n+2} \theta_{\min}(t),
\] 
for every \(t \leq \mathcal{T}_{\boldsymbol{L}_{0,\beta}} \wedge \min_{1 \le i \le r} \mathcal{T}_\varepsilon^{(i)} \wedge T_0\). To this end, a sufficient condition is given by 
\[
4 K r^{5/2} \frac{\gamma_1}{N} e^{2(1+C_0r) \beta \sqrt{M} \lambda^2 t}  \frac{e^{2(1+C_0r) \beta \sqrt{M} \lambda^2 t}-1}{1+C_0r} + \frac{2r(r+1)}{N} \leq \frac{2 C_0 \beta \sqrt{M} \lambda^2}{n+2}\frac{\gamma_2}{N} e^{4(1-C_0 r) \beta \sqrt{M} \lambda^2 t},
\] 
for every \(t \leq \mathcal{T}_{\boldsymbol{L}_{0,\beta}} \wedge \min_{1 \le i \le r} \mathcal{T}_\varepsilon^{(i)} \wedge T_0\). We easily verify that the inequality holds at \(t=0\) and at \(t=T_0\), provided 
\[
\sqrt{M} \gtrsim \frac{r^{5/2} \gamma_1 (n+2)}{C_0 \beta \lambda^2 \gamma_2} N^{\frac{2 C_0r}{1-C_0r}}.
\]
Therefore, we have the third term is controlled for every \(t \leq \mathcal{T}_{\boldsymbol{L}_{0,\beta}} \wedge \min_{1 \le i \le r} \mathcal{T}_\varepsilon^{(i)} \wedge T_0\).
\end{enumerate}
We therefore have that on the event \(\mathcal{A}\), \(\min_i \mathcal{T}_\varepsilon^{(i)} \wedge T_0 \leq  \mathcal{T}_{\boldsymbol{L}_{0,\beta}}\) with \(\mathbb{Q}_{\boldsymbol{X}}\)-probability at least \(1 - K_2 \exp(-\gamma_0^2/(K_2 T_0))\) and with \(\mathbb{P}\)-probability at least \(1 -\exp(-KN)\). In particular, under the event \(\mathcal{C}_{0,\beta}^{(N)}(n,\gamma_0) \cap \mathcal{C}_1'(\gamma_1,\gamma_2)\) we have
\[
\min_{1 \le i \le r} \mathcal{T}_\varepsilon^{(i)} \leq T_0 \lesssim \frac{\log(N)}{(n+2) \gamma_0^2 \gamma_1 N^{\frac{1}{2(n+1)} \vee \frac{2C_0r}{1-C_0r}}},
\]
with \(\mathbb{Q}_{\boldsymbol{X}}\)-probability at least \(1 - K_2 \exp(- \gamma_1 \gamma_0^4 (n+2) N^{\frac{1}{2(n+1)} \vee \frac{2C_0r}{1-C_0r}} /(K_2 \log(N))) - r(r+1) \exp(- \gamma^2/(4r^2))\) and with \(\mathbb{P}\)-probability at least \(1 -\exp(-KN)\). This completes the proof of Lemma~\ref{lem: weak recovery largest eig Langevin}.
\end{proof}

It remains to prove Proposition~\ref{thm: strong recovery isotropic Langevin p=2}. 

\begin{proof}[\textbf{Proof of Proposition~\ref{thm: strong recovery isotropic Langevin p=2}}]
We recall from Lemma~\ref{lem: weak recovery largest eig Langevin} the set \(E(\varepsilon, C_0)\) for every \(\varepsilon>0\) and \(C_0 \in (0, \frac{1}{r})\) given by
\[
E(\varepsilon,C_0) = \left \{ \boldsymbol{X} \colon \theta_{\max}(\boldsymbol{X}) \ge \varepsilon \enspace \textnormal{and} \enspace \theta_{\min}(\boldsymbol{X}) \gtrsim N^{-\frac{2C_0r}{1+C_0r}}\right\}.
\]
For every \(T > T_0 \gtrsim \frac{\log(N)}{(n+2) \gamma_0^2 \gamma_1 N^{\frac{1}{2(n+1)} \vee \frac{2C_0r}{1-C_0r} }} \), we have by the strong Markov property that
\[
\begin{split}
& \int_{\mathcal{S}_{N,r}} \mathbb{Q}_{\boldsymbol{X}} \left ( \inf_{t \in [T_0,T]} \theta_{\min}(\boldsymbol{X}_t^\beta) \geq 1 - \varepsilon \right ) d \mu_N^0(\boldsymbol{X}) \\
& \geq \inf_{\boldsymbol{X} \in E(\varepsilon,C_0)} \mathbb{Q}_{\boldsymbol{X}} \left ( \inf_{t \in [T_0,T]} \theta_{\min}(\boldsymbol{X}_t^\beta) \geq 1 - \varepsilon \right ) \times \int_{\mathcal{S}_{N,r}} \mathbb{Q}_{\boldsymbol{X}} \left (\mathcal{T}_{E(\varepsilon,C_0)} \lesssim \frac{\log(N)}{(n+2) \gamma_0^2 \gamma_1 N^{\frac{1}{2(n+1)} \vee \frac{2C_0r}{1-C_0r} }} \right) d\mu_N^0(\boldsymbol{X}).
\end{split}
\]
We estimate the integral according to Lemma~\ref{lem: weak recovery largest eig Langevin} since it holds that
\[
\begin{split}
& \int_{\mathcal{S}_{N,r}} \mathbb{Q}_{\boldsymbol{X}} \left (\mathcal{T}_{E(\varepsilon,C_0)} \gtrsim \frac{\log(N)}{(n+2) \gamma_0^2 \gamma_1 N^{\frac{1}{2(n+1)} \vee \frac{2C_0r}{1-C_0r} }} \right) d\mu_N^0(\boldsymbol{X}) \\
& \leq \mu_N^0 (\mathcal{C}_{0,\beta}^{(N)}(n,\gamma_0)^\textnormal{c}) +  \mu_N^0 (\mathcal{C}_1'(\gamma_1,\gamma_2)^\textnormal{c}) \\
& + \int_{\mathcal{S}_{N,r}} \mathbb{Q}_{\boldsymbol{X}} \left (\mathcal{T}_{E(\varepsilon,C_0)} \lesssim \frac{\log(N)}{(n+2) \gamma_0^2 \gamma_1 N^{\frac{1}{2(n+1)} \vee \frac{2C_0r}{1-C_0r} }} \right) \textbf{1}_{\{\mathcal{C}_{0,\beta}^{(N)}(n,\gamma_0) \cap \mathcal{C}_1'(\gamma_1,\gamma_2)\}} (\boldsymbol{X}) d\mu_N^0(\boldsymbol{X}).
\end{split}
\]
Therefore, it remains to estimate the probability of strong recovery of all eigenvalues starting from \(\boldsymbol{X}_0 \in E(\varepsilon,C_0)\). It follows from Corollary~\ref{cor: evolution equation for G} and Weyl's inequality that for every \(i \in [r]\),
\[
\left | \theta_i(t) - \mu_i \left (\boldsymbol{G}(0) + 4 \beta \sqrt{M} \lambda^2 \int_0^t \boldsymbol{G}(s) (\boldsymbol{I}_r - \boldsymbol{G}(s)) ds\right )\right | \leq \norm{M_t^{\boldsymbol{G}}}_{\textnormal{op}} + r \int_0^t \norm{\boldsymbol{L}_{0,\beta} \boldsymbol{G}(s)}_{\textnormal{op}} ds,
\]
where \(\mu_i(\boldsymbol{A})\) stands for the \(i\)th eigenvalue of \(\boldsymbol{A}\). Now, for every \(\varepsilon' > \varepsilon > 0\) such that \(\varepsilon \leq \frac{1}{2}\), according to Lemma~\ref{lem: integral inequality matrix case} and especially to~\eqref{eq: integral inequality eigenvalues 2}, we have
\[
\mu_i \left (\boldsymbol{G}(0) + 4 \beta \sqrt{M} \lambda^2 \int_0^t \boldsymbol{G}(s) (\boldsymbol{I}_r - \boldsymbol{G}(s)) ds\right ) \geq \theta_i(0) + 4 \beta \sqrt{M} \lambda^2 \int_0^t \theta_{\min}(s) \left (1 - \theta_{\min}(s)\right) ds,
\]
for every \(t \leq  \mathcal{T}_{\varepsilon'}^{(\max)} \wedge \mathcal{T}_{\varepsilon}^{(\min)}\). In particular, we assume that \(\theta_{\max} \in (\varepsilon,\varepsilon')\). We then obtain that
\[
\theta_{\min}(t) \geq \theta_{\min}(0) + 4 \beta \sqrt{M} \lambda^2 \int_0^t \theta_{\min}(s)ds - \norm{M_t^{\boldsymbol{G}}}_{\textnormal{op}} - r \int_0^t \norm{\boldsymbol{L}_{0,\beta} \boldsymbol{G}(s)}_{\textnormal{op}} ds,
\]
for \(t \leq  \mathcal{T}_{\varepsilon'}^{(\max)} \wedge \mathcal{T}_{\varepsilon}^{(\min)}\). According to~\eqref{eq: bound norm noise generator G}, the operator norm of \(\boldsymbol{L}_{0,\beta} \boldsymbol{G}\) is bounded by 
\[
\norm{\boldsymbol{L}_{0,\beta} \boldsymbol{G}}_{\textnormal{op}} \leq 2 \norm{\boldsymbol{M}}_{\textnormal{op}} \norm{L_{0,\beta }\boldsymbol{M}}_{\textnormal{op}} + \frac{2r}{N} + \frac{2}{N} \norm{\boldsymbol{G}}_{\textnormal{op}} \leq 2 r^2 \max_{i,j} \norm{L_{0,\beta} m_{ij}}_\infty + \frac{2r(r+1)}{N}.
\]
We recall from Lemma~\eqref{eq: bound norm L_0m_ij} that for every \(n \ge 1\), there exists a constant \(\Lambda = \Lambda(p, n, \{\lambda_i\}_{i=1}^2)\) such that \(\norm{L_{0,\infty} m_{ij}}_\infty  \leq \Lambda\) with \(\mathbb{P}\)-probability at least \(1 - \exp(-K N)\). It then follows that
\[
r \int_0^t \norm{\boldsymbol{L}_{0,\beta} \boldsymbol{G}(s)}_{\textnormal{op}} ds \leq 4C_0r \beta \sqrt{M} \lambda^2 \int_0^t \theta_{\min}(s) ds, 
\]
since 
\[
\begin{split}
4C_0r \beta \sqrt{M} \lambda^2 \int_0^t \theta_{\min}(s) ds & \geq 4C_0r \beta \sqrt{M} \lambda^2 t N^{-\frac{2C_0r}{1+C_0r}} \\
& > 4rt (n+2) \gamma_0^2 \gamma_1 N^{\frac{4C_0^2 r^2}{1-C_0^2r^2}} \\
& > 2rt \left (r^2 \Lambda + \frac{r(r+1)}{N}\right ) \geq \norm{\boldsymbol{L}_{0,\beta} \boldsymbol{G}}_{\textnormal{op}}.
\end{split}
\]
Furthermore, since \(\int_0^t \norm{\boldsymbol{M}(s)}_{\textnormal{op}}^2 ds \leq r^2 t\), it follows from Lemma~\ref{lem: Doob max inequality for operator norm M} with \(f(t) = r^2 t\) that
\[ 
\sup_{\boldsymbol{X} \in \mathcal{S}_{N,r}} \mathbb{Q}_{\boldsymbol{X}} \left(\sup_{t\in [0,T]}  \norm{\boldsymbol{M}_t^{\boldsymbol{G}}}_{\textnormal{op}} \ge \gamma \right) \leq r(r+1) e^{- N\gamma^2/(4r^4T)}.
\]
Combining the above results yields
\[
\theta_{\min}(t) \geq \frac{1}{2}N^{-\frac{2C_0r}{1+C_0r}} + 4(1-C_0r) \beta \sqrt{M} \lambda^2 \int_0^t \theta_{\min}(s)ds,
\]
for every \(t \leq  \mathcal{T}_{\varepsilon'}^{(\max)} \wedge \mathcal{T}_{\varepsilon}^{(\min)}\), with \(\mathbb{Q}_{\boldsymbol{X}}\)-probability at least \(1 - r(r+1)\exp(- N^{1-\frac{4C_0r}{1+C_0r}}/(8r^4 \mathcal{T}_{\varepsilon}^{(\min)}))\) and with \(\mathbb{P}\)-probability at least \(1- \exp(-KN)\). Grönwall's inequality from Lemma~\ref{lem: Gronwall} implies that
\[
\theta_{\min}(t) \geq \frac{1}{2}N^{-\frac{2C_0r}{1+C_0r}} \exp (4(1-C_0r) \beta \sqrt{M} \lambda^2 t),
\]
for every \(t \leq  \mathcal{T}_{\varepsilon'}^{(\max)} \wedge \mathcal{T}_{\varepsilon}^{(\min)}\), with \(\mathbb{Q}_{\boldsymbol{X}}\)-probability at least \(1 - r(r+1)\exp(- N^{1-\frac{4C_0r}{1+C_0r}}/(8r^4 \mathcal{T}_{\varepsilon}^{(\min)}))\) and with \(\mathbb{P}\)-probability at least \(1- \exp(-KN)\). A simple computation shows that 
\[
\mathcal{T}_\varepsilon^{(\min)} \lesssim \frac{C_0^2 \log(N)}{(n+2) \gamma_0^2 \gamma_1 N^{\frac{1}{2(n+1)} \vee \frac{2C_0r}{1-C_0r}}}.
\]
Showing strong recovery of \(\theta_{\min}\) from weak recovery is straightforward and we direct the reader to the proof of Lemma~\ref{lem: weak implies strong recovery Langevin p>2}.
\end{proof}

\appendix

\section{Concentration properties of the uniform measure on the Stiefel manifold} \label{appendix: invariant measure}

The concentration and anti-concentration properties of the uniform measure \(\mu_{N \times r}\) on the normalized Stiefel manifold \(\mathcal{S}_{N,r}\) were studied in~\cite[Appendix A]{benarous2024gradientflow}. In particular, it was shown there that \(\mu_{N \times r}\) satisfies Condition \(1\) on the typical scale \(N^{-1/2}\) of the initial correlations 
\[
m_{ij}^{(N)} (\boldsymbol{X}) = \frac{1}{N} \left ( \boldsymbol{V}^\top \boldsymbol{X} \right)_{ij} = \frac{1}{N} \langle \boldsymbol{v}_i, \boldsymbol{x}_j \rangle. 
\]
In this appendix, we first prove that \(\mu_{N \times r}\) satisfies Condition \(1'\) on the typical scale \(N^{-1}\) of the eigenvalues of \(\boldsymbol{G}\). We then show that \(\mu_{N \times r}\) weakly satisfies Condition \(0\) at level \(\infty\) (for inverse temperature \(\beta\)).

\begin{lem} \label{lem: concentration eigenvalues}
Let \(\boldsymbol{X} \sim \mu_{N \times r}\), and define 
\[
\boldsymbol{G} = \boldsymbol{M}^\top \boldsymbol{M}, \quad \textnormal{with} \enspace \boldsymbol{M} = \frac{1}{N} \boldsymbol{V}^\top \boldsymbol{X}.
\]
Let \(\theta_{\max} (\boldsymbol{G})\) and \(\theta_{\min} (\boldsymbol{G})\) denote the largest and smallest eigenvalues of \(\boldsymbol{G}\). Then, there exist constants \(C(r), C'(r)\), and \(c(r) >0\), depending only on \(r\), such that for every \(t >0\), 
\[
\mu_{N \times r} \left( \theta_{\max}(\boldsymbol{G})  > \frac{r}{N} + t\right) \leq C(r)\exp(-c(r)Nt),
\]
and for every \(t \in (0, \frac{1}{2})\),
\[
\mu_{N \times r} \left( \theta_{\min} (\boldsymbol{G}) < \frac{t}{N}\right) \leq C'(r)\sqrt{t}+C(r)\exp\left(-c(r) N^{1/4} t\right).
\]
\end{lem}

\begin{proof}
We begin by recalling the the operator norm of \(\boldsymbol{M}^\top \boldsymbol{M}\) and \(\boldsymbol{M} \boldsymbol{M}^\top\) coincide. Thus it suffices to study deviations of \(\boldsymbol{M} \boldsymbol{M}^\top\). By the standard representation of the uniform measure on the Stiefel manifold (see, e.g., \cite[Theorem 2.2.1]{chikuse2012statistics}), if \(\boldsymbol{X} \sim \mu_{N \times r}\), then
\[
\boldsymbol{X} = \boldsymbol{Z} \left(\frac{1}{N} \boldsymbol{Z}^\top \boldsymbol{Z}\right)^{-1/2},
\]
where \(\boldsymbol{Z} \in \R^{N \times r}\) is a random matrix filled with i.i.d.\ standard normal entries. Hence,
\[
\boldsymbol{M} \boldsymbol{M}^\top = \left (\frac{1}{N} \boldsymbol{V}^\top \boldsymbol{Z} \right) \left( \frac{1}{N}\boldsymbol{Z}^\top \boldsymbol{Z}\right)^{-1} \left( \frac{1}{N} \boldsymbol{Z}^\top \boldsymbol{V} \right ).
\]
We now introduce \(\widetilde{\boldsymbol{M}} \coloneqq  \frac{1}{\sqrt{N}} \boldsymbol{V}^\top \boldsymbol{Z} \in \mathbb{R}^{r \times r}\) which has i.i.d.\ standard normal entries. Rewriting the previous expression gives
\begin{equation} \label{decomposition}
\boldsymbol{M} \boldsymbol{M}^\top = \frac{1}{N} \widetilde{\boldsymbol{M}} \widetilde{\boldsymbol{M}}^\top + \left (\frac{1}{N} \boldsymbol{V}^\top \boldsymbol{Z} \right) \left(\left(\frac{1}{N}\boldsymbol{Z}^\top \boldsymbol{Z}\right)^{-1} - \boldsymbol{I}_r\right) \left( \frac{1}{N}\boldsymbol{Z}^\top\boldsymbol{V} \right).
\end{equation}
Using the triangle inequality, we have
\begin{equation} \label{triangle inequality}
\begin{split}
& \mu_{N \times r} \left(\norm{\boldsymbol{M} \boldsymbol{M}^\top-\frac{r}{N} \boldsymbol{I}_r }_{\textnormal{op}} > t\right) \\
& \leq \mu_{N \times r} \left(\norm{\frac{1}{N} \widetilde{\boldsymbol{M}}^\top \widetilde{\boldsymbol{M}}-\frac{r}{N}\boldsymbol{I}_r}_{\textnormal{op}} > \frac{t}{2} \right) \\
& \quad + \mu_{N \times r} \left( \norm{\frac{1}{N}(\boldsymbol{V}^\top\boldsymbol{Z}) \left(\left(\frac{1}{N}\boldsymbol{Z}^\top\boldsymbol{Z}\right)^{-1}-\boldsymbol{I}_r\right)\frac{1}{N} (\boldsymbol{Z}^\top\boldsymbol{V})}_{\textnormal{op}} > \frac{t}{2}\right) .
\end{split}
\end{equation}
The first probability concerns a centered Wishart matrix. By (A.1) in~\cite{benarous2024gradientflow}, we find
\[
\begin{split}
\mu_{N \times r} \left(\norm{\frac{1}{N} \widetilde{\boldsymbol{M}}^\top \widetilde{\boldsymbol{M}} -\frac{r}{N} \boldsymbol{I}_r}_{\textnormal{op}} > \frac{t}{2}\right) & = \mu_{N \times r} \left(\norm{\frac{1}{r} \widetilde{\boldsymbol{M}}^\top
\widetilde{\boldsymbol{M}} -\boldsymbol{I}_r}_{\textnormal{op}} > \frac{N t}{2r}\right) \\
& \leq 2 \exp\left(-r\left(\frac{1}{c}\left(\frac{Nt}{2r}\right)^{1/2}-1\right)^2 \right),
\end{split}
\]
For the second term in~\eqref{triangle inequality}, we use submultiplicativity:
\[
\norm{\frac{1}{N}(\boldsymbol{V}^\top\boldsymbol{Z}) \left(\left(\frac{1}{N}\boldsymbol{Z}^\top\boldsymbol{Z}\right)^{-1}-\boldsymbol{I}_r\right)\frac{1}{N} (\boldsymbol{Z}^\top\boldsymbol{V})}_{\textnormal{op}} \le \norm{\frac{1}{N}(\boldsymbol{V}^\top\boldsymbol{Z})}_{\textnormal{op}}^2 \norm{\left( \frac{1}{N} \boldsymbol{Z}^\top \boldsymbol{Z}\right)^{-1} - \boldsymbol{I}_r }_{\textnormal{op}}.
\] 
The deviations of each factor are exactly the ones estimated in the proof of~\cite[Lemma A.1]{benarous2024gradientflow}.

For the lower bound on the smallest eigenvalue of \(\boldsymbol{G}\), we first note that by Weyl's inequality and the decomposition~\eqref{decomposition}, 
\[
\theta_{\min}(\boldsymbol{G}) \geq a- b,
\]
where
\[
a = \lambda_{\min} \left( \frac{1}{N} \widetilde{\boldsymbol{M}} \widetilde{\boldsymbol{M}}^\top \right), \quad b= \lambda_{\max} \left(\left(\frac{1}{N} \boldsymbol{V}^\top \boldsymbol{Z} \right) \left( \left( \frac{1}{N} \boldsymbol{Z}^\top \boldsymbol{Z}\right)^{-1} - \boldsymbol{I}_r \right) \left( \frac{1}{N} \boldsymbol{Z}^\top \boldsymbol{V}\right)\right).
\]
Thus, for every \(t>0\),
\[
\mathbb{P}\left(\theta_{\min}(\boldsymbol{G}) \ge \frac{t}{N} \right) \geq \mathbb{P}\left(a-b \geq \frac{t}{N} \right) \geq \mathbb{P}\left(a \geq \frac{2t}{N} \cap b \leq \frac{t}{N} \right), \\
\]
so that 
\begin{equation} \label{0}
\mathbb{P}\left(\theta_{\min}(\boldsymbol{G}) < \frac{t}{N} \right) \leq \mathbb{P}\left(a < \frac{2t}{N} \cup b > \frac{t}{N} \right) \leq \mathbb{P} \left(a < \frac{2t}{N} \right) + \mathbb{P}\left(b > \frac{t}{N}\right).
\end{equation}
We first focus on \(\mathbb{P}(a < 2t/N)\). The matrix \(\widetilde{\boldsymbol{M}} \widetilde{\boldsymbol{M}}^\top\) is Wishart and its smallest eigenvalue density is explicitly known in~\cite[Theorem 3.1]{edelman1988eigenvalues}. If \(f_{\theta_{\min}}\) denotes its density function, for every \(t \in (0,1)\), there exists a constant \(C'(r)\) such that 
\[
f_{\theta_{\min}}(t) \leq  \frac{C'(r)}{\sqrt{t}}.
\]
Integrating over \([0, 2t/N]\) yields, for \(t \in (0,1/2)\),
\begin{equation} \label{1}
\mathbb{P} \left( a \leq \frac{2t}{N} \right) \leq 2 C'(r) \sqrt{t}.
\end{equation}
Turning to the second term \(\mathbb{P}(b > t / N)\), we see that 
\[
\begin{split}
b & \leq \norm{\frac{1}{N} \boldsymbol{V}^\top \boldsymbol{Z}}_{\textnormal{op}}^2 \norm{\left( \frac{1}{N} \boldsymbol{Z}^\top \boldsymbol{Z} \right)^{-1} }_{\textnormal{op}} \norm{\frac{1}{N} \boldsymbol{Z}^\top \boldsymbol{Z} -\boldsymbol{I}_r}_{\textnormal{op}}.
\end{split}
\]
For every \(t_1, t_2, t_3 \geq 0\) we then have that 
\[
\begin{split}
\mathbb{P}\left(\norm{\frac{1}{N}\boldsymbol{V}^\top\boldsymbol{Z}}_{\textnormal{op}}^2 > t_1 \right) & = \mathbb{P}\left(\norm{\frac{1}{N}\boldsymbol{V}^\top\boldsymbol{Z}}_{\textnormal{op}} > \sqrt{t_1} \right) \leq 2 \exp\left(-\left(\frac{\sqrt{t_1 N}}{C} - 2\sqrt{r}\right)^2 \right),\\
\mathbb{P} \left( \norm{\left(\frac{1}{N} \boldsymbol{Z}^\top\boldsymbol{Z}\right)^{-1}}_{\textnormal{op}} > \frac{1}{t_2} \right) & \leq 2\exp\left(-\left(\frac{\sqrt{N}}{C}(1-t_2)-\sqrt{r}\right)^2\right),\\
\mathbb{P} \left(\norm{\frac{1}{N} \boldsymbol{Z}^\top \boldsymbol{Z}-\boldsymbol{I}_r}_{\textnormal{op}} >  \sqrt{t_3} \right) & \leq 2\exp\left(-\left(\frac{\sqrt{t_3 N}}{C}-\sqrt{r}\right)^2 \right).
\end{split}
\]
Choosing \(t_1 = \frac{\sqrt{t}}{\sqrt{2}N^{3/4}}\), \(\sqrt{t_3} = \frac{\sqrt{t}}{\sqrt{2}N^{1/4}}\) and \(t_2 = \frac{1}{2}\), we deduce that there exist constants \(C(r),c(r)\) such that 
\begin{equation} \label{2}
\mathbb{P} \left(b> t/N \right) \leq C(r)\exp\left(-c(r) N^{1/4} t\right).
\end{equation}
Combining~\eqref{0},~\eqref{1}, and~\eqref{2} gives
\[
\mathbb{P}\left(\theta_{\min}(\boldsymbol{G}) < \frac{t}{N} \right)  \le 2 C'(r) \sqrt{t} + C(r) \exp\left(-c(r) N^{1/4} t\right),
\]
as claimed.
\end{proof}

It remains to prove the following concentration estimate, which implies that \(\mu_{N \times r}\) weakly satisfies Condition 0 at level \(\infty\) (for inverse temperature \(\beta\)).

\begin{lem} \label{lem: concentration cond infinity Langevin}
For every \(\beta \in (0,\infty)\), \(T > 0\), and \(1 \leq i,j \leq r\), there exist constants \(C_1, C_2, K > 0\), depending only on \(p,r,\{\lambda_i\}_{i=1}^r\), such that for every \(\gamma>0\),
\[
\mu_{N \times r} \left(\sup_{t \leq T} \vert e^{t L_{0,\beta}} L_{0,\beta} m_{ij}^{(N)} (\boldsymbol{X}) \vert \geq \gamma \right) \leq C_{1} N T \exp\left(-C_{2}\gamma^2N\right),
\]
with \(\mathbb{P}\)-probability at least \(1-\mathcal{O}(e^{-KN})\).
\end{lem}

The proof follows the same structure as in the gradient-flow case treated in~\cite{benarous2024gradientflow}, which in turn adapts ideas from the argument of~\cite[Theorem~6.2]{arous2020algorithmic}. Let \(\widehat{\boldsymbol{X}}_t\) denote the Langevin dynamics generated by \(L_{0,\beta}\) (see~\eqref{eq: generator Langevin noise}). The first step is to establish the rotational invariance properties of this dynamics. For \(\boldsymbol{X} \in \mathcal{S}_{N,r}\), let \(R^N_{\boldsymbol{X}} \colon T_{\boldsymbol{X}} \mathcal{S}_{N,r} \to \mathcal{S}_{N,r}\) denote the polar retraction, defined for \(\boldsymbol{U} \in T_{\boldsymbol{X}} \mathcal{S}_{N,r}\) by 
\[
R^N_{\boldsymbol{X}}(\boldsymbol{U}) = \left (\boldsymbol{X} +\boldsymbol{U} \right) \left(\boldsymbol{I}_r + \frac{1}{N} \boldsymbol{U}^\top \boldsymbol{U}\right)^{-1/2},
\]
which satisfies \(\left(R^N_{\boldsymbol{X}} (\boldsymbol{U})\right)^\top R^N_{\boldsymbol{X}} (\boldsymbol{U}) = N \boldsymbol{I}_r\). 

\begin{lem} \label{lem: noise invariance}
Let \(\widehat{\boldsymbol{X}}_0 \sim \mu_{N \times r}\). Then, for every \(t \ge 0\), both \(\widehat{\boldsymbol{X}}_t\) and \(R^N_{\boldsymbol{X}} \left(\nabla H_0(\widehat{\boldsymbol{X}}_t)\right)\) take values in \(\mathcal{S}_{N,r}\), and their laws are invariant under left multiplication by any orthogonal matrix.  
\end{lem}

\begin{proof}
A similar argument to that used in proving~\cite[Lemma A.5]{benarous2024gradientflow} applies, upon noting that Brownian motion is invariant under left rotations.
\end{proof}

We now provide the proof of Lemma~\ref{lem: concentration cond infinity Langevin}.

\begin{proof}[\textbf{Proof of Lemma~\ref{lem: concentration cond infinity Langevin}}]
We write \(\E_{\mathbb{Q}_{\boldsymbol{X}}}\) for the expectation with respect to the Langevin dynamics \(\widehat{\boldsymbol{X}}_t\) generated by \(L_{0,\beta}\) started at \(\boldsymbol{X}\). By definition of the semigroup generated by \(L_{0,\beta}\), for every \(t \ge 0\) and every \(1 \leq i,j \leq r\),
\[
e^{tL_{0,\beta}} L_{0,\beta} m_{ij}^{(N)} (\widehat{\boldsymbol{X}}_0) = \E_{\mathbb{Q}_{\widehat{\boldsymbol{X}}_0}} \left [ L_{0,\beta} m_{ij}^{(N)}(\widehat{\boldsymbol{X}}_t) \right ] .
\]
Using the expression of \(L_{0,\beta} \) in~\eqref{eq: generator Langevin noise} together with the formula for the Laplacian on the normalized Stiefel manifold~\eqref{eq: stiefel laplace}, we obtain
\[
L_{0,\beta} m_{ij}^{(N)}(\widehat{\boldsymbol{X}}_t) = \Delta m_{ij}^{(N)}(\widehat{\boldsymbol{X}}_t) - \frac{\beta}{N}  \langle \nabla H_0 (\widehat{\boldsymbol{X}}_t), [\boldsymbol{v}_i]_j  \rangle = - \frac{N-1}{N^2} \langle \widehat{\boldsymbol{X}}_t, [\boldsymbol{v}_i]_j  \rangle - \frac{\beta}{N}  \langle \nabla H_0 (\widehat{\boldsymbol{X}}_t), [\boldsymbol{v}_i]_j  \rangle,
\]
where \([\boldsymbol{v}_i]_j \coloneqq [\boldsymbol{0}, \ldots, \boldsymbol{0}, \boldsymbol{v}_i, \boldsymbol{0}, \ldots, \boldsymbol{0}] \in \R^{N \times r}\) denotes the matrix with all zero columns except the \(j\)th, which is \(\boldsymbol{v}_i\). Thus, for every \(t\ge 0\),
\begin{equation}\label{eq:semigroup-L0m}
e^{tL_{0,\beta}} L_{0,\beta} m _{ij}^{(N)}(\widehat{\boldsymbol{X}}_0) = -\frac{N-1}{N^2}  \E_{\mathbb{Q}_{\widehat{\boldsymbol{X}}_0}}  \left [ \langle \widehat{\boldsymbol{X}}_t, [\boldsymbol{v}_i]_j  \rangle \right ]  - \frac{\beta}{N}  \E_{\mathbb{Q}_{\widehat{\boldsymbol{X}}_0}}  \left [  \langle \nabla H_0 (\widehat{\boldsymbol{X}}_t), [\boldsymbol{v}_i]_j  \rangle\right].
\end{equation}

Let \(\boldsymbol{H} \in \R^{r \times r}\) be a matrix sampled from the Haar measure on \(O(r)\). For \(t \geq 0\), set 
\[
\boldsymbol{Z}_t \coloneqq R^N_{\widehat{\boldsymbol{X}}_0} (\nabla H_0(\widehat{\boldsymbol{X}}_t)) \boldsymbol{H}, \quad  \widetilde{\boldsymbol{Z}}_t  \coloneqq \widehat{\boldsymbol{X}}_t \boldsymbol{H}.
\]
By Lemma~\ref{lem: noise invariance} and invariance of the Haar measure, for each fixed \(t \ge 0\), the random matrices \(\boldsymbol{Z}_t\) and \(\widetilde{\boldsymbol{Z}}_t\) take values in \(\mathcal{S}_{N,r}\) and their laws are invariant under left and right multiplication by orthogonal matrices. Since this property uniquely characterizes the uniform measure on \(\mathcal{S}_{N,r}\), we conclude that for every \(t \ge 0\), both \(\boldsymbol{Z}_t\) and \(\hat{\boldsymbol{Z}}_t\) are distributed according to \(\mu_{N \times r}\). In particular, for any fixed \((i,j)\) and any \(t\ge 0\), the one-dimensional projections
\[
 \frac{1}{N}\langle \boldsymbol{Z}_t, [\boldsymbol{v}_i]_j \rangle   \quad \text{and} \quad \frac{1}{N} \langle \widetilde{\boldsymbol{Z}}_t, [\boldsymbol{v}_i]_j\rangle
\]
have exactly the same distribution as \(m_{ij}^{(N)}(\boldsymbol{X})\) under \(\boldsymbol{X}\sim\mu_{N\times r}\).

The remainder of the proof follows a similar approach to that used in~\cite[Lemma A.4]{benarous2024gradientflow}. Specifically, we establish uniform-in-time deviation inequalities over one-dimensional projections of \(\boldsymbol{Z}_t\) and \(\widetilde{\boldsymbol{Z}}_t\) by discretizing the interval \([0,T]\) and relating these inequalities to the quantities appearing in \(L_{0,\beta} m_{ij}^{(N)}(\widehat{\boldsymbol{X}}_t)\). By Lemma~\ref{lem: regularity H0} there exist constants \(\Gamma = \Gamma (p, r, \{\lambda_i\}_{i=1}^r)\) and \(K = K(p, r, \{\lambda_i\}_{i=1}^r)\) such that
\[
\mathcal{E} \coloneqq \left\{\norm{H_0}_{\mathcal{G}^2} \le \Gamma N \right\}
\]
satisfies
\[
\mathbb{P} (\mathcal{E}^\textnormal{c}) \le e^{-KN},
\]
where we remind the reader of the \(\mathcal{G}^n\)-norm on \(\mathcal{S}_{N,r}\) in Definition~\ref{def: G norm}. According to Definition~\ref{def: G norm} and the ladder relations of Lemma~\ref{lem: ladder relations}, we also notice that, on the event \(\mathcal{E}\), 
\begin{equation}\label{eq:regularity-bounds}
\|\nabla H_0\|_\infty \lesssim \Gamma \sqrt{N}, \quad \|\vert\nabla^2 H_0\vert_{\mathrm{op}}\|_\infty \le \Gamma.
\end{equation}
From now on we work on the event \(\mathcal E\) and all bounds are understood to hold \(\mathbb P\)-a.s.\ on \(\mathcal E\).\\

We first control the increments of the process \((\widetilde{\boldsymbol{Z}}_t)_{t \ge 0} = (\widehat{\boldsymbol{X}}_t \boldsymbol{H})_{t \ge 0}\). Define 
\[
\tilde{m}_{ij}^{(N)} (t) \coloneqq \frac{1}{N} \langle \widetilde{\boldsymbol{Z}}_t, [\boldsymbol{v}_i]_j\rangle = \frac{1}{N} \langle \widehat{\boldsymbol{X}}_t \boldsymbol{H}, [\boldsymbol{v}_i]_j\rangle.
\]
By Itô's formula, we obtain, for \(0 \le s \le t\),
\[
\tilde{m}_{ij}^{(N)} (t) - \tilde{m}_{ij}^{(N)} (s) = \frac{1}{N} \int_s^t \langle \boldsymbol{H}^\top [\boldsymbol{v}_i]_j, d\boldsymbol{B}_u \rangle + \int_s^t L_{0,\beta} \tilde{m}_{ij}^{(N)}(\widehat{\boldsymbol{X}}_u) du,
\]
where \((\boldsymbol{B}_t)_{t \ge 0}$ is a Brownian motion on \(\mathcal{S}_{N,r}\). Taking the expectation \(\mathbb{E}_{\mathbb{Q}_{\widehat{\boldsymbol{X}}_0}} \) and applying the Burkholder-Davis-Gundy inequality (see e.g.~\cite[Theorem 5.16]{legall}) to the martingale part, we obtain 
\[
\left \vert \mathbb{E}_{\mathbb{Q}_{\widehat{\boldsymbol{X}}_0}} \left[ \tilde{m}_{ij}^{(N)} (t) - \tilde{m}_{ij}^{(N)} (s) \right] \right \vert \leq r \sqrt{t-s} + \norm{L_{0,\beta}\tilde{m}_{ij}^{(N)}}_\infty(t-s).
\]
By Lemma~\ref{lem: ladder relations} and the bounds~\eqref{eq:regularity-bounds}, the norm \(\norm{L_{0,\beta}\tilde{m}_{ij}^{(N)}}_\infty\) is bounded by a constant depending only on \((\beta, p, r, \{\lambda_i\}_{i \le r})\). Thus, there exists \(C=C (\beta, p, r, \{\lambda_i\}_{i \le r})\) such that
\begin{equation} \label{eq:increment-tildeZ-final}
\left \lvert \mathbb{E}_{\mathbb{Q}_{\widehat{\boldsymbol{X}}_0}} \left [ \tilde{m}_{ij}^{(N)} (t) - \tilde{m}_{ij}^{(N)} (s) \right ] \right \vert \le C \left (\sqrt{t-s} + (t-s)\right),
\end{equation}
for every \(0\le s\le t\).

We now turn to the process \((\boldsymbol{Z}_t)_{t \ge 0} = \left ( R^N_{\widehat{\boldsymbol{X}}_0} (\boldsymbol{U}_t) \boldsymbol{H} \right)_{t \ge 0}\) with \(\boldsymbol{U}_t = \nabla H_0\left(\widehat{\boldsymbol{X}}_t\right)\). From the identity proved in~\cite[Lemma A.4]{benarous2024gradientflow}, we have
\[
\norm{\boldsymbol{Z}_t - \boldsymbol{Z}_s}_{\textnormal{F}} \leq \frac{r}{\sqrt{N}} \norm{\widehat{\boldsymbol{X}}_0+\boldsymbol{U}_t}_{\textnormal{F}} \norm{\boldsymbol{U}_t^\top\boldsymbol{U}_t - \boldsymbol{U}_s^\top\boldsymbol{U}_s}^{1/2}_{\textnormal{F}} + r \norm{\boldsymbol{U}_t-\boldsymbol{U}_s}_{\textnormal{F}},
\]
for every \(0\le s\le t\). Applying Itô's formula to the vector field \(\boldsymbol{U}_t\) yields 
\[
\boldsymbol{U}_t - \boldsymbol{U}_s = \int_s^t \nabla^2 H_0 (d\boldsymbol{B}_u, \cdot ) + \int_s^t L_{0,\beta} \nabla H_0(\widehat{\boldsymbol{X}}_u) du,
\]
where we used the shorthand \(L_{0,\beta}\nabla H_0(\widehat{\boldsymbol{X}}_u)\) to denote the vector field on \(\mathcal{S}_{N,r}\) whose coordinates are given, for any \(1 \leq k \leq N, 1 \leq \ell \leq r\), by
\[
[L_{0,\beta}\nabla H_0(\widehat{\boldsymbol{X}}_u)]_{k \ell} = L_{0,\beta} [\nabla H_0(\widehat{\boldsymbol{X}}_u)]_{k,\ell}.
\]
Using again the Burkholder-Davis-Gundy inequality, we obtain
\begin{equation}\label{eq:U-increment-bound}
\begin{split}
\E_{\mathbb{Q}_{\widehat{\boldsymbol{X}}_0}} \left [ \norm{\boldsymbol{U}_t-\boldsymbol{U}_s}_{\text{F}}\right ] & \leq \left(\int_s^t\norm{\nabla^2H_0}_{\text{F}}^2ds\right)^{1/2}+\norm{L_{0,\beta} \nabla H_0}_\infty(t-s) \\
& \leq \sqrt{Nr}\Gamma\sqrt{t-s}+\norm{L_{0,\beta} \nabla H_0}_\infty(t-s) \\
& \le C_1 (\Gamma,r) \sqrt{N} \sqrt{t-s} + C_2(\Gamma,r) \sqrt{N} (1+\sqrt{N}) (t-s),
\end{split}
\end{equation}
since
\begin{align*}
\norm{L_{0,\beta}\nabla H_0}_\infty & \leq \norm{ \vert \nabla^2H_0 \vert_{\textnormal{op}}}_\infty\norm{\vert \nabla H_0 \vert_{\textnormal{op}}}_\infty + \sqrt{Nr}\left(\sup_{1 \leq i \leq N \atop 1 \leq j \leq r} \left \lVert \Delta \left (\frac{\partial H_0}{\partial x_{ij}} \right) \right \rVert ^2_\infty\right)^{1/2} \\
&\leq \Gamma^2\sqrt{N} + \Gamma N\sqrt{r},
\end{align*}
where we used the ladder relation~\eqref{eq: ladder relation 1} in the last line. A similar computation applied to \(\boldsymbol{U}_t^\top \boldsymbol{U}_t -\boldsymbol{U}_s^\top\boldsymbol{U}_s\) gives
\begin{align*}
\boldsymbol{U}_t^\top \boldsymbol{U}_t -\boldsymbol{U}_s^\top\boldsymbol{U}_s & = 2\int_s^t\nabla^2H_0 \left(d\boldsymbol{B}_u,\nabla H_0(\widehat{\boldsymbol{X}}_u)\right) + 2\int_s^t L_{0,\beta} \nabla H_0(\widehat{\boldsymbol{X}}
_u)^\top\nabla H_0(\widehat{\boldsymbol{X}}_u) du,
\end{align*}
so that, applying Burkholder-Davis-Gundy inequality again for the martingale part, we obtain
\[
\mathbb{E}_{\mathbb{Q}_{\widehat{\boldsymbol{X}}_0}}\left[\norm{\boldsymbol{U}_t^\top \boldsymbol{U}_t -\boldsymbol{U}_s^\top\boldsymbol{U}_s}_{\text{F}}\right] \leq 2\Gamma^2\sqrt{N}\sqrt{t-s} + 2 \norm{L_{0,\beta} \nabla H_0(\widehat{\boldsymbol{X}}_u)^\top\nabla H_0(\widehat{\boldsymbol{X}}_u)}_\infty (t-s).
\]
Finally, to bound \(\norm{L_{0,\beta} \nabla H_0(\widehat{\boldsymbol{X}}_u)^\top\nabla H_0(\widehat{\boldsymbol{X}}_u)}_\infty\), we again use the ladder relation~\eqref{eq: ladder relation 1} again to control the Laplacian (applied coordinate-wise):
\[
\begin{split}
\norm{L_{0,\beta} \nabla H_0(\widehat{\boldsymbol{X}}_u)^\top\nabla H_0(\widehat{\boldsymbol{X}}_u)}_\infty & \leq \norm{\nabla^2 H_0(\nabla H_0,\nabla H_0)}_\infty + Nr \left(\sup_{1 \leq i \leq N \atop 1 \leq j \leq r} \left \lVert \Delta \left(\frac{\partial H_0}{\partial x_{ij}}\right)^2 \right \rVert_\infty\right)^{1/2} \\
&\leq \Gamma^3 N+\Gamma Nr
\end{split}
\]
Therefore, 
\[
\mathbb{E}_{\mathbb{Q}_{\widehat{\boldsymbol{X}}_0}}\left[\norm{\boldsymbol{U}_t^\top \boldsymbol{U}_t -\boldsymbol{U}_s^\top\boldsymbol{U}_s}_{\text{F}}\right] \leq 2\Gamma^2 \sqrt{N} \sqrt{t-s} + C_3 (\Gamma, r) N (t-s).
\]
Combining all together and using the fact that \(\norm{\widehat{\boldsymbol{X}}_0 + \boldsymbol{U}_t}_{\textnormal{F}} \le (1+\Gamma) \sqrt{N}\), we obtain
\[
\mathbb{E}_{\mathbb{Q}_{\widehat{\boldsymbol{X}}_0}} \left [ \norm{\boldsymbol{Z}_t - \boldsymbol{Z}_s}_{\textnormal{F}}  \right ] \le \tilde{C}_1 (\Gamma,r) \sqrt{N} \sqrt{t-s} + \tilde{C}_2(\Gamma,r) \sqrt{N} (1+\sqrt{N}) (t-s) + \tilde{C}_3 (\Gamma,r) N^{1/4} (t-s)^{1/4}.
\]

To conclude the proof, we follow the approach outlined in the proof of~\cite[Lemma A.4]{benarous2024gradientflow} for gradient flow dynamics. Specifically, we discretize the interval \([0,T]\) with a sufficiently fine partition to handle the polynomial increments in $N$. We then apply exponential concentration inequalities for one-dimensional projections of matrices sampled from the invariant measure on \(\mathcal{S}_{N,r}\) and conclude with a union bound.
\end{proof}
\printbibliography
\end{document}